%% file: arxiv_version.tex
\documentclass{article} %
\usepackage[margin=1in]{geometry}

\usepackage{iclr2023_conference,times}

\input{math_commands.tex}

\usepackage{hyperref}
\usepackage{url}
\usepackage{booktabs}       %
\usepackage{amsfonts}       %
\usepackage{nicefrac}       %
\usepackage{microtype}      %
\usepackage{xcolor}         %
\usepackage{enumitem}
\usepackage{bbold}
\usepackage{graphicx}
\usepackage{booktabs}
\usepackage{todonotes}
\usepackage{comment}
\usepackage{caption}
\usepackage{subcaption}
\usepackage{cleveref}
\usepackage{amssymb}
\usepackage{amsmath}
\usepackage{pifont}
\usepackage{wrapfig}
\title{BARTSmiles: Generative Masked Language \\Models for Molecular Representations}

\author{Gayane Chilingaryan$^1$\footnotemark[1]\ , Hovhannes Tamoyan$^1$\footnotemark[1]\ , Ani Tevosyan$^1$\thanks{Joint First Authors. Correspondence: \texttt{armenag@meta.com}}\ , Nelly Babayan$^{2,3}$, \vspace{-2cm} 
\And
Lusine Khondkaryan$^{2,3}$, Karen Hambardzumyan$^1$, Zaven Navoyan$^3$, Hrant Khachatrian$^{1,4}$, \vspace{-2cm}
\And Armen Aghajanyan$^5$ \\ \\
$^1$ YerevaNN, Yerevan, Armenia %
$^2$ Institute of Molecular Biology, NAS RA, Yerevan, Armenia \\
$^3$ Toxometris.ai, Yerevan, Armenia %
$^4$ Yerevan State University, Yerevan, Armenia \\
$^5$ Meta AI Research, Seattle, Washington, USA \vspace{-0.5cm}
}

\newcommand{\MODEL}{\texttt{BARTSmiles}}

\iclrfinalcopy %
\begin{document}

\maketitle
\begin{abstract}
We discover a robust self-supervised strategy tailored towards molecular representations for generative masked language models through a series of tailored, in-depth ablations. Using this pre-training strategy, we train \MODEL{}, a BART-like model with an order of magnitude more compute than previous self-supervised molecular representations. In-depth evaluations show that \MODEL{} consistently outperforms other self-supervised representations across classification, regression, and generation tasks setting a new state-of-the-art on 11 tasks. We then quantitatively show that when applied to the molecular domain, the BART objective learns representations that implicitly encode our downstream tasks of interest. For example, by selecting seven neurons from a frozen \MODEL{}, we can obtain a model having performance within two percentage points of the full fine-tuned model on task Clintox. Lastly, we show that standard attribution interpretability methods, when applied to \MODEL{}, highlight certain substructures that chemists use to explain specific properties of molecules. The code and the pretrained model are publicly available.
\end{abstract}

\begin{figure}[h!]
    \includegraphics[width=0.87\textwidth]{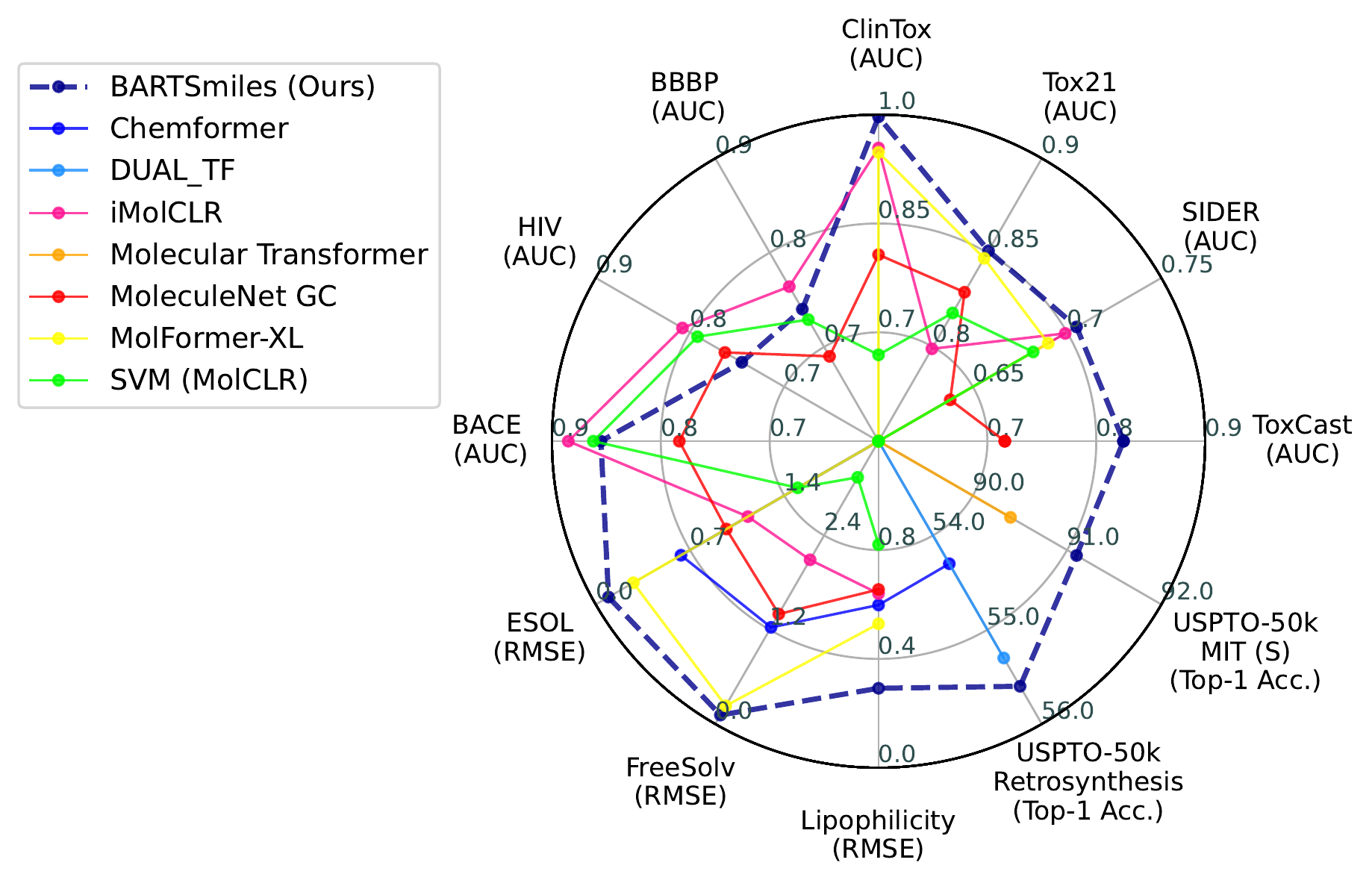}
    \caption{Performance of \MODEL{} on a variety of tasks compared to the current state-of-the-art.}
\end{figure}

\section{Introduction}
Recent advancements in large-scale representation learning have significantly improved downstream performance on virtually every modality, from images \citet{CM3, BEIT, MAE} to text \citet{ROBERTA, MARGE, HTLM1} to speech \citet{XLSR, WHISPER}. Meanwhile, the domain of molecular representations has only seen a fraction of the scale of these other domains. In this paper, we show that pre-training with masked language modeling objective on a large collection of molecules allows us to learn generalizable representations that perform well on a wide range of computational chemistry benchmarks.

We train \MODEL{}, a general purpose, pre-trained generative masked language model for molecules trained with an order of magnitude more compute than previous pre-trained molecular representations over all available molecules from the ZINC20 dataset (over 1.7 billion molecules \citep{ZINC20}).

To validate the efficacy of our representations, we do in-depth evaluations across a large amount of classification, regression, and generation tasks achieving a new state of the art on 11 tasks. We then explore the representations from a domain-expert perspective by contrasting the output of an attribution interpretability method with structural alerts, a rule-based system developed by chemists to explain particular molecular properties. In particular, for a model trained to predict the molecular toxicity, the Integrated Gradients method usually highlights atoms known to be part of the substructures responsible for toxicity \citep{integrated-gradients}.

Furthermore, our pre-trained model, in a completely unsupervised fashion, learned representations for specific tasks, i.e., certain neurons correlated highly with downstream tasks. For specific tasks, we see that a linear combination of frozen 7 to 15 neurons is enough to reach more than 90\% of the full fine-tuning performance.

To summarize our contributions are the following:
\begin{itemize}
    \item A robust pre-training strategy tailored towards molecular representations discovered through a series of in-depth ablations for generative masked language models.
    \item We present \MODEL{} pre-trained model for molecules trained with an order of magnitude more compute than previous pre-trained molecular representations. The model is publicly released at \url{https://github.com/YerevaNN/BARTSmiles/}.
    \item We have fine-tuned \MODEL{} on multiple chemical property prediction, chemical reaction prediction and retrosynthesis tasks, and set state-of-the-art results on 11 of them.
    \item We present a quantitative analysis of molecular properties learned during the pre-training process by analyzing individual neurons in the representations that correlate highly with downstream tasks. 
    \item A qualitative analysis using attribution interpretability methods finds that the model internally learns patterns known to in-domain experts to correlate with certain chemical properties.
\end{itemize}

\section{Related work}
\paragraph{Chemical Property Prediction:}
Machine learning has been applied to chemical property prediction since the 1960s \citep{qsar-history}. Initially, molecules were represented by fingerprints such as Morgan fingerprints \citep{morgan-fingerprints}, circular fingerprints \citep{circular-fingerprints} or extended-connectivity circular fingerprints (ECFP) \citep{ECFP-fingerprints}, which were treated as inputs to various machine learning models. Deep learning has been applied to fingerprints since at least 2014 \citep{NN-QSAR-2014}. \citet{CNN-on-molecules} was the first to apply convolutional networks directly on the graphs by generalizing the computation of circular fingerprints. In contrast, \citet{GCN-on-molecules} introduced graph convolutions on molecular graphs.

\citet{moleculenet} introduced the MoleculeNet benchmark suite, which enabled direct comparisons among machine learning algorithms for several classification and regression tasks. It became the standard evaluation system for most of the subsequent work. On the generative modeling front \citet{lowe-USPTO-2012} extracted a large set of chemical reactions from US patents, becoming a standard benchmark for evaluating the generation abilities of machine learning methods. 

\paragraph{Self-Supervised Learning for Molecular Representations:}
\citet{mol2vec} introduced Mol2vec, an adaptation of word2vec for molecules. Within the realm of self-supervised research, \citet{SMILES_BERT} was the first to apply BERT-like models on SMILES strings but unfortunately did not evaluate on MoleculeNet benchmarks. \citet{chemberta} further scaled up BERT-like models for molecules showing consistent improvement with scale. On the other hand, MolCLR \citet{MolCLR} and iMolCLR \citet{iMolCLR} use graph neural networks coupled with contrastive learning on 10 million unlabeled molecules to learn fine-tunable representations. Recently, \citet{reaction-aware-molRL} improved molecular representations by adding additional inductive biases through constraints on the sums of the representations of reactants. 

The work most related to ours is ChemFormer \citet{chemformer}, which also trains a BART-like model but, unlike our paper, does so with a fraction of the compute and scale. Additionally, through ablations, our paper explores the tweaks needed to the generative masked language model objective to make it optimal in the domain of molecular representations. We also present a unified fine-tuning recipe that removes the need for complex hyper-parameter tuning while consistently outperforming previous state-of-the-art models.

\section{Pretraining}
Below we present the pre-training setting used for training \MODEL{}.
\paragraph{Dataset:} We use the Zinc20 dataset from \citet{ZINC20} for all of our pre-training ablations and experiments. We deduplicate the data based on the hashes of canonicalized SMILES using RDKit \citep{RDKIT}, leaving us with a total of slightly north of $1.7$ billion samples from which we reserve 10000 samples as our validation set.

\paragraph{Implementation:}
\label{sec:model_implementation}
In all of our experiments, we parameterize our masked language model with the standard pre-layer norm Transformer architecture, precisely the BART-Large model proposed in \citet{BART}. We have a maximum sequence length per sample of 128 tokens for all models. We train using the FairSeq framework \citet{fairseq} with PyTorch \citet{pytorch} as the underlying framework. For our larger models, we use the fully sharded data-parallel implementation available in FairScale \citep{fairscale}. All experiments use automatic mixed-precision available in FairSeq. We use the Aim ecosystem \citep{AIM} for experiment tracking.

\paragraph{Choice Of Objective:}
When choosing an objective, there are a certain number of pre-conditions to consider depending on the end goal of the model. \citet{lm_objective_architecture} argues that for zero-shot or k-shot prompting, causal language models with uni-directional attention is optimal, while bidirectionality (in both context and attention masks) is the primary driver of success in the fine-tuning setting.

Within the fine-tuning setting, there are different objectives conditioned on model type. For encoder-based models, the masked language modeling objective as initially proposed in \citet{bert} and further refined in \citet{ROBERTA}. The downside of encoder models is the inability to do generative fine-tuning, which led to the introduction of the denoising model for the encoder-decoder models \citet{BART}. Decoder causal models are problematic because they are not bidirectional, although recently proposed objectives such as causal masking in \citet{CM3} are bidirectional in context but not in attention.

For \MODEL{} we select both the denoising objective and architecture from \citet{BART} while previous works have focused on encoder-only \citep{SMILES_BERT, chemberta}.

\subsection{Ablation}
Given the novelty of the molecular representation domain for pre-training, we aim to do an in-depth ablation analysis on how to successfully pre-train within this domain.

\begin{wrapfigure}{r}{0.6\textwidth}
  \centering
    \includegraphics[width=0.6\textwidth]{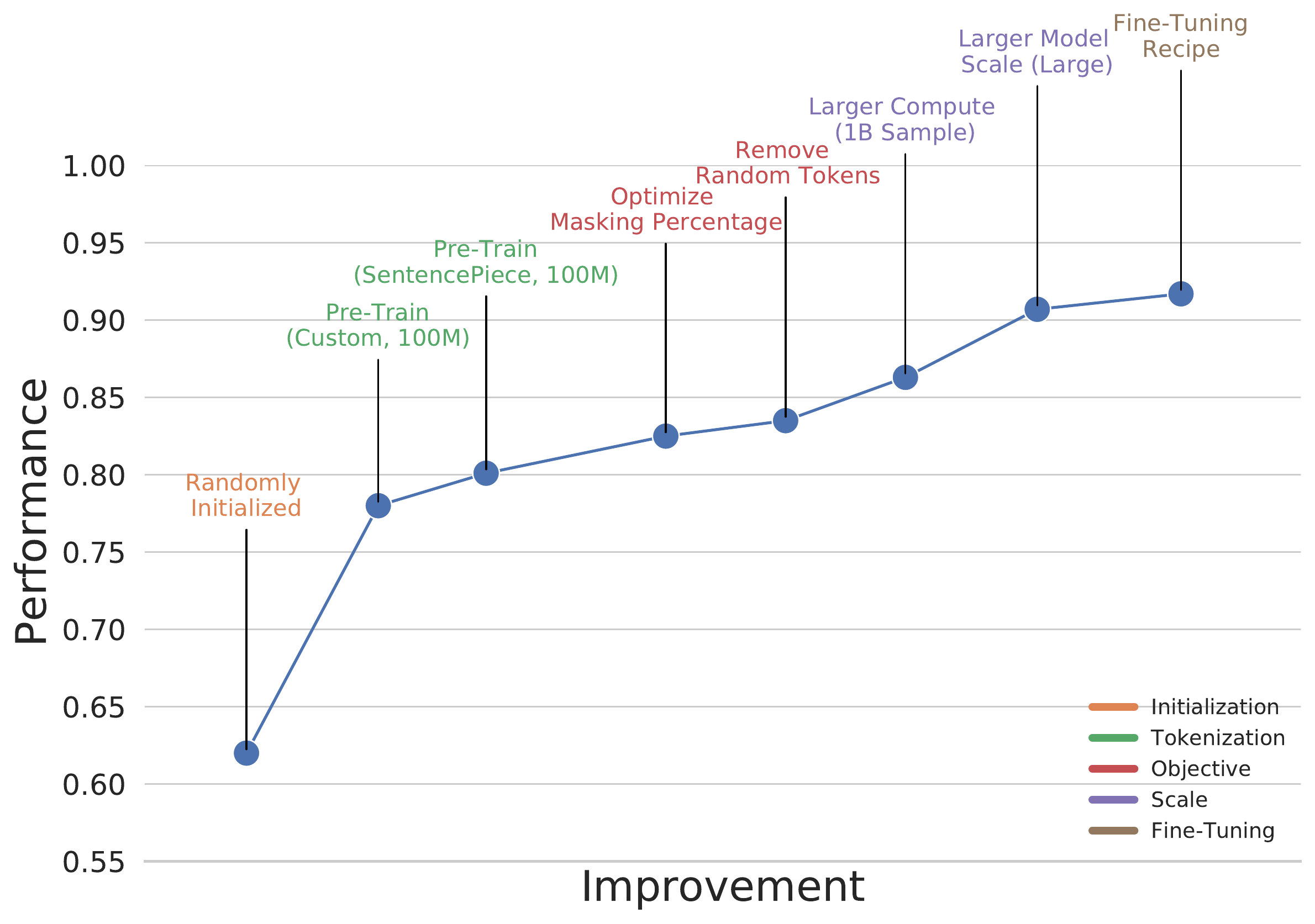}
    \caption{A summary of decision made when pre-training \MODEL{}, as well as their absolute gain on our ablation benchmark. All numbers are AUC-ROC on the three datasets.}
    \label{fig:ablation_summary}
\end{wrapfigure}

We measure the quality of learned representations by fine-tuning with a fixed set of hyper-parameters on three datasets, HIV from \citet{moleculenet}, BBBP from \citet{bbbp}, and ClinTox from \citet{moleculenet}, reporting the average of AUC-ROC scores. Note that we have selected these datasets before our final fine-tuning experiments. For the exact fixed hyper-parameters please refer to Table \ref{tab:hyperparams-pretrain} in Appendix~\ref{sec:hp}. We provide an overarching summary of our ablation in Figure~\ref{fig:ablation_summary}.

\paragraph{Choice of Tokenization:}
Previous work, such as \citet{SMILES_BERT} uses pre-defined tokenization rules on SMILES usually separating individual logical elements such as atoms (i.e., \textit{C}, \textit{N}, \textit{H}) and chemical bond symbols like (i.e., \textit{\#}, \textit{-}, \textit{=}). These fundamental units for molecules are analogous to using characters in natural language. Recent advancements in natural language pre-training have ubiquitously used sub-word tokenization as a compromise between the full character coverage of character level tokenization and the need for large vocabularies for word-based tokenization. We train a unigram tokenizer on a random sample of 10 Million samples from our training set with a 1021 vocabulary size using SentencePiece \citep{sentencepiece}. 

To ablate the benefit of learned tokenization vs. hand-crafted tokenization, we train two equivalent models (\S\ref{sec:model_implementation}), using the \MODEL{}-Base architecture. We fix the amount of compute for each run to the amount needed for one epoch, using learned tokenization with \MODEL{}-Base on 100M samples. Since our hand-crafted tokenization produces more tokens, given a fixed compute budget, the model trained with this tokenization sees roughly 92M samples. Each model was trained on 128 A100 GPUs. As a baseline, we also fine-tune a randomly initialized model of the same architecture and size as \MODEL{}-Base. We present our results in Table~\ref{table:tokenization_ablation}.

\begin{table}[h]
\centering
\begin{tabular}{@{}llll@{}}
\toprule
            & Custom Tokenization (Random) & Custom Tokenization & Learned Tokenization \\ \midrule
Performance & 0.628            & 0.779                     &   \textbf{0.801}           \\ \bottomrule
\end{tabular}
\caption{Performance (average AUC-ROC over three datasets) on ablation benchmark dataset over a randomly initialized network, a network pre-trained with hand-crafted tokenization rules, and a network pre-trained with SentencePiece tokenization.}
\label{table:tokenization_ablation}
\end{table}

We significantly improve pre-training representations by using a learned tokenization method, and given that learned tokenization is more efficient (due to smaller context lengths) and pre-trains better, we select using a learned tokenizer for the rest of our experiments.

\paragraph{Masking Strategy in Objective:}
We selected the denoising objective using encoder-decoder models from \citep{BART} to learn fine-tunable representations for classification, regression, and generative tasks. Fundamentally the denoising objective has the following hyper-parameters \texttt{random\_mask} (percent of tokens flipped to \texttt{<mask>}), $\lambda$ (Poisson distribution for length of masked span), \texttt{randomize\_tokens} (whether or not to randomize specific tokens, if true, 50\% of \texttt{random\_mask} is used for this). In conjunction with a \texttt{mask\_token\_budget} (the percent of tokens per sample allowed to be changed or masked), these three parameters cover the surface area of our ablation.

Given the computational cost of individual pre-training runs, even on the scale of \MODEL{}-Base, we are unable to do a proper grid search; therefore, we aimed first to find an optimal \texttt{mask\_token\_budget} (Table~\ref{table:ablation_mask_token}), then for that given budget, find optimal \texttt{random\_mask} and $\lambda$ (Table~\ref{table:ablation_lambda}), and finally given all the previous parameters we ablate whether or not to include to randomize specific tokens (Table~\ref{table:ablation_randomize}).

\input{sections/pretrain-ablation-tables}

In general, we see that the standard hyper-parameter of 0.3 for the \texttt{mask\_token\_budget} as proposed in \citet{BART} is suboptimal by a significant margin, and the use of randomizing tokens is suboptimal as well. The type of masking beyond that seems to not be of importance.
\paragraph{Importance of Pre-Training Scale}
As the last step in our ablation, we significantly increased the amount of compute to do a single pass over the full Zinc20 dataset, which required 20 hours on 1024 A100 GPUs. The previous model utilizing the most compute used 3330 V100 hours \cite{MolFormer}, while our model was trained for 20480 A100 hours. Training on the full dataset improves performance by five absolute points compared to the same setting trained on 100 million samples.

\section{Fine-tuning}
\subsection{Datasets}

To evaluate our pre-trained models, we fine-tune on multiple classification and regression datasets, mostly from the MoleculeNet benchmark suite \citep{moleculenet}.

MoleculeNet is a collection of 17 commonly used datasets for evaluating chemical representations. MoleculeNet defines splitting methods and evaluation metrics for all datasets. We have run experiments on 10 of them (excluding quantum mechanical datasets and the huge datasets PCBA and MUV). We use the AUC-ROC score to evaluate the classification models. Three of the selected datasets are regression tasks evaluated with RMSE. Four out of seven classification tasks have multiple target labels for each sample. We treat each target variable as a separate task and report the mean of the AUC scores. The remaining three are single-label classification tasks, and scaffolds of the molecules (the two-dimensional structural frameworks of the molecules as defined by \citet{scaffold1996}) define their training-test splits. This implies some distribution shift in the test set, which introduces additional difficulties in model selection.

We additionally perform fine-tuning on two datasets developed for toxicology. The first one is called the Ames dataset \citet{Ames}, which holds the results of a bacterial reverse mutation test, commonly known as the Ames test. And the second one holds the results of another genotoxicity test, namely the micronucleus (MN) assay~\citep{Micronucleus}. MN assay is a cytogenetic test to assess the genotoxicity of chemicals and physical factors on a chromosomal level.  %
We use cross-validation for the MN assay dataset to select the best hyperparameters and apply the best model to an external dataset of 65 compounds to obtain the final score. This makes our results comparable to the baselines introduced by \citep{Ames}. 

\begin{table}[]
\vspace{-3em}
\centering
\scriptsize
\begin{tabular}{llllllll}
\toprule
              & \multicolumn{1}{l}{SIDER} & \multicolumn{1}{l}{ClinTox} & \multicolumn{1}{l}{Tox21} & \multicolumn{1}{l}{ToxCast}        & \multicolumn{1}{l}{HIV}                   & \multicolumn{1}{l}{BACE}                  & \multicolumn{1}{l}{BBBP}                \\
\midrule
RF MolCLR           & 0.684                     & 0.713                       & 0.769                     & -                                  & 0.781                 & 0.867                 & 0.714                 \\
SVM  MolCLR         & 0.682                     & 0.669                       & 0.818                     & -                                  & 0.792                 & 0.862                 & 0.729                 \\ \midrule
MoleculeNet GC      & 0.638 & 0.807     & 0.829     &   0.716     &   0.763   & 0.783   & 0.690  \\
D-MPNN        & 0.646                     & 0.894                 & 0.845                           & 0.737                              & *0.816                 & *0.875                 & *0.913                 \\
Attentive FP  & 0.637                     & 0.940                        & \textbf{0.858}            & 0.805          & *0.832            & *0.850                 & *0.920              \\
3D Infomax    & 0.534                     & 0.594                        & 0.745                    & 0.644          & 0.761             & 0.794                  & 0.691              \\ \midrule
ChemBERTa 10M & \multicolumn{1}{c}{-}     & \multicolumn{1}{c}{-}       & \multicolumn{1}{c}{-}     & -                                  & 0.622                 & \multicolumn{1}{c}{-} & 0.643                 \\
MoLFormer-XL  & 0.690                     & 0.948                       &  0.847  & -                                  & *0.822                & *0.882                 & *0.937                 \\
GIN           & 0.627                     & 0.726                       & 0.781                     &  0.657                                   & \textbf{0.799}         & 0.845                 & 0.687                 \\
GROVER large  & 0.658                     & 0.944                       & 0.831                     & 0.737          & -                     & *0.894                 & *0.940                 \\
MolCLR        & 0.680                     & 0.932                       & 0.798                     & -                                  & \textbf{0.806}        & \textbf{0.890}            & 0.736                 \\
iMolCLR       & \textbf{0.699}            & 0.954                       & 0.799                     & -                                  & \textbf{0.808}        & \textbf{0.885}            & \textbf{0.764}         \\\midrule
\MODEL{} & \textbf{0.705}                     & \textbf{0.997}              & \textbf{0.851}                     & \textbf{0.825} & 0.745                 & 0.855                 & 0.74                 \\ \bottomrule
\end{tabular}
\caption{Results on the classification datasets from MoleculeNet. All numbers are AUC-ROC, higher is better. We bold the state-of-the-art numbers as well as all numbers within a 0.01 range. * indicates a non-standard split.}
\label{tab:results-classification}
\end{table}
\subsubsection{Recipe}

We fine-tune our pre-trained model with Fairseq~\citep{fairseq}, a sequence modeling toolkit based on PyTorch. We use the recommended hyperparameters for the GLUE benchmark with a few modifications \citep{GLUE}. In particular, we set maximum input and output lengths to $128$, changed dropout values, and set the gradient clipping norm to $0.1$. We skip inputs longer than $128$ and set the initial fp16 scaling to $128$. We use batch size $16$. We refer to Table~\ref{tab:hyperparams-finetune} in Appendix~\ref{sec:hp} for the rest of the hyperparameters.

We train for ten epochs and use linear warmup, with the peak occurring at 16 percent of the training. After the peak. For each dataset, we do a fixed grid search over dropout and learning rate parameters with the values $\left[0.1, 0.2, 0.3\right]$ and $\left[5\mathrm{e}{-6}, 1\mathrm{e}{-5}, 3\mathrm{e}{-5}\right]$, respectively. Finally, we apply Stochastic Weight Averaging (SWA) on three sets of four checkpoints: around the checkpoint having the best validation loss, the best validation accuracy, and the last checkpoint \citep{SWA}. We select the best one according to the performance on the validation set.

For multi-task problems, we train individual models for each task. We perform the grid search only on the first four tasks, pick the best hyperparameters and apply them to all tasks. We also tried to train a single multi-task classification model for each dataset to cover all tasks simultaneously, but the results were significantly worse (e.g. by 0.25 points on Tox21 and by 0.1 points on SIDER). 

\begin{table}[h]
\centering\scriptsize
\begin{subfigure}{.56\textwidth}
\centering
\begin{tabular}{lcccc}
\toprule
        & ESOL       & FreeSolv   & Lipophilicity   & Avg \\ \midrule
RF MolCLR            & 1.070      & 2.030      & 0.880          & 1.327    \\
SVM MolCLR           & 1.500      & 3.140      & 0.820          & 1.820    \\ \midrule
MoleculeNet GC      & 0.97      & 1.40       & 0.655  & 1.008       \\
D-MPNN      & 0.980      & 2.180      & 0.650            & 1.270    \\
Attentive FP      & 0.503      & 0.736      & 0.578           & 0.606  \\
3D Infomax  & 0.894      & 2.337      & 0.695    & 1.309      \\
\midrule
Chemformer  & 0.633      & 1.230      & 0.598           & 0.820     \\
MoLFormer-XL      & 0.279      & 0.231      & 0.529           & 0.346 \\ GROVER large      & 0.831      & 1.544      & 0.560           & 0.978 \\
MolCLR            & 1.110      & 2.200      & 0.650           & 1.320    \\
iMolCLR           & 1.130      & 2.090      & 0.640           & 1.287    \\ \midrule
\MODEL{}     &  \textbf{0.095}          &  \textbf{0.114}          &   \textbf{0.292}           & \textbf{0.167} \\ \bottomrule
\end{tabular}
\caption{Regression tasks from MoleculeNet. All numbers are RMSE, lower is better. }
\label{tab:results-regression}
\end{subfigure}\qquad
\begin{subfigure}{.4\textwidth}
\centering
\begin{tabular}{lr}
        \toprule
                           & AUC   \\ \midrule
        FP\_Pubchem\_SVM   & 0.948 \\
        FP\_MACCS\_RF      & 0.947 \\
        Descriptor\_SVM            & \textbf{0.952} \\
        Descriptor\_RF             & 0.933 \\ \midrule
        \MODEL{} & 0.914 \\
        \MODEL{} (Frozen) + ECFP + SVM & \textbf{0.948} \\ \bottomrule
\end{tabular}
\caption{Micronucleus assay dataset} 
        \begin{tabular}{lc}
        \toprule
                                  & \multicolumn{1}{c}{{AUC}} \\ \midrule
        SVM                       & 0.86  \\
        GP                         & 0.84    \\
        RF                         & 0.83    \\
        k-NN                       & 0.79    \\ \midrule
        \MODEL{}                   & 0.830   \\
        \MODEL{} (Frozen) + ECFP + SVM        & \textbf{0.869}   \\ \bottomrule          
        \end{tabular}
\caption{Ames dataset (cross-validation)}\label{tab:results-tox}
\end{subfigure}

\caption{Results on MoleculeNet regression tasks and two toxicology datasets.}
\end{table}

\subsection{Baselines}
MoleculeNet paper introduced its own set of baselines using multiple machine learning methods. MoleculeNet did not apply most baselines to all datasets. Among the ones applied to all datasets, the convolutional graph baseline performed the best, and we have included it as our baseline.
The other graph-based supervised baselines include %
D-MPNN~\citep{DMPNN-original}, Attentive FP~\citep{AttentiveFP} and the 3D InfoMax method~\citep{3dinfomaxGNN}.

We also compare with baselines that use self-supervised pretraining. We skip SMILES-BERT, as SMILES-BERT did not test it on MoleculeNet. ChemBERTa~\citep{chemberta}, Chemformer~\citep{chemformer} and MolFormer-XL~\citep{MolFormer} are based on regular text-based transformers. Graph Isomorphism Networks (GIN) with Context Prediction from \cite{GIN}, GROVER-Large from \citet{GROVER}, MolCLR~\citep{MolCLR} and iMolCLR~\citep{iMolCLR} are self-supervised methods based on graph neural networks. For Micronucleus Assay and Ames datasets we compare with the baselines introduced in the original papers.

\vspace{-0.5em}
\subsubsection{Results}

Our model produced state-of-the-art results on all three regression tasks of the MoleculeNet suite, as seen in Table \ref{tab:results-regression}. %
On classifications tasks the results are mixed (Table \ref{tab:results-classification}). On Clintox and Toxcast datasets, our model beats all previous models. Toxcast is a huge dataset with 617 tasks, so many papers do not include results. On SIDER and Tox21, the performance of our models are comparable to the state-of-the-art. %

The other three classification tasks (HIV, BACE, and BBBP) require specific scaffold splits and are sensitive to the choice of the split. Unfortunately, many papers do not specify whether they used the split recommended by MoleculeNet.  %
\citet{GNN-drug-comparison} claim D-MPNN and Attentive FP results are based on random splits. MolFormer and GROVER-large use scaffold splits, but the splits are different.
We explored the split's impact on the BBBP dataset's results.

Molecule scaffolds induce 1024 clusters of 1940 molecules in the BBBP dataset. The split is organized so that all non-singleton clusters belong to the training set, while the validation and test sets contain only molecules with unique scaffolds. Most of the errors of our model are concentrated on roughly 30\% of the singleton clusters in the test set. This implies that even a scaffold split with a different random seed can significantly affect the results. In Table~ \ref{tab:results-classification}, we have marked the results of these three classification datasets with an asterisk if we believe the authors used a different split. 

The results for the two toxicology datasets that direct fine-tuning did not beat the baselines based on fingerprints and descriptors. To test whether the fingerprints still contain helpful information not covered by \MODEL{}, we fine-tuned a classifier head that takes 2048-dimensional ECFP4 fingerprints \citep{ECFP-fingerprints} as an input in addition to the [CLS] vector. We also froze \MODEL{}, averaged the last layers' representations per molecule, concatenated them with ECFP fingerprints, and ran the default SVM implementation from \citet{scikit}. This helped close the gap with the baselines, as seen in Table~\ref{tab:results-tox}. Fundamentally this implies that fingerprints carry some information that is not trivially recoverable from our pre-trained representation. We do a deeper analysis of our pre-training dataset in \S~\ref{section:dataset_rep_space} and hypothesize that the Zinc20 dataset is not representative of the downstream tasks computational chemistry is interested in.

\subsection{Generative tasks}
We follow a straightforward recipe for generative fine-tuning for all generative tasks, mainly derived from \citet{RXF}. Similarly to our results on SIDER, we found that training in full 32-bit precision (fp32) significantly impacted downstream results. We use the R3F method from \citet{RXF}, running a sweep over the noise types and $\lambda$ regularization term for each task (See Table~\ref{tab:hyperparams-generative} in Appendix~\ref{sec:hp}). Consistent with our fine-tuning recipe for classification and regression tasks, we utilize SWA and do not do any augmentation on the data. For additional stability, we initialize our fine-tuning optimizer (Adam) with the pre-training moving averages of the gradient and squared gradient from our pre-training. We also use a polynomial decay strategy where the peak learning rate is at $0.06*\text{max\_updates}$ where the max number of updates is the number of updates needed to do ten epochs for the respective dataset.

During the evaluation, we implement two different strategies for sequence generation. The first runs a straightforward beam search with a beam size of ten. The second approach samples 128 samples from the natural distribution of tokens from the model (temperature of 1.0) and then re-ranks the list of 128 samples based on the perplexity of the entire predicted SMILES.

We select two common generative molecular tasks; retrosynthesis and chemical reaction prediction.
\subsubsection{Retrosynthesis}
Retrosynthesis is a chemical synthesis technique involving the deconstruction of a target molecule into its starting materials to assess the best synthetic route. Retrosynthesis is a cornerstone of modern organic product synthesis, traditionally the domain of human experts' knowledge. Due to the algorithmic advances and the availability of large chemical reaction collections, computational approaches have been suggested to solve the problems in retrosynthesis \citep{davey2018retrosynthesis,watson2019retrosynthetic}. We apply our generative fine-tuning strategy on the USPTO dataset \citep{lowe-USPTO-2012}. Table~\ref{table:retrosynthesis} shows the ratio of molecules for which the correct reactants were among the top K predictions generated by the respective models. Our approach has the best result for the Top-1 metric.

\begin{table}
\centering%
\begin{tabular}{@{}llll@{}}
\toprule
& Top-1 & Top-5 & Top-10 \\ \midrule
RetroComposer \citep{retrocomposer} & 53.3  & 80.9  & 85.0  \\
ATx100 \citep{augmented_transformer} & 53.5 & \textbf{81.0} & \textbf{85.7} \\
GraphRetro \citep{graph_retro} & 53.7 & 72.2 & 75.5 \\
ChemFormer \citep{chemformer} & 54.3 & 62.3 & 63.0 \\
Dual-TF \citep{retrosynthesis_energy} & \textbf{55.3} & 73.0 & 75.0 \\ \midrule
\MODEL{} Beam-10 & \textbf{55.2} & 68.5 & 73.2 \\ 
\MODEL{} Sample-128 + PPL ReRank & \textbf{55.6} & 74.2 & 80.9 \\\midrule
\end{tabular}
\caption{We present our experimental results on the USPTO-50k retrosynthesis task across varying sampling strategies. We bold the best performing numbers within a 0.5 range.}\label{table:retrosynthesis}
\end{table}

\subsubsection{Chemical Reaction Prediction}
The dual problem of retrosynthesis is chemical reaction prediction which is the automatic construction of the target molecule given the set of reactant molecules. This can also be treated as a sequence-to-sequence task, as first shown by \citet{reaction-prediction-seq2seq}, and can be performed by fine-tuning \MODEL{} on a relevant dataset. We follow the setup presented by \citet{molecular-transformer} and apply our generative fine-tuning recipe to several subsets of the USPTO dataset. As seen in Table~\ref{table:chemical_reaction_prediction}, \MODEL{} outperforms all baselines across all subsets.

\begin{table}[h]
\scriptsize
\centering
\begin{tabular}{@{}lllllll@{}}
\toprule
USPTO Split       & MIT (S)  & MIT (M) & LEF (S)  & LEF (M) & STEREO (S) & STEREO (M) \\ \midrule
S2S \citep{reaction-prediction-seq2seq}     & 80.3 &     &      &     & 65.4   &        \\
WLDN \citep{jin2017predicting}    & 79.6 & 74  & 84.0 &     &        &        \\
ELECTRO \citep{bradshaw2018generative} &      &     & 87.0 &     &        &        \\
GTPN \citep{do2019graph}    & 82.4 &     & 87.4 &     &        &        \\
WLDN5 \citep{coley2019graph}   & 85.6 &     & 88.3 &     &        &        \\
Transformer \citep{molecular-transformer}     & 90.4 & 88.6  & \textbf{92.0} & 90.3  & 78.1 & 76.2 \\ \midrule
\MODEL{} Beam-10 & \textbf{91.8} & \textbf{89.1} & \textbf{92.1} & \textbf{92.8} & \textbf{82.5} & \textbf{82.1}\\
\MODEL{} Sample + ReRank & 91.1 & \textbf{89.1} & 91.8 & 90.2 & 80.4 & 81.5 \\\bottomrule
\end{tabular}
\caption{We present our experimental results on the USPTO chemical prediction task across varying sampling strategies and three data subsets (MIT, LEF and STEREO). (S) signifies the ``split" and (M) signifies the ``mixed" sampling strategy from \citet{molecular-transformer}). We bold the best performing numbers within a 0.5 range. The baselines scores are taken from \citet{molecular-transformer}. }\label{table:chemical_reaction_prediction}
\end{table}

\section{Interpretability}

\subsection{Uncovering the key features}
\citet{radford2017learning} showed that a large language model trained to generate text had a particular neuron correlated with the sentence's sentiment, colloquially dubbed the sentiment neuron.  Further work showed that self-supervised representations had a very low intrinsic dimension for downstream tasks, arguing that self-supervised representations were capable of learning complex tasks implicitly \citep{intrinsic_dimensionality_finetuning}. For each dataset, we extracted the representations of each molecule by taking the average of the last layer representations of all tokens given by the pre-trained model and trained an L1-regularized logistic regression model to predict the label. We indirectly control the number of selected features by changing the regularization strength. The results are shown on Fig. \ref{fig:num_features_vs_auc_roc}. %

Clintox tasks are among the easiest ones. As seen in Figure~\ref{fig:num_features_vs_auc_roc}, a single unsupervised neuron can predict the label of its first subtask with a $0.77$ ROC-AUC score, while with seven neurons, one can achieve $0.987$ score. It is interesting to note that for the regularization parameter $C \in \{2^{-6},2^{-7},2^{-8}\} $, both subtasks of Clintox select the same set of features.

This might indicate that there are features learned in a completely unsupervised fashion that are highly correlated with the labels of these datasets, at least on the corresponding validation sets of the datasets.

\subsection{Datasets in the Representation Space}
\label{section:dataset_rep_space}

\begin{figure}[h]
\centering
\begin{minipage}[t]{0.45\columnwidth}
    \centering
    \includegraphics[width=0.9\columnwidth]{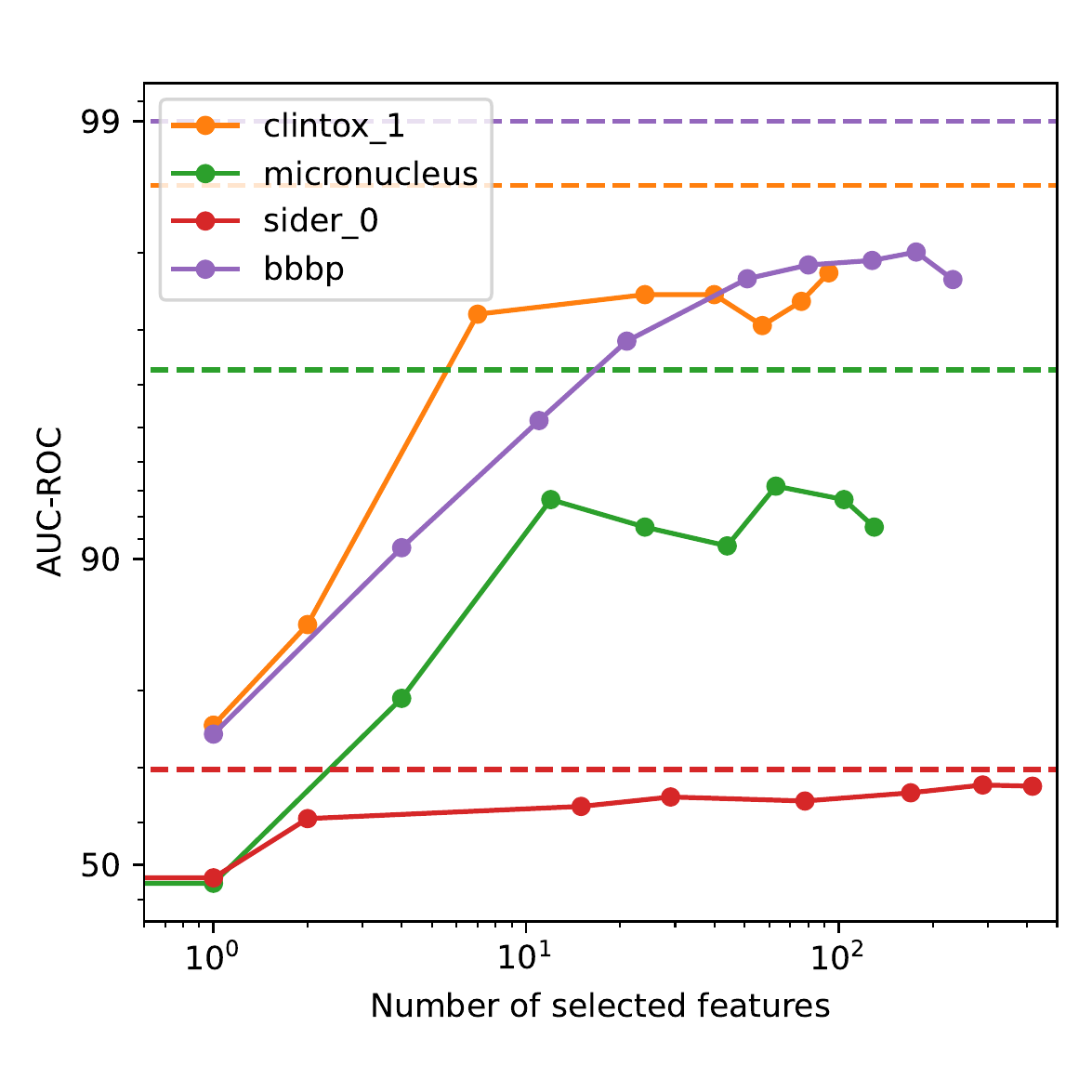}
    \caption{\small The AUC-ROC score plotted against the number of self-supervised features selected from \MODEL{}. Dashed lines indicate the performance achieved on the same dataset with full fine-tuning of the model.}
    \label{fig:num_features_vs_auc_roc}
\end{minipage}%
\qquad
\begin{minipage}[t]{0.45\columnwidth}
    \centering
    \subfloat {{\includegraphics[width=0.9\columnwidth]{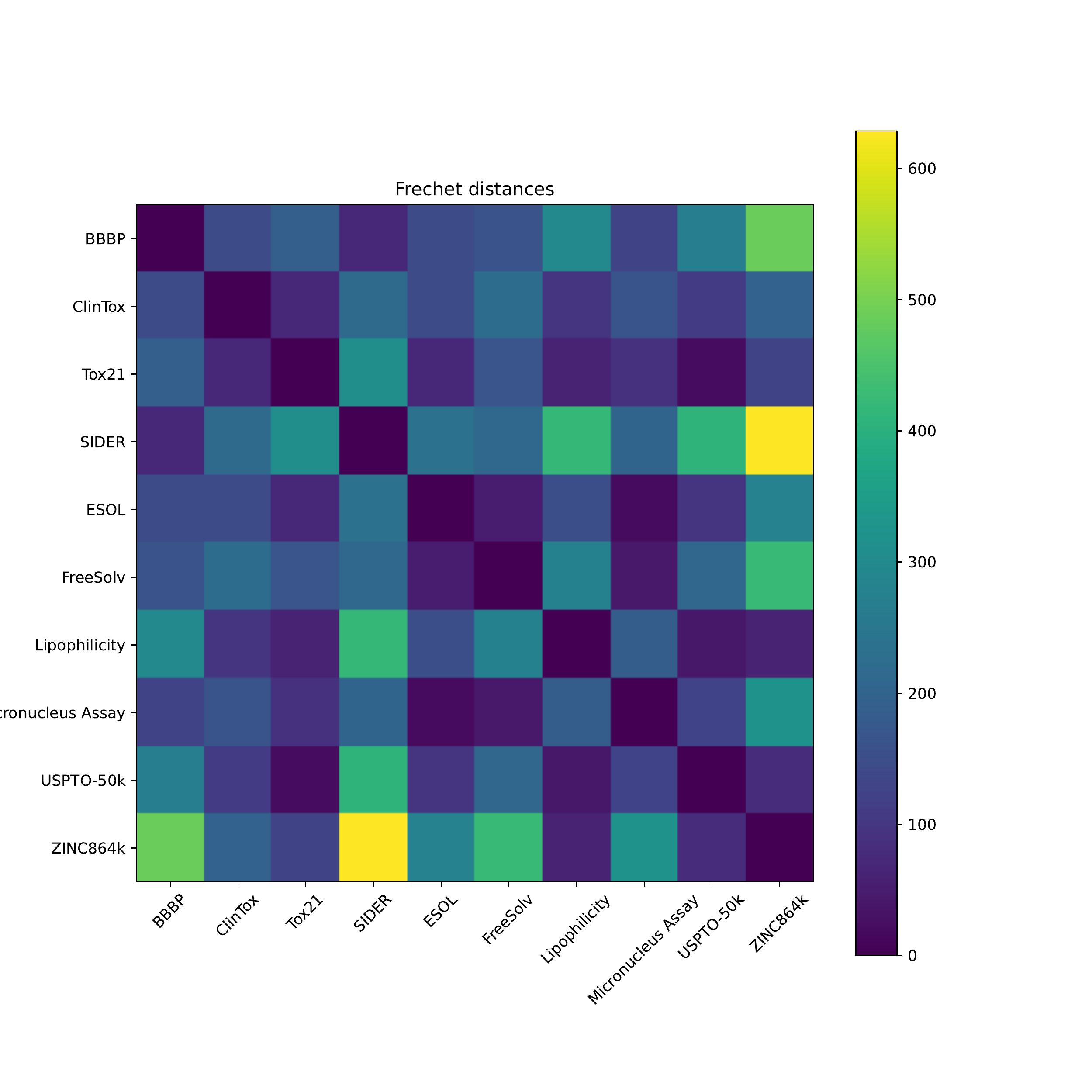}
    }}%
    \caption{\small Frechet distances between datasets in the representation space of \MODEL{}.}
    \label{fig:frechet}
\end{minipage}
\end{figure}

To explore the relative positions of various datasets in the \MODEL{} space, we compute the $1024$-dimensional representations of all molecules by taking the average of token representations, and compute Frechet distance~\citep{FrechetChemnet} between the datasets. As the ZINC dataset is vast, we uniformly sample $0.05\%$ of the molecules. As seen in Fig. \ref{fig:frechet}, several datasets like BBBP and SIDER are further from ZINC, which might explain a relatively poor performance of \MODEL{} compared to the models like iMolCLR, which do not use ZINC for pretraining. 

\subsection{Interpreting Fine-Tuned Models}\label{sec:captum}
\begin{figure}[h]
    \centering
    \begin{subfigure}[b]{0.35\textwidth} 
      \centering 
      \includegraphics[width=\textwidth]{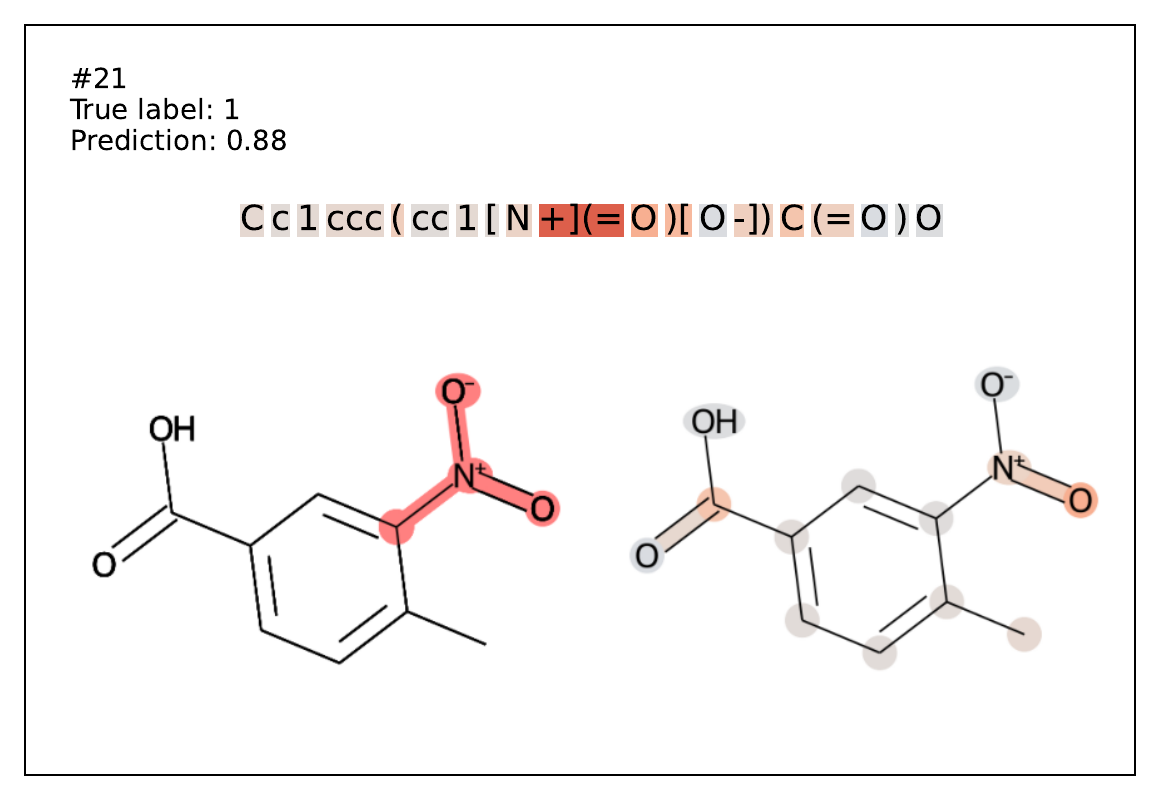} 
    \end{subfigure}\begin{subfigure}[b]{0.35\textwidth} 
      \centering 
      \includegraphics[width=\textwidth]{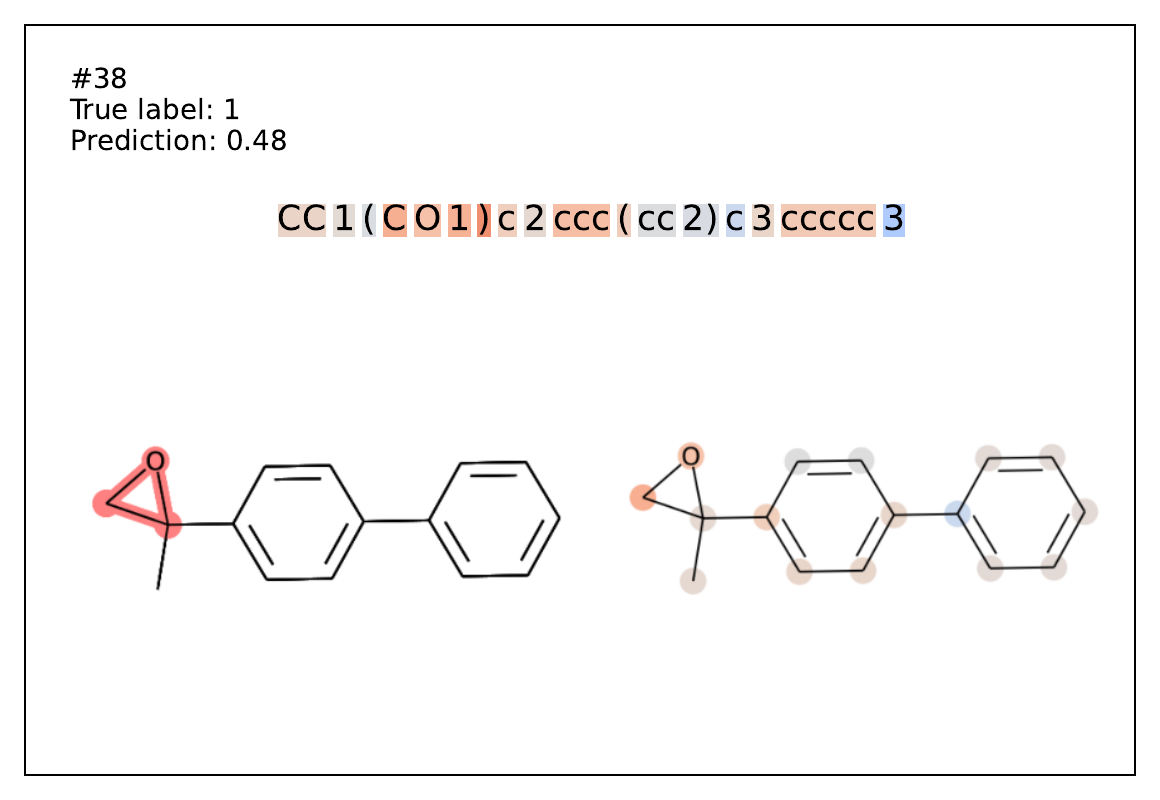} 
    \end{subfigure}
    \begin{subfigure}[b]{0.35\textwidth} 
      \centering 
      \includegraphics[width=\textwidth]{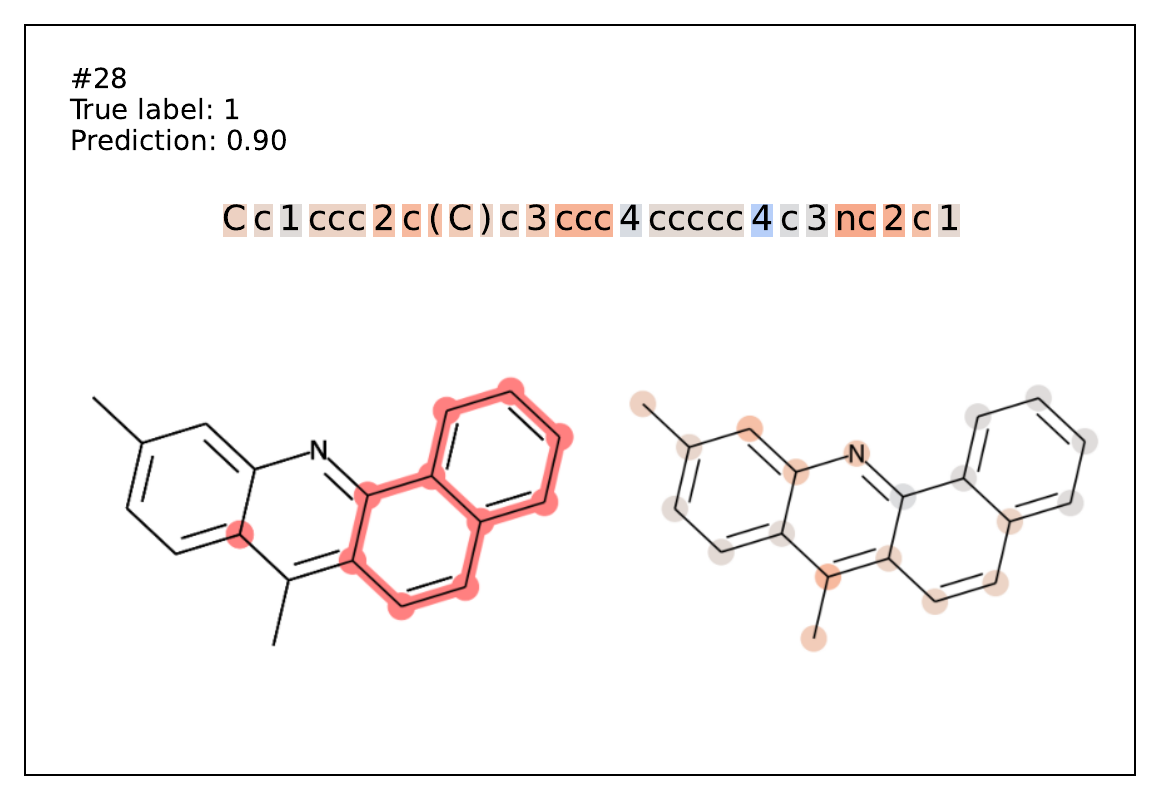} 
    \end{subfigure}\begin{subfigure}[b]{0.35\textwidth} 
      \centering 
      \includegraphics[width=\textwidth]{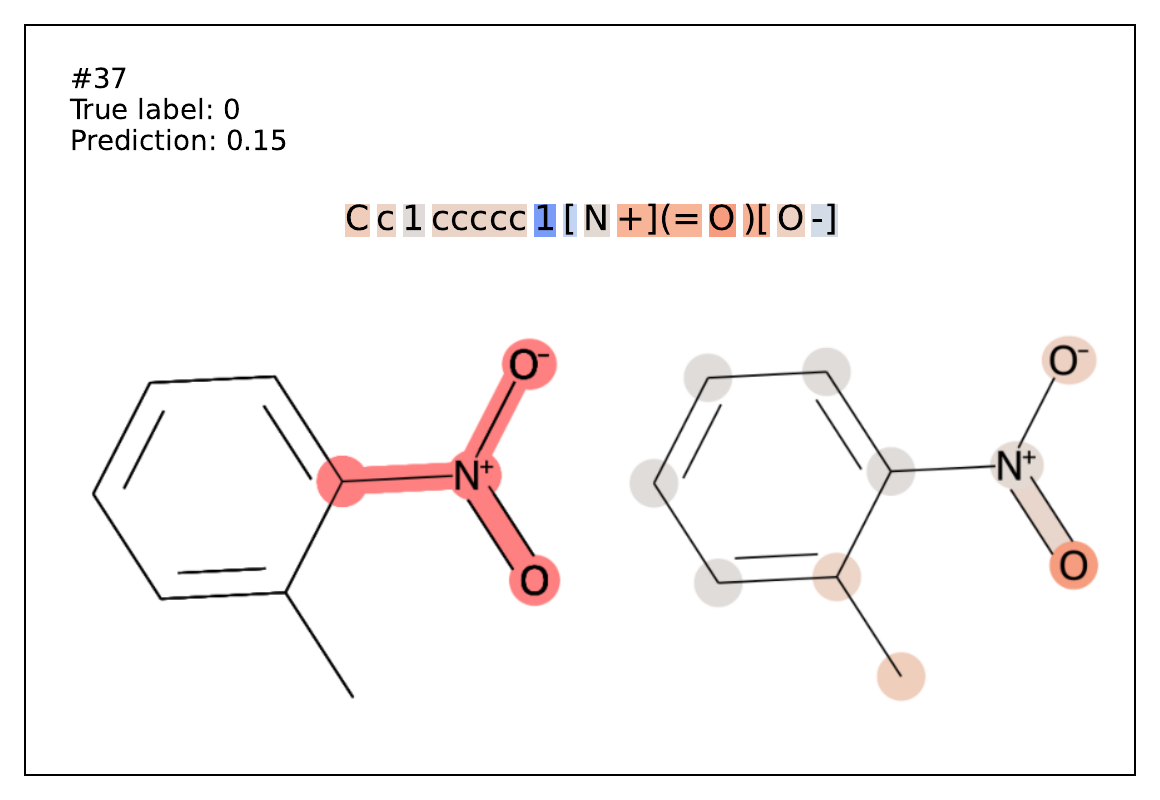} 
    \end{subfigure}
    \caption{The contributions of each token and each atom to the prediction of our model fine-tuned on the Ames dataset, according to the Integrated Gradient algorithm.}
    \label{fig:captum-ames-main}
\end{figure}
\paragraph{Methodology:}
In this section, we explore the contribution of each token of the given SMILES to the predictions of the fine-tuned model. We use the Integrated Gradients method \citep{integrated-gradients} implemented in Captum library \citep{captum} to get the contributions of the tokens. Note that due to the tokenization we used in pretraining, there is no one-to-one mapping between tokens and the atoms. Nevertheless, we attempt to visualize the contributions of each atom on the molecular graph. If a token contains multiple symbols, we assume the contributions are additive and split the contribution of the token equally for all symbols in the token. Thus, we get the contribution of each atom and highlight the atom in the graph accordingly. Note that the SMILES string contains many non-atom symbols like brackets. We do not visualize the contributions of the brackets. We also visualize the contribution of a double-bond symbol $=$ by highlighting the bond in the graph. 
\paragraph{Ames Dataset:}
\vspace{-1em}
For the Ames dataset, we fine-tuned another model on a random split, and explored the contributions of the atoms of the molecules from the validation set. We compared the highlighted atoms in the graph with the known molecular substructures related to the mutagenic properties of molecules, so-called structural alerts~\citep{tox-structural-alerts}. Figure \ref{fig:captum-ames-main} shows a couple of examples. On each of the plots, the left image highlights the structural pattern while the right one shows the contributions of each atom according to the Integrated Gradients method. Many more examples and a more thorough discussion can be found in Appendix \ref{sec:app-captum-ames}. 

The compound \#21 in Fig. \ref{fig:captum-ames-main} is a nitroaromatic compound representing organic molecules that consist of at least one nitro group (-NO2) attached to an aromatic ring. The structural alert in this molecule is the nitro group (highlighted on the left part of the figure), whose components have been highlighted %
as a contribution in most cases of this class of compounds. The compound \#38 on the same Figure is an epoxy group-bearing molecule. The epoxy group consists of an oxygen atom joined by single bonds to two adjacent carbon atoms, which form the three-membered epoxide ring ~\citep{tox-structural-alerts} and is a well-known structural alert. It is worth noting that in the case of nitroaromatic compounds, the components of the structural alerts have been detected, along with the high probability ($>0.8$) of correct prediction.

To summarize, the Integrated Gradients method, in most cases, recognizes %
contributions of non-complex substructures. In the case of more complex substructures, it provides explanations currently not understood through the lens of structural alerts. Additionally, the recognition of known substructures having the contribution to specific labels does not interfere with correct opposite label prediction, evidencing that the model's learning ability is much deeper than estimating simple correlations between molecular substructures and their effects.

\begin{figure}[h]
    \centering
    \begin{subfigure}[b]{0.49\textwidth}\label{fig:esol-stats}
      \centering 
      \includegraphics[width=\textwidth]{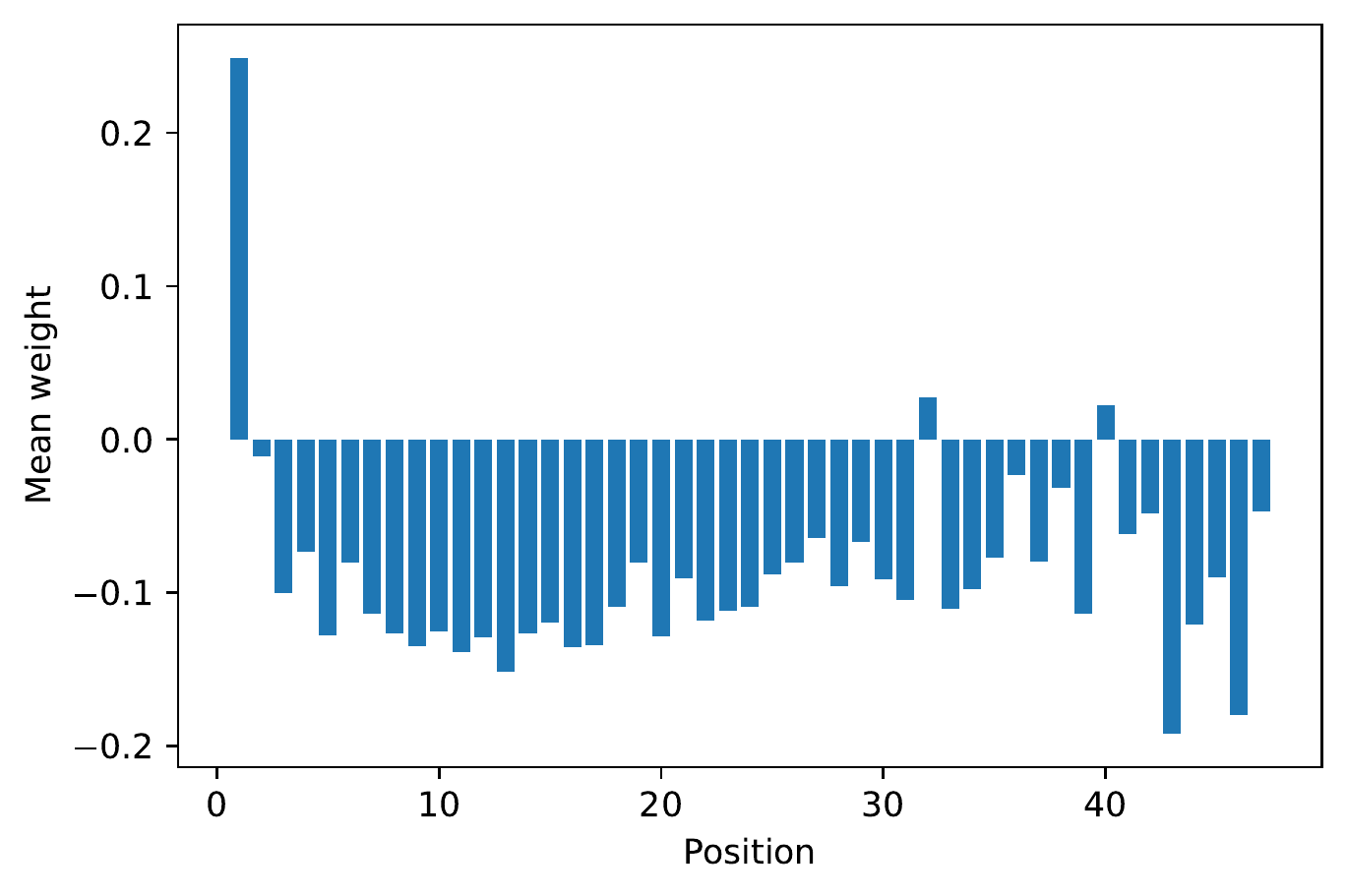} 
    \end{subfigure}\begin{subfigure}[b]{0.49\textwidth}\label{fig:esol12}
      \centering 
      \includegraphics[width=\textwidth]{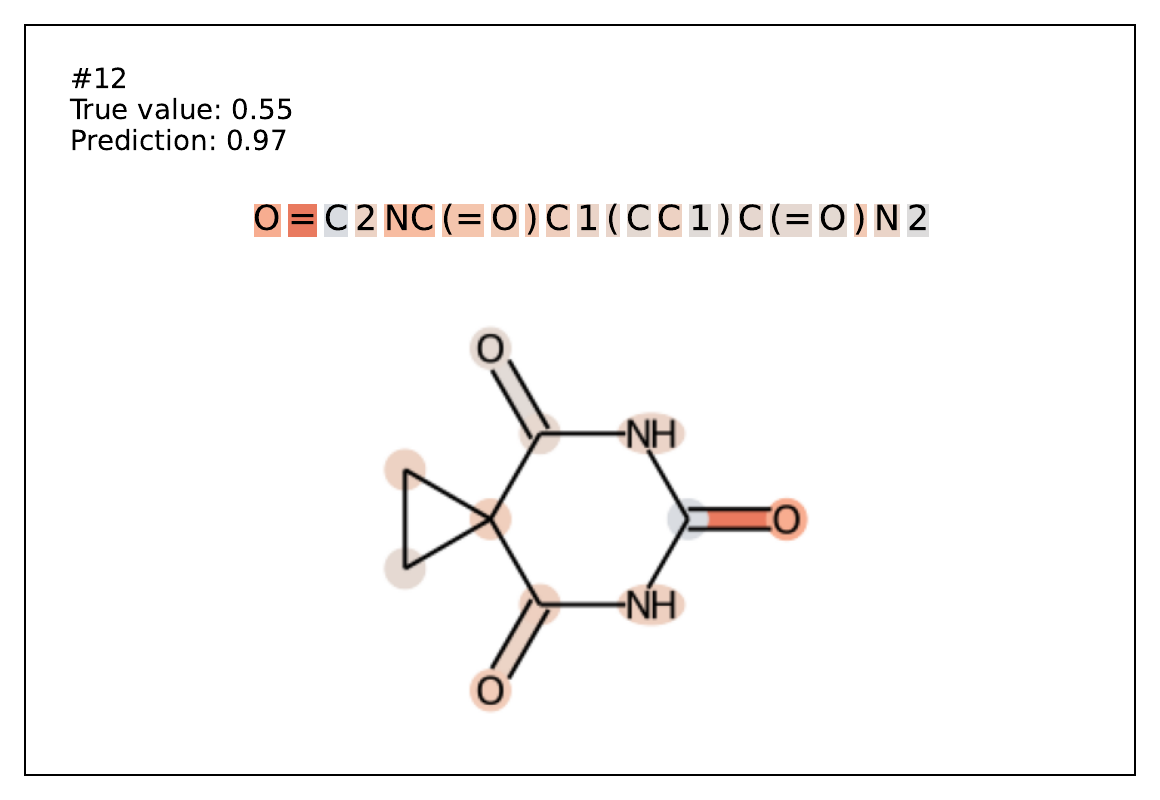} 
    \end{subfigure}
    \begin{subfigure}[b]{0.49\textwidth}\label{fig:esol19}
      \centering 
      \includegraphics[width=\textwidth]{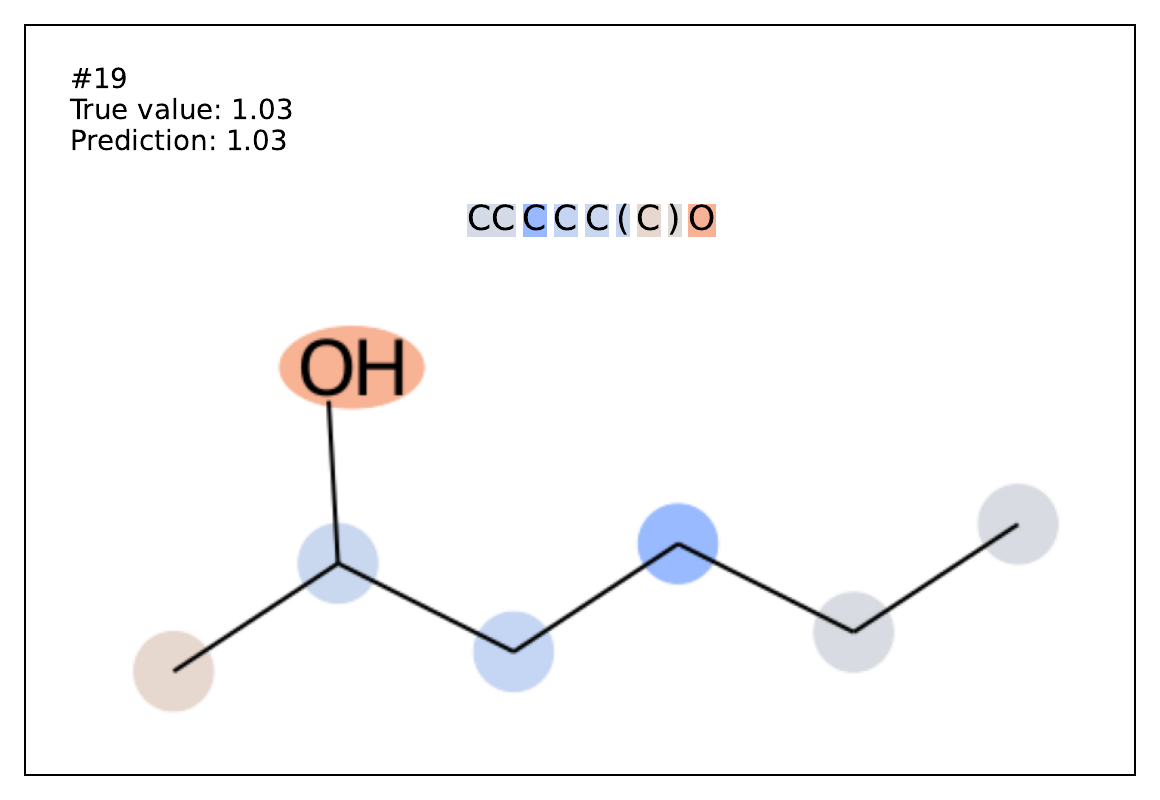} 
    \end{subfigure}\begin{subfigure}[b]{0.49\textwidth}\label{fig:esol51}
      \centering 
      \includegraphics[width=\textwidth]{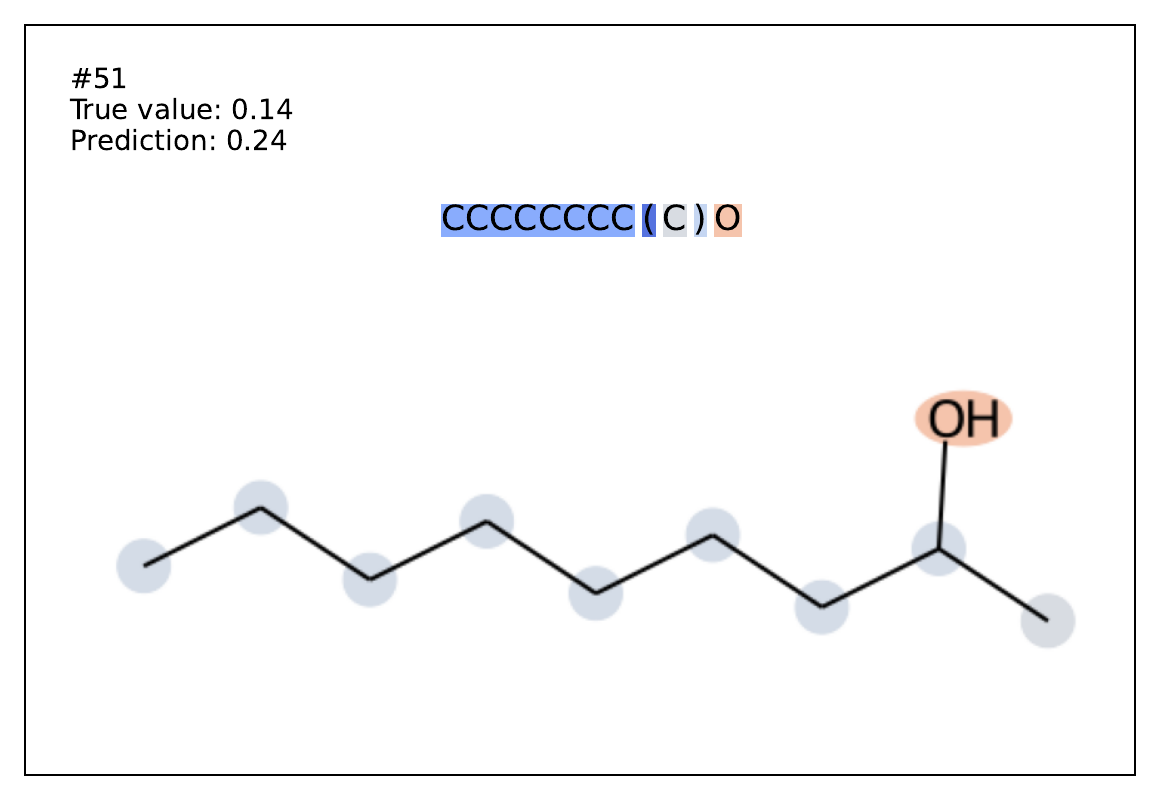} 
    \end{subfigure}
    \caption{Top-left: the average attribution score of a token depending on its position in the tokenized SMILES. There is a significant positive bias on the first token. Others: the contributions of each token and each atom to the prediction of our model fine-tuned on the ESOL dataset, according to the Integrated Gradient algorithm.}
    \label{fig:captum-esol-main}
\end{figure}

\paragraph{ESOL dataset:}
Next, we analyze the attribution scores Integrated Gradients gave for one of the regression tasks, ESOL, which aims to measure the solubility of the molecules in water. 
We first noted a strange pattern: the first token of each molecule tended to receive a positive score, regardless of the molecule's label or structure. We confirmed this by measuring the mean attribution score across the tokens with respect to their positions (Fig. \ref{fig:captum-esol-main}, top-left). We decided to ``normalize'' the attribution scores by subtracting those mean attributions from the scores given by the Integrated Gradients method.

Since water is a polar solvent, polar groups, such as hydroxyl group (-OH) and carbonyl group (C=O), will contribute to the solubility of the molecule \citep{book-review-organic-functional-groups}. The Integrated Gradients method highlighted almost all instances of hydroxyl groups, as seen in the bottom row of Fig. \ref{fig:captum-esol-main}, and most instances of carbonyl groups, especially if the molecule is more soluble (top-right image of Fig. \ref{fig:captum-esol-main}). It is known that long chains of hydrocarbon compounds are known to decrease the solubility of the molecules. This is captured by the negative attributions of "C" tokens in such molecules, as seen in the bottom row of Fig. \ref{fig:captum-esol-main}. 
More examples and analysis is presented in Appendix \ref{sec:app-captum-esol}. 

We hope the interpretability methods developed for deep learning, our visualization toolkit, and the strong predictive models made possible by the large pre-trained models will help chemists uncover new correlations between substructures and target variables or even causal links.

\section{Conclusion}
In this paper, we introduce \MODEL{}, a large pre-trained model for molecular representations which sets state-of-the-art results across many classification, regression, and generation tasks. We showed the power of self-supervised training through a series of quantitative and qualitative analyses of representations showing that interpretability methods applied to these representations mirror that of chemists. There is still room for investigation of the impact of data on pre-training, i.e., ZINC contains more than a billion molecules, but its distribution might be irrelevant for many downstream tasks. Further collaboration between machine learning and chemistry experts might allow for the discovery of new molecular substructures linked to chemical phenomena.  

\section{Acknowledgements}
We would like to thank Luke Zettlemoyer, Lucas Allan, Garik Petrosyan and Garegin Papoian for useful feedback. 
This work was partially supported by the RA MES State Committee of Science, in the framework of the research project No. 20TTCG-1F004.

\bibliography{chemistry-nlp}
\bibliographystyle{iclr_conference}

\clearpage
\appendix
\section{Dataset Descriptions}

\begin{table}[h]
\begin{tabular}{llllll}
\toprule

Dataset       & Task type      & \# of tasks & \# of compounds & Split type & Metric  \\ \midrule
ESOL          & Regression     & 1              & 1128               & Random     & RMSE    \\
FreeSolv      & Regression     & 1              & 642                & Random     & RMSE    \\
Lipophilicity & Regression     & 1              & 4200               & Random     & RMSE    \\ \midrule
HIV           & Classification & 1              & 41127              & Scaffold   & ROC-AUC \\
BACE          & Classification & 1              & 1513               & Scaffold   & ROC-AUC \\ BBBP          & Classification & 1              & 2039               & Scaffold   & ROC-AUC \\
\midrule
Tox21         & Classification & 12             & 7831               & Random     & ROC-AUC \\
ToxCast       & Classification & 617            & 8575               & Random     & ROC-AUC \\
SIDER         & Classification & 27             & 1427               & Random     & ROC-AUC \\
ClinTox       & Classification & 2              & 1478               & Random     & ROC-AUC \\
\midrule
Ames          & Classification & 1              & 6512               & Random     & ROC-AUC \\
Micronucleus Assay & Classification & 1              & 641                & Random     & ROC-AUC

    \\ \bottomrule
    
\end{tabular}
\caption{Dataset statistics for our fine-tuning evaluations.}

\end{table}

\section{Ames Interpretability: Detailed Analysis}\label{sec:app-captum-ames}

\input{sections/appendix-ames}

\section{ESOL Interpretability: Detailed Analysis}\label{sec:app-captum-esol}

\input{sections/appendix-esol}

\section{Hyperparameters}

\begin{table}[h]
\centering
\begin{tabular}{@{}ll@{}}
\toprule
Hyper parameter             & Value                     \\ \midrule
batch-size                  & 32                    \\
task                        & sentence\_prediction  \\
add-prev-output-tokens      & True                  \\
layernorm-embedding         & True                  \\
share-all-embeddings        & True                  \\
share-decoder-input-output-embed         & True     \\
skip-invalid-size-inputs-valid-test      & True     \\
required-batch-size-multiple             & 1        \\
criterion                   & cross\_entropy  \\
arch                        & bart\_large           \\
total-num-update            & 50000   \\
max-update                  & 50000   \\
warmup-updates              & 1500  \\
max-target-positions        & 128                   \\
max-source-positions        & 128                   \\
best-checkpoint-metric      & loss                  \\
Dropout                     & 0.1     \\
attention-dropout           & 0.2                   \\
relu-dropout                & 0.1                   \\
clip-norm                   & 1.0                   \\
max-source-positions        & 128                   \\
weight-decay                & 0.01                  \\
fp16                        & True        \\
fp32                        & True        \\
fp16-scale-window           & 128        \\
threshold-loss-scale        & 1           \\
Optimizer                   & Adam                      \\
Adam-betas                  & (0.9, 0.999)              \\
Adam-eps                    & 1e-08                 \\
LR Scheduler                & polynomial decay          \\
Learning Rate               & 1e-05                     \\
mask	& 0.2 \\
mask\_length	& span-poisson \\
mask\_random	& 0.1 \\
GPU	& 1024 \\
\bottomrule
\end{tabular}
\quad

\caption{Table shows the hyperparameters used for pre-training.}
\label{tab:hyperparams-pretrain}
\end{table}

\label{sec:hp}
\begin{table}[h]
\centering
\begin{tabular}{@{}ll@{}}
\toprule

Hyper parameter             & Value                     \\ \midrule
batch-size                  & 16                    \\
task                        & sentence\_prediction  \\
add-prev-output-tokens      & True                  \\
layernorm-embedding         & True                  \\
share-all-embeddings        & True                  \\
share-decoder-input-output-embed         & True     \\
reset-optimizer             & True                  \\
reset-dataloader            & True                  \\
reset-meters                & True                  \\
skip-invalid-size-inputs-valid-test      & True     \\
required-batch-size-multiple             & 1        \\
criterion                   & sentence\_prediction  \\
arch                        & bart\_large           \\
total-num-update            & dataset\_size*0.8/batch\_size*epoch   \\
max-update                  & total\_num\_update   \\
warmup-updates              & total\_num\_update*0.16  \\
max-target-positions        & 128                   \\
max-source-positions        & 128                   \\
regression-target           & \{True, False\}       \\
best-checkpoint-metric      & loss                  \\
Dropout                     & \{0.1, 0.2, 0.3\}     \\
attention-dropout           & 0.2                   \\
relu-dropout                & 0.1                   \\
clip-norm                   & 0.1                   \\
max-source-positions        & 128                   \\
num-classes                 & \{1, 2\}              \\
weight-decay                & 0.01                  \\
fp16                        & \{True, null\}        \\
fp32                        & \{True, null\}        \\
fp16-scale-window           & 128        \\
threshold-loss-scale        & \{1, null\}           \\
Optimizer                   & Adam                      \\
Adam-betas                  & (0.9, 0.999)              \\
Adam-eps                    & 1e-08                 \\
LR Scheduler                & polynomial decay          \\
Learning Rate               & \{5e-06, 1e-05, 3e-05\}                     \\

\bottomrule
\end{tabular}
\quad

\caption{Table shows the hyperparameters used for fine-tuning experiments for classification and regression tasks. We set the total number of updates to the number of steps required to do ten epochs and the number of warmup updates being 16\% of the total number of updates.}
\label{tab:hyperparams-finetune}
\end{table}

\begin{table}[h]
\centering
\begin{tabular}{@{}ll@{}}
\toprule
Hyper parameter & Value              \\ \midrule
Optimizer       & Adam               \\
Adam-betas      & (0.9, 0.98)        \\
Adam-eps        & 1e-8               \\
LR Scheduler    & polynomial decay   \\
Learning Rate   & 3e-05              \\

\bottomrule
\end{tabular}
\quad
\begin{tabular}{@{}ll@{}}
\toprule
Hyper parameter & Value           \\ \midrule
$\lambda$          & [0.001, 0.01, 0.1] \\
Noise Types     & [$\mathcal{U}$, $\mathcal{N}$] \\
$\sigma$        & $1e-5$\\ 
Dropout         & 0.1                \\
Weight Decay    & 0.01               \\
Clip Norm       & 0.1                \\
\bottomrule
\end{tabular}
\caption{Hyper parameters for the use of R3F on all generative fine-tuning experiments. All other hyper-parameters are in accordance with Table~\ref{tab:hyperparams-finetune}}
\label{tab:hyperparams-generative}.
\end{table}

\end{document}

%% file: math_commands.tex
\usepackage{amsmath,amsfonts,bm}

\def\eqref#1{equation~\ref{#1}}

\def\1{\bm{1}}

\DeclareMathAlphabet{\mathsfit}{\encodingdefault}{\sfdefault}{m}{sl}
\SetMathAlphabet{\mathsfit}{bold}{\encodingdefault}{\sfdefault}{bx}{n}

%% file: sections/pretrain-ablation-tables.tex
\begin{table}[!htb]
\footnotesize
    \begin{subtable}{.33\linewidth}
      \centering
        \begin{tabular}{@{}lr@{}}
        \toprule
        \texttt{mask\_token} & AUC \\
        \texttt{\_budget} &  \\ \midrule
        0.10                  &  0.753      \\
        0.15                  &  0.812      \\
        0.20                  &  \textbf{0.821}      \\
        0.25                  &  0.808      \\
        0.30                  &  0.801      \\
        0.35                  &  0.760      \\
        0.40                  &  0.701      \\ \bottomrule
        \end{tabular}
        \caption{}
        \label{table:ablation_mask_token}
    \end{subtable}%
    \begin{subtable}{.33\linewidth}
      \centering
        
    \end{subtable}%
    \begin{subtable}{.33\linewidth}
      \centering
        \begin{tabular}{@{}lll@{}}
            \toprule
            \texttt{random} & $\lambda$ & AUC \\ 
            \texttt{\_mask} &   &  \\ \midrule
            0.05               & 1.5    &  0.814       \\ 
            0.05               & 2.5    &  0.818       \\ 
            0.05               & 3.5    &  0.813       \\ 
            0.10               & 1.5    &  0.820      \\ 
            0.10               & 2.5    &  0.821      \\ 
            0.10               & 3.5    &  \textbf{0.821}  \\ \bottomrule
            \end{tabular}
            \caption{}
            \label{table:ablation_lambda}
    \end{subtable}%
    \begin{subtable}{.3\linewidth}
      \centering
        \begin{tabular}{@{}lr@{}}
        \toprule
        \texttt{randomize} & AUC  \\ 
        \texttt{\_tokens} &  \\ \midrule
        \ding{51} &  0.821      \\ 
        \ding{55} &  \textbf{0.835}      \\ \bottomrule
        \end{tabular}
        \caption{}
        \label{table:ablation_randomize}
    \end{subtable} 
    \caption{All the ablations ran using the \MODEL{}-Base
    model on 100 million samples from ZINC20. In (b) we ablate the impact of the random mask percentage and the poisson $\lambda$ parameter which controls the length of the individual masks. For this we fix \texttt{mask\_token\_budget} to 0.20, the best forming configuration from (a). %
    In (c) we ablate the impact of randomizing tokens in the input, using the best configurations found in (a) and (b) %
    All numbers are the average of AUC-ROC scores on HIV, BBBP and ClinTox datasets.
    } 
    \label{tab:ablation-all}
\end{table}

%% file: sections/appendix-ames.tex
\begin{figure}
    \centering
\begin{subfigure}[b]{0.33\textwidth} 
  \centering 
  \includegraphics[width=\textwidth]{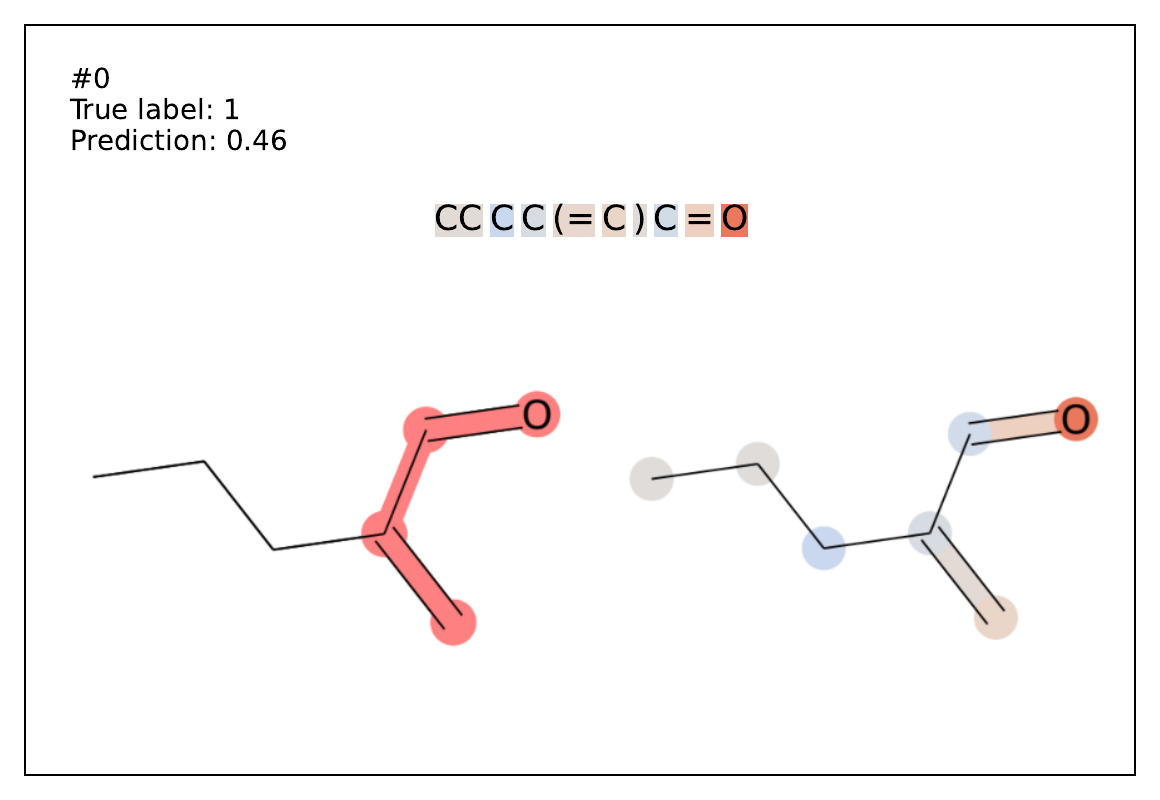} 
\end{subfigure}\begin{subfigure}[b]{0.33\textwidth} 
  \centering 
  \includegraphics[width=\textwidth]{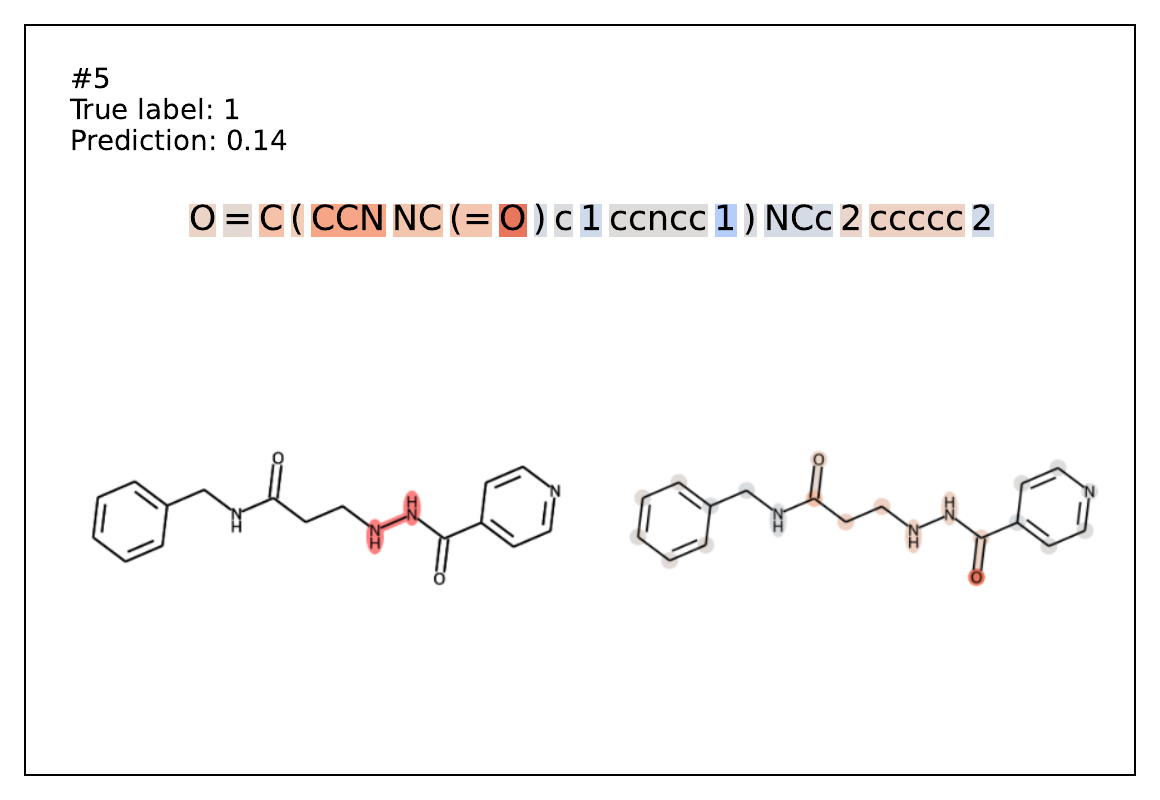} 
\end{subfigure}\begin{subfigure}[b]{0.33\textwidth} 
  \centering 
  \includegraphics[width=\textwidth]{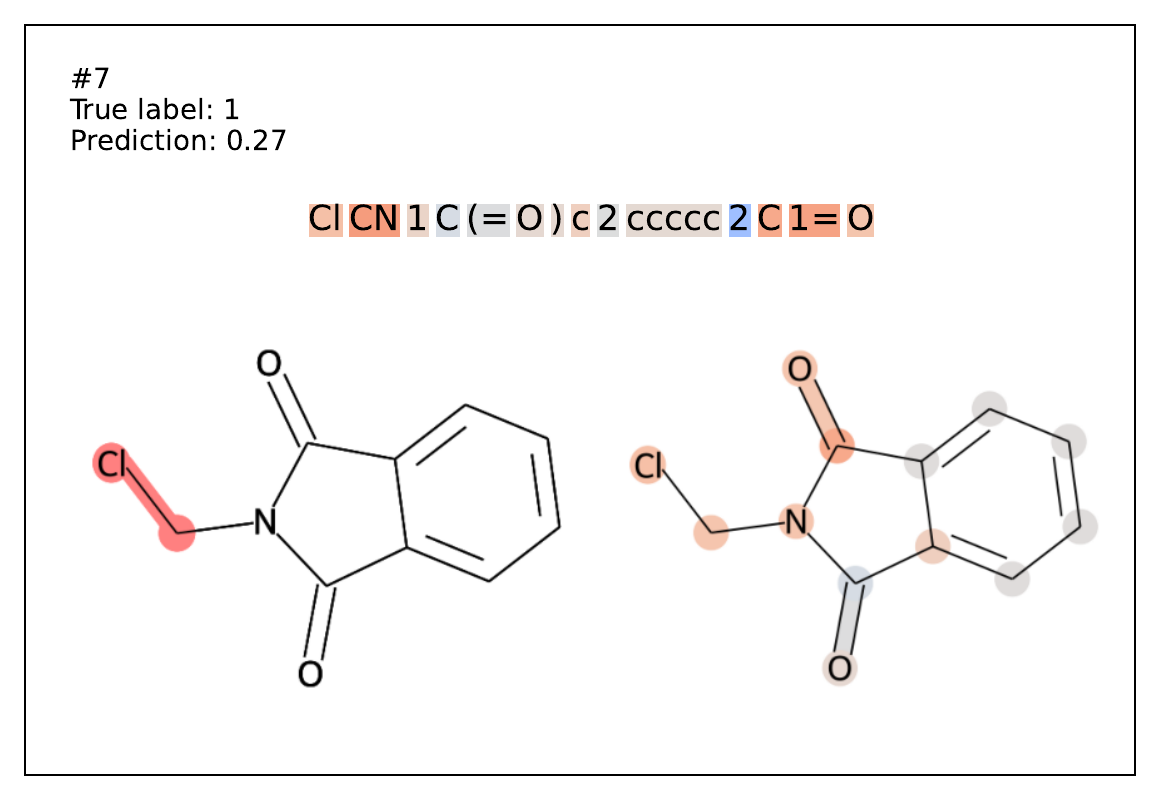} 
\end{subfigure} 
\begin{subfigure}[b]{0.33\textwidth} 
  \centering 
  \includegraphics[width=\textwidth]{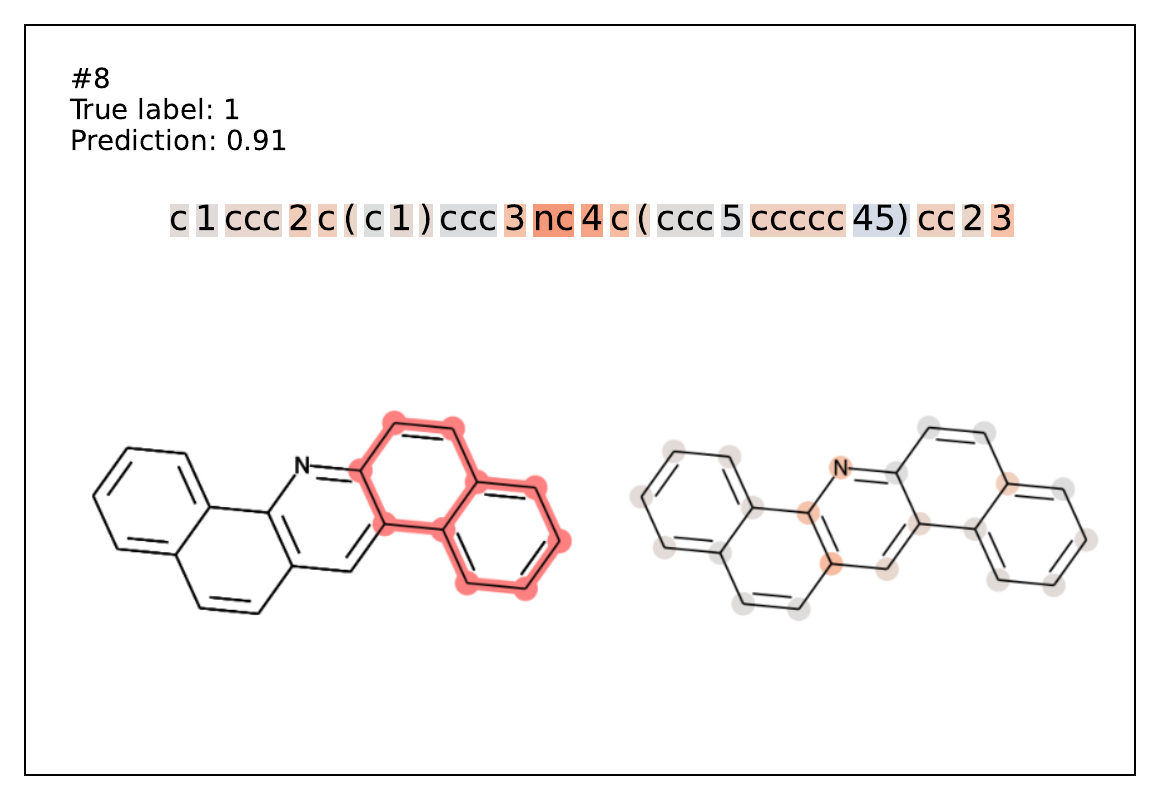} 
\end{subfigure}\begin{subfigure}[b]{0.33\textwidth} 
  \centering 
  \includegraphics[width=\textwidth]{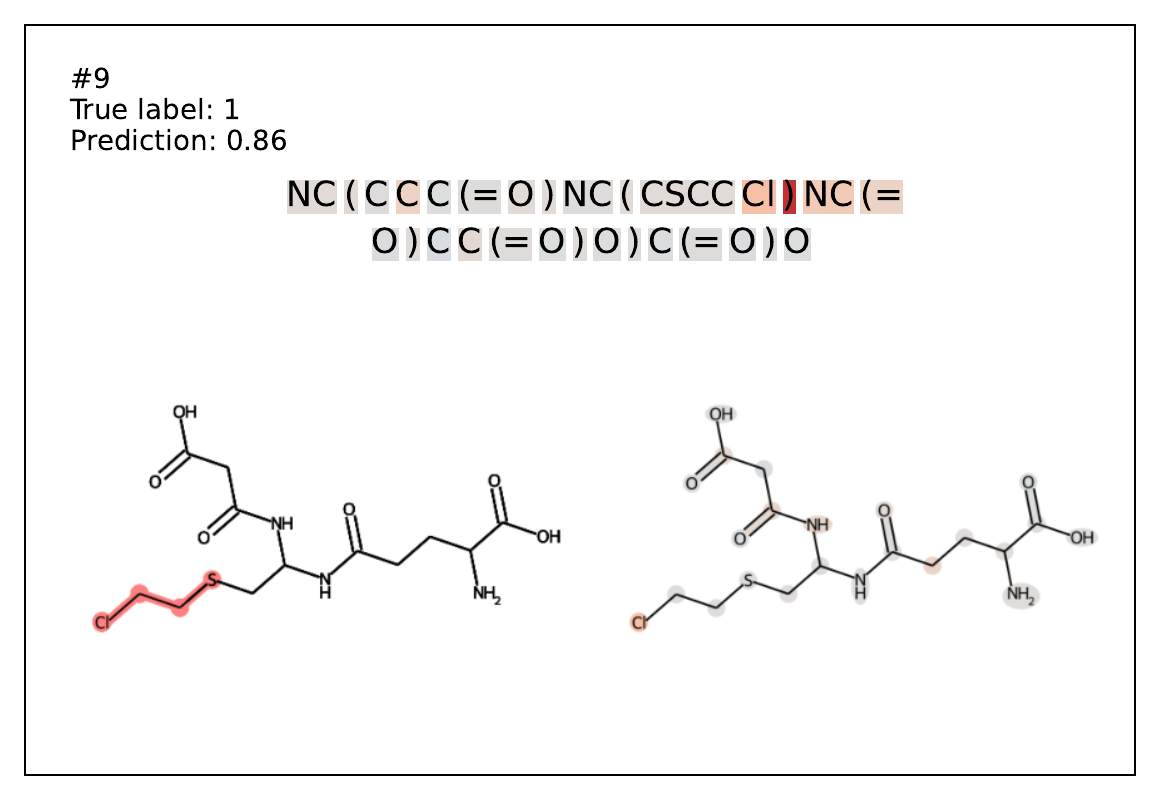} 
\end{subfigure}\begin{subfigure}[b]{0.33\textwidth} 
  \centering 
  \includegraphics[width=\textwidth]{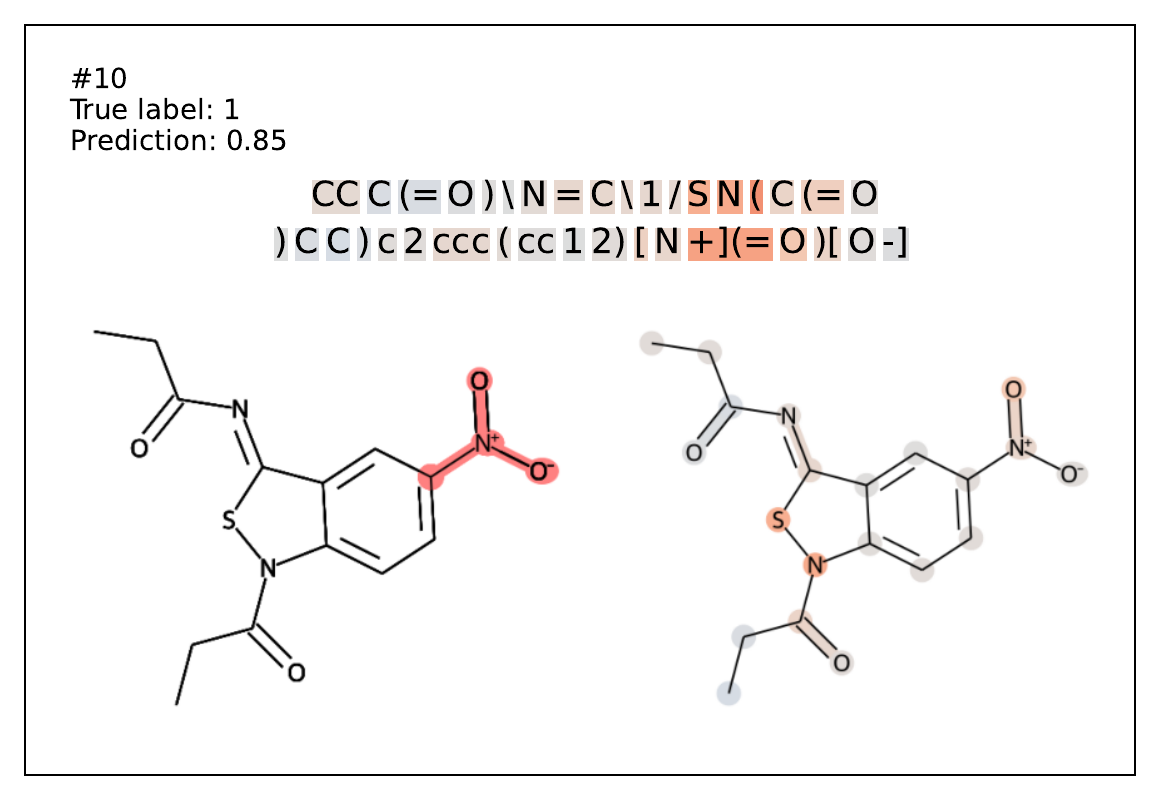} 
\end{subfigure} 
\begin{subfigure}[b]{0.33\textwidth} 
  \centering 
  \includegraphics[width=\textwidth]{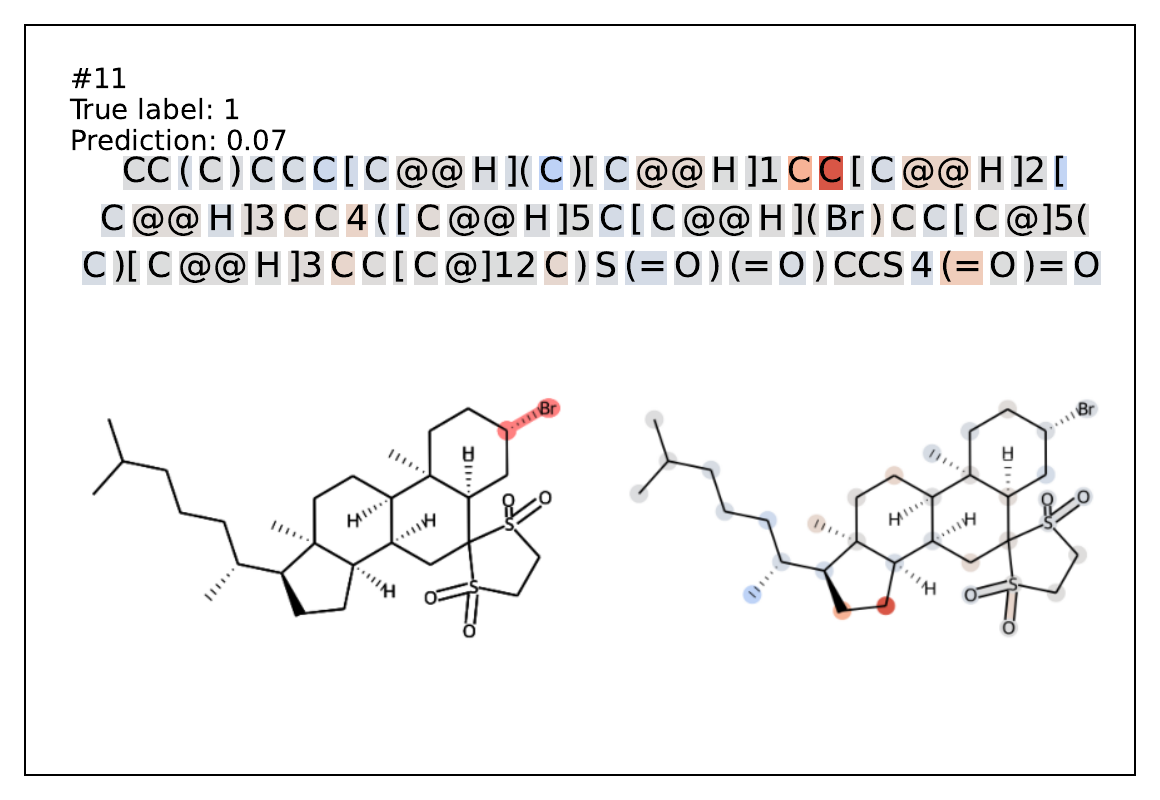} 
\end{subfigure}\begin{subfigure}[b]{0.33\textwidth} 
  \centering 
  \includegraphics[width=\textwidth]{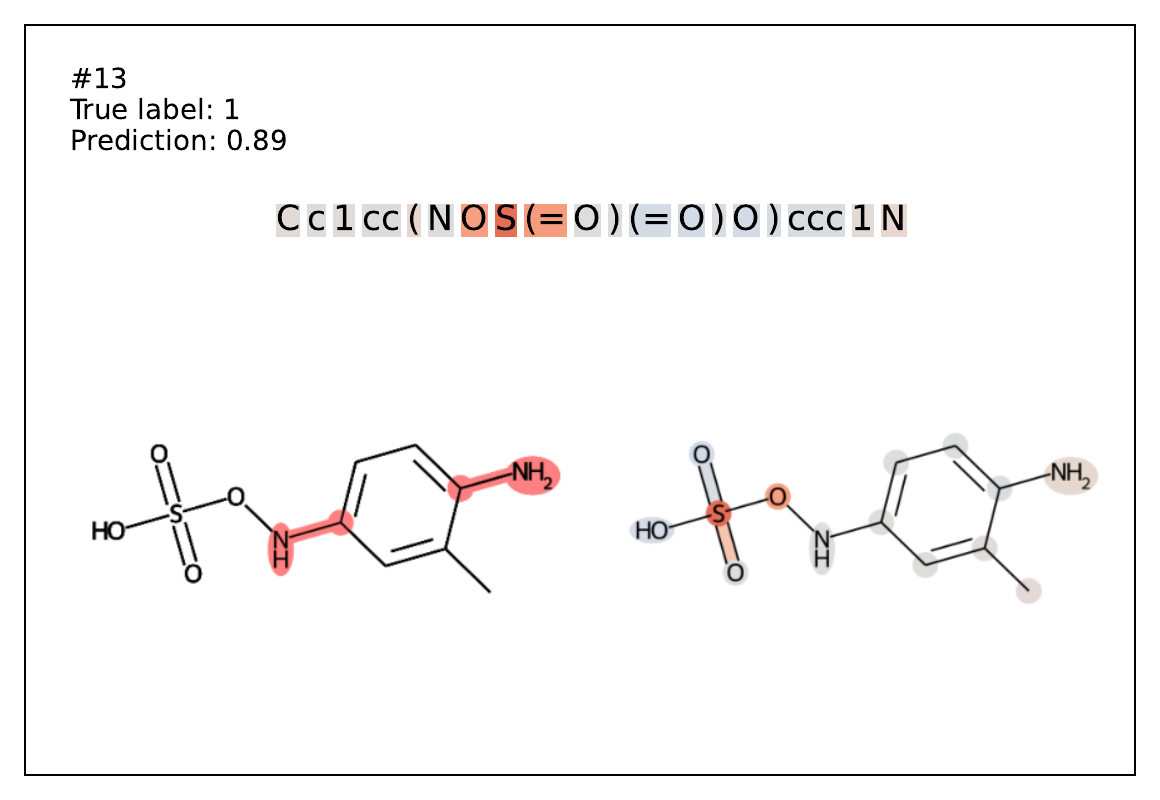} 
\end{subfigure}\begin{subfigure}[b]{0.33\textwidth} 
  \centering 
  \includegraphics[width=\textwidth]{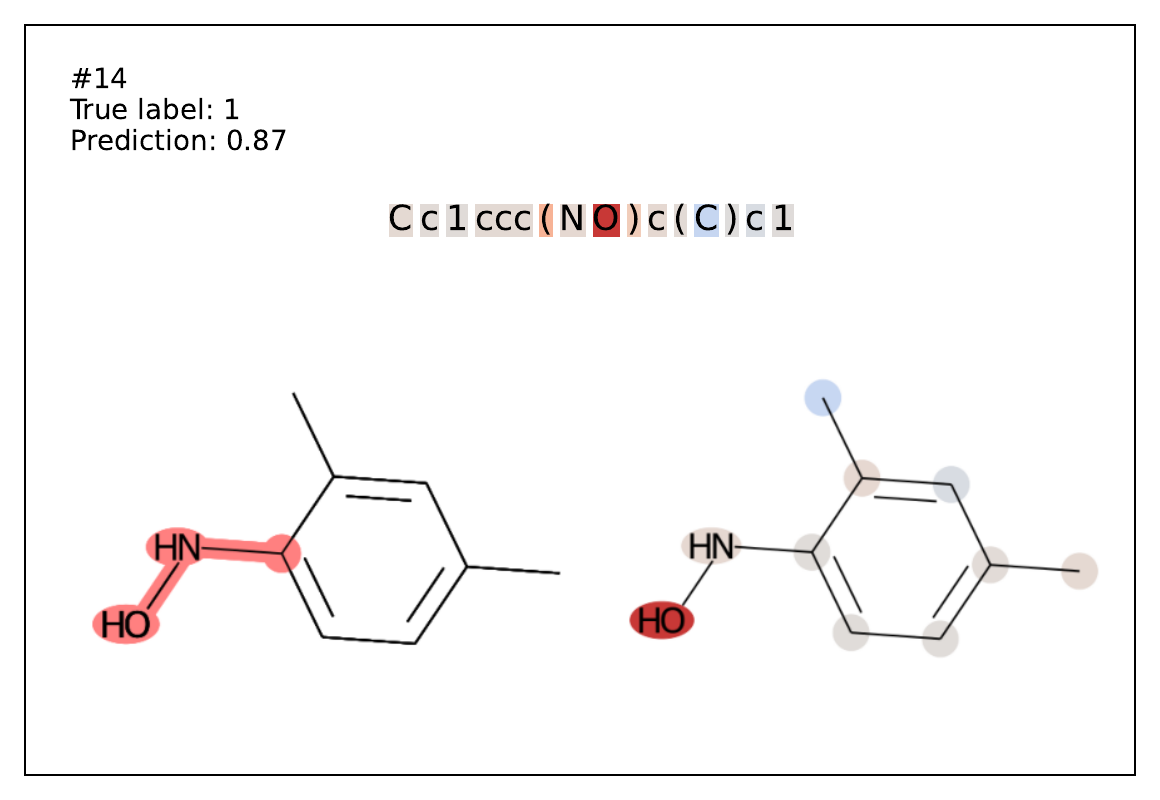} 
\end{subfigure} 
\begin{subfigure}[b]{0.33\textwidth} 
  \centering 
  \includegraphics[width=\textwidth]{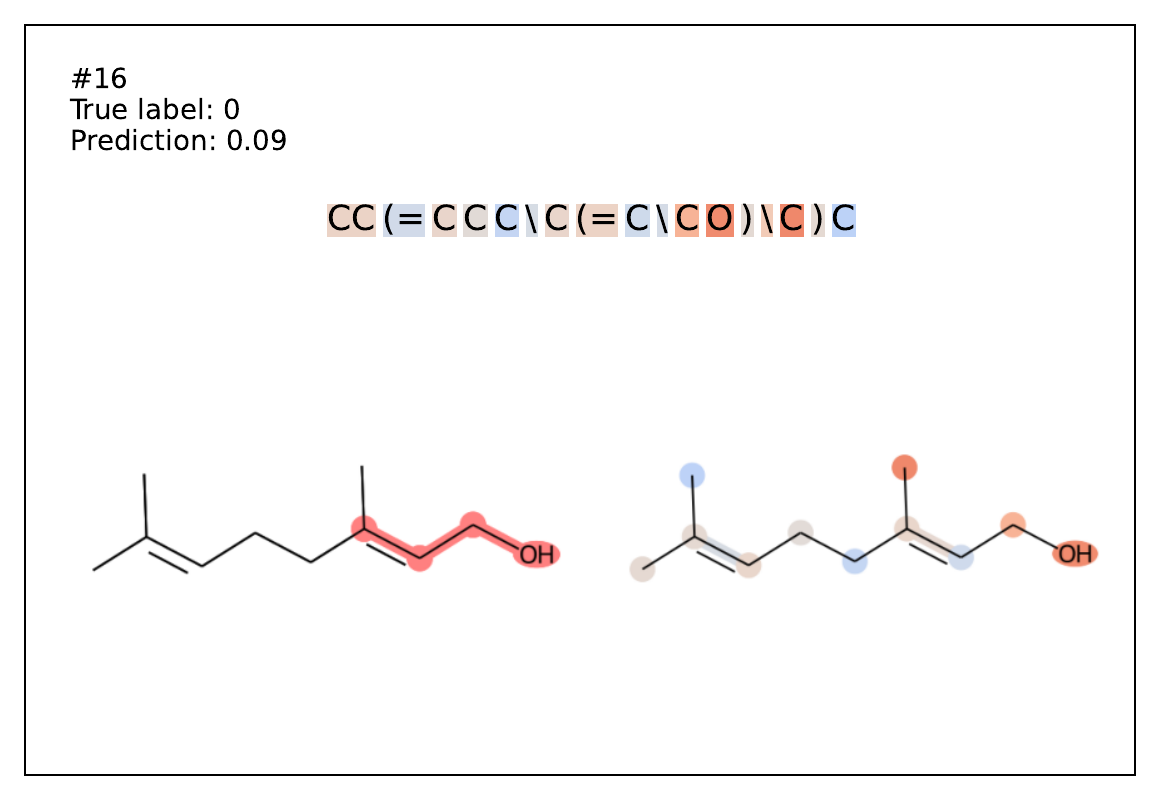} 
\end{subfigure}\begin{subfigure}[b]{0.33\textwidth} 
  \centering 
  \includegraphics[width=\textwidth]{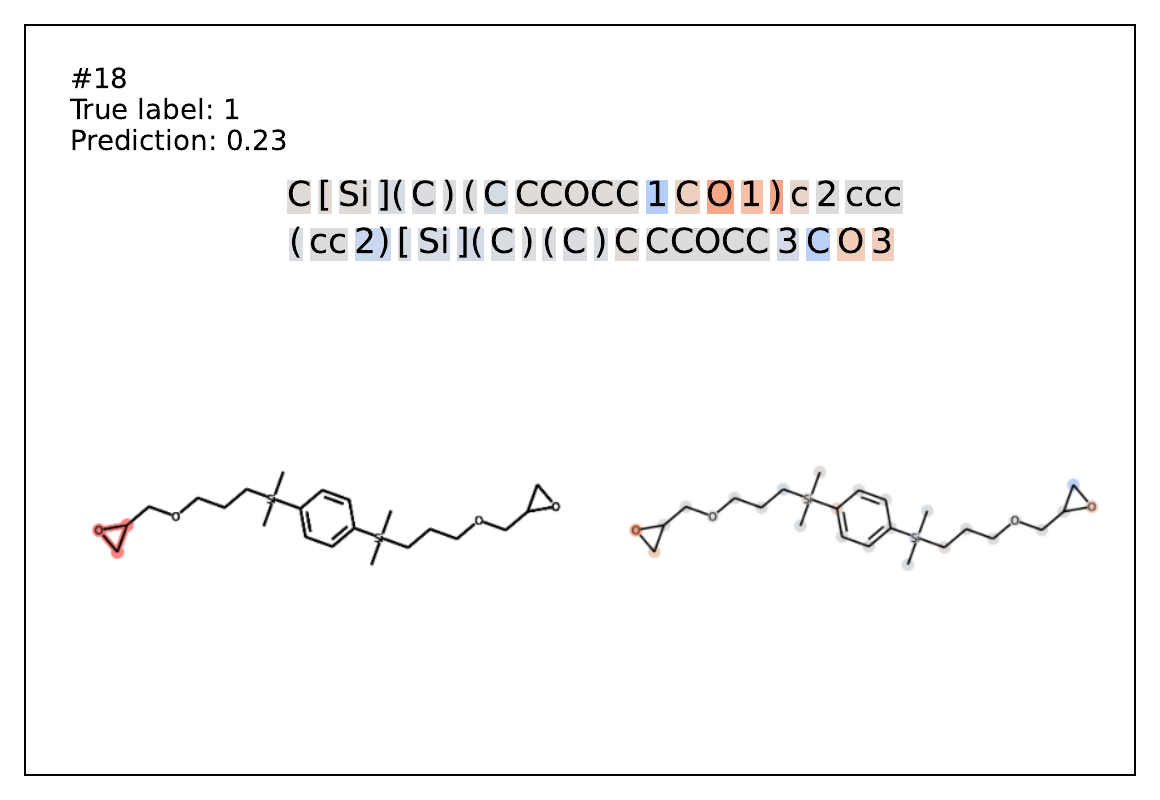} 
\end{subfigure}\begin{subfigure}[b]{0.33\textwidth} 
  \centering 
  \includegraphics[width=\textwidth]{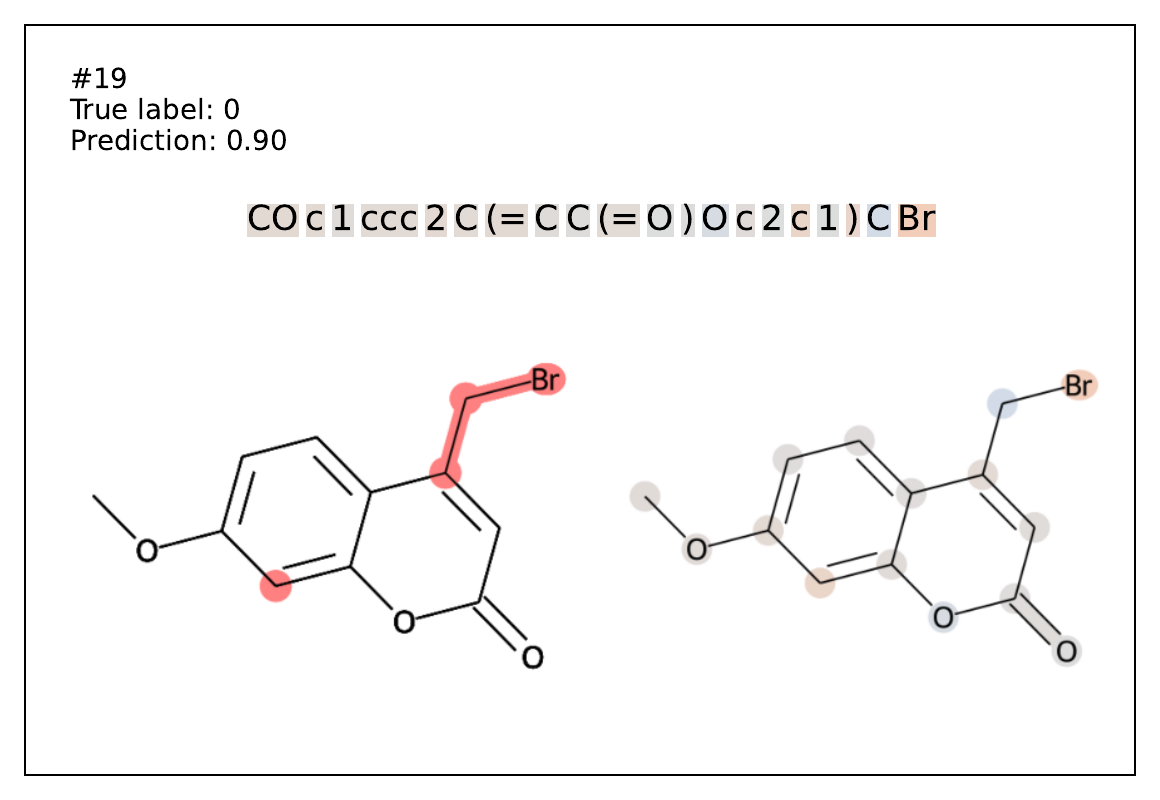} 
\end{subfigure} 
\begin{subfigure}[b]{0.33\textwidth} 
  \centering 
  \includegraphics[width=\textwidth]{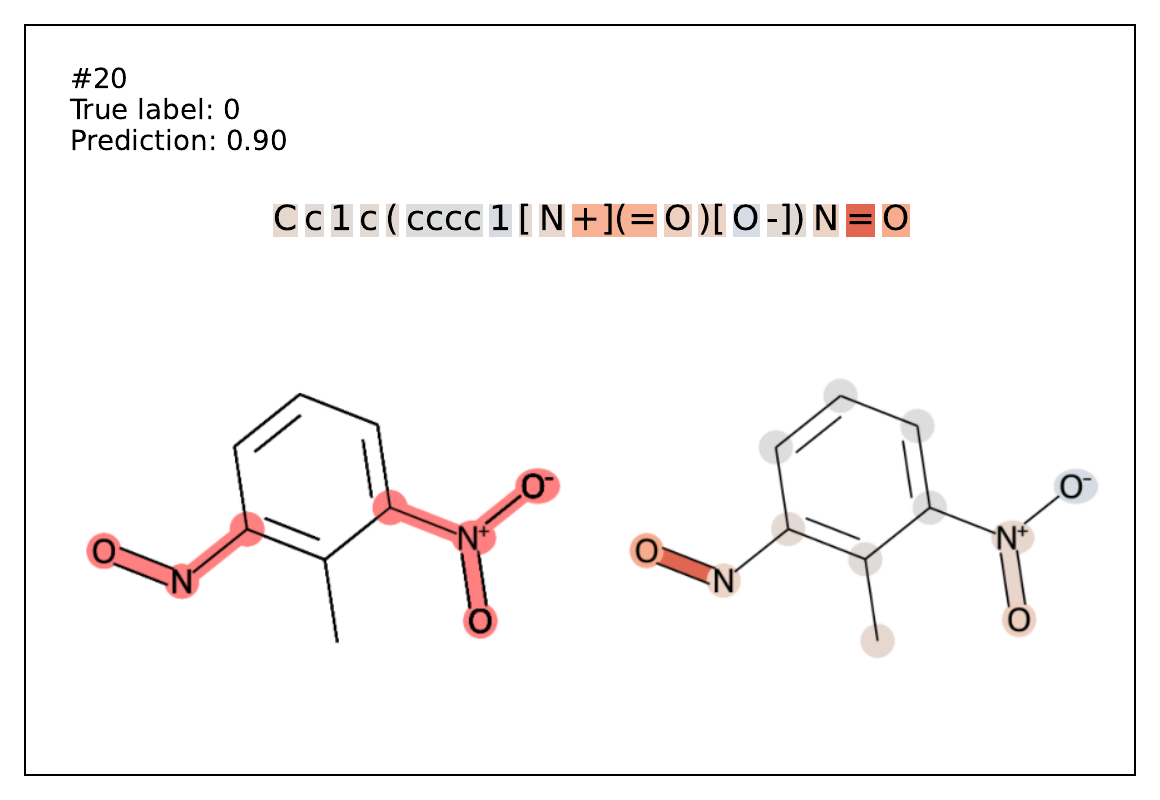} 
\end{subfigure}\begin{subfigure}[b]{0.33\textwidth} 
  \centering 
  \includegraphics[width=\textwidth]{figures/ames/ames21.pdf} 
\end{subfigure}\begin{subfigure}[b]{0.33\textwidth} 
  \centering 
  \includegraphics[width=\textwidth]{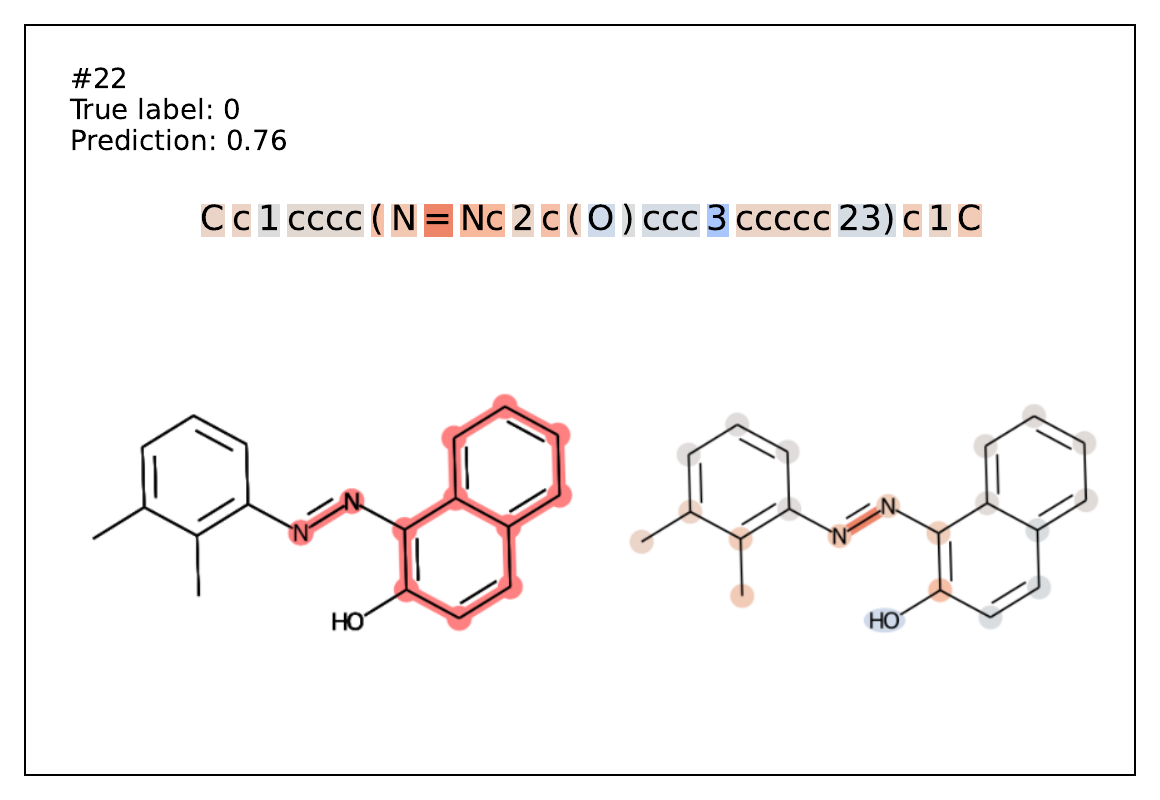} 
\end{subfigure} 
\begin{subfigure}[b]{0.33\textwidth} 
  \centering 
  \includegraphics[width=\textwidth]{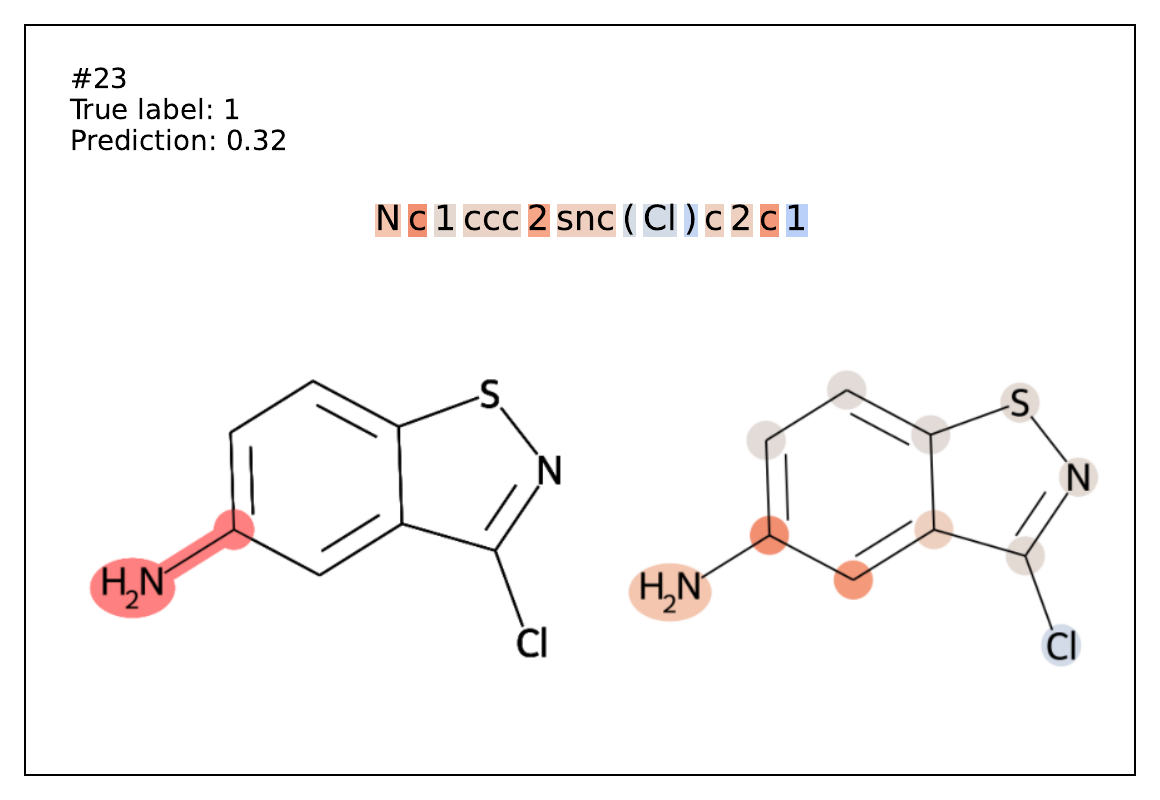} 
\end{subfigure}\begin{subfigure}[b]{0.33\textwidth} 
  \centering 
  \includegraphics[width=\textwidth]{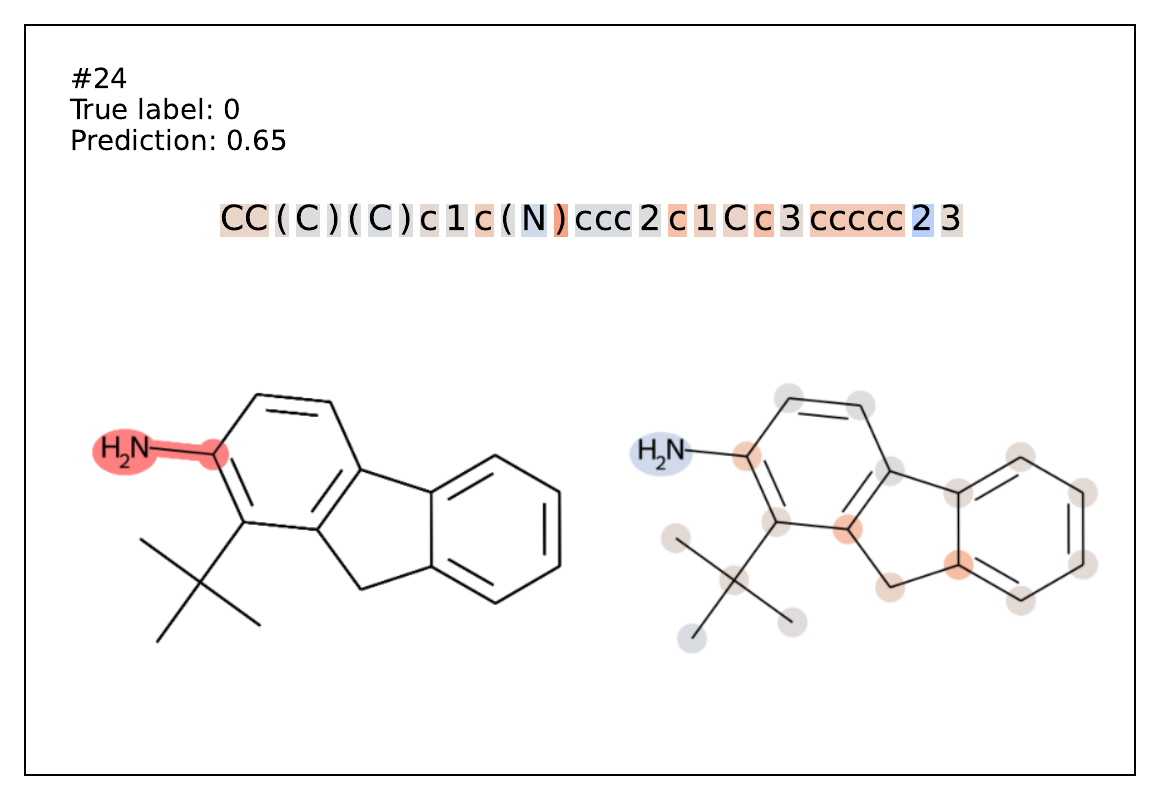} 
\end{subfigure}\begin{subfigure}[b]{0.33\textwidth} 
  \centering 
  \includegraphics[width=\textwidth]{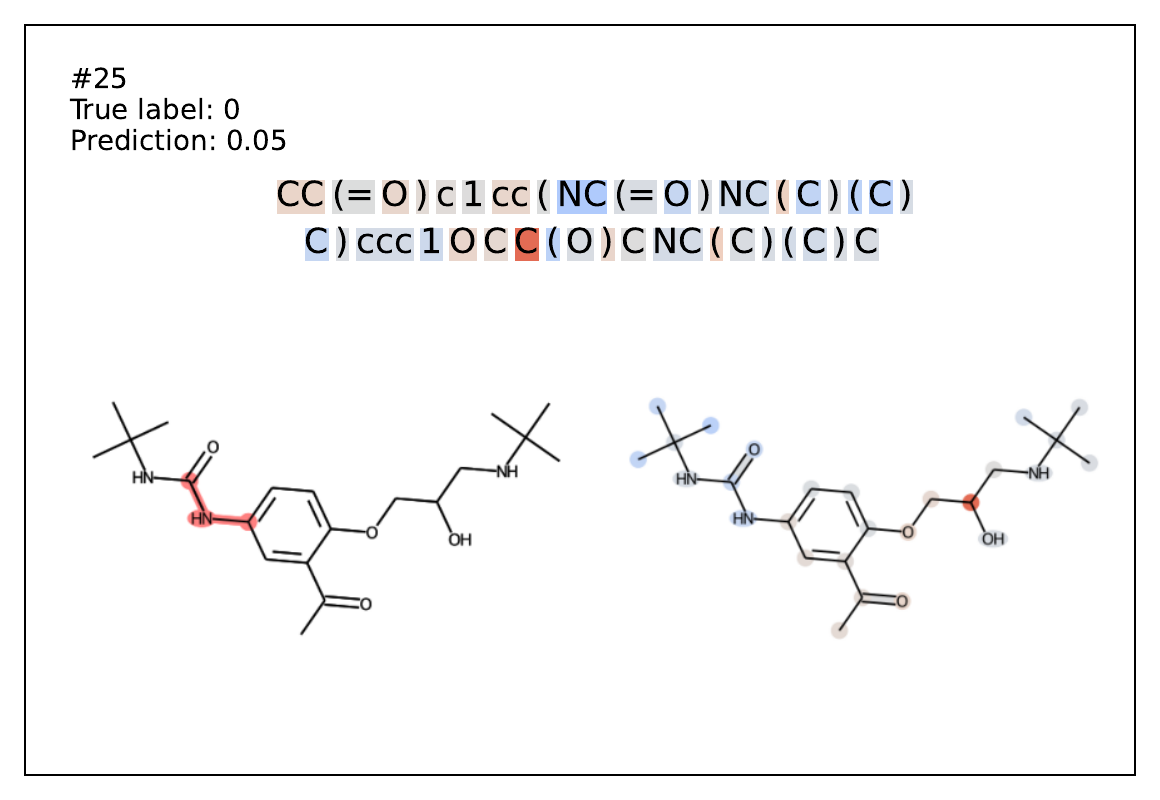} 
\end{subfigure} 
\begin{subfigure}[b]{0.33\textwidth} 
  \centering 
  \includegraphics[width=\textwidth]{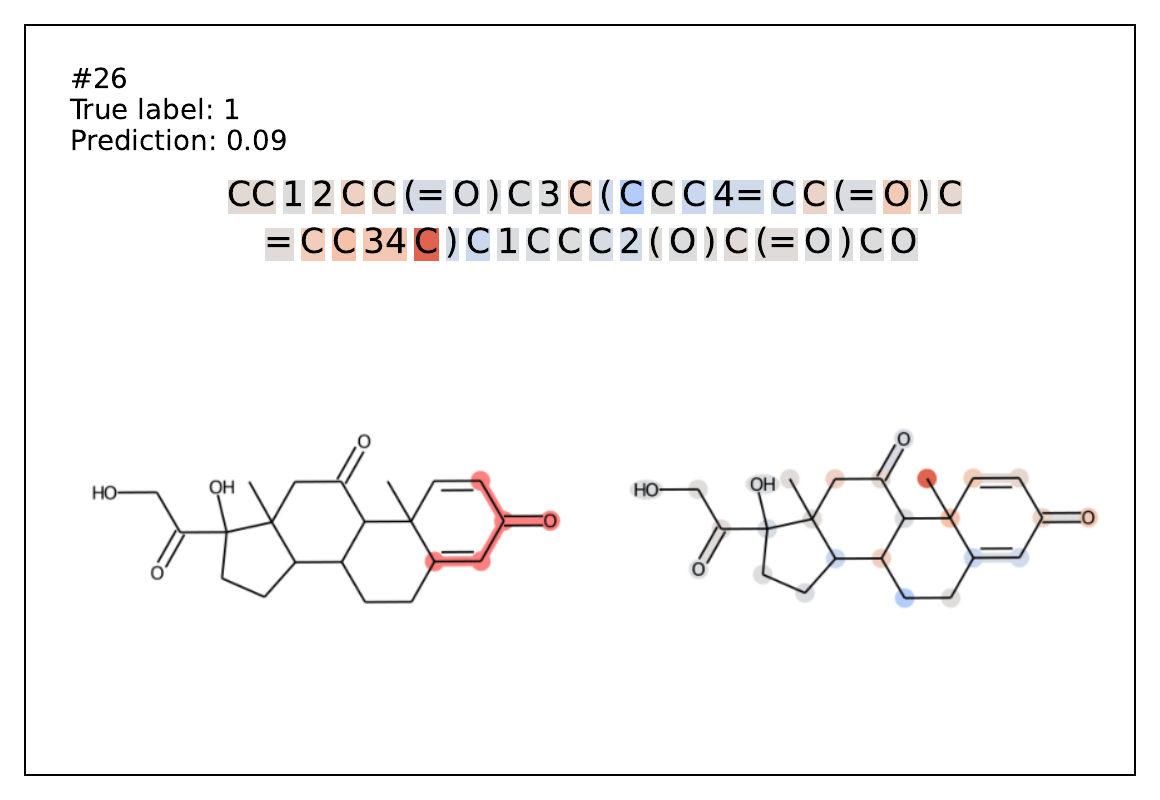} 
\end{subfigure}\begin{subfigure}[b]{0.33\textwidth} 
  \centering 
  \includegraphics[width=\textwidth]{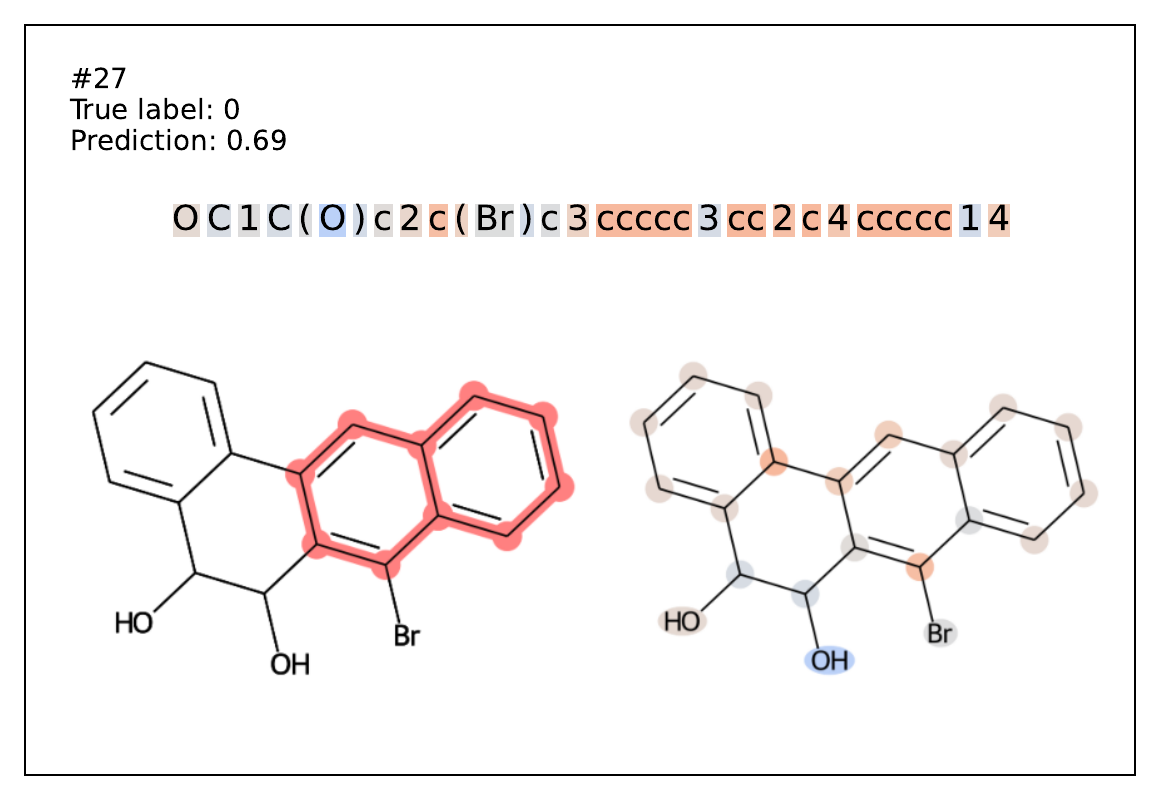} 
\end{subfigure}\begin{subfigure}[b]{0.33\textwidth} 
  \centering 
  \includegraphics[width=\textwidth]{figures/ames/ames28.pdf} 
\end{subfigure} 
\caption{Explaining predictions of the fine-tuned model on Ames dataset. See Section \ref{sec:captum}. Part 1/5}
\label{fig:captum-ames-1}
\end{figure}

\begin{figure}
\centering
\begin{subfigure}[b]{0.33\textwidth} 
  \centering 
  \includegraphics[width=\textwidth]{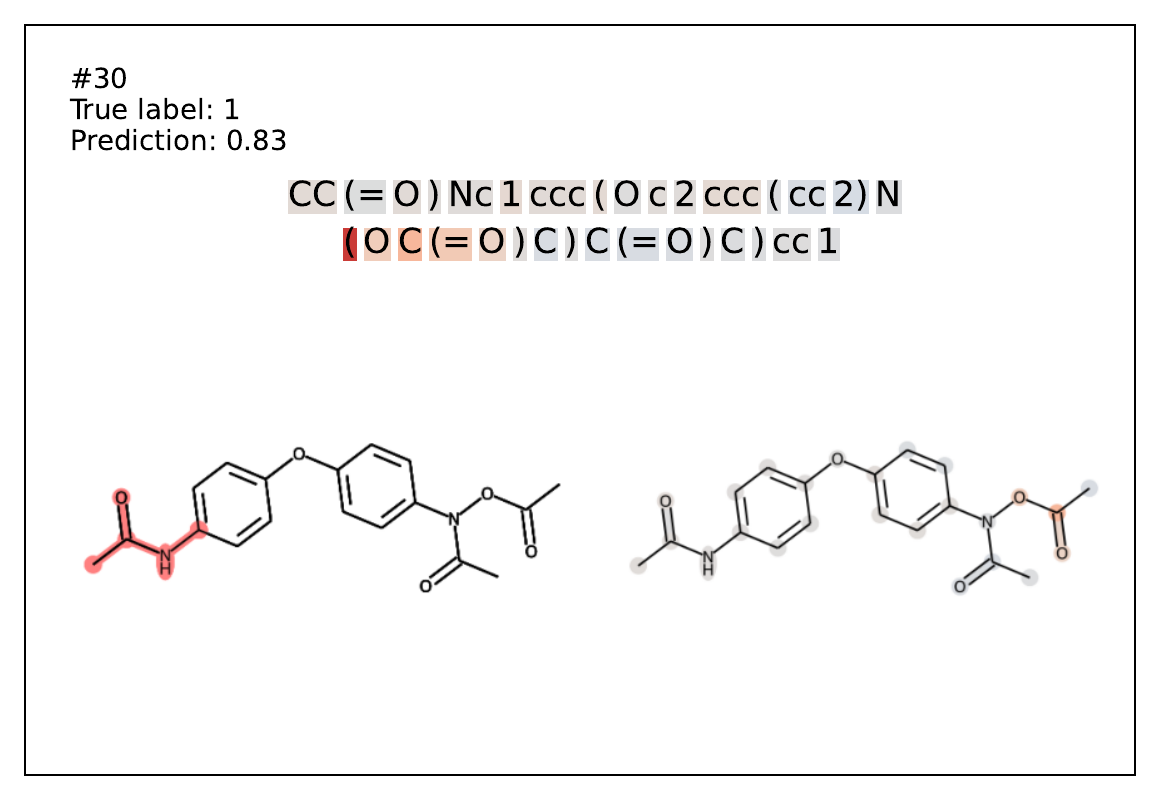} 
\end{subfigure}\begin{subfigure}[b]{0.33\textwidth} 
  \centering 
  \includegraphics[width=\textwidth]{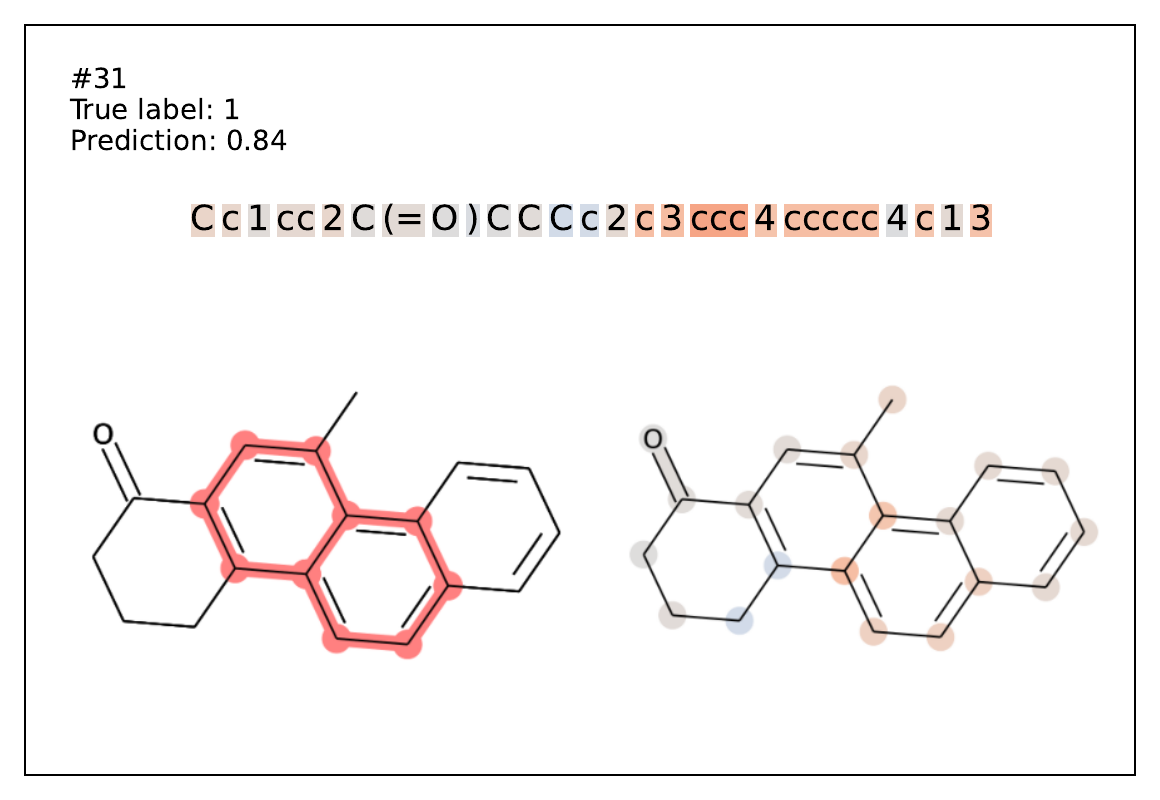} 
\end{subfigure}\begin{subfigure}[b]{0.33\textwidth} 
  \centering 
  \includegraphics[width=\textwidth]{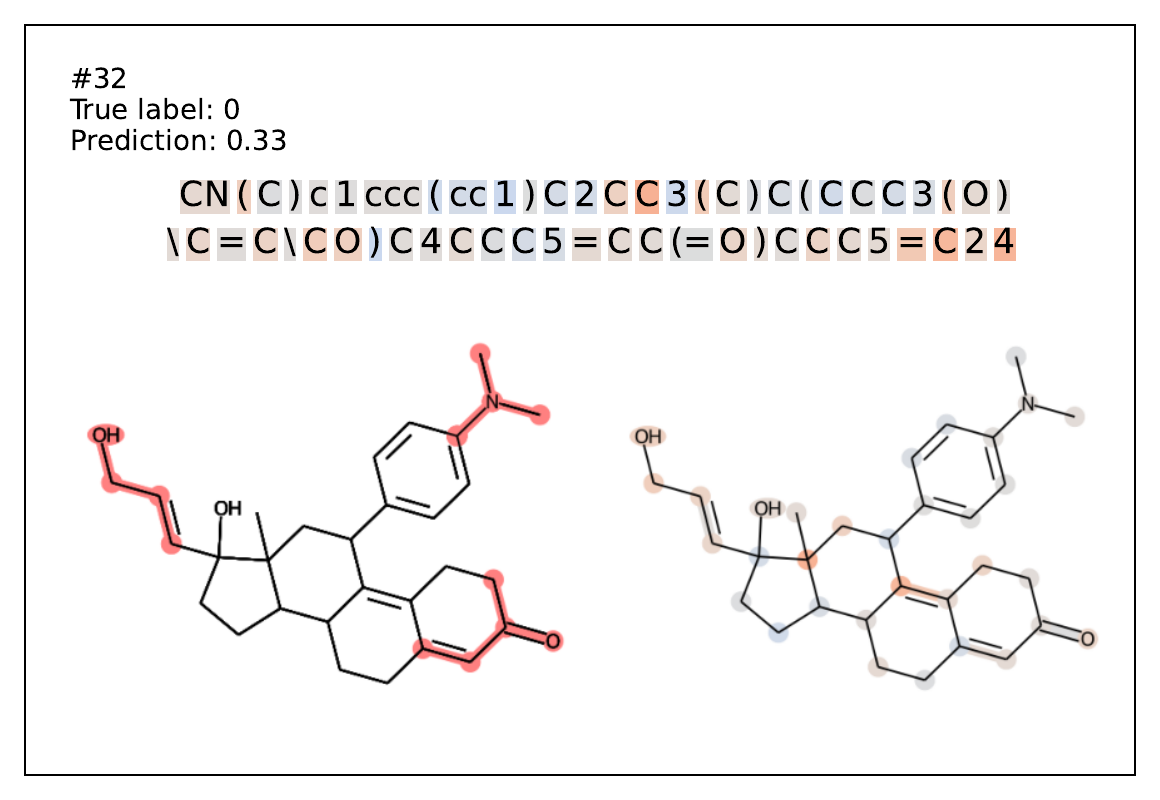} 
\end{subfigure} 
\begin{subfigure}[b]{0.33\textwidth} 
  \centering 
  \includegraphics[width=\textwidth]{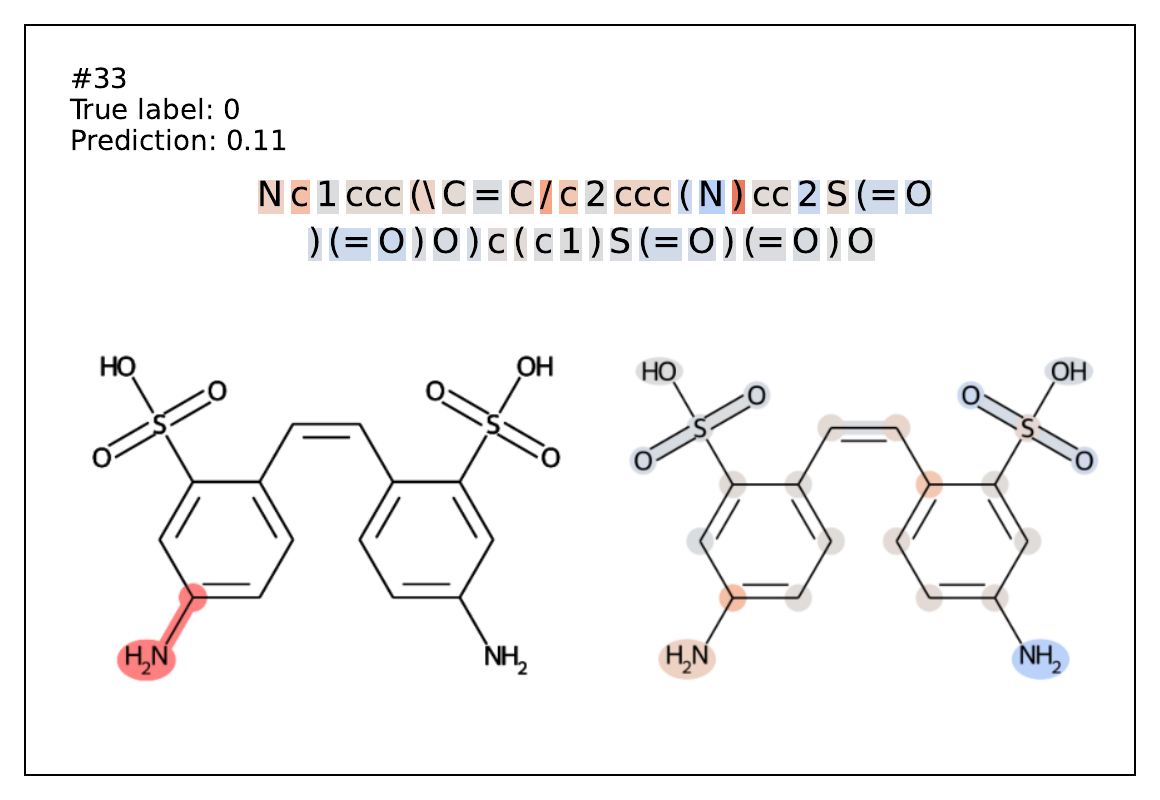} 
\end{subfigure}\begin{subfigure}[b]{0.33\textwidth} 
  \centering 
  \includegraphics[width=\textwidth]{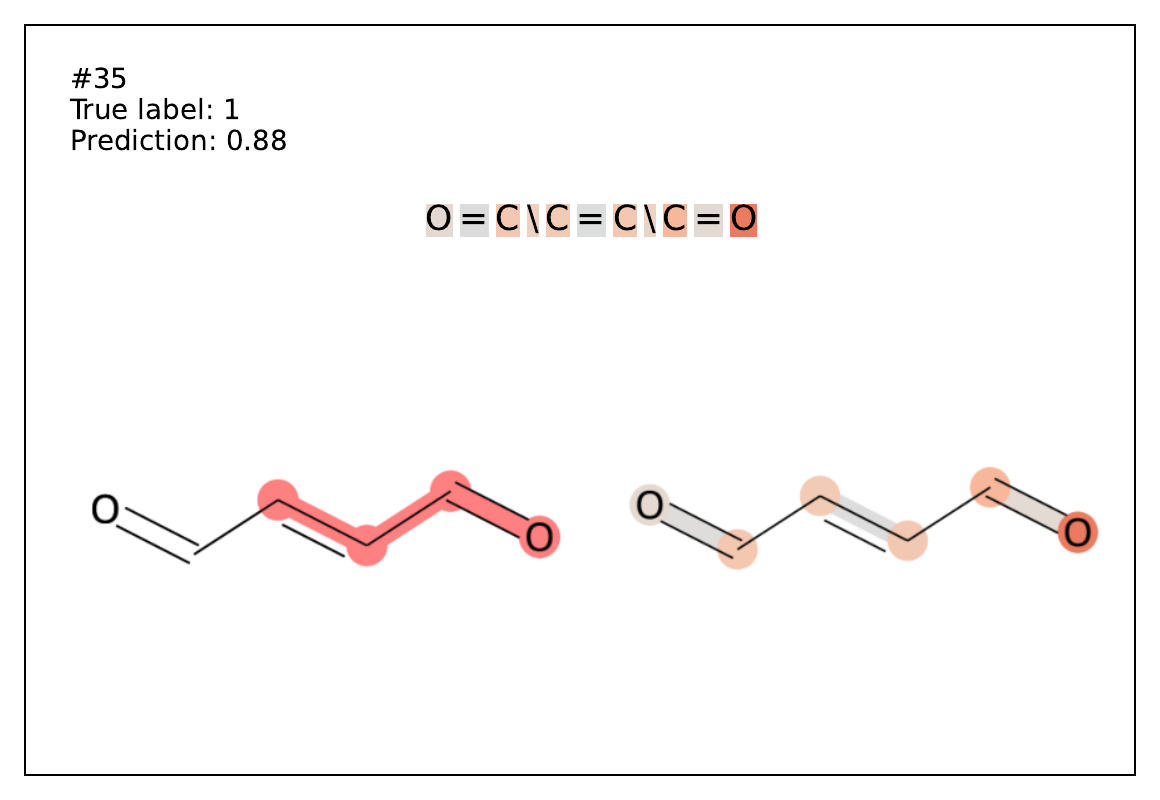} 
\end{subfigure}\begin{subfigure}[b]{0.33\textwidth} 
  \centering 
  \includegraphics[width=\textwidth]{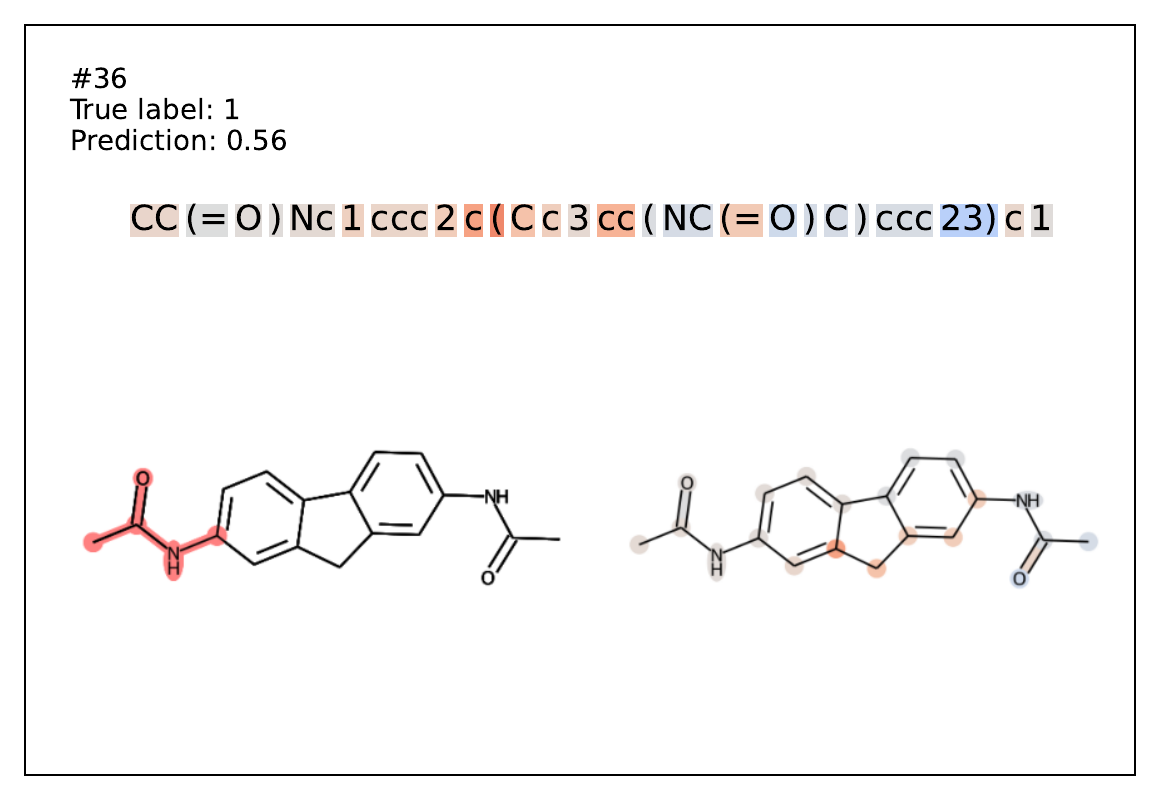} 
\end{subfigure} 
\begin{subfigure}[b]{0.33\textwidth} 
  \centering 
  \includegraphics[width=\textwidth]{figures/ames/ames37.pdf} 
\end{subfigure}\begin{subfigure}[b]{0.33\textwidth} 
  \centering 
  \includegraphics[width=\textwidth]{figures/ames/ames38.pdf} 
\end{subfigure}\begin{subfigure}[b]{0.33\textwidth} 
  \centering 
  \includegraphics[width=\textwidth]{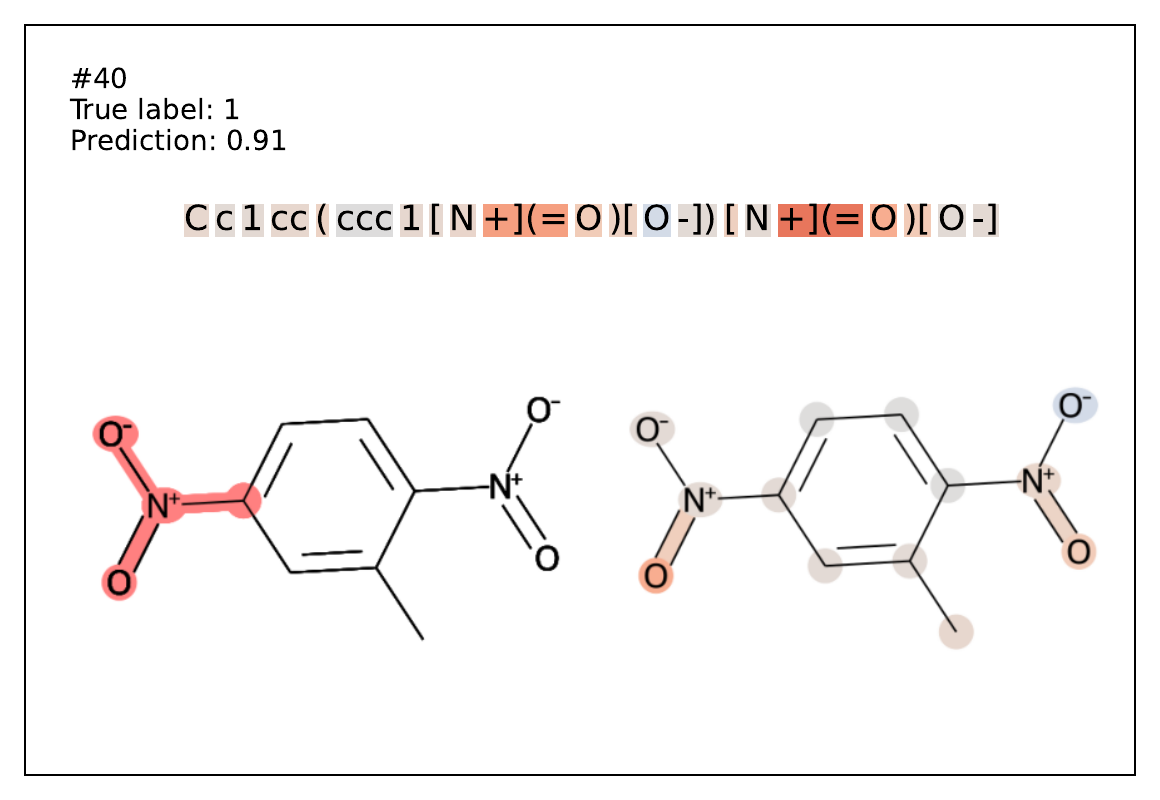} 
\end{subfigure} 
\begin{subfigure}[b]{0.33\textwidth} 
  \centering 
  \includegraphics[width=\textwidth]{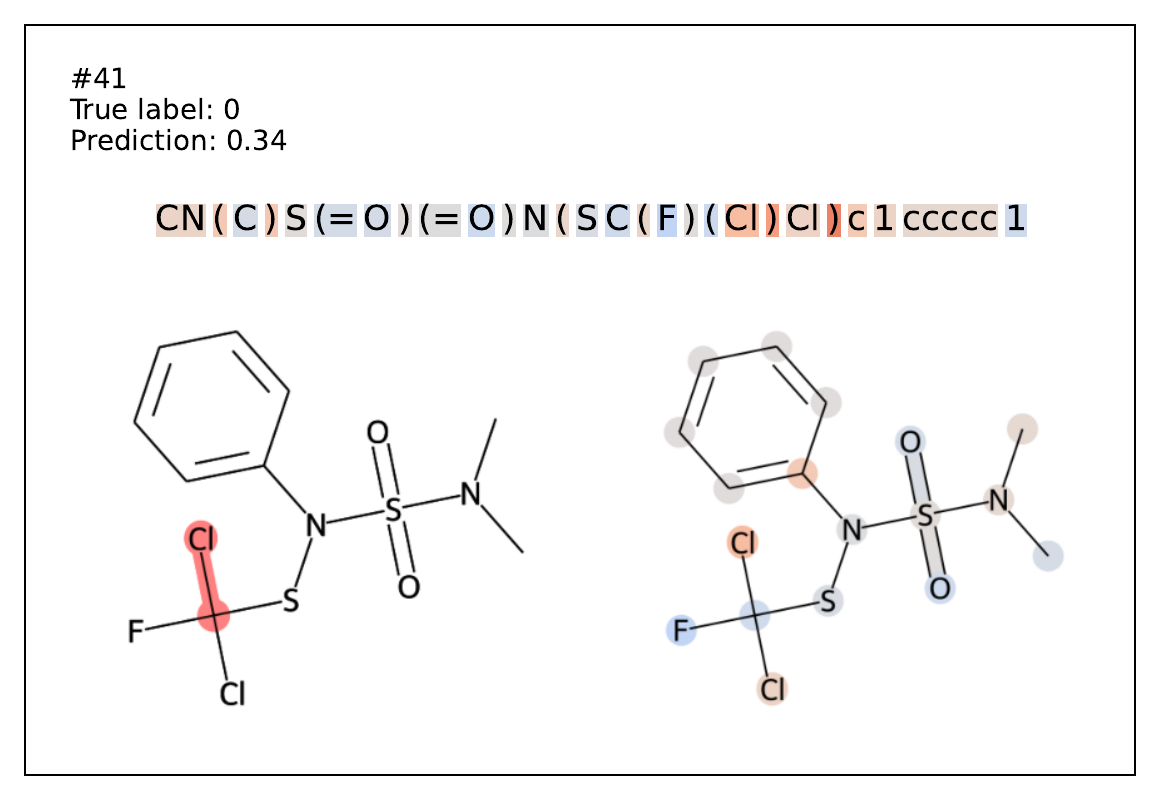} 
\end{subfigure}\begin{subfigure}[b]{0.33\textwidth} 
  \centering 
  \includegraphics[width=\textwidth]{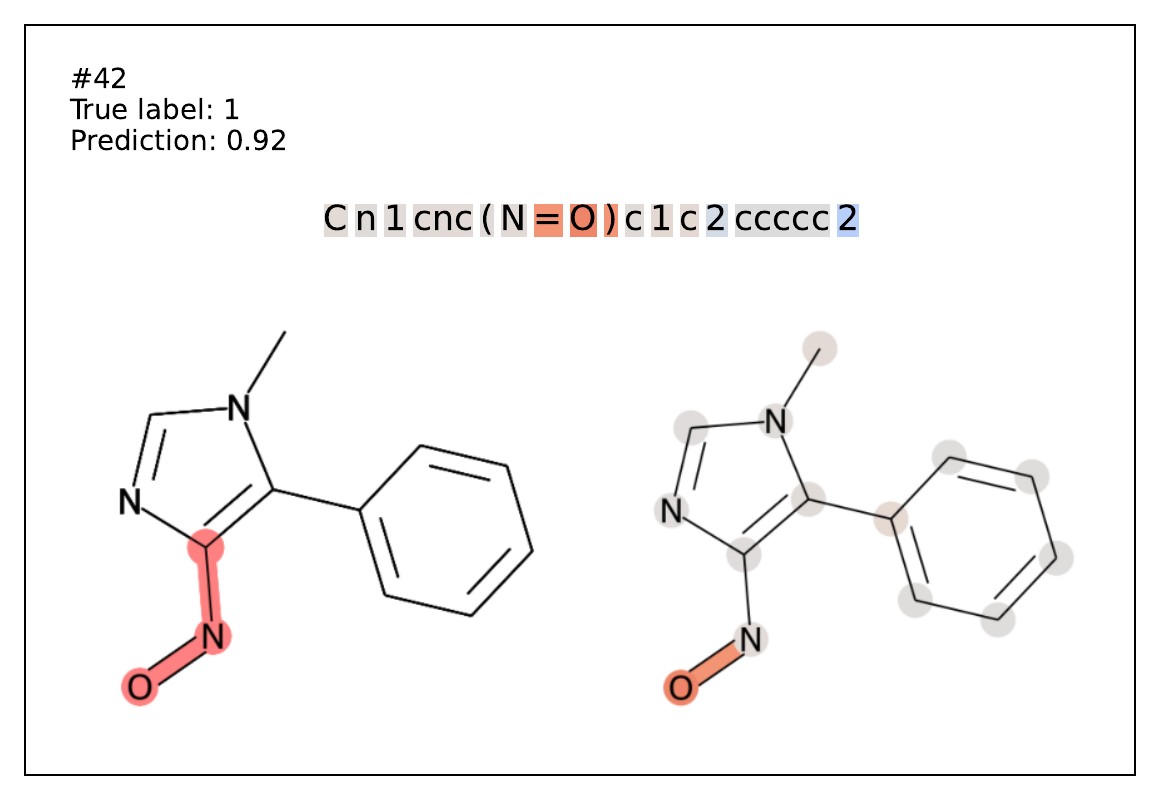} 
\end{subfigure}\begin{subfigure}[b]{0.33\textwidth} 
  \centering 
  \includegraphics[width=\textwidth]{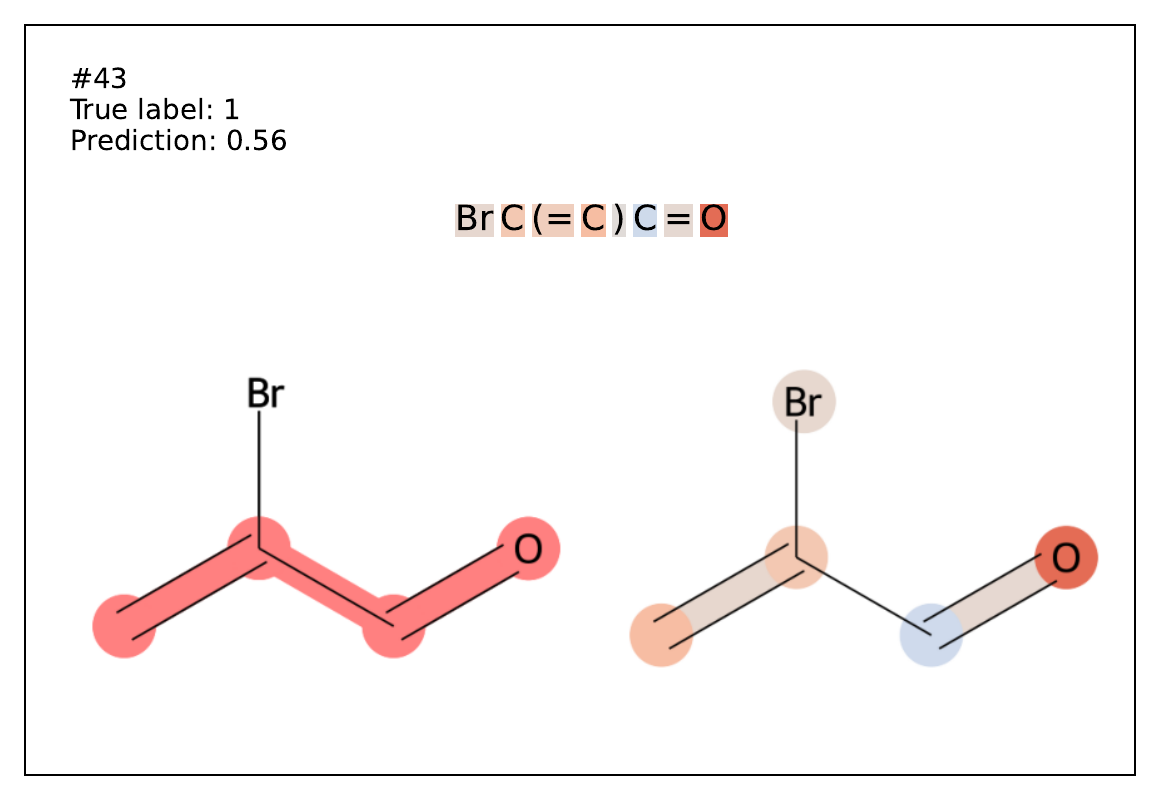} 
\end{subfigure} 
\begin{subfigure}[b]{0.33\textwidth} 
  \centering 
  \includegraphics[width=\textwidth]{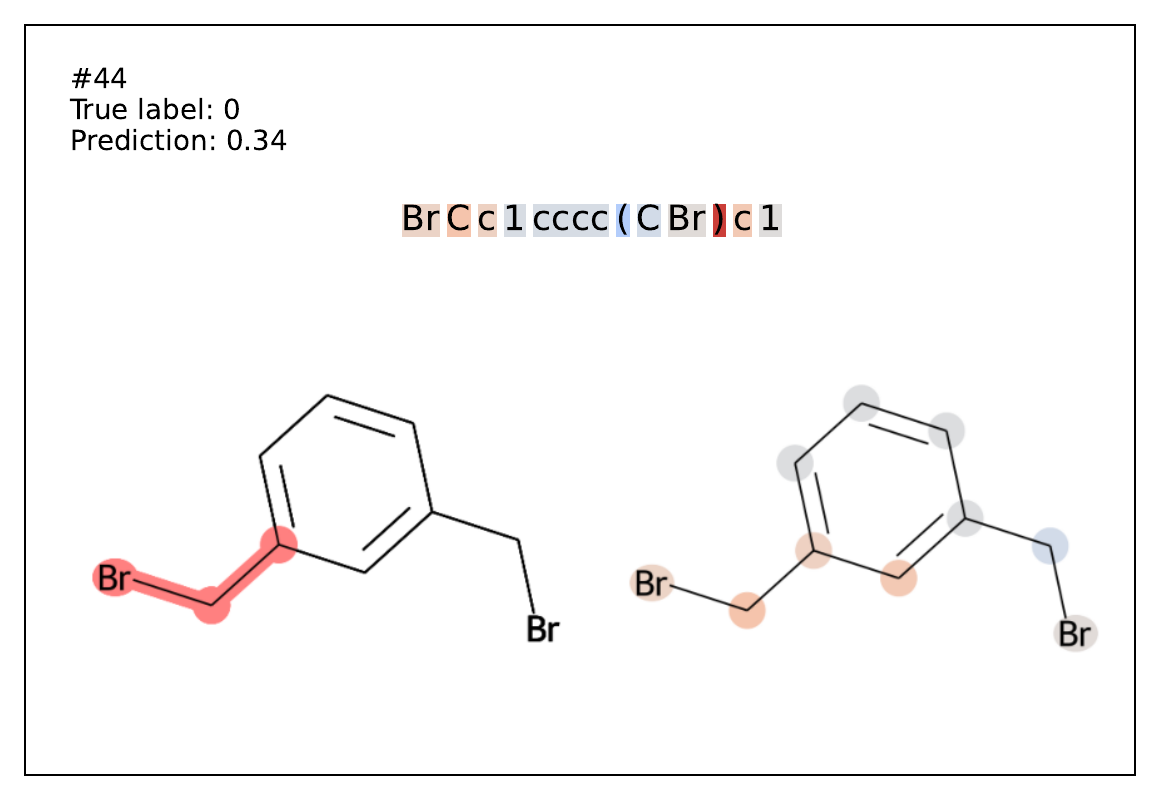} 
\end{subfigure}\begin{subfigure}[b]{0.33\textwidth} 
  \centering 
  \includegraphics[width=\textwidth]{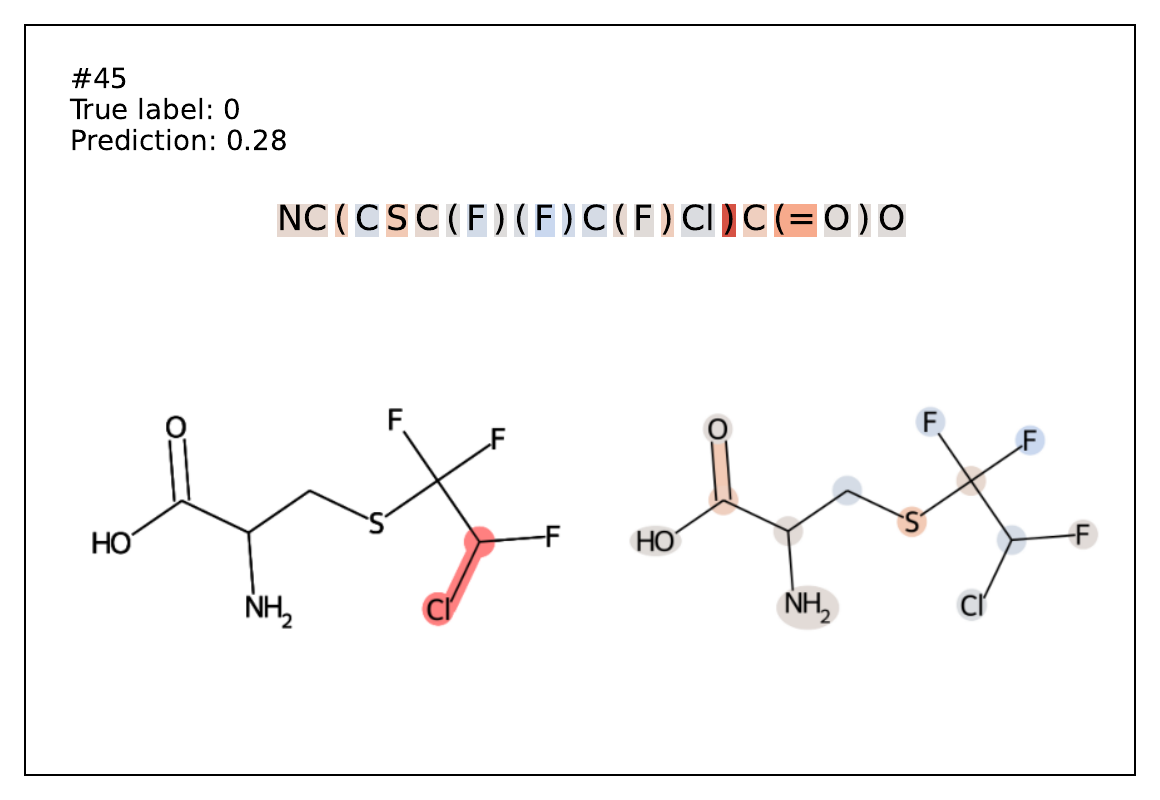} 
\end{subfigure}\begin{subfigure}[b]{0.33\textwidth} 
  \centering 
  \includegraphics[width=\textwidth]{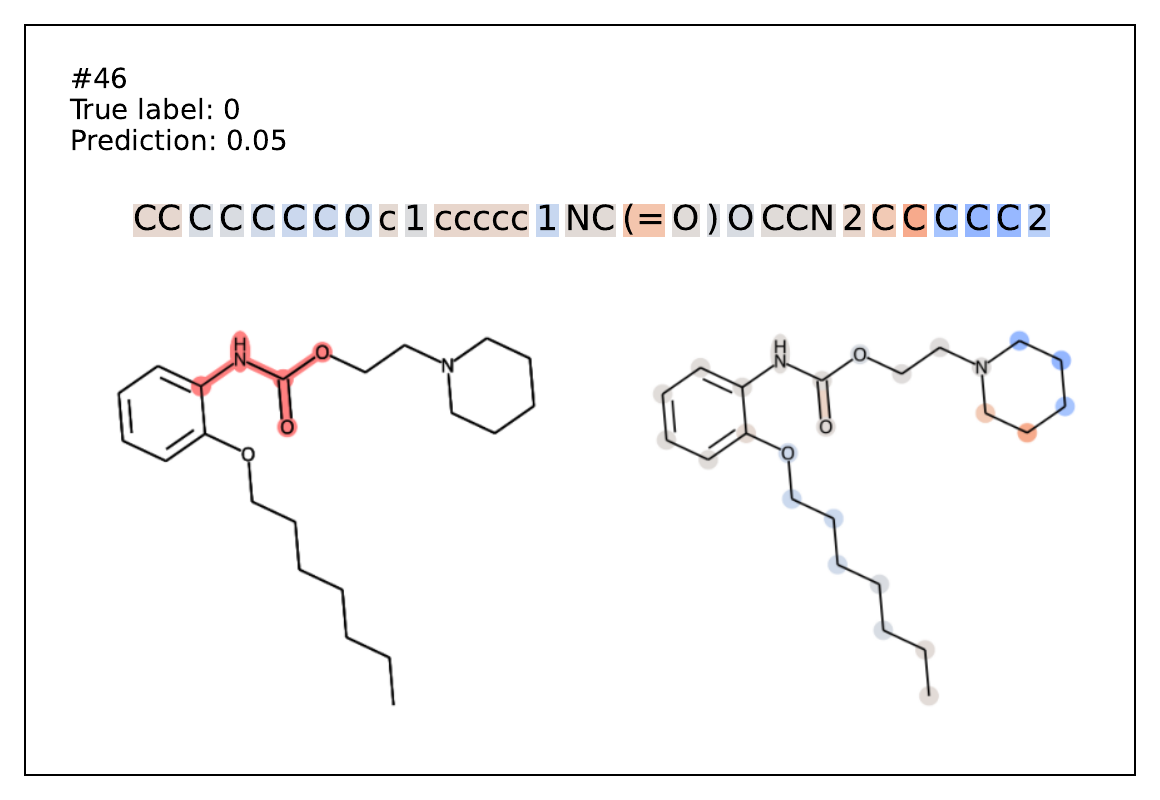} 
\end{subfigure} 
\begin{subfigure}[b]{0.33\textwidth} 
  \centering 
  \includegraphics[width=\textwidth]{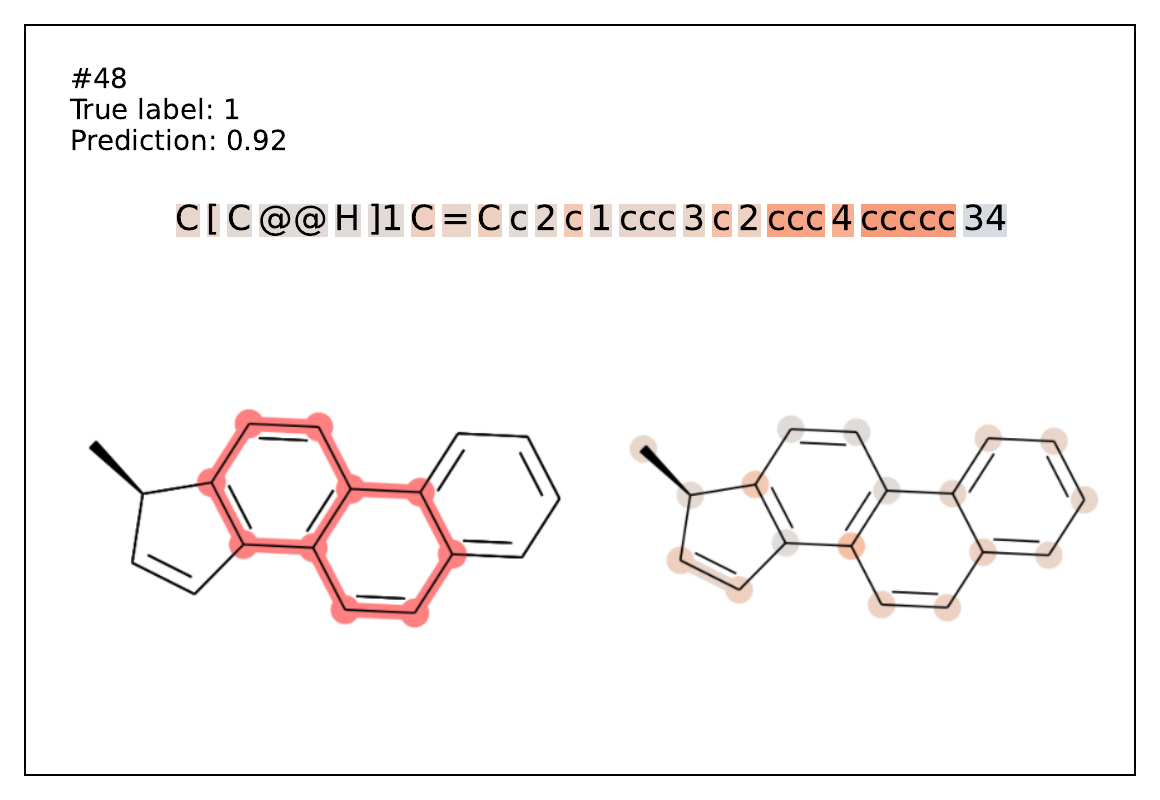} 
\end{subfigure}\begin{subfigure}[b]{0.33\textwidth} 
  \centering 
  \includegraphics[width=\textwidth]{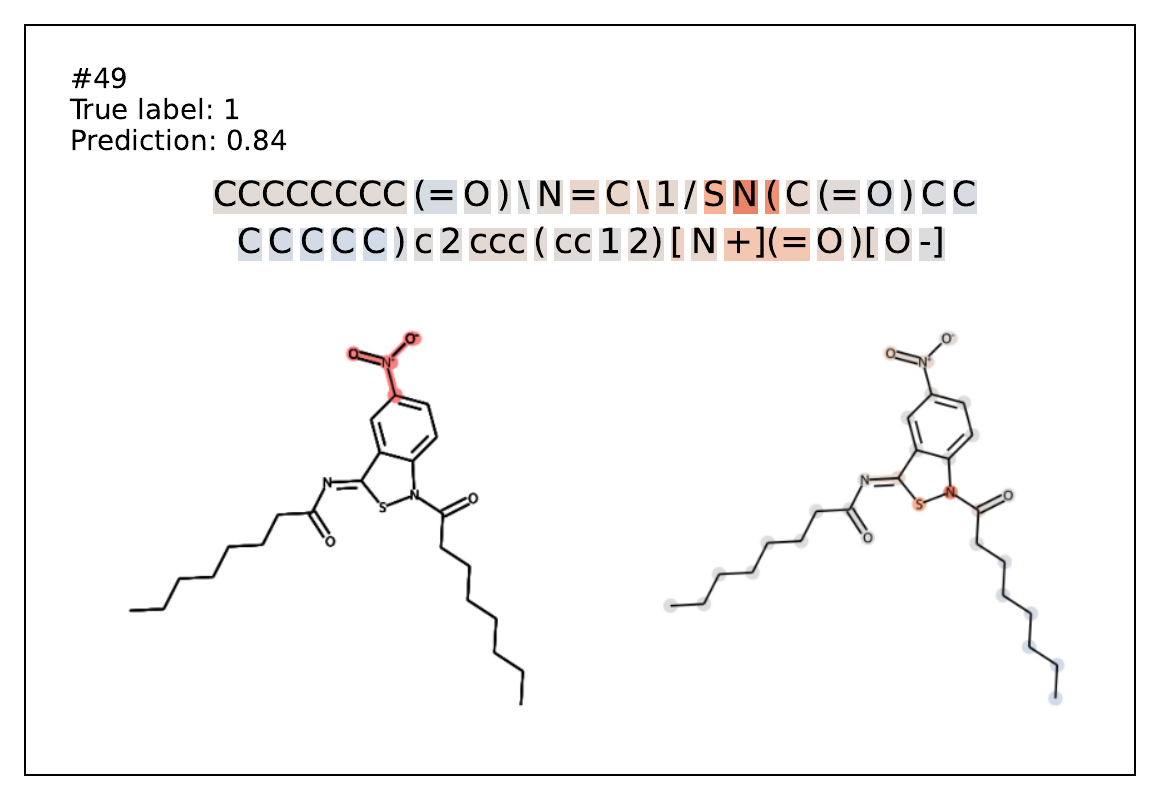} 
\end{subfigure}\begin{subfigure}[b]{0.33\textwidth} 
  \centering 
  \includegraphics[width=\textwidth]{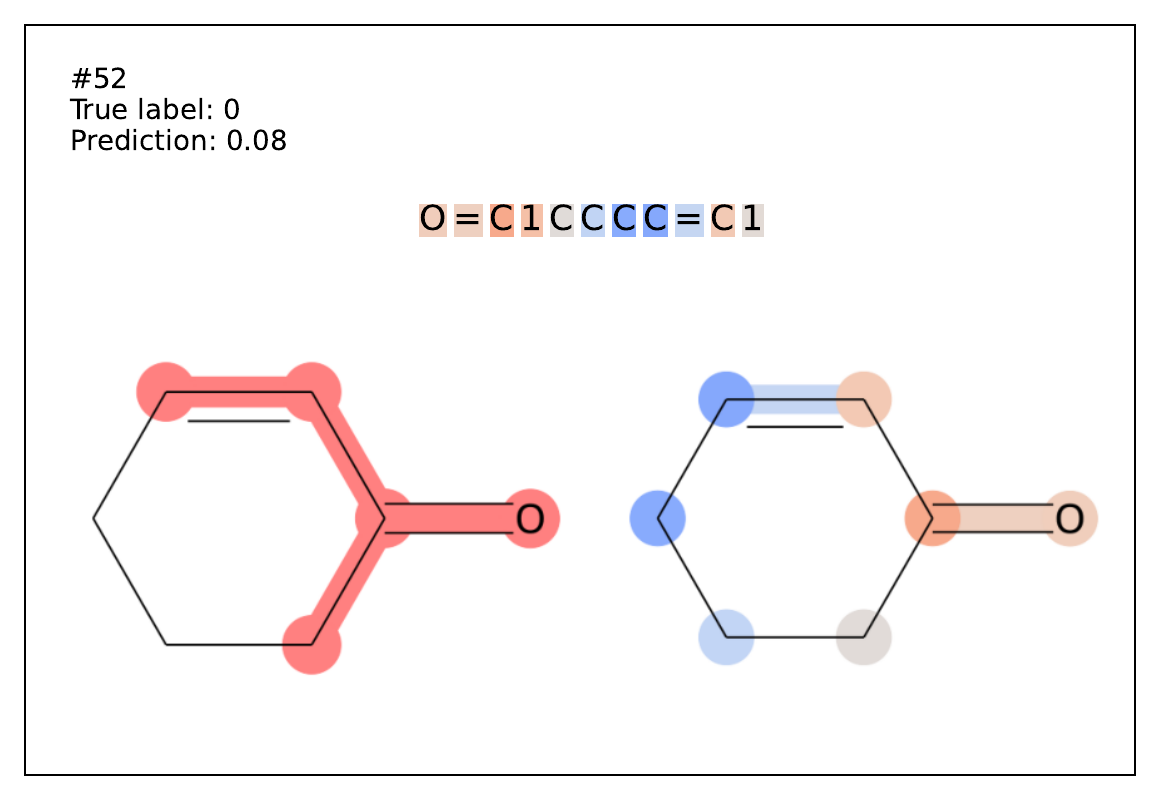} 
\end{subfigure} 
\begin{subfigure}[b]{0.33\textwidth} 
  \centering 
  \includegraphics[width=\textwidth]{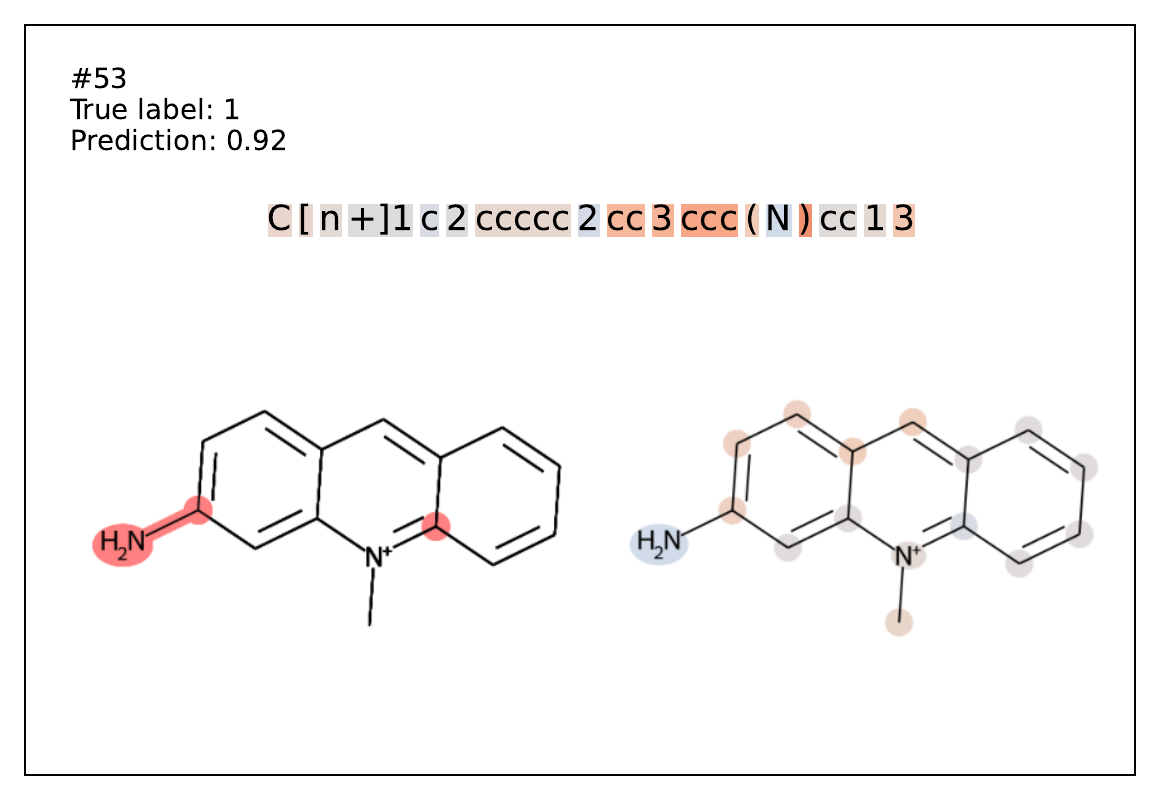} 
\end{subfigure}\begin{subfigure}[b]{0.33\textwidth} 
  \centering 
  \includegraphics[width=\textwidth]{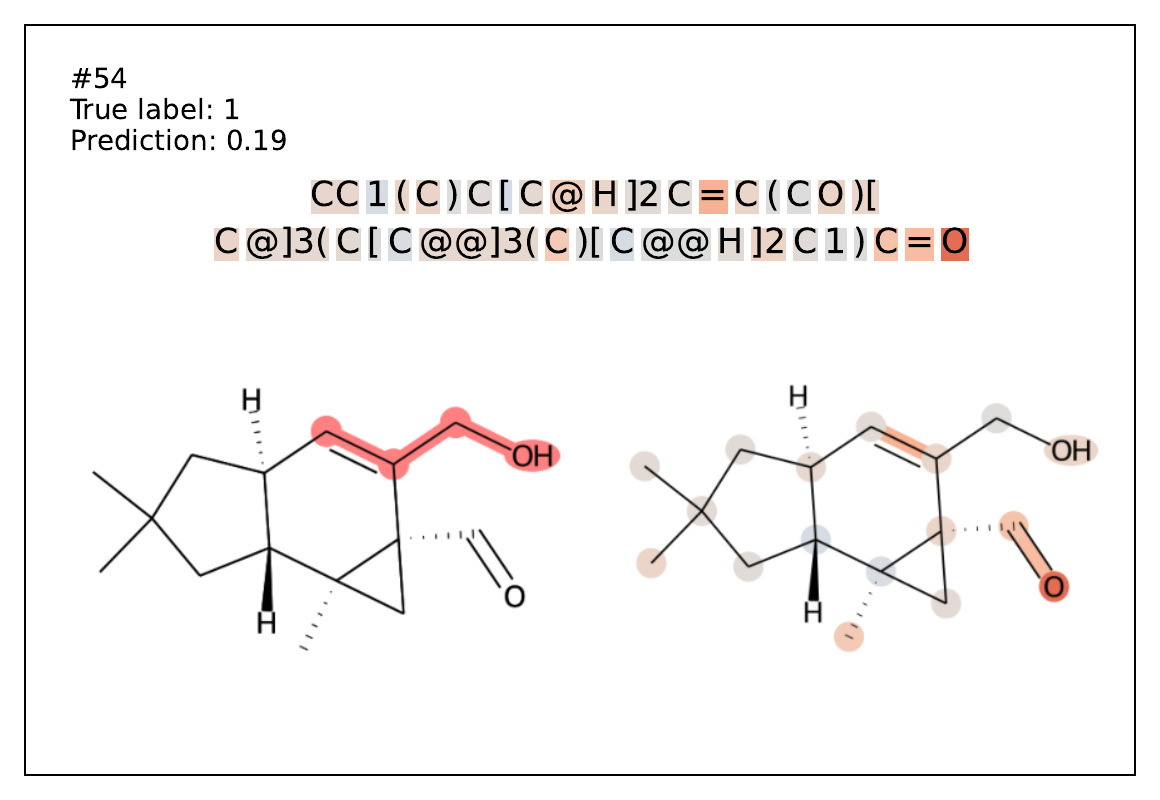} 
\end{subfigure}\begin{subfigure}[b]{0.33\textwidth} 
  \centering 
  \includegraphics[width=\textwidth]{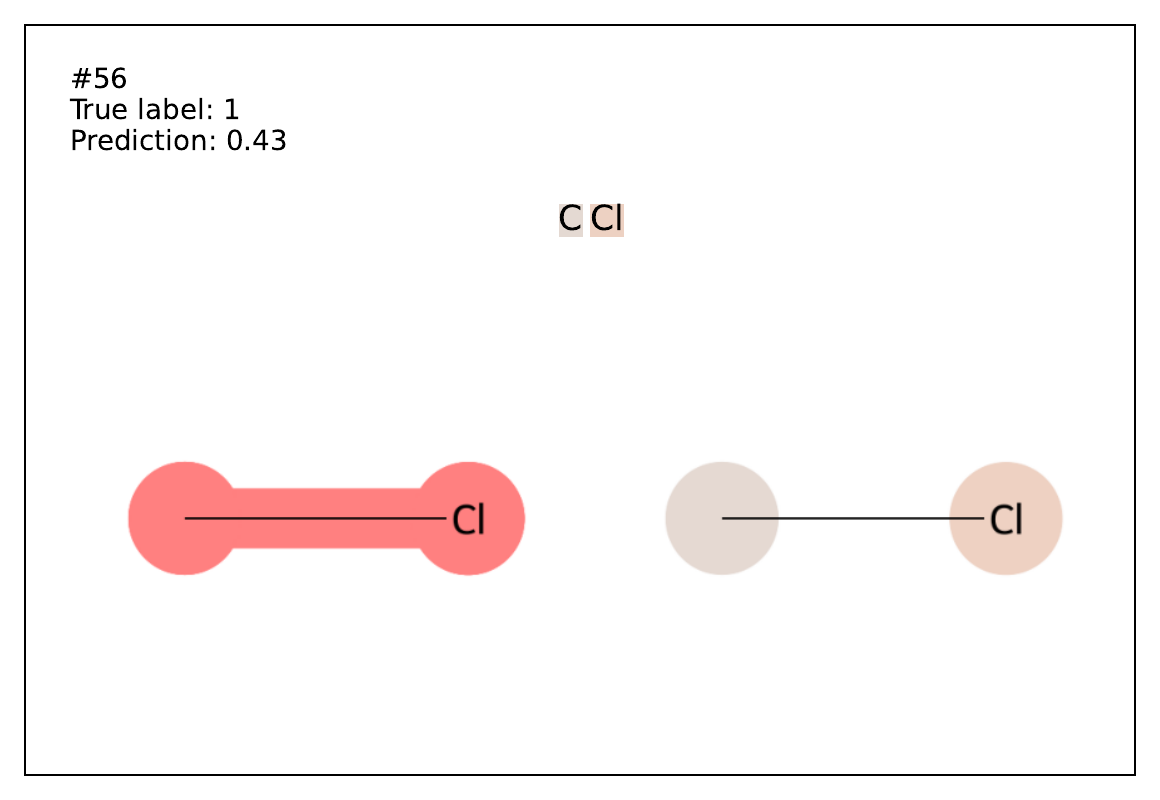} 
\end{subfigure} 
\caption{Explaining predictions of the fine-tuned model on Ames dataset. See Section \ref{sec:captum}. Part 2/5}
\label{fig:captum-ames-2}
\end{figure}

\begin{figure}
\centering
\begin{subfigure}[b]{0.33\textwidth} 
  \centering 
  \includegraphics[width=\textwidth]{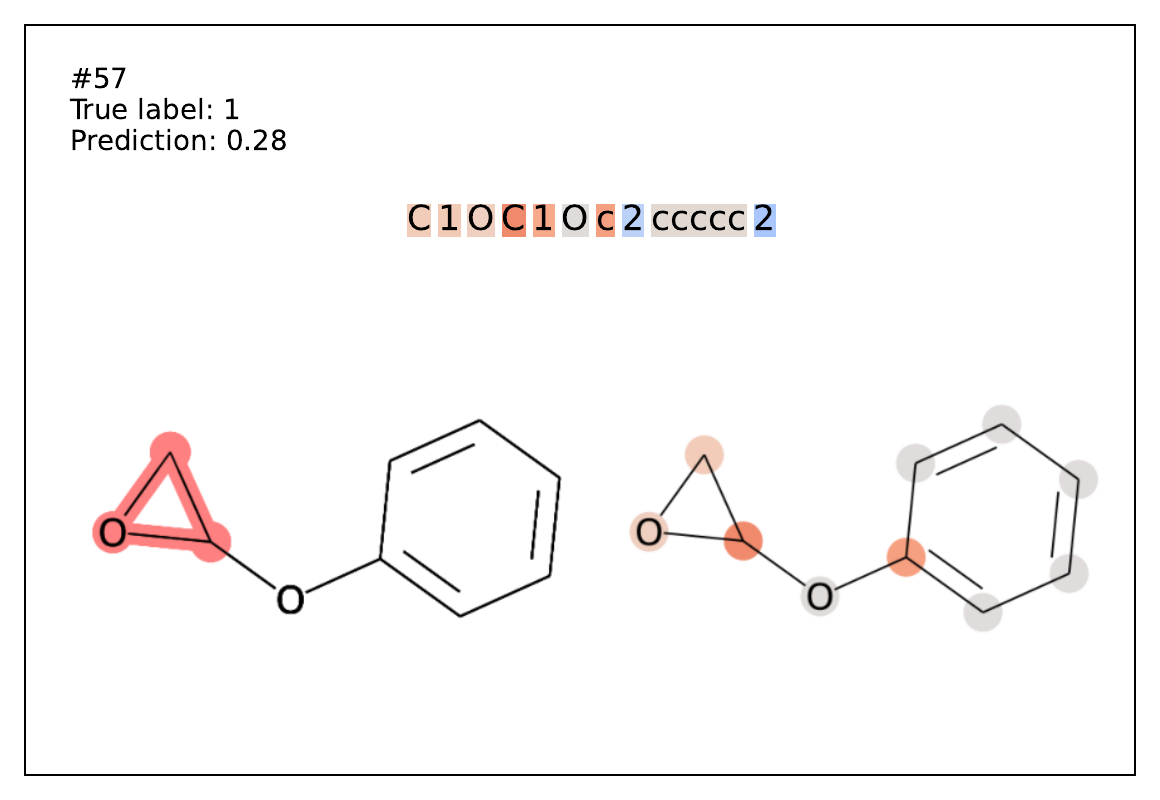} 
\end{subfigure}\begin{subfigure}[b]{0.33\textwidth} 
  \centering 
  \includegraphics[width=\textwidth]{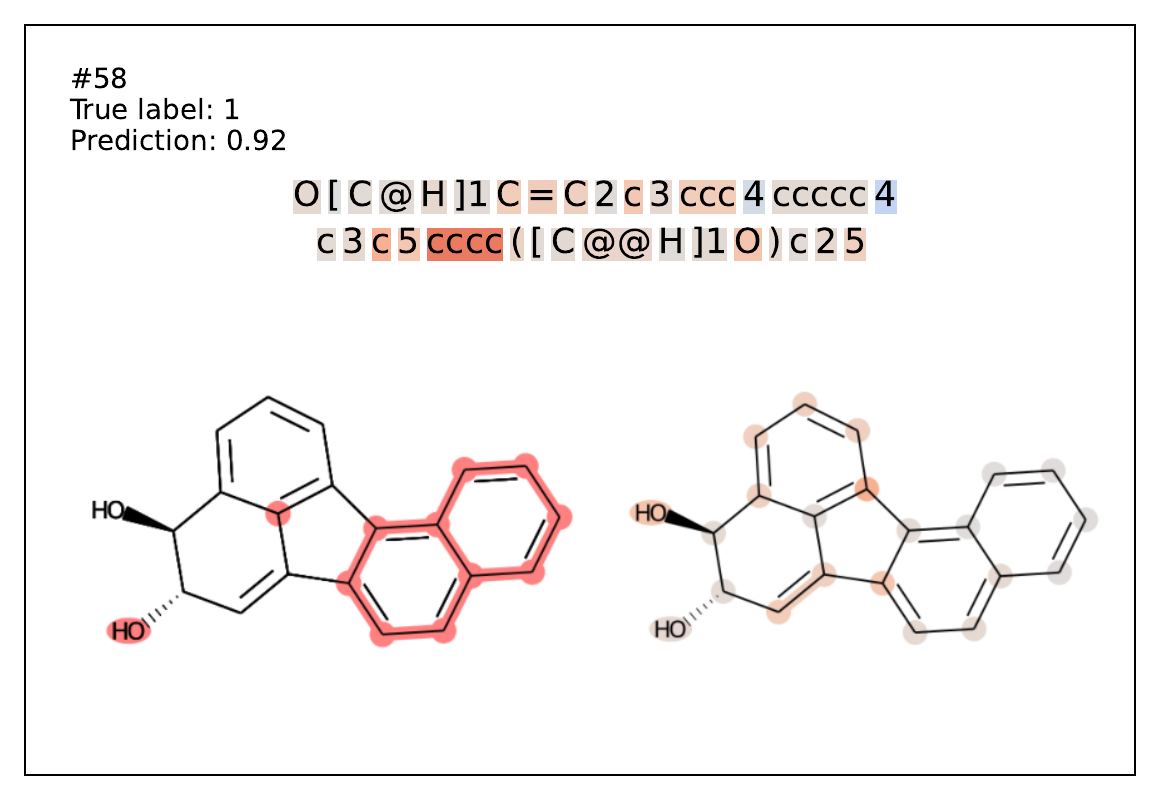} 
\end{subfigure}\begin{subfigure}[b]{0.33\textwidth} 
  \centering 
  \includegraphics[width=\textwidth]{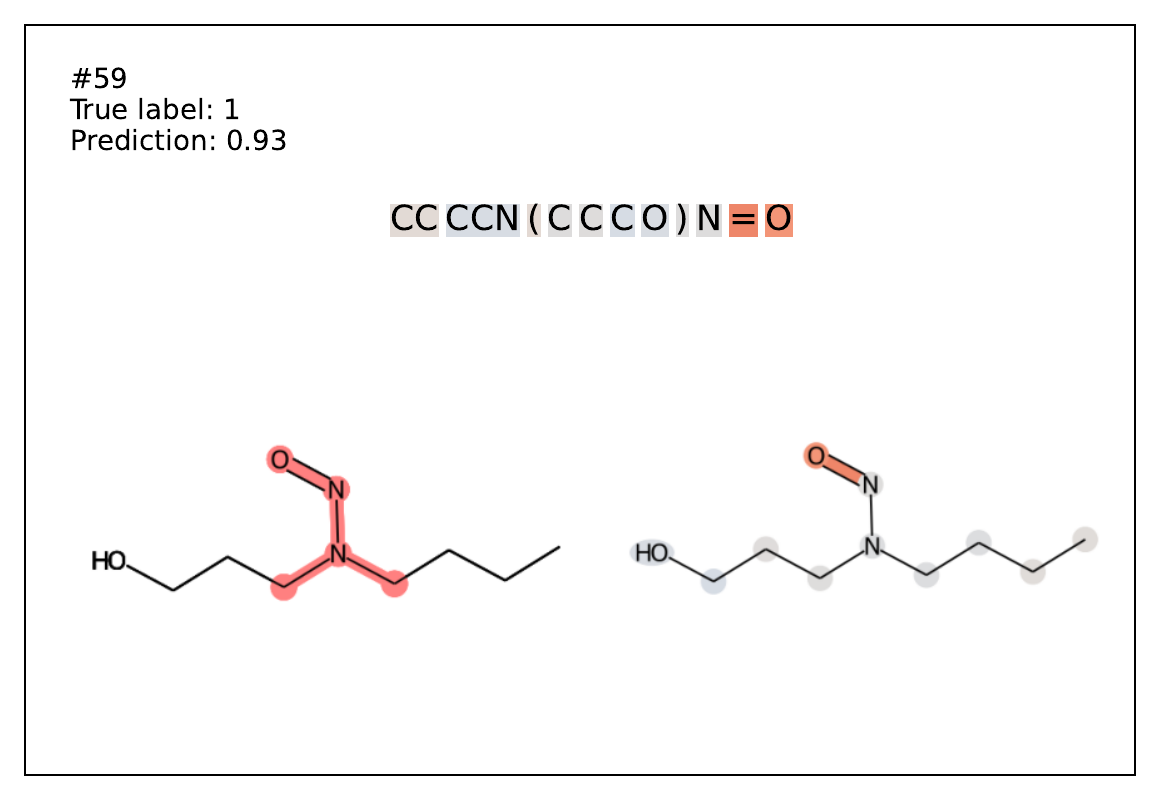} 
\end{subfigure} 
\begin{subfigure}[b]{0.33\textwidth} 
  \centering 
  \includegraphics[width=\textwidth]{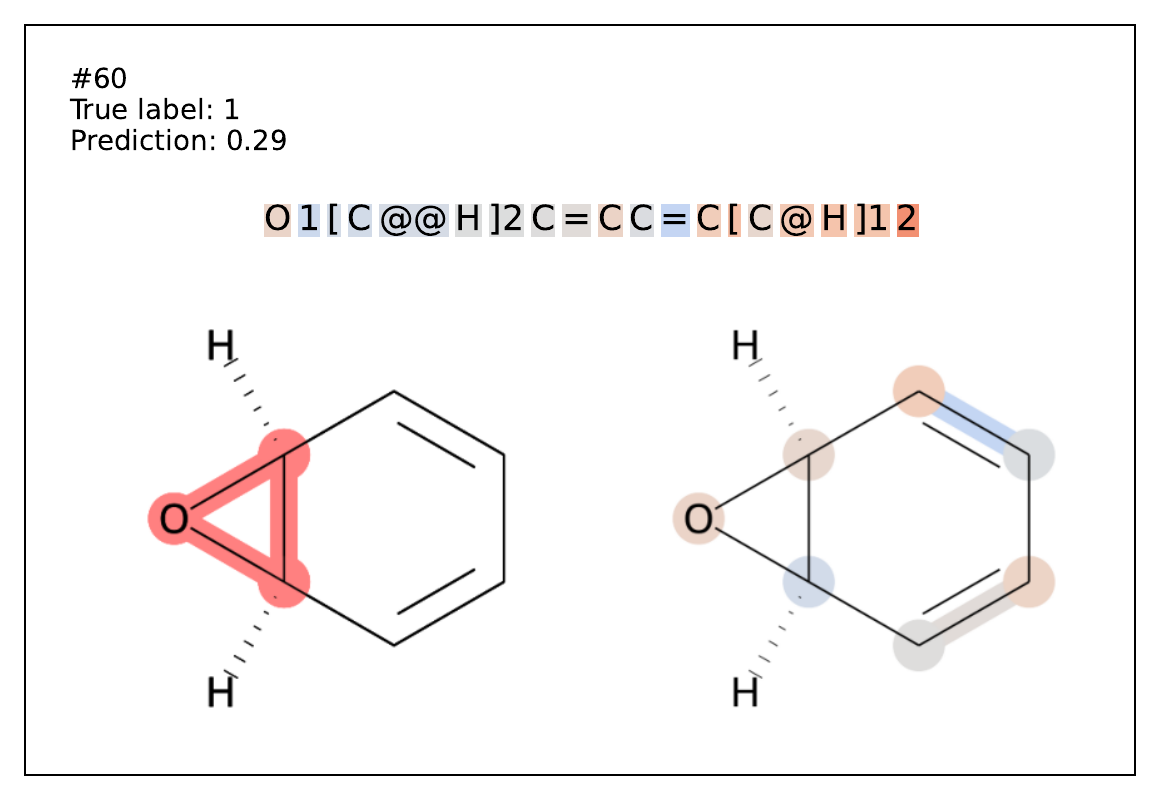} 
\end{subfigure}\begin{subfigure}[b]{0.33\textwidth} 
  \centering 
  \includegraphics[width=\textwidth]{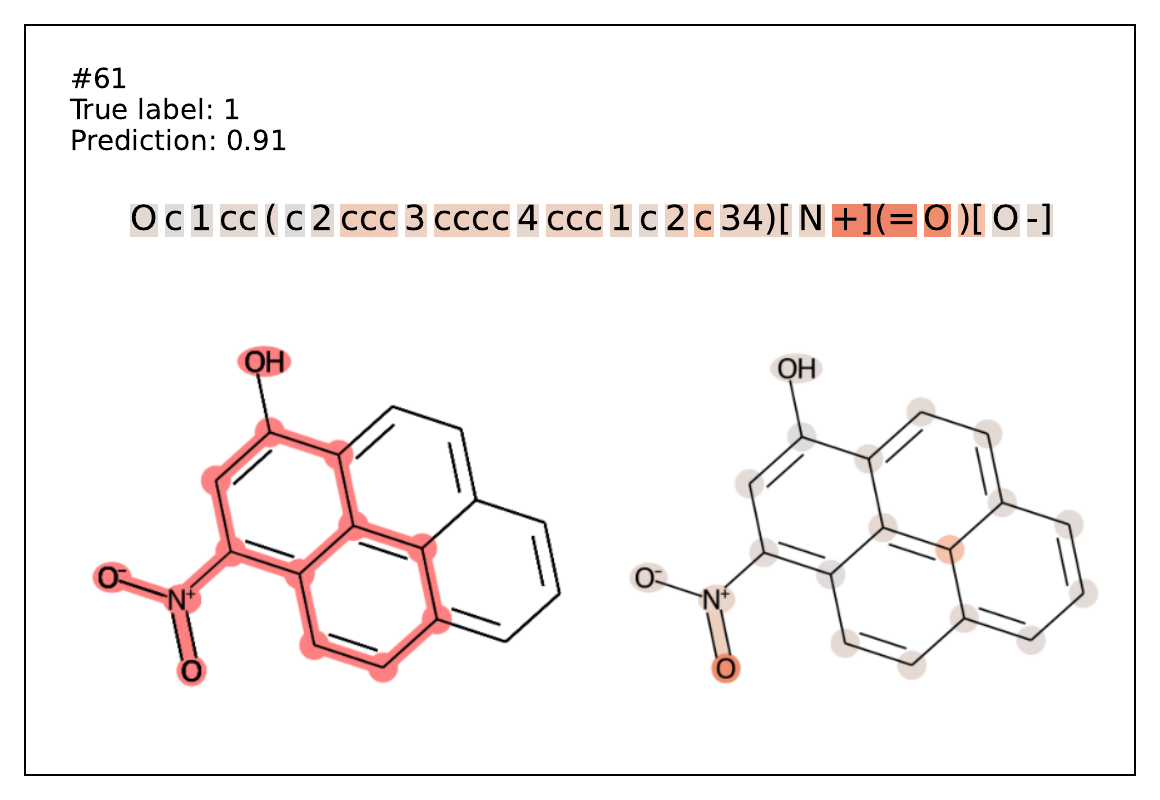} 
\end{subfigure}\begin{subfigure}[b]{0.33\textwidth} 
  \centering 
  \includegraphics[width=\textwidth]{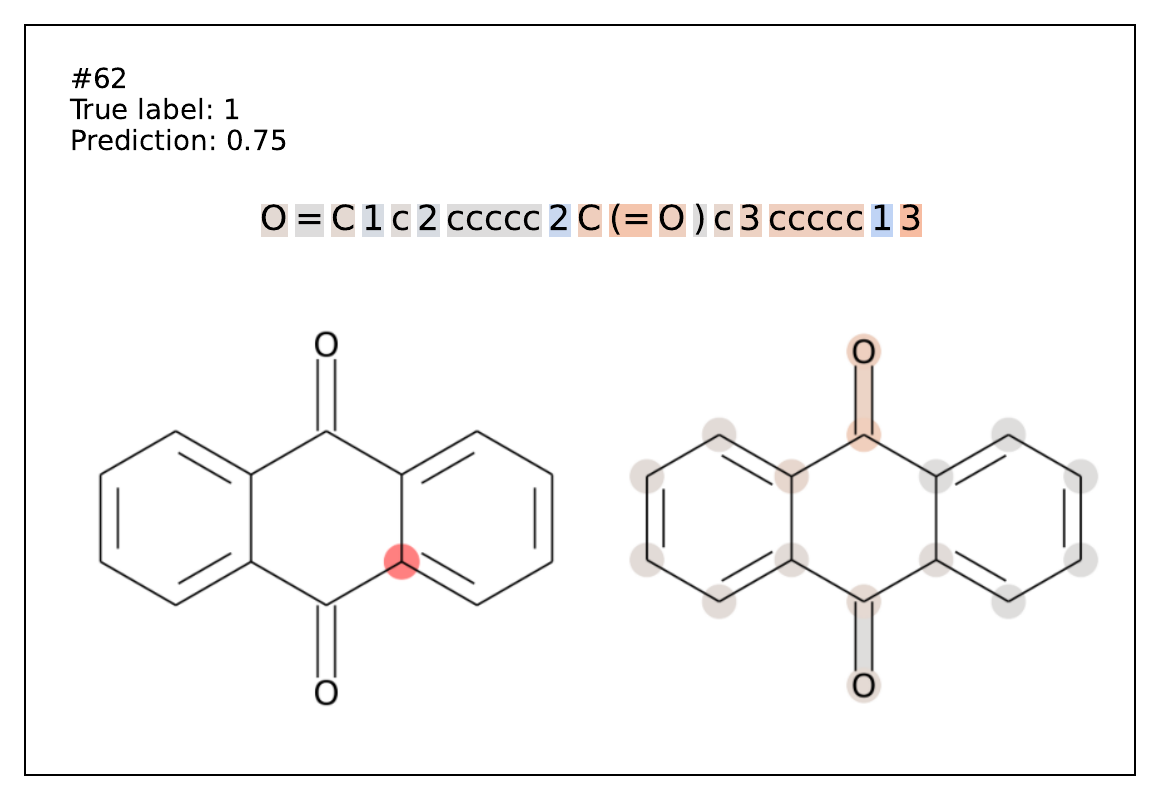} 
\end{subfigure} 
\begin{subfigure}[b]{0.33\textwidth} 
  \centering 
  \includegraphics[width=\textwidth]{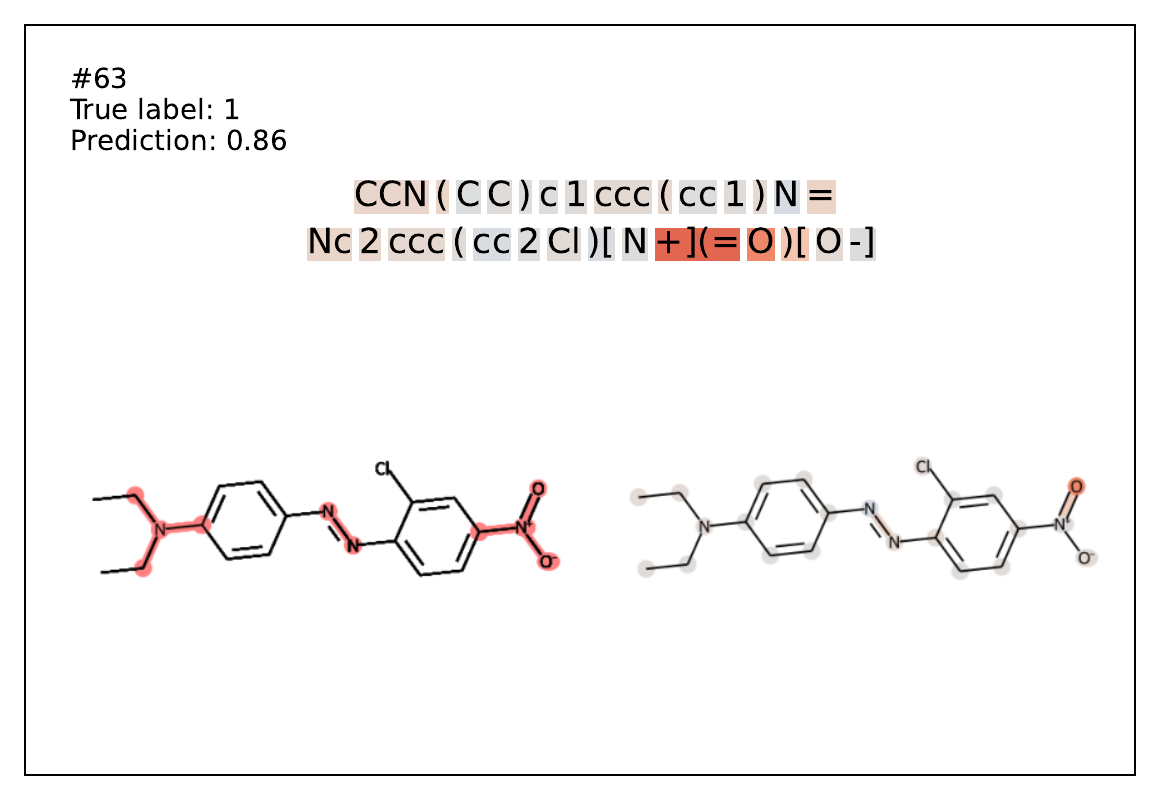} 
\end{subfigure}\begin{subfigure}[b]{0.33\textwidth} 
  \centering 
  \includegraphics[width=\textwidth]{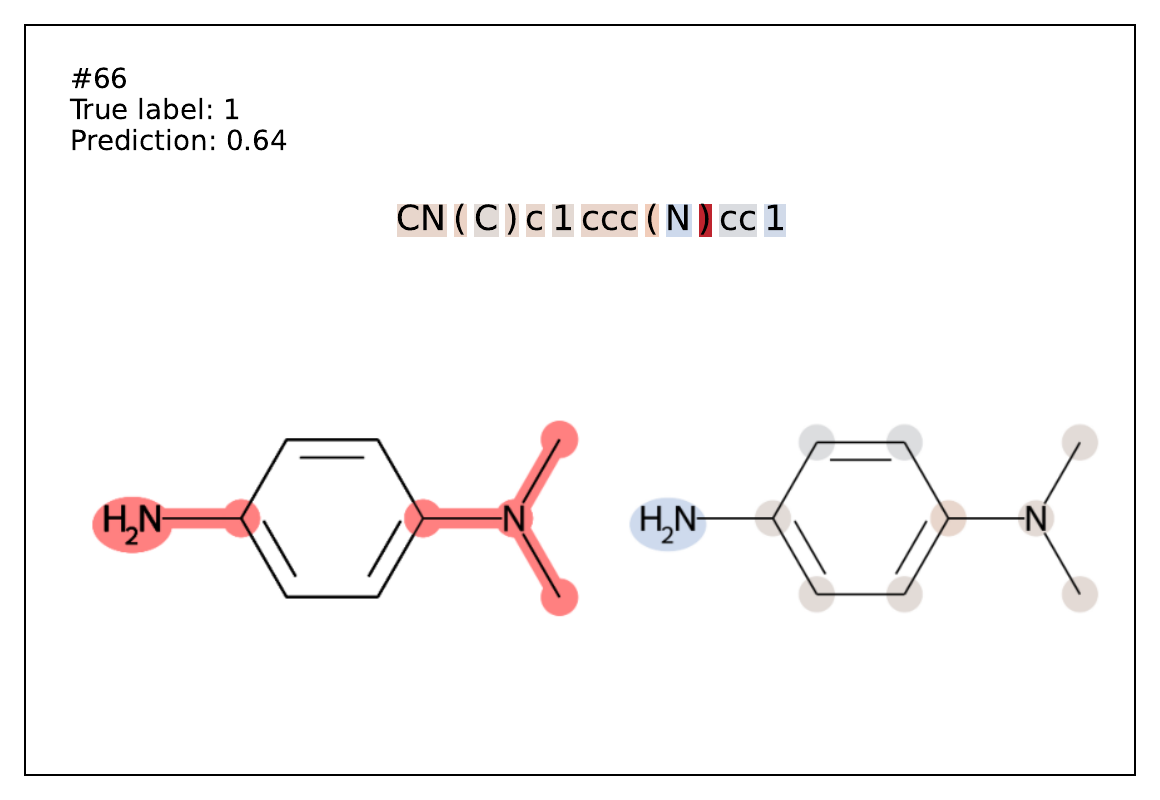} 
\end{subfigure}\begin{subfigure}[b]{0.33\textwidth} 
  \centering 
  \includegraphics[width=\textwidth]{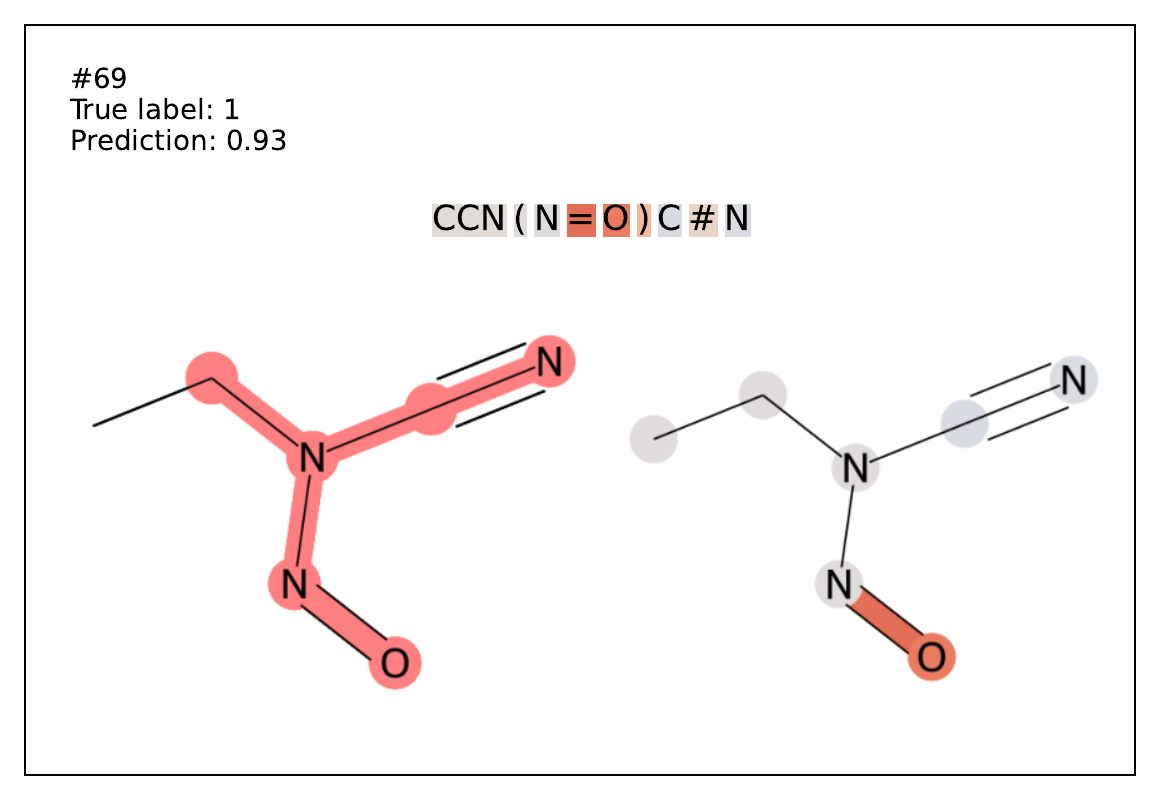} 
\end{subfigure} 
\begin{subfigure}[b]{0.33\textwidth} 
  \centering 
  \includegraphics[width=\textwidth]{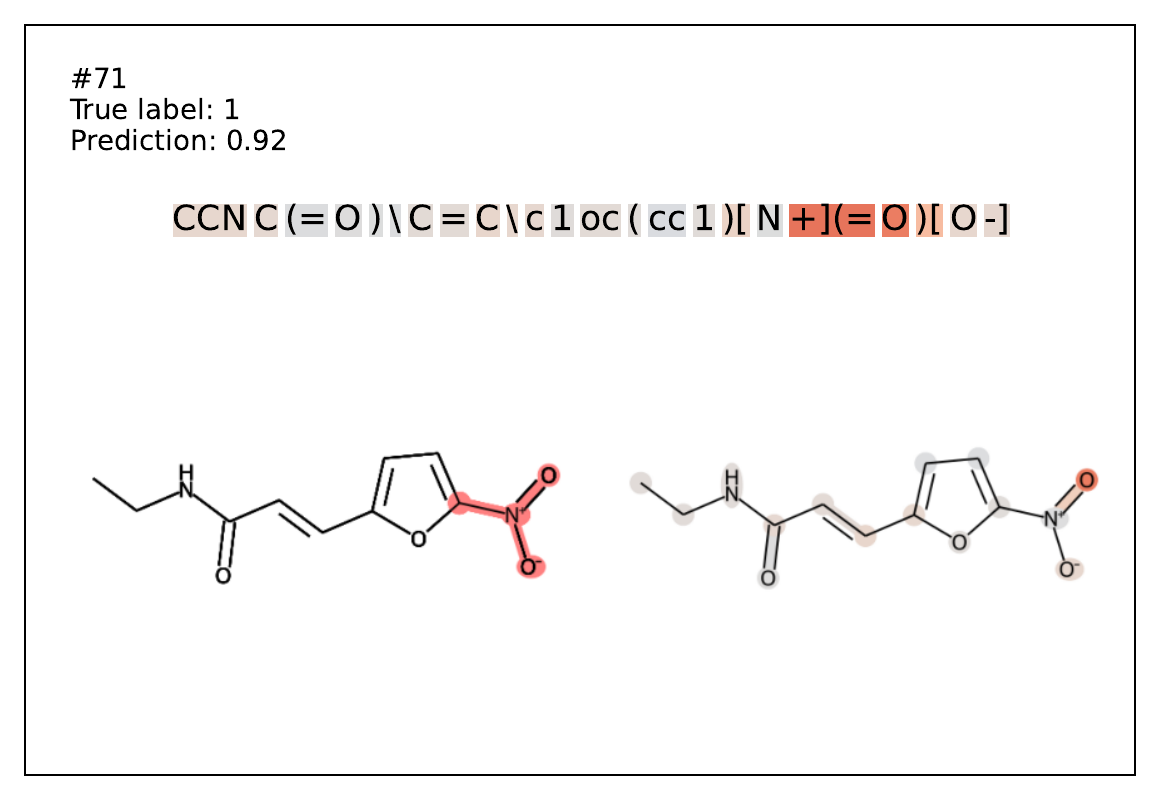} 
\end{subfigure}\begin{subfigure}[b]{0.33\textwidth} 
  \centering 
  \includegraphics[width=\textwidth]{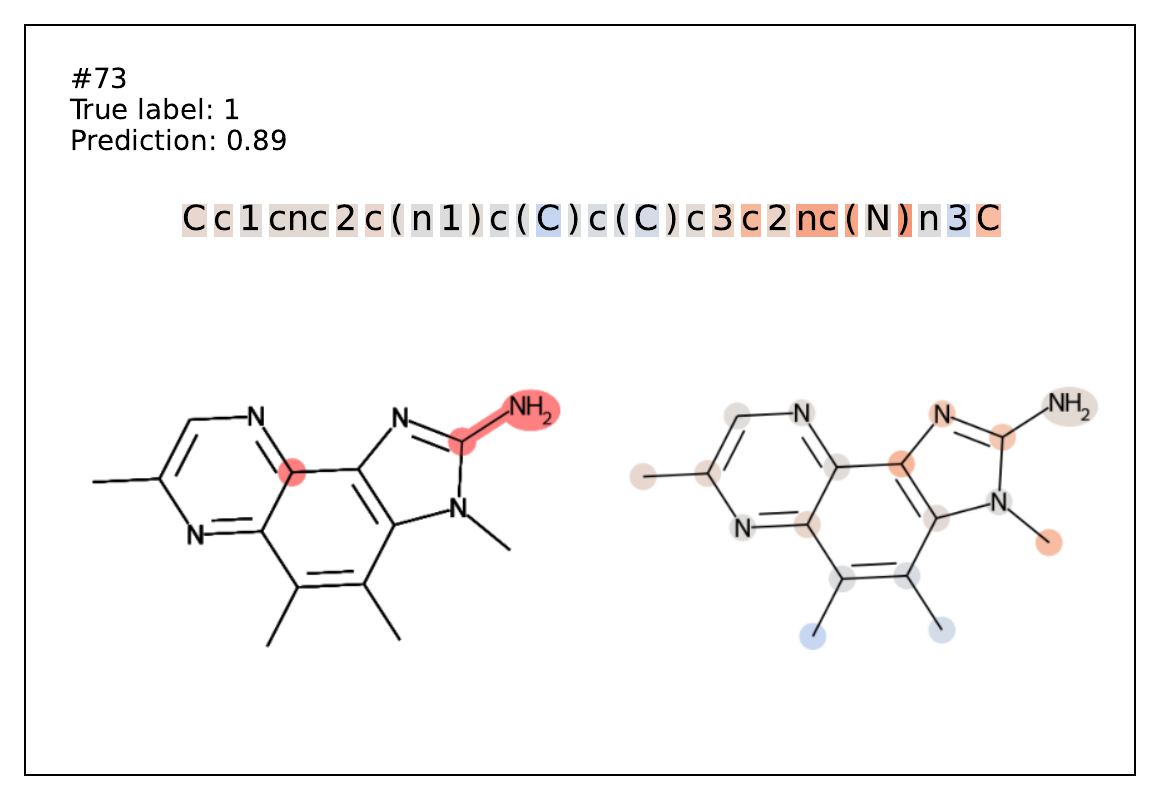} 
\end{subfigure}\begin{subfigure}[b]{0.33\textwidth} 
  \centering 
  \includegraphics[width=\textwidth]{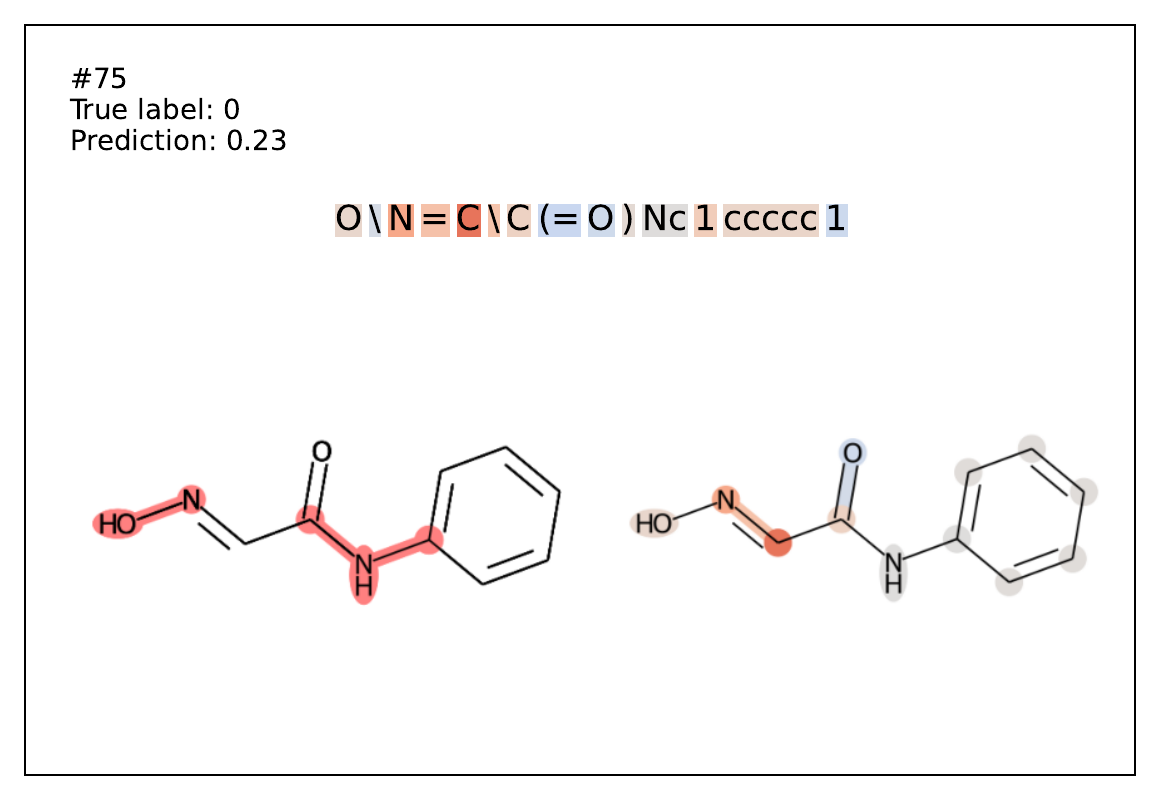} 
\end{subfigure} 
\begin{subfigure}[b]{0.33\textwidth} 
  \centering 
  \includegraphics[width=\textwidth]{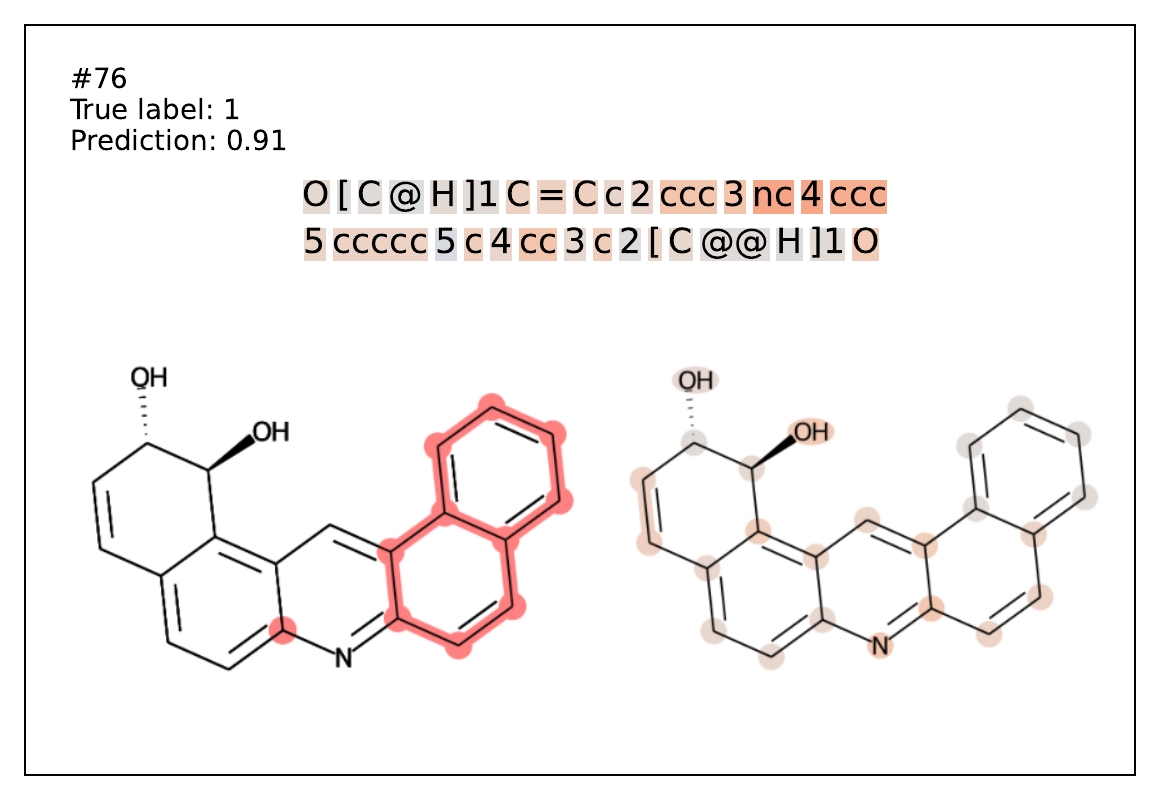} 
\end{subfigure}\begin{subfigure}[b]{0.33\textwidth} 
  \centering 
  \includegraphics[width=\textwidth]{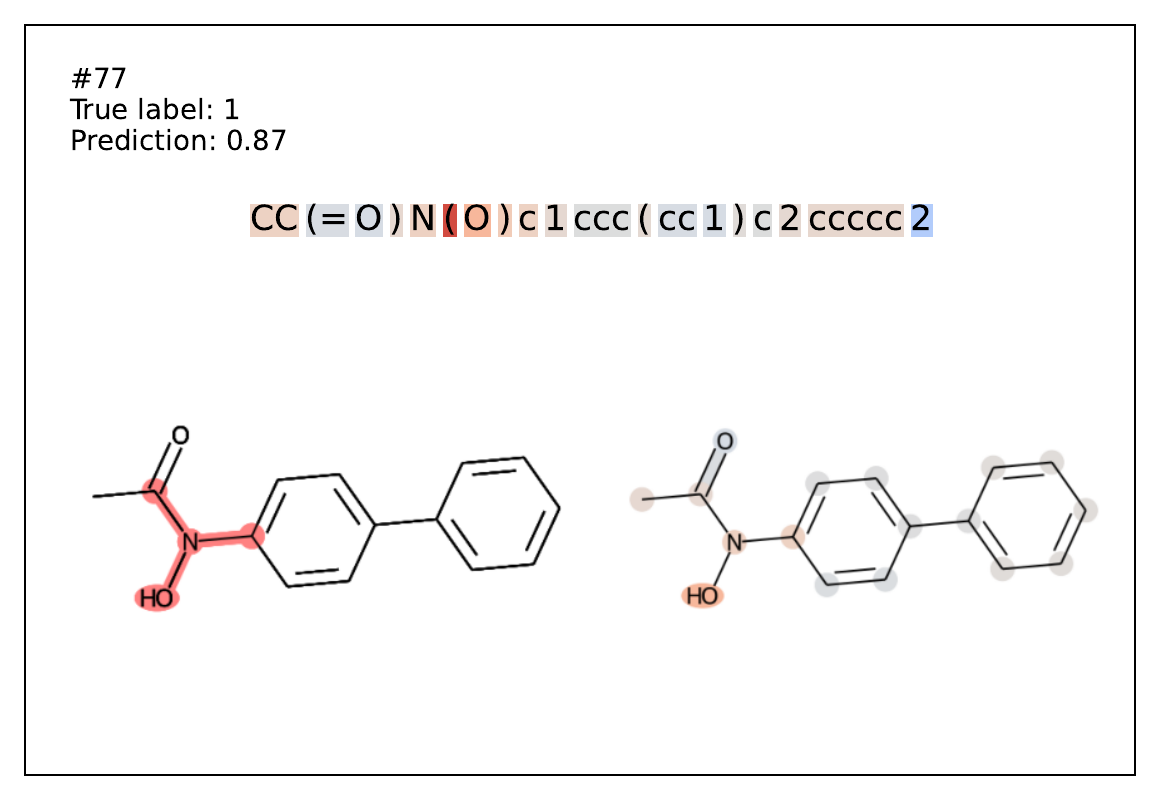} 
\end{subfigure}\begin{subfigure}[b]{0.33\textwidth} 
  \centering 
  \includegraphics[width=\textwidth]{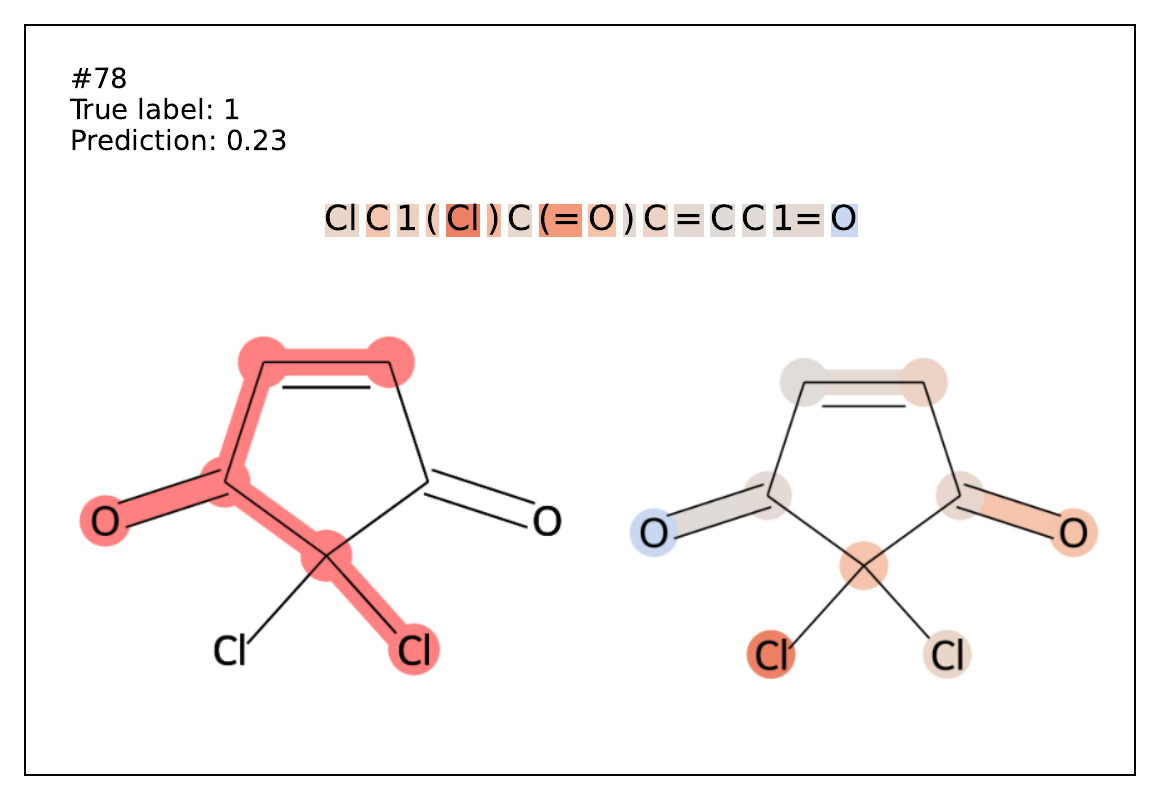} 
\end{subfigure} 
\begin{subfigure}[b]{0.33\textwidth} 
  \centering 
  \includegraphics[width=\textwidth]{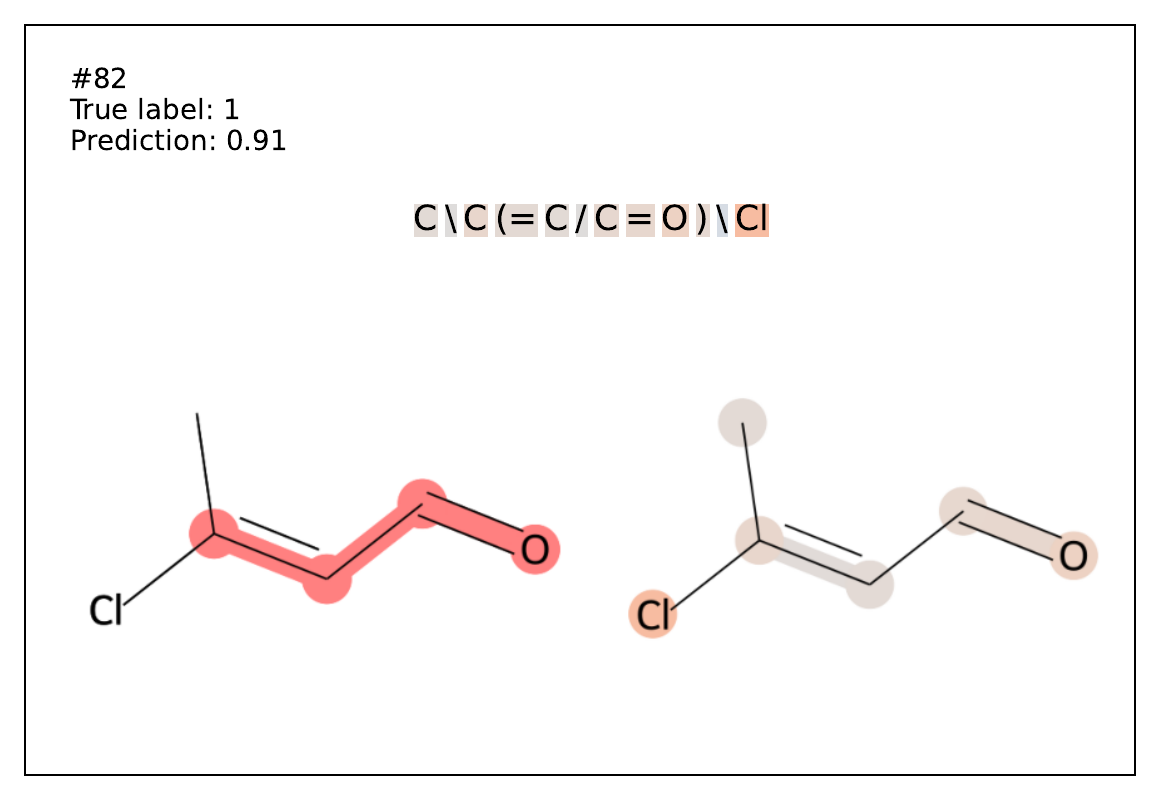} 
\end{subfigure}\begin{subfigure}[b]{0.33\textwidth} 
  \centering 
  \includegraphics[width=\textwidth]{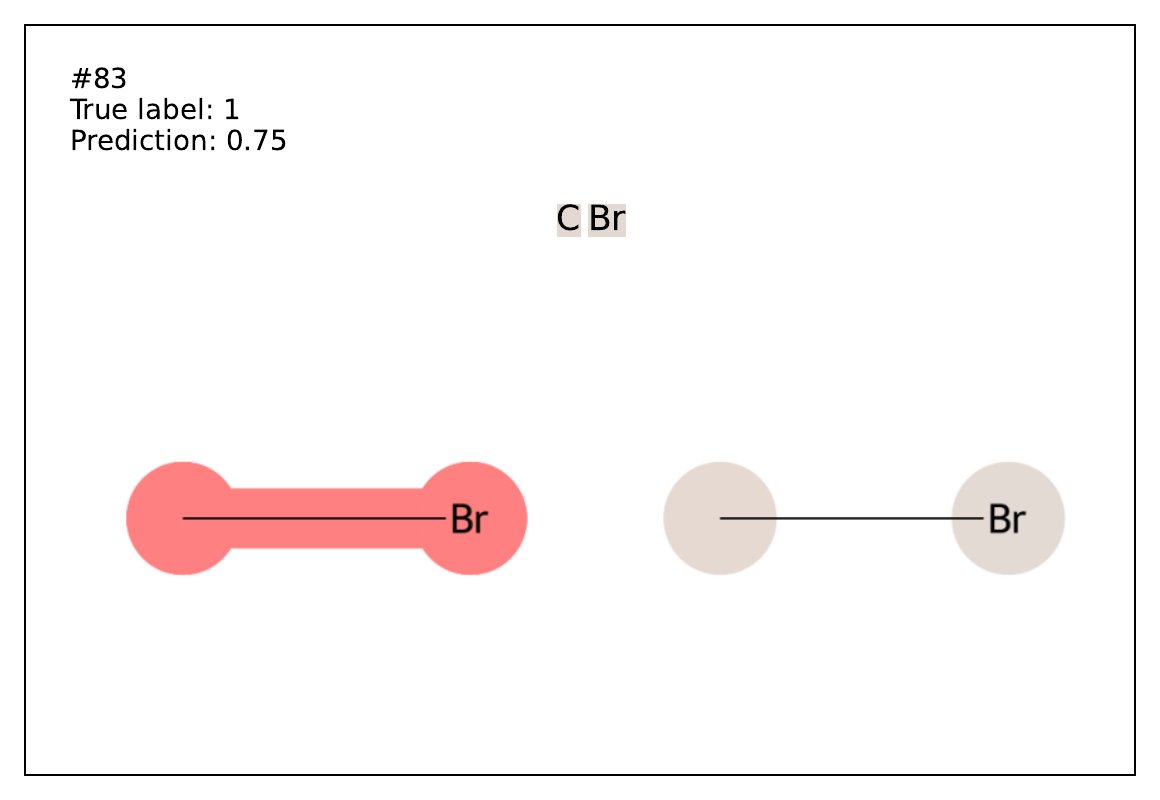} 
\end{subfigure}\begin{subfigure}[b]{0.33\textwidth} 
  \centering 
  \includegraphics[width=\textwidth]{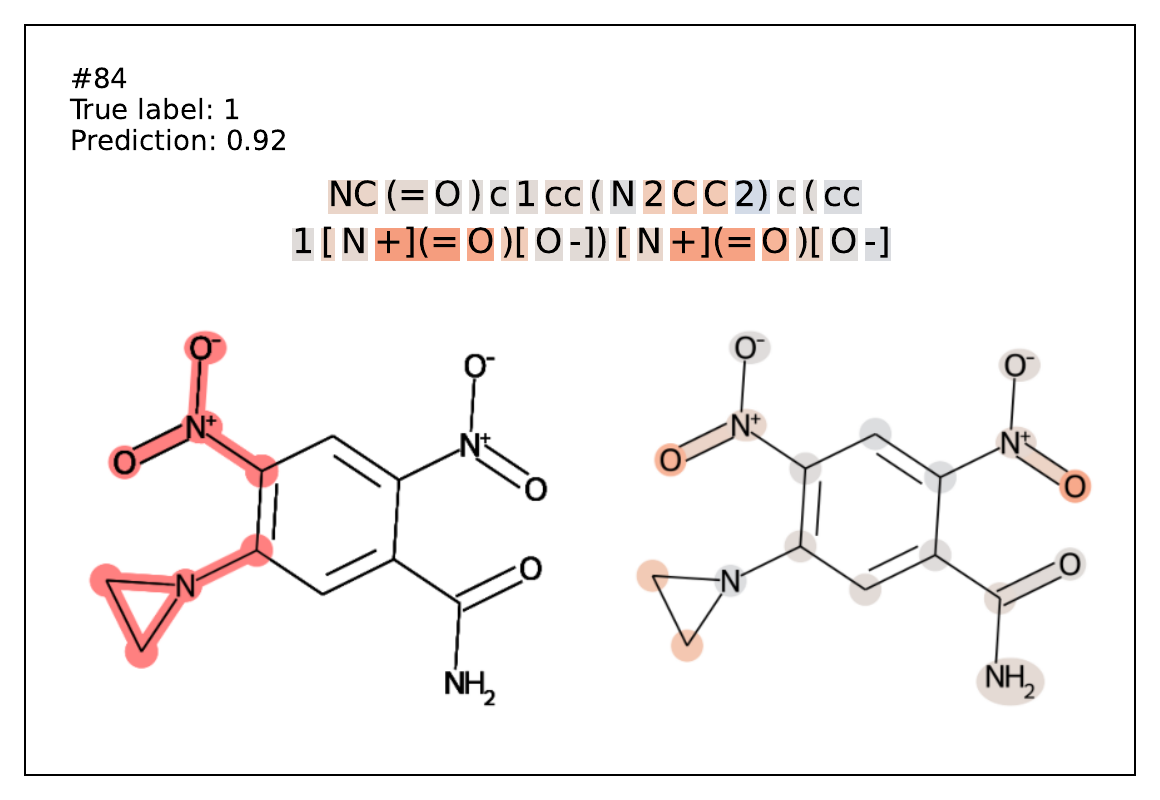} 
\end{subfigure} 
\begin{subfigure}[b]{0.33\textwidth} 
  \centering 
  \includegraphics[width=\textwidth]{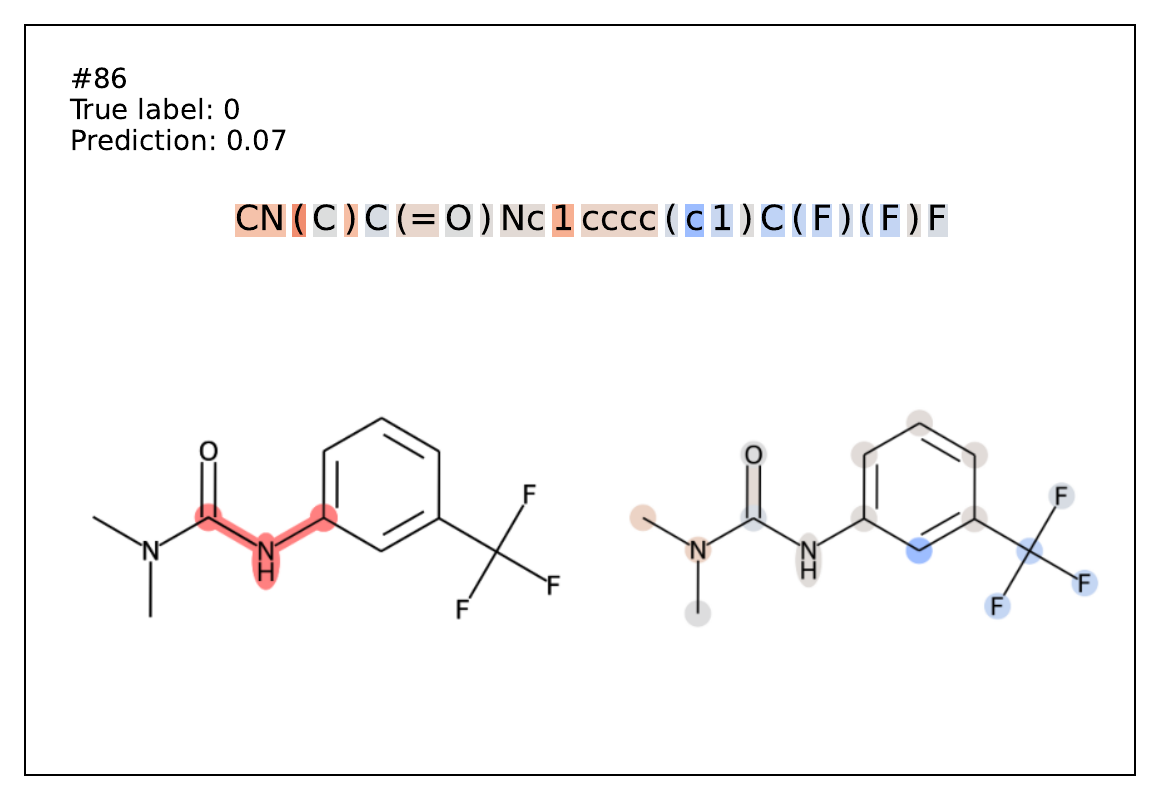} 
\end{subfigure}\begin{subfigure}[b]{0.33\textwidth} 
  \centering 
  \includegraphics[width=\textwidth]{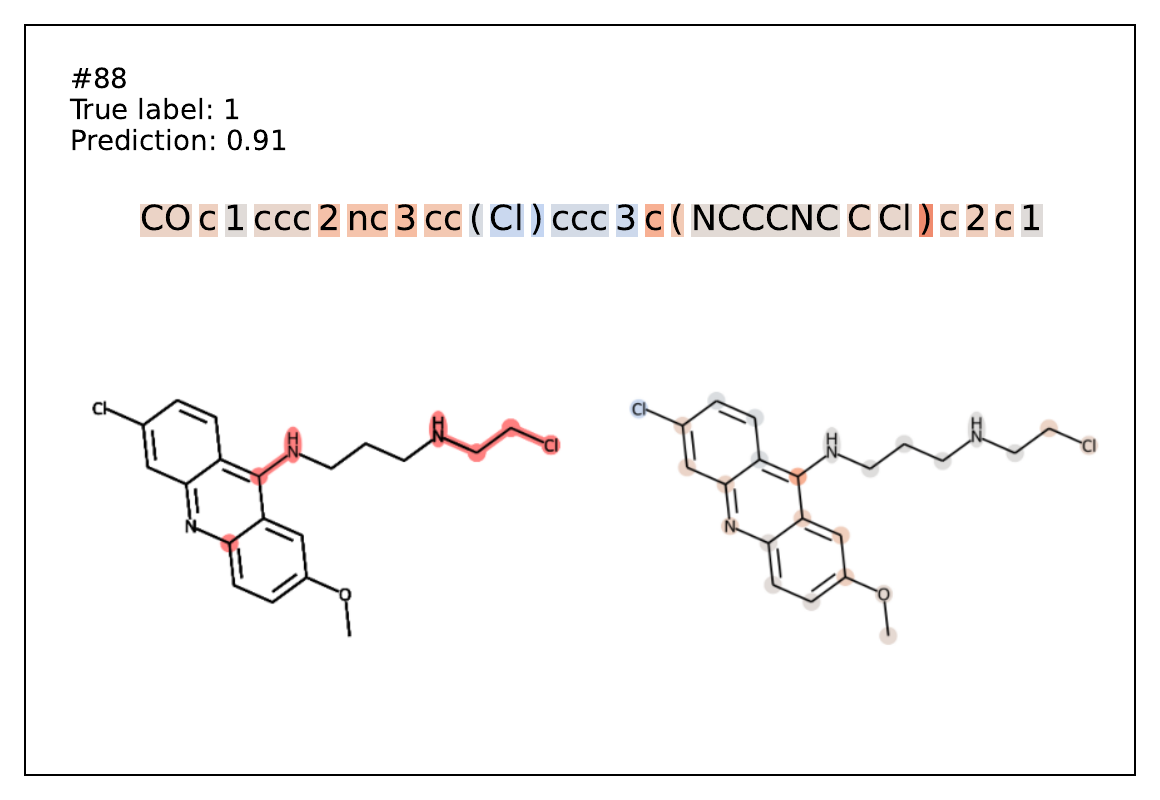} 
\end{subfigure}\begin{subfigure}[b]{0.33\textwidth} 
  \centering 
  \includegraphics[width=\textwidth]{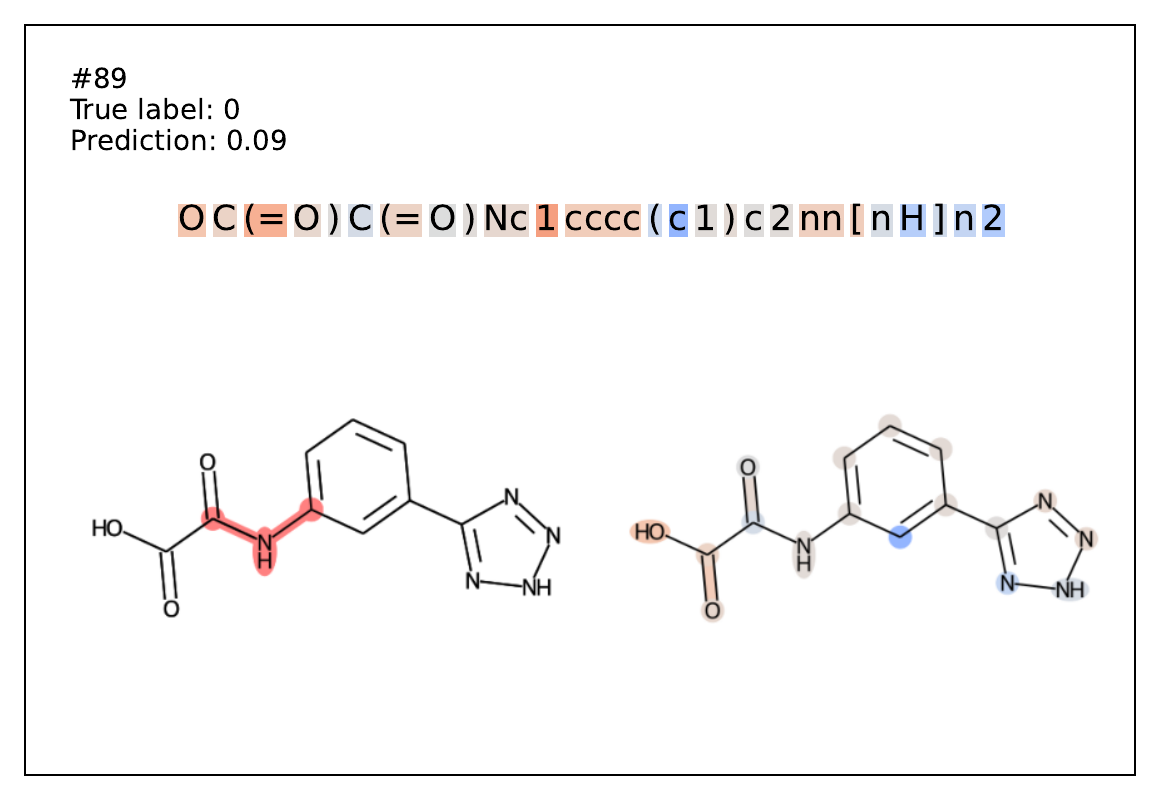} 
\end{subfigure} 
\caption{Explaining predictions of the fine-tuned model on Ames dataset. See Section \ref{sec:captum}. Part 3/5}
\label{fig:captum-ames-3}
\end{figure}

\begin{figure}
\centering
\begin{subfigure}[b]{0.33\textwidth} 
  \centering 
  \includegraphics[width=\textwidth]{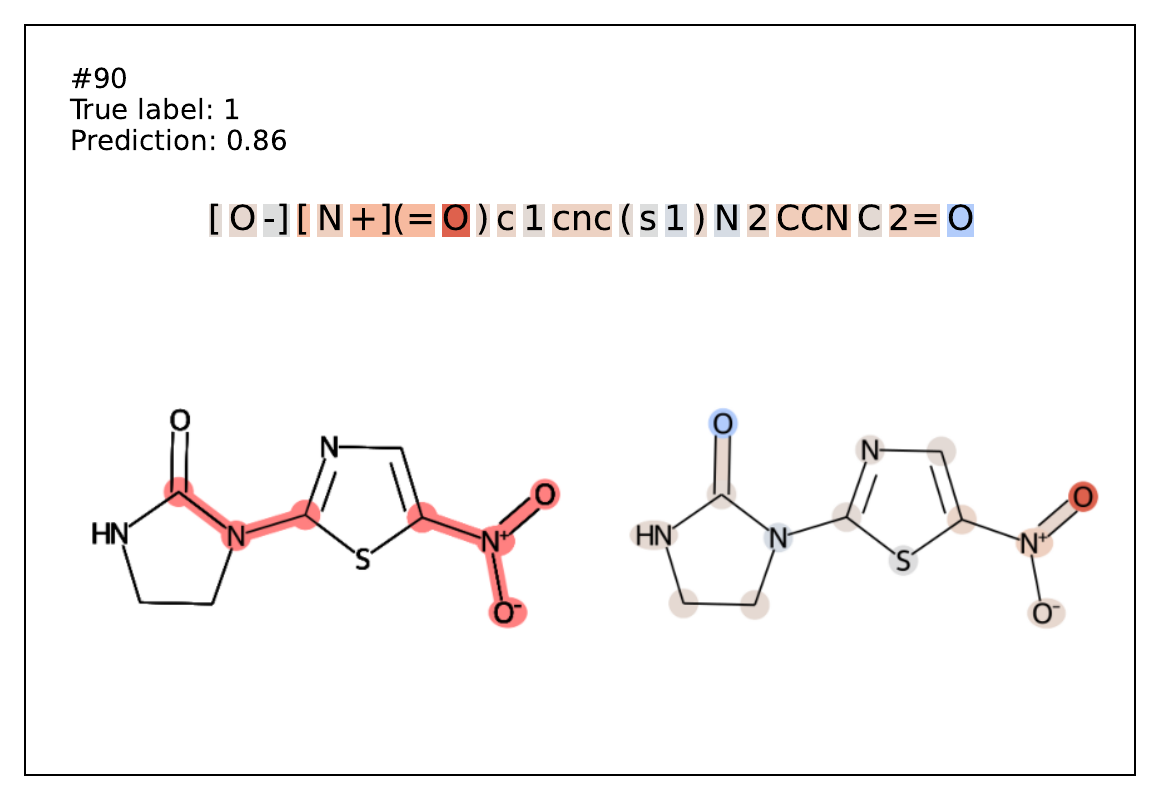} 
\end{subfigure}\begin{subfigure}[b]{0.33\textwidth} 
  \centering 
  \includegraphics[width=\textwidth]{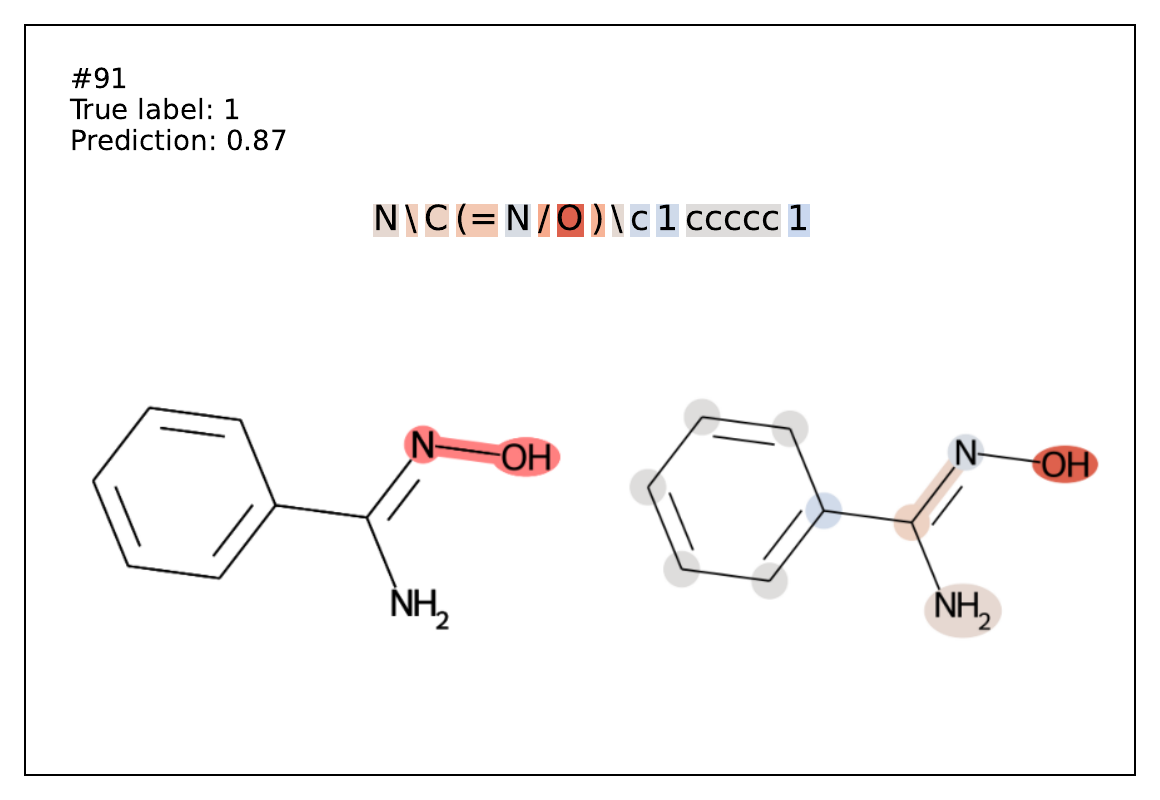} 
\end{subfigure}\begin{subfigure}[b]{0.33\textwidth} 
  \centering 
  \includegraphics[width=\textwidth]{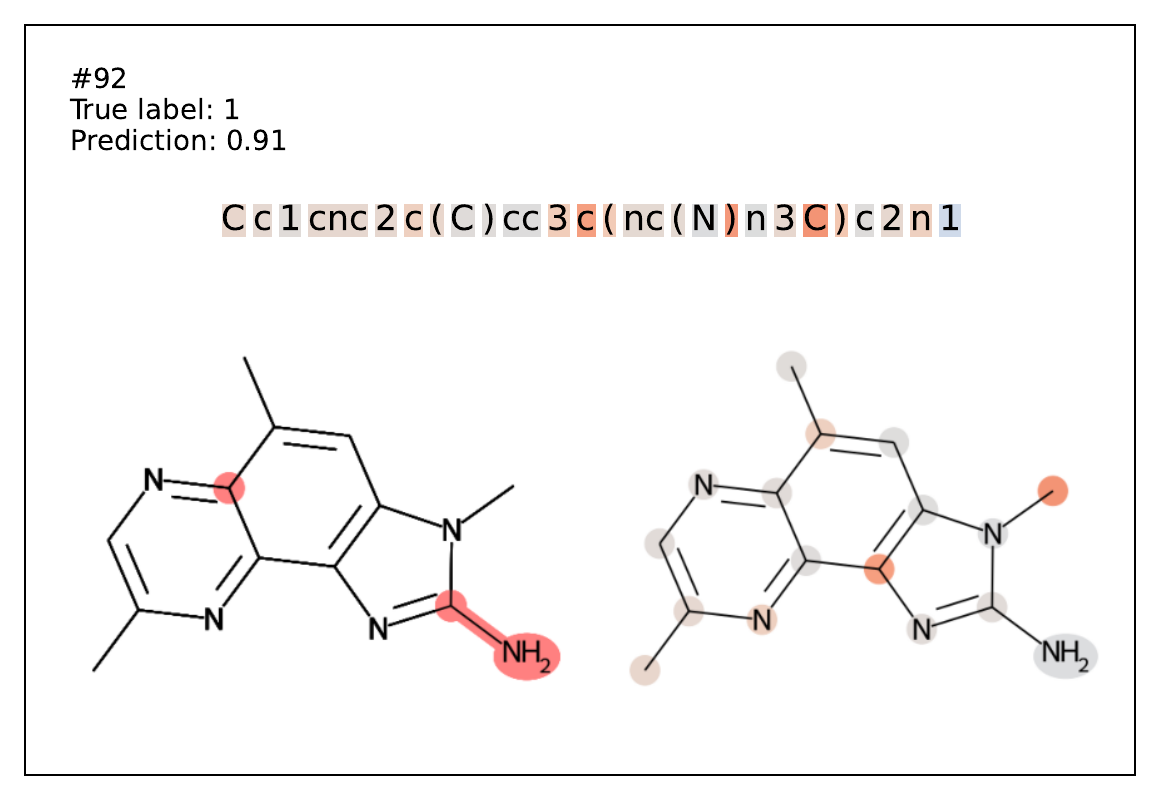} 
\end{subfigure} 
\begin{subfigure}[b]{0.33\textwidth} 
  \centering 
  \includegraphics[width=\textwidth]{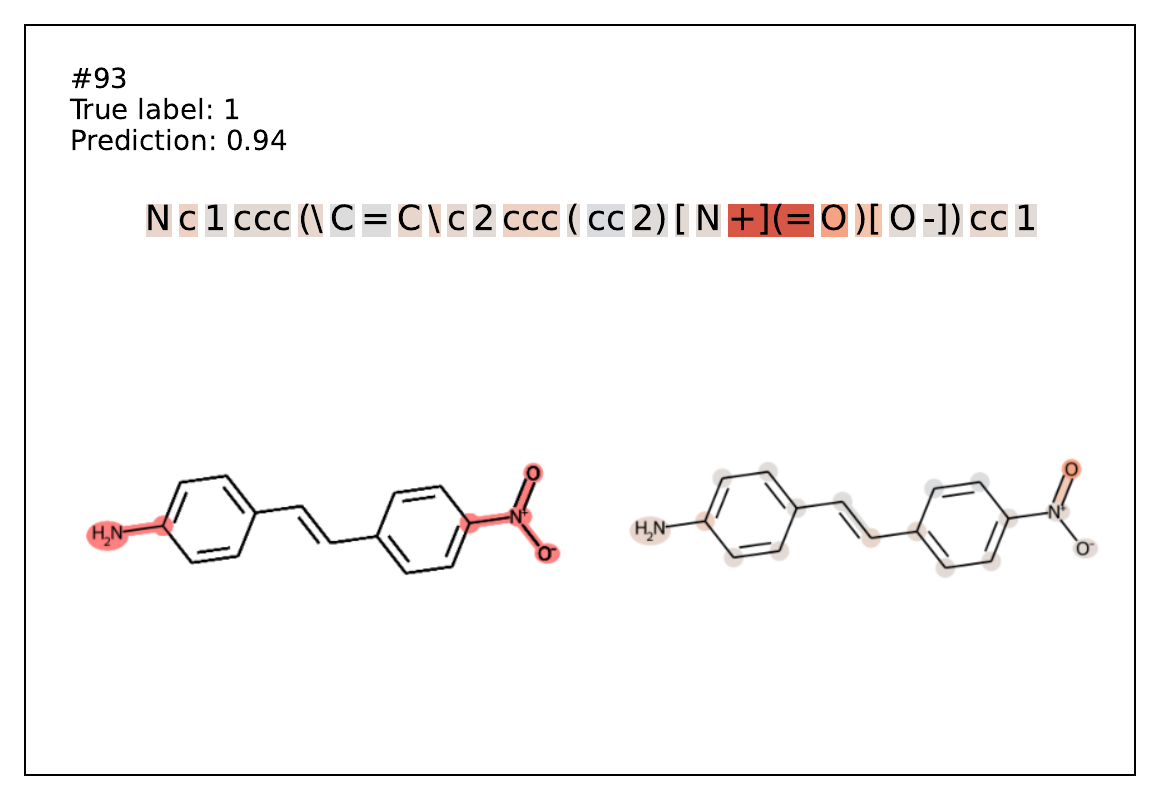} 
\end{subfigure}\begin{subfigure}[b]{0.33\textwidth} 
  \centering 
  \includegraphics[width=\textwidth]{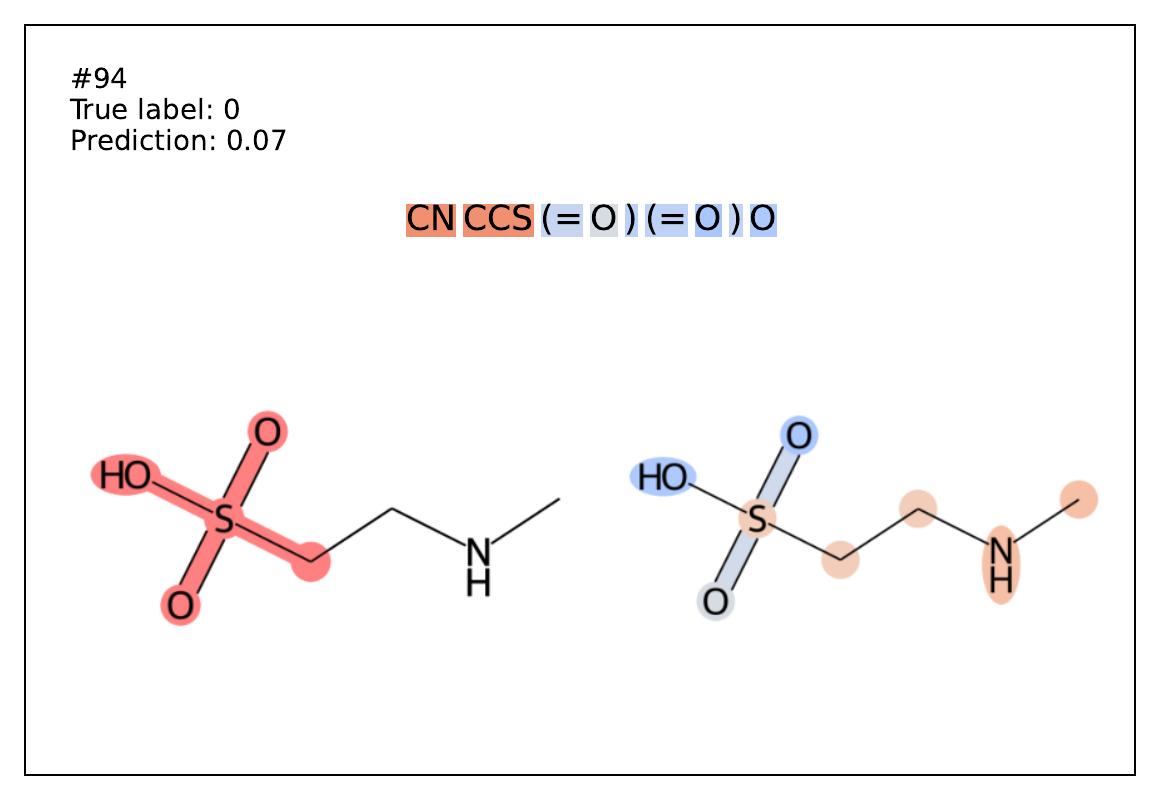} 
\end{subfigure}\begin{subfigure}[b]{0.33\textwidth} 
  \centering 
  \includegraphics[width=\textwidth]{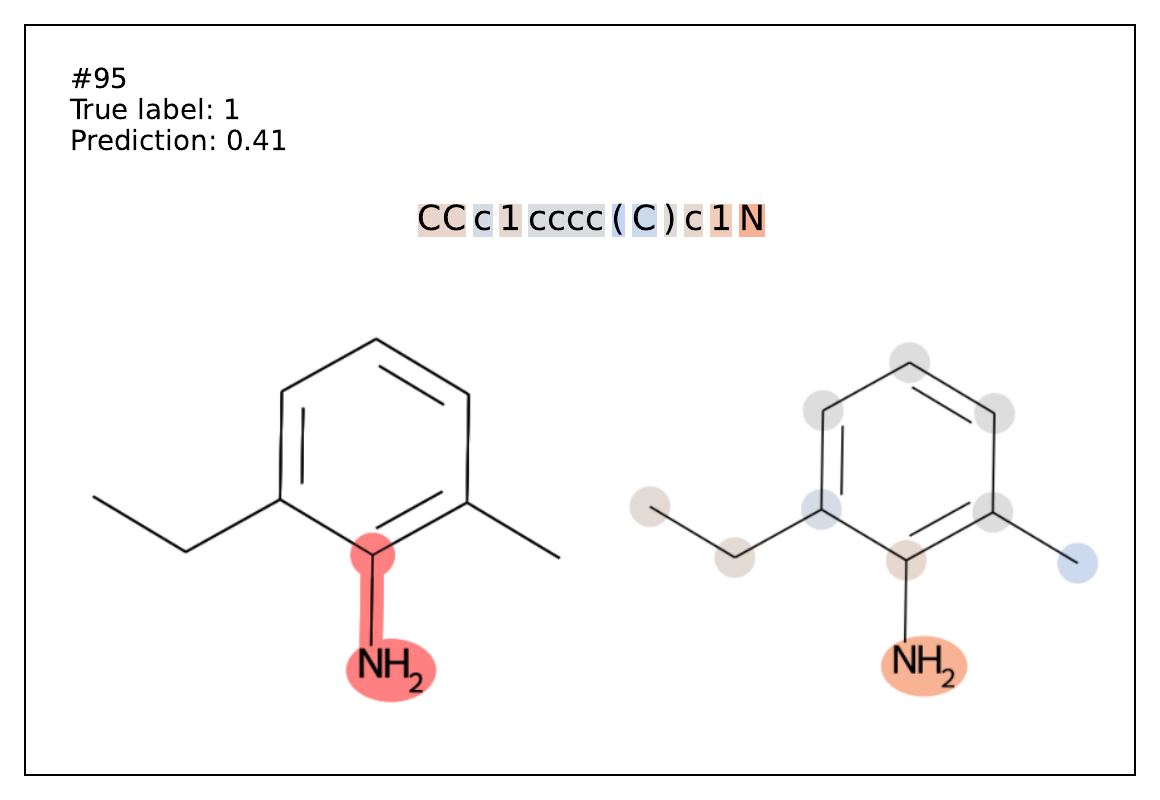} 
\end{subfigure} 
\begin{subfigure}[b]{0.33\textwidth} 
  \centering 
  \includegraphics[width=\textwidth]{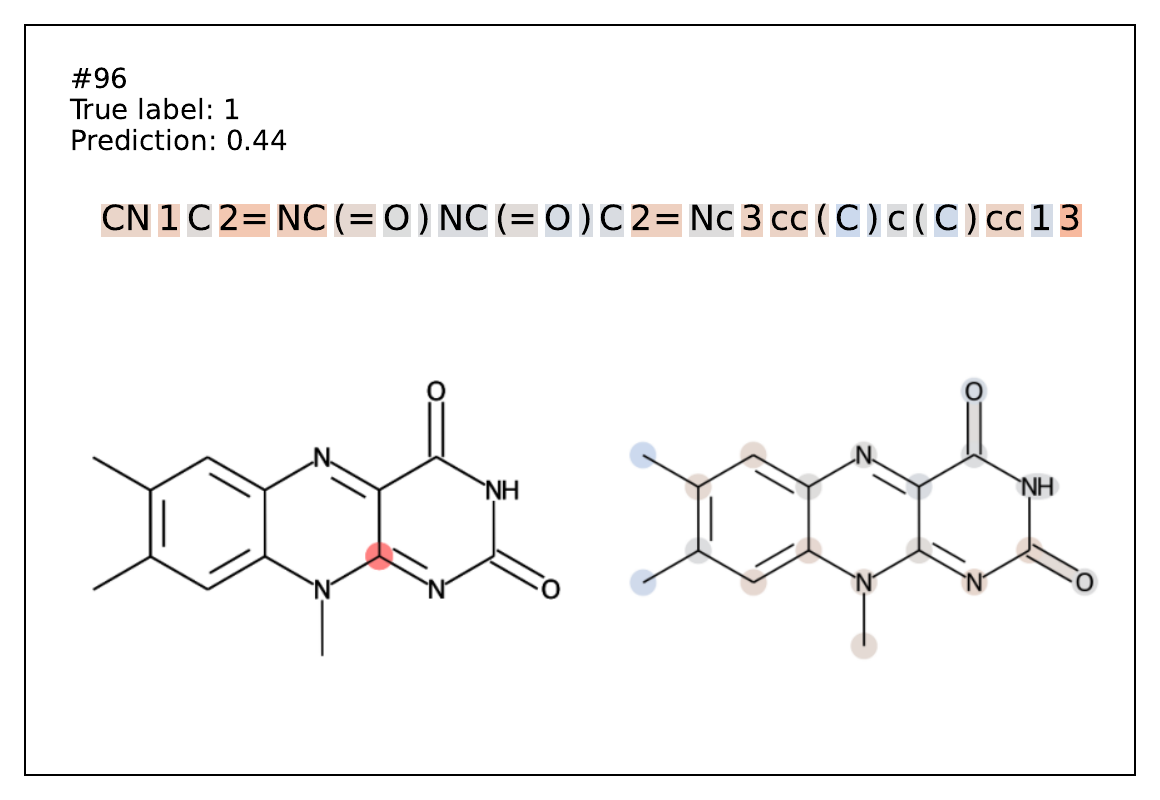} 
\end{subfigure}\begin{subfigure}[b]{0.33\textwidth} 
  \centering 
  \includegraphics[width=\textwidth]{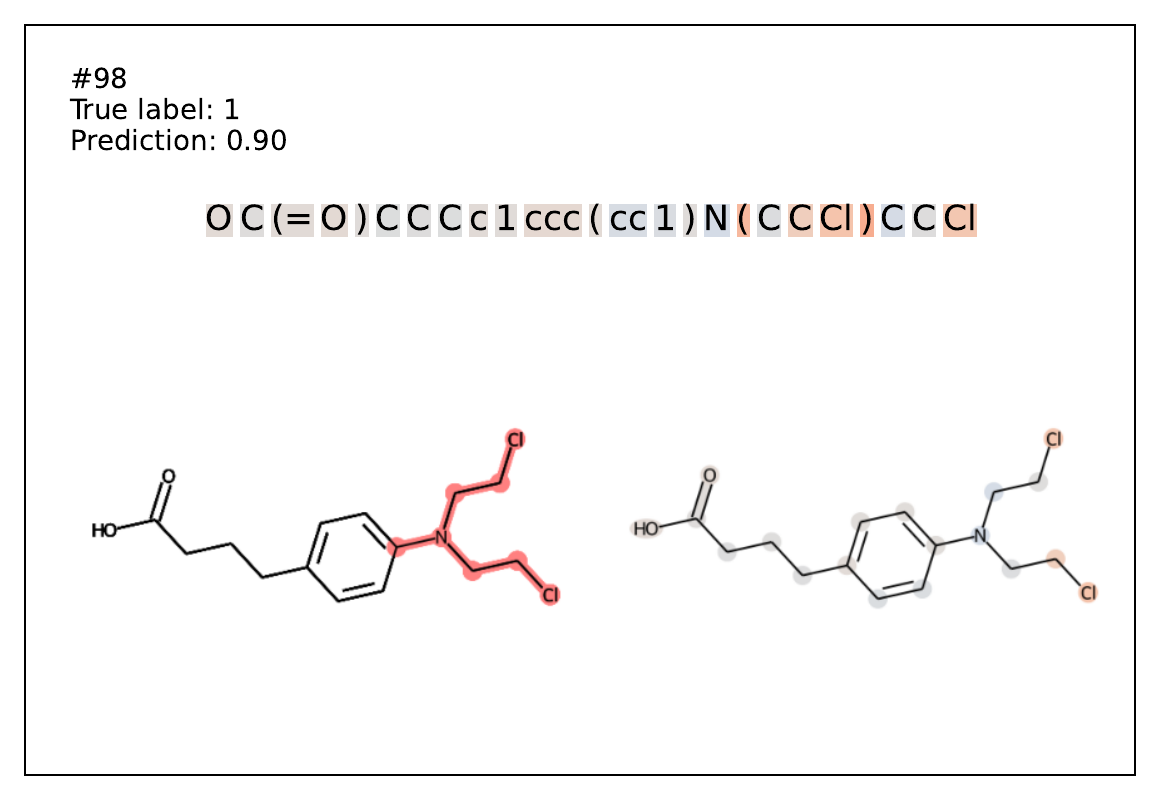} 
\end{subfigure}\begin{subfigure}[b]{0.33\textwidth} 
  \centering 
  \includegraphics[width=\textwidth]{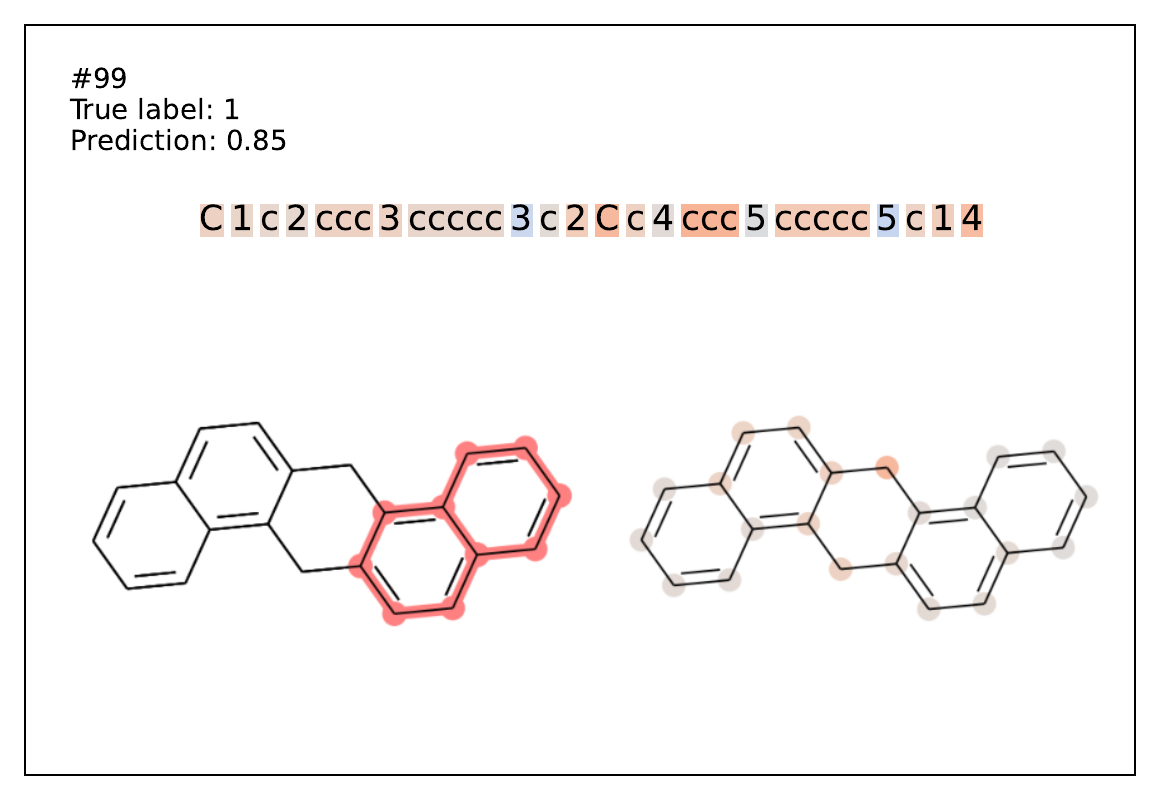} 
\end{subfigure} 
\begin{subfigure}[b]{0.33\textwidth} 
  \centering 
  \includegraphics[width=\textwidth]{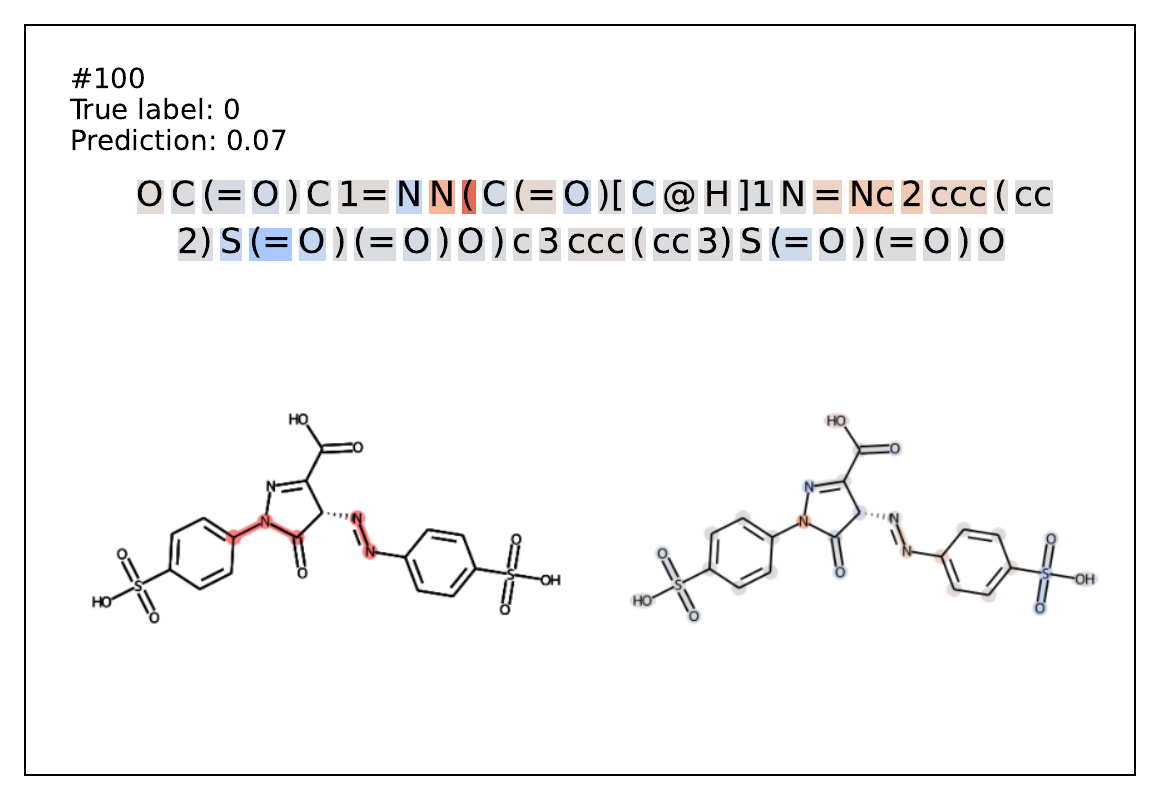} 
\end{subfigure}\begin{subfigure}[b]{0.33\textwidth} 
  \centering 
  \includegraphics[width=\textwidth]{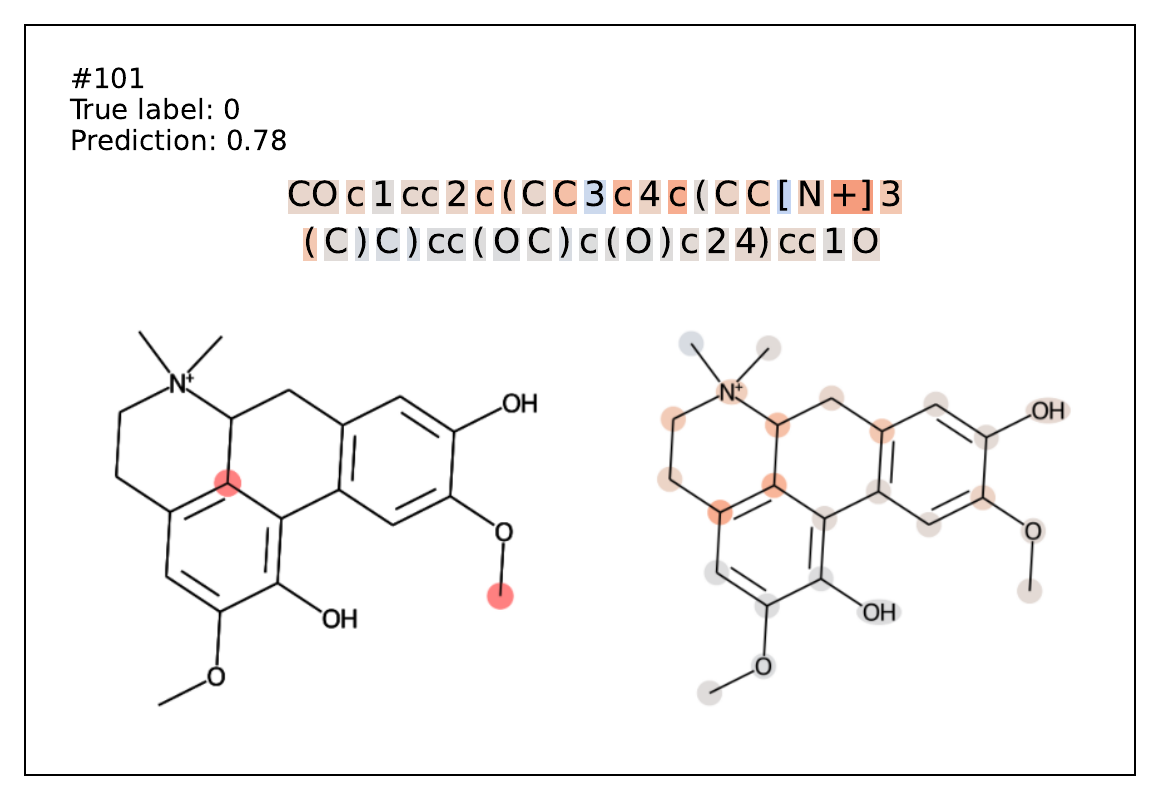} 
\end{subfigure}\begin{subfigure}[b]{0.33\textwidth} 
  \centering 
  \includegraphics[width=\textwidth]{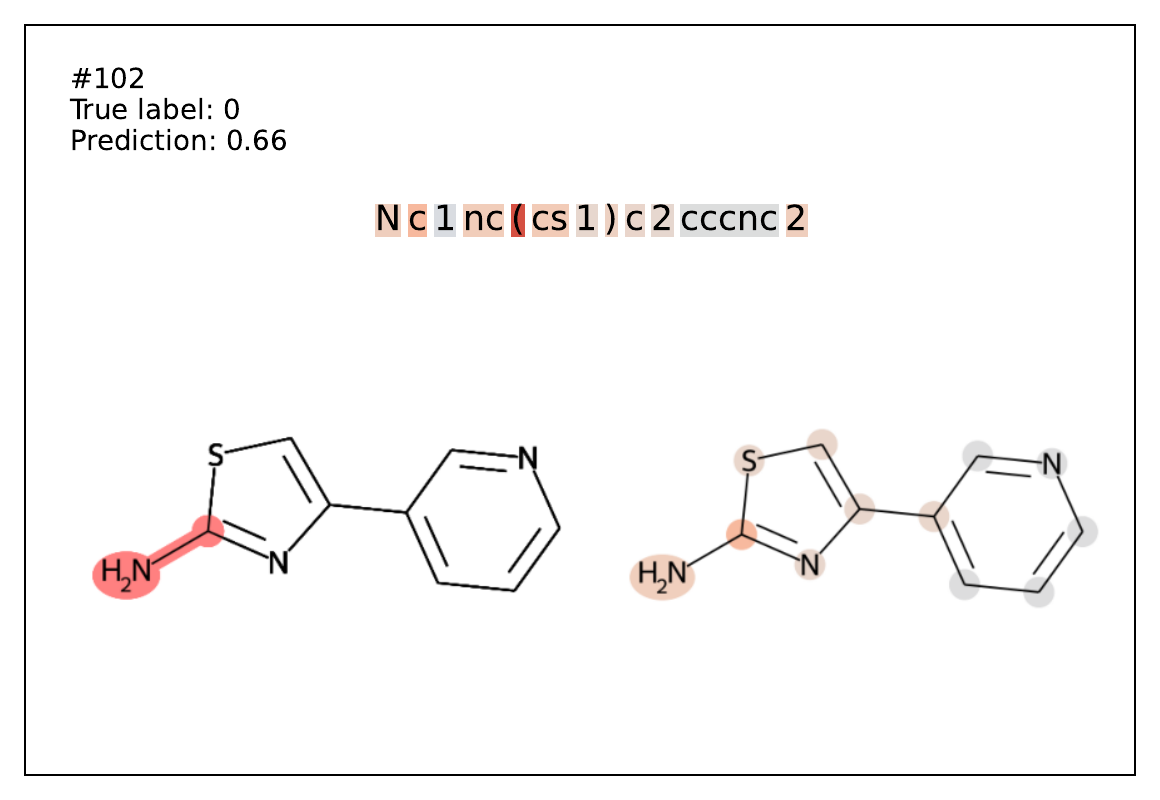} 
\end{subfigure} 
\begin{subfigure}[b]{0.33\textwidth} 
  \centering 
  \includegraphics[width=\textwidth]{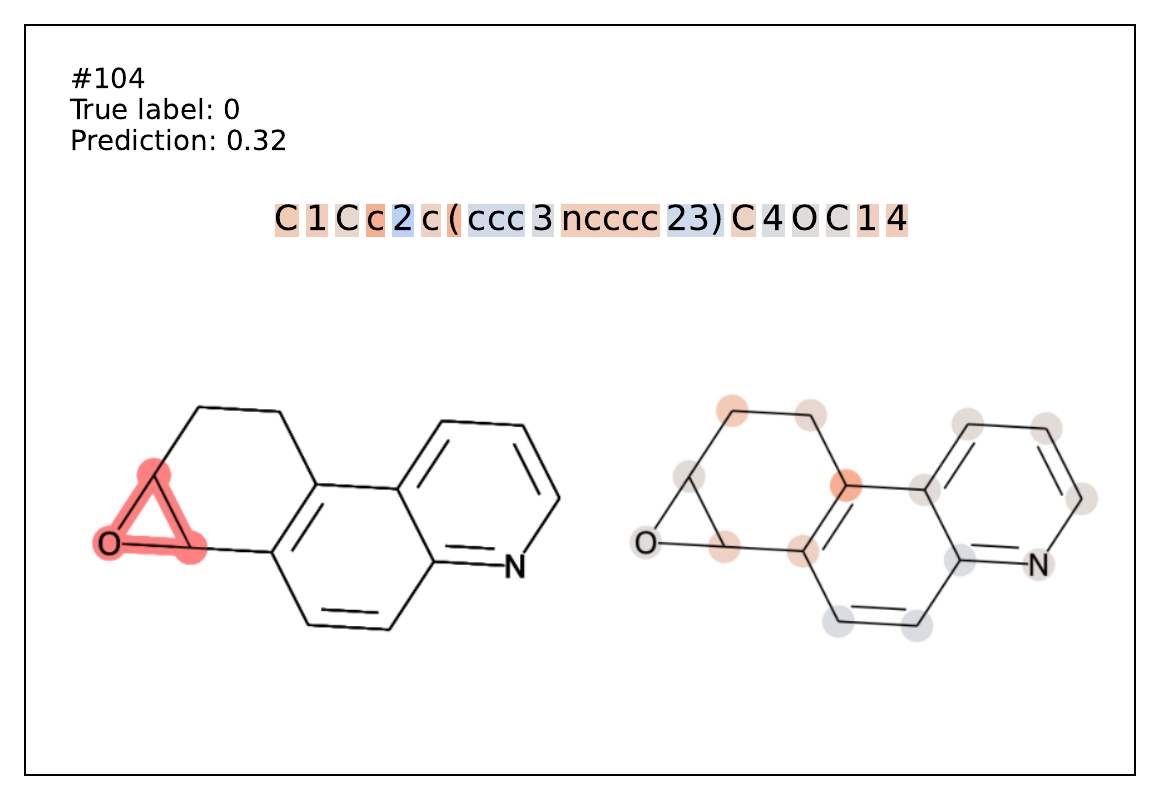} 
\end{subfigure}\begin{subfigure}[b]{0.33\textwidth} 
  \centering 
  \includegraphics[width=\textwidth]{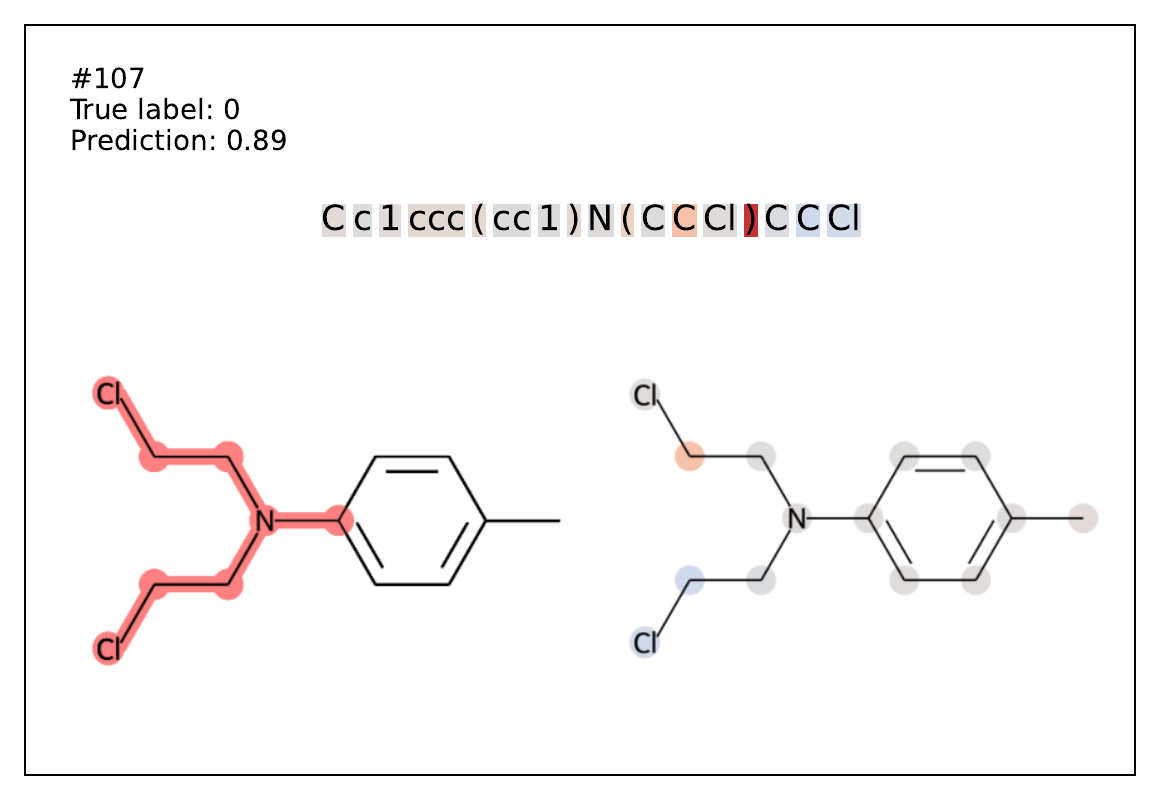} 
\end{subfigure}\begin{subfigure}[b]{0.33\textwidth} 
  \centering 
  \includegraphics[width=\textwidth]{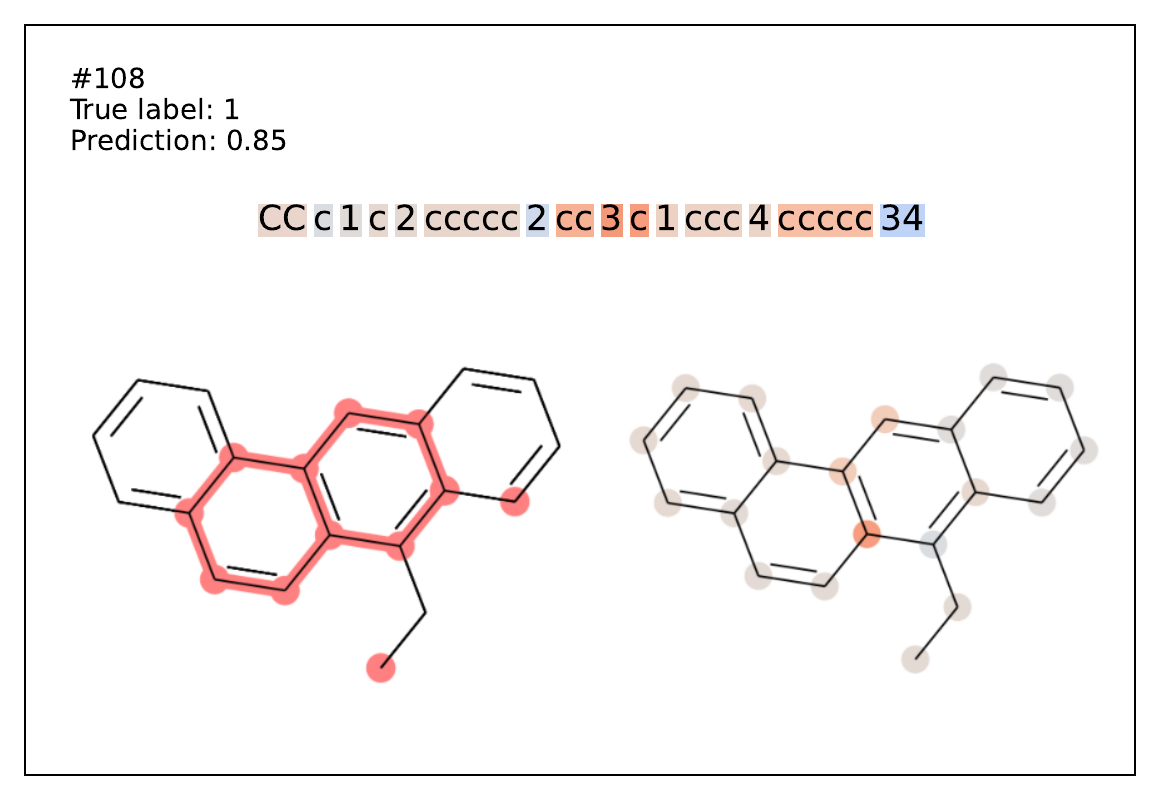} 
\end{subfigure} 
\begin{subfigure}[b]{0.33\textwidth} 
  \centering 
  \includegraphics[width=\textwidth]{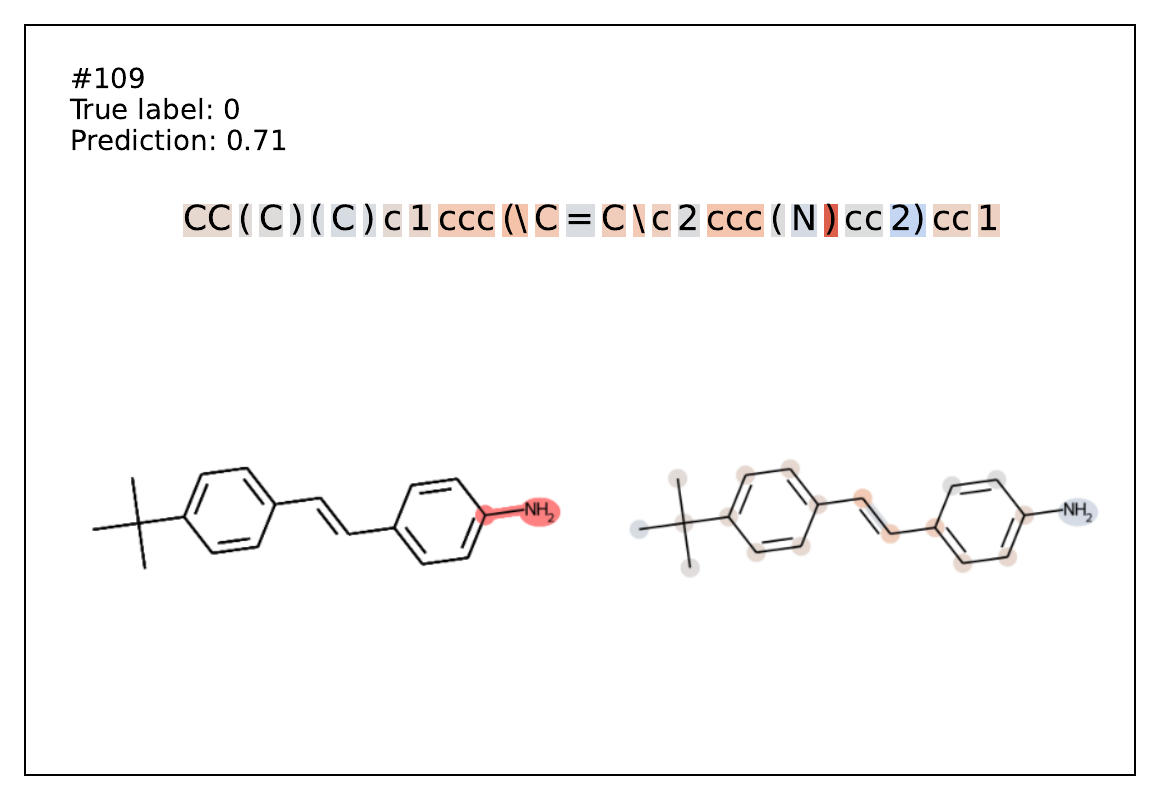} 
\end{subfigure}\begin{subfigure}[b]{0.33\textwidth} 
  \centering 
  \includegraphics[width=\textwidth]{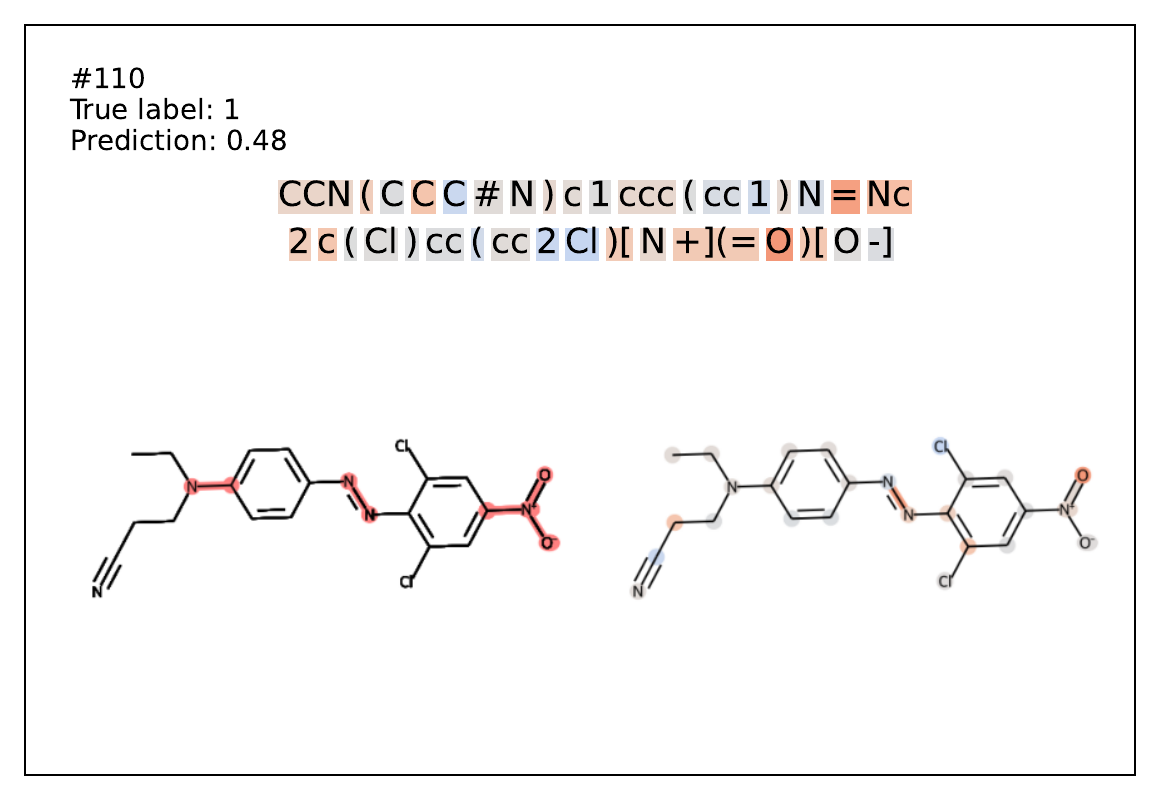} 
\end{subfigure}\begin{subfigure}[b]{0.33\textwidth} 
  \centering 
  \includegraphics[width=\textwidth]{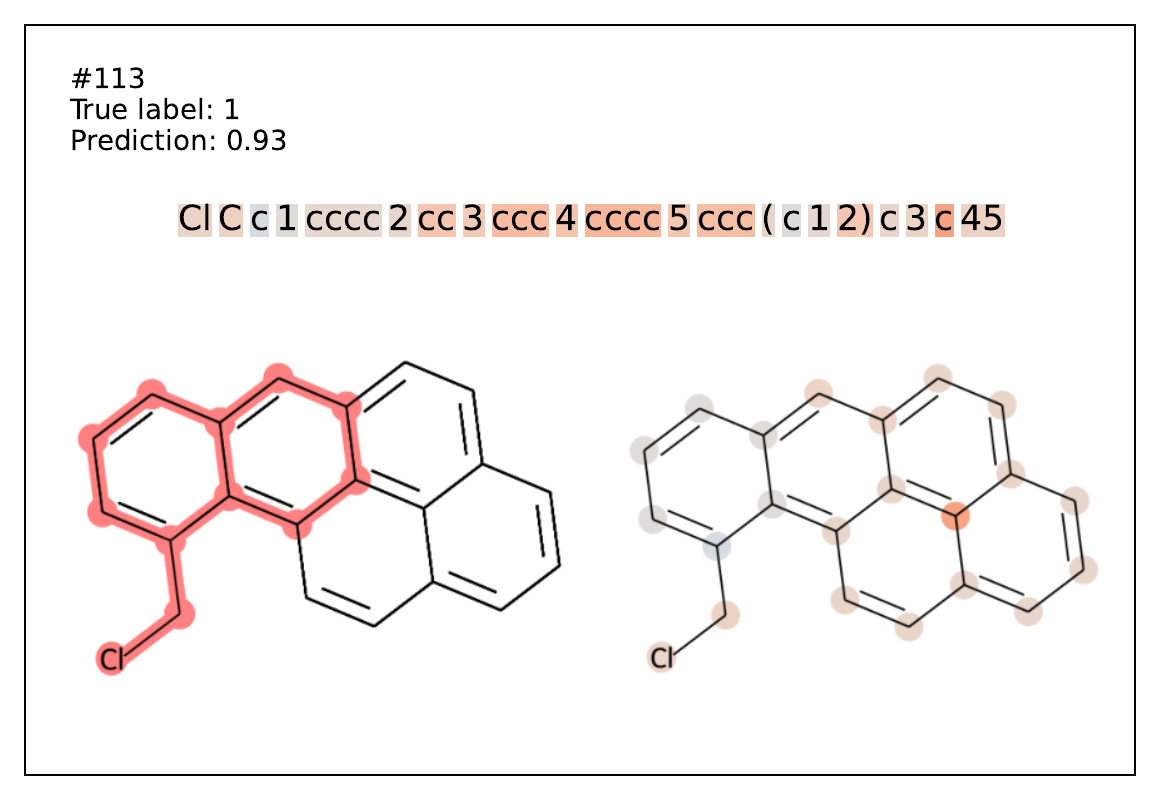} 
\end{subfigure} 
\begin{subfigure}[b]{0.33\textwidth} 
  \centering 
  \includegraphics[width=\textwidth]{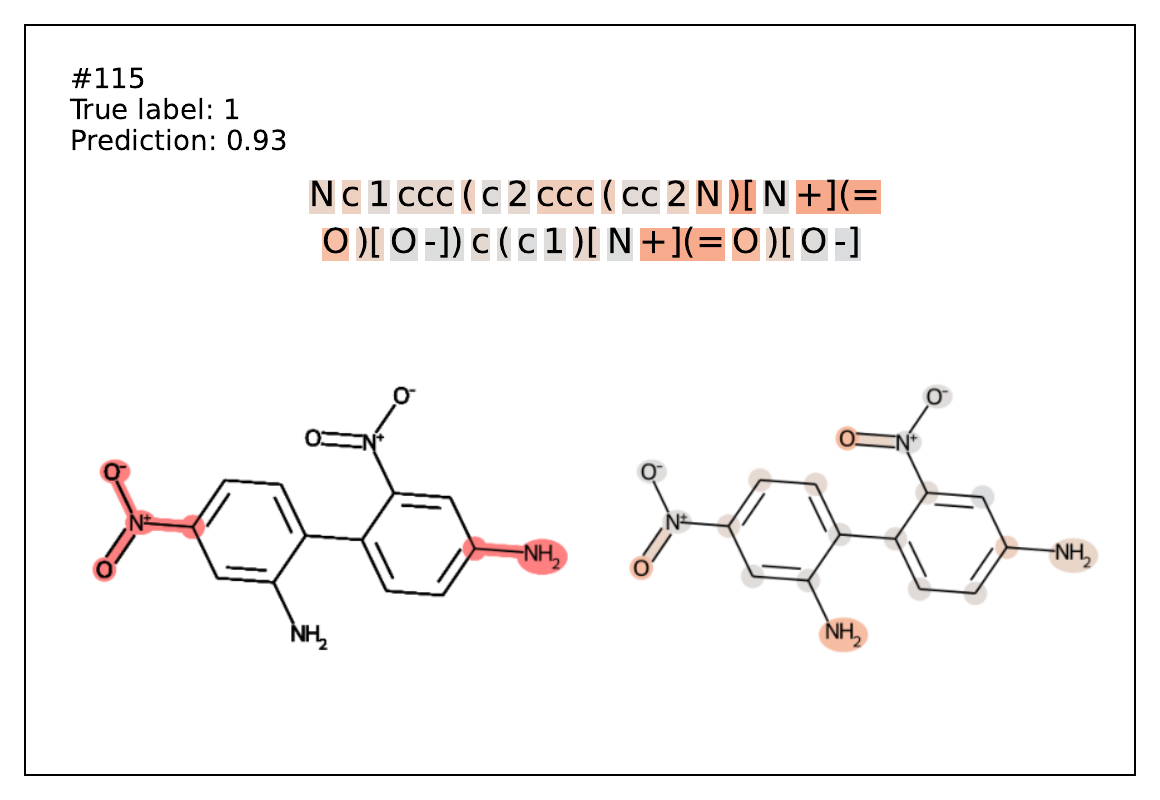} 
\end{subfigure}\begin{subfigure}[b]{0.33\textwidth} 
  \centering 
  \includegraphics[width=\textwidth]{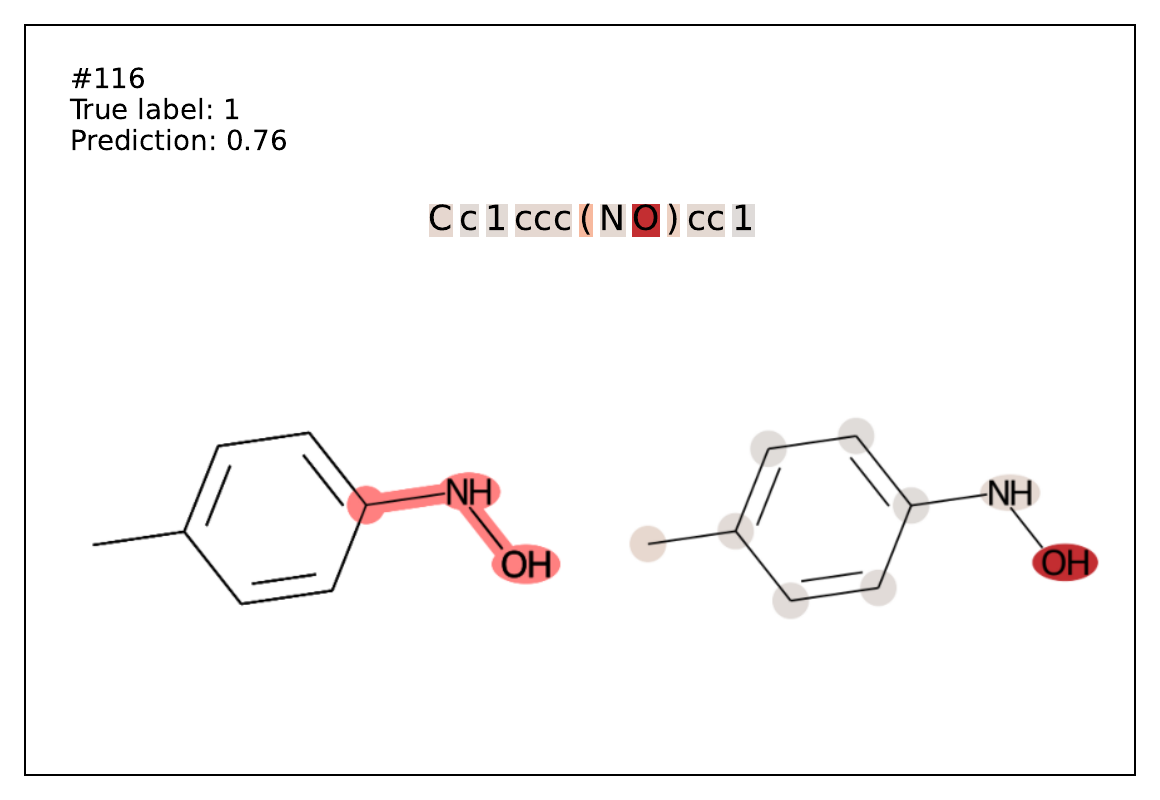} 
\end{subfigure}\begin{subfigure}[b]{0.33\textwidth} 
  \centering 
  \includegraphics[width=\textwidth]{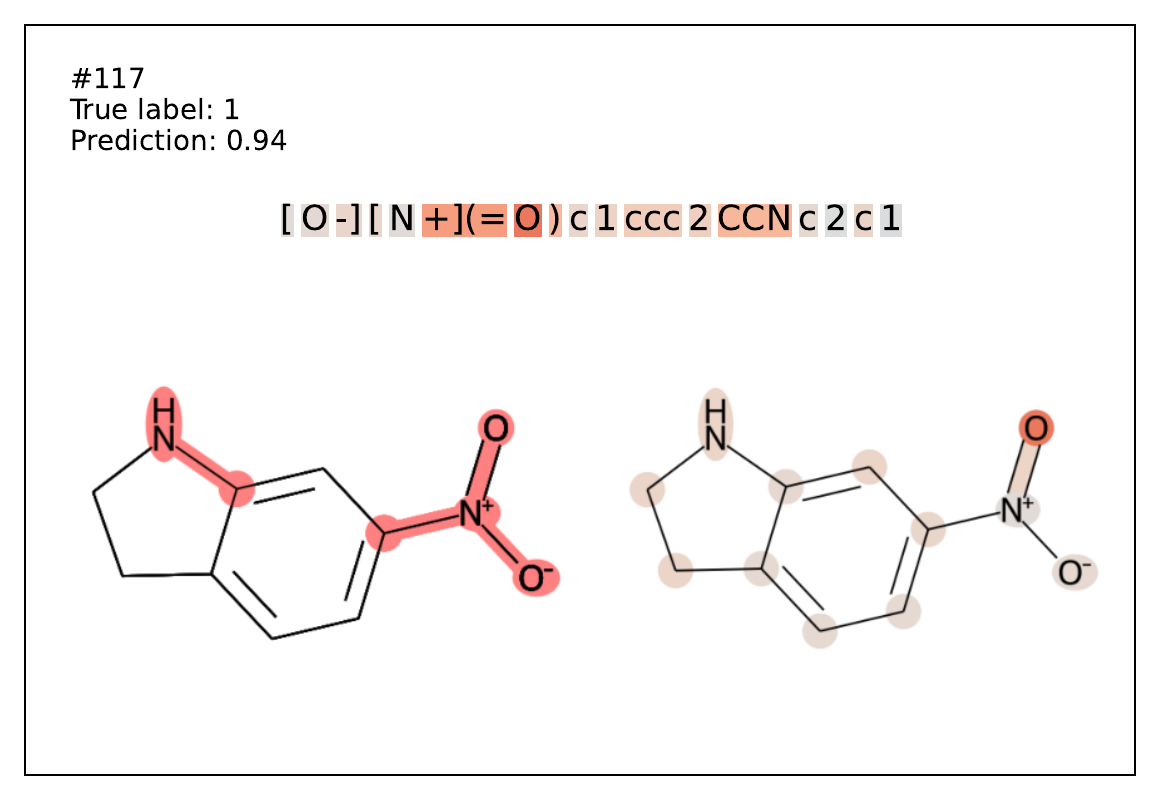} 
\end{subfigure} 
\caption{Explaining predictions of the fine-tuned model on Ames dataset. See Section \ref{sec:captum}. Part 4/5}
\label{fig:captum-ames-4}
\end{figure}

\begin{figure}
\centering
\begin{subfigure}[b]{0.33\textwidth} 
  \centering 
  \includegraphics[width=\textwidth]{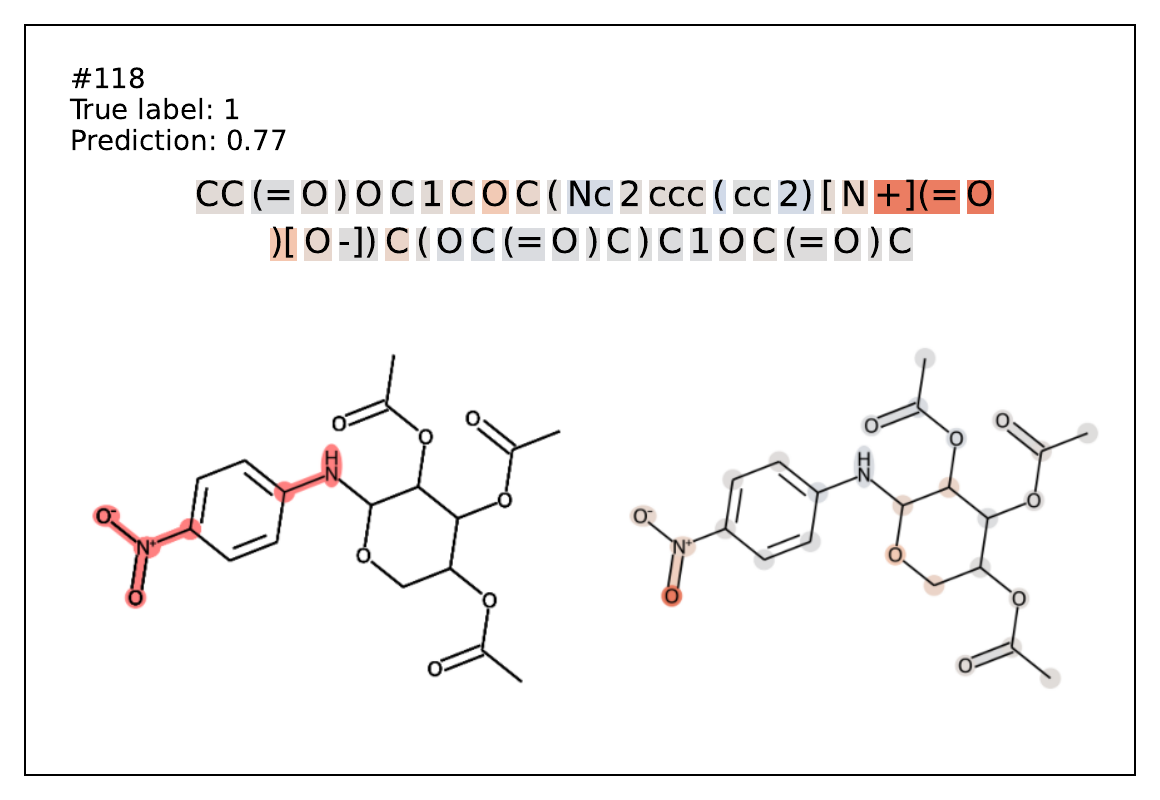} 
\end{subfigure}\begin{subfigure}[b]{0.33\textwidth} 
  \centering 
  \includegraphics[width=\textwidth]{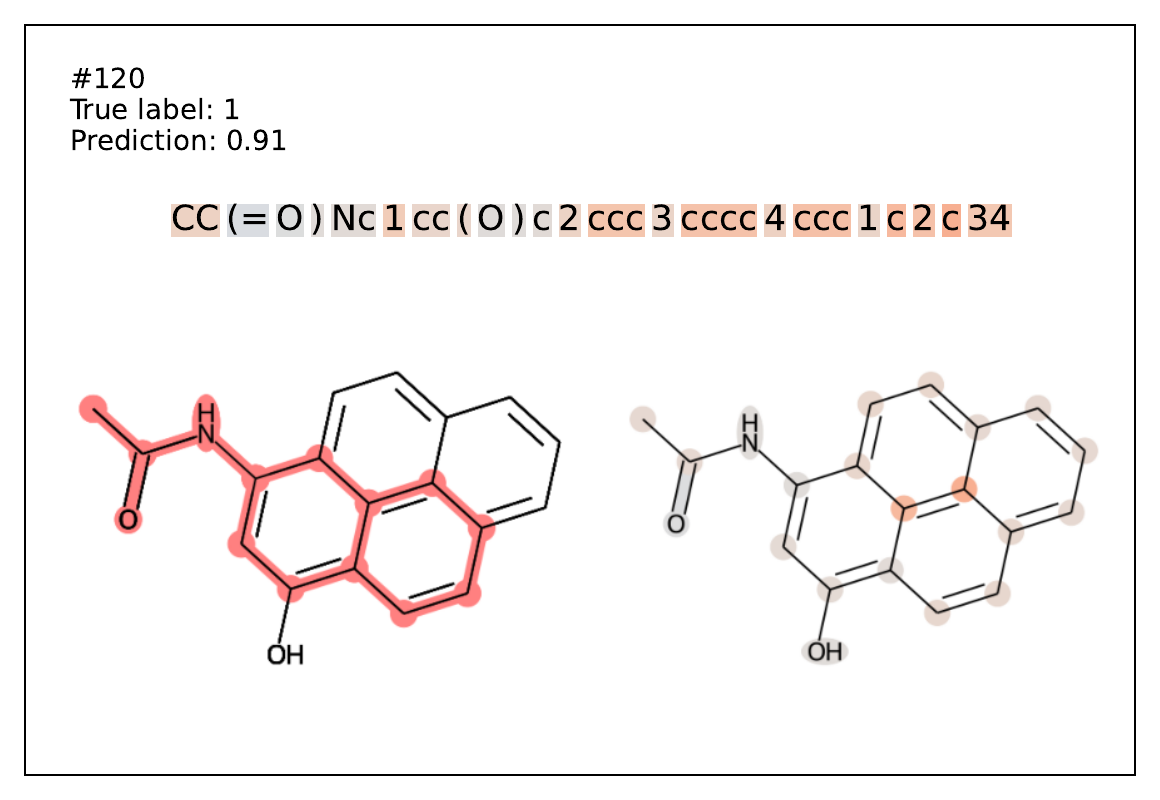} 
\end{subfigure}\begin{subfigure}[b]{0.33\textwidth} 
  \centering 
  \includegraphics[width=\textwidth]{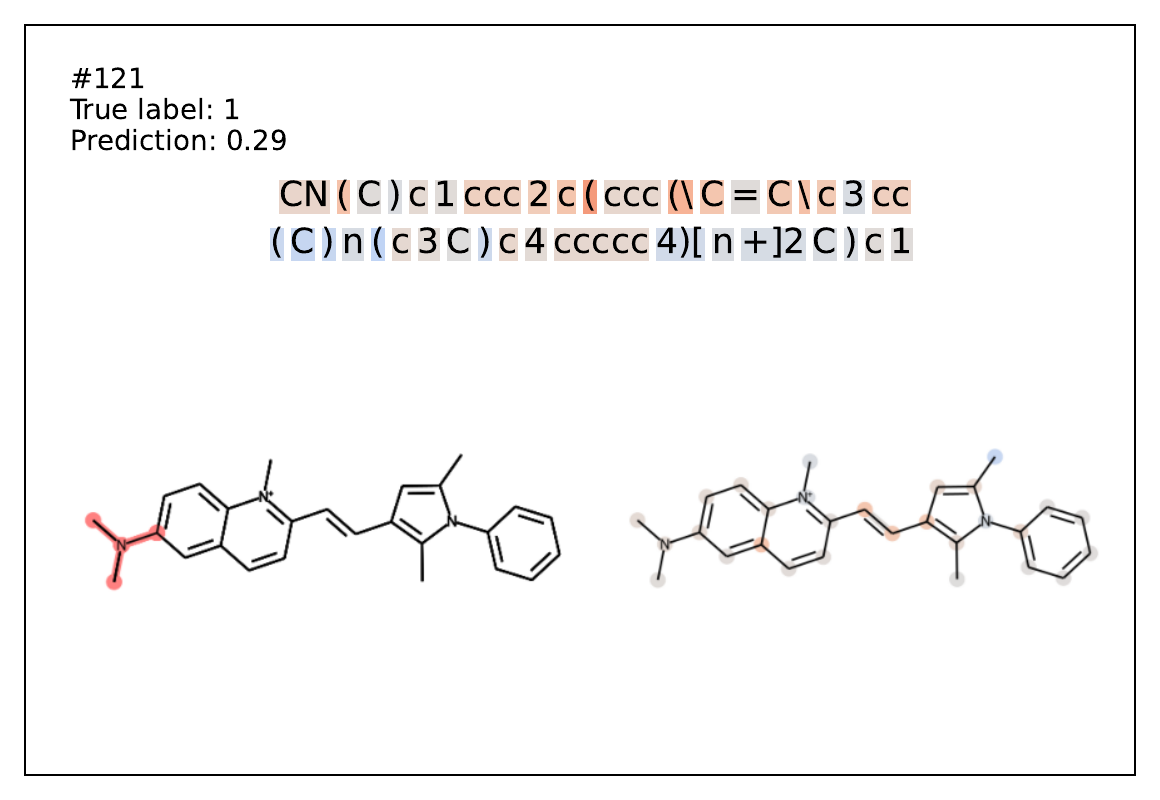} 
\end{subfigure} 
\begin{subfigure}[b]{0.33\textwidth} 
  \centering 
  \includegraphics[width=\textwidth]{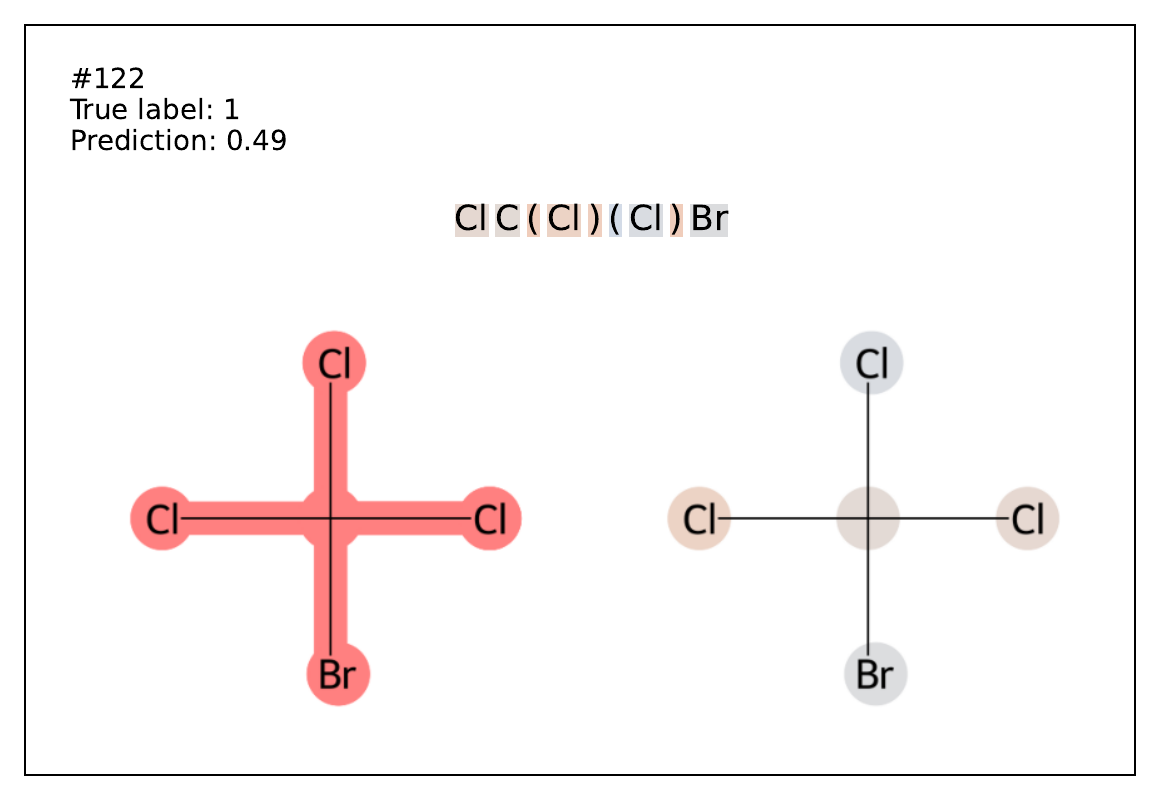} 
\end{subfigure}\begin{subfigure}[b]{0.33\textwidth} 
  \centering 
  \includegraphics[width=\textwidth]{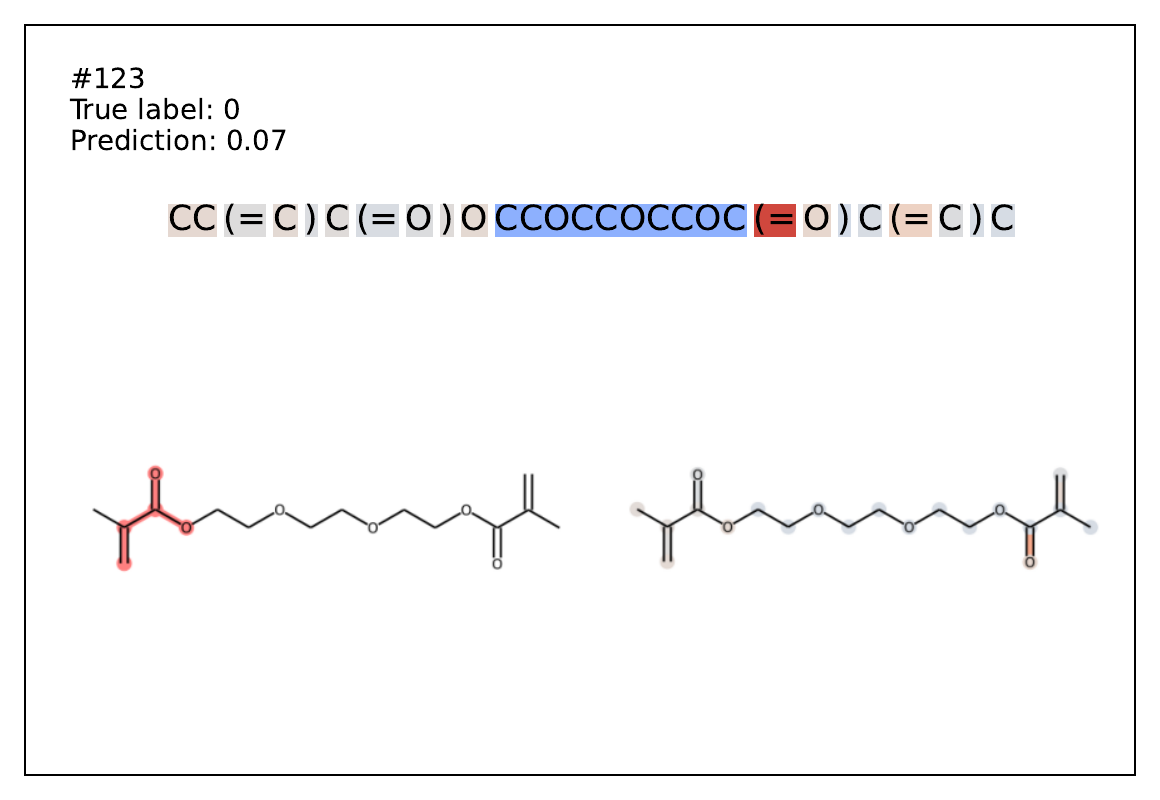} 
\end{subfigure}\begin{subfigure}[b]{0.33\textwidth} 
  \centering 
  \includegraphics[width=\textwidth]{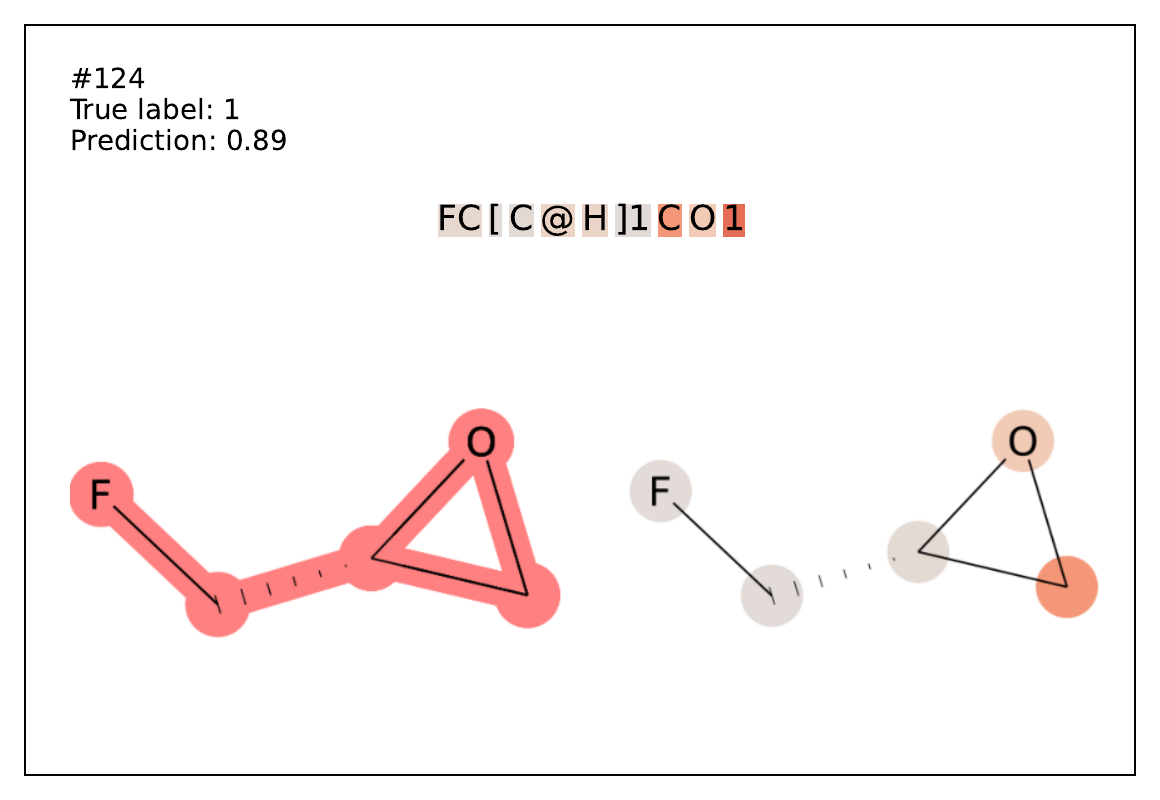} 
\end{subfigure} 
\begin{subfigure}[b]{0.33\textwidth} 
  \centering 
  \includegraphics[width=\textwidth]{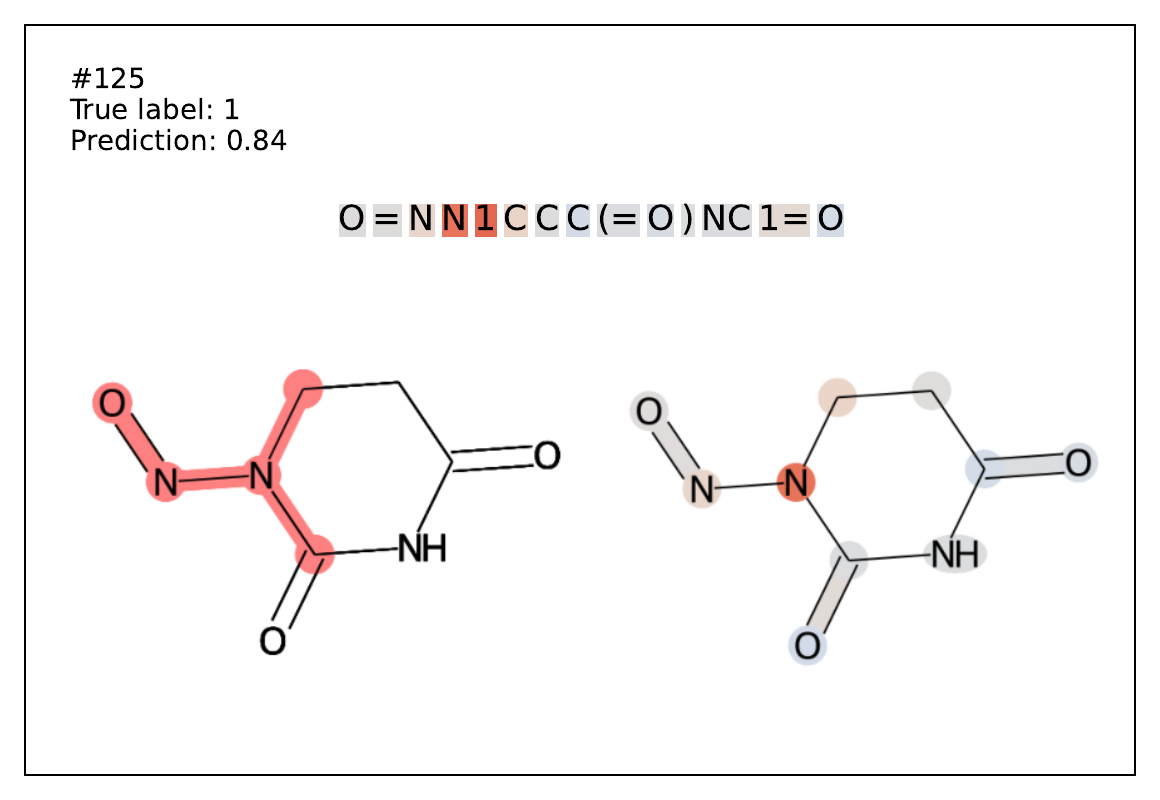} 
\end{subfigure}\begin{subfigure}[b]{0.33\textwidth} 
  \centering 
  \includegraphics[width=\textwidth]{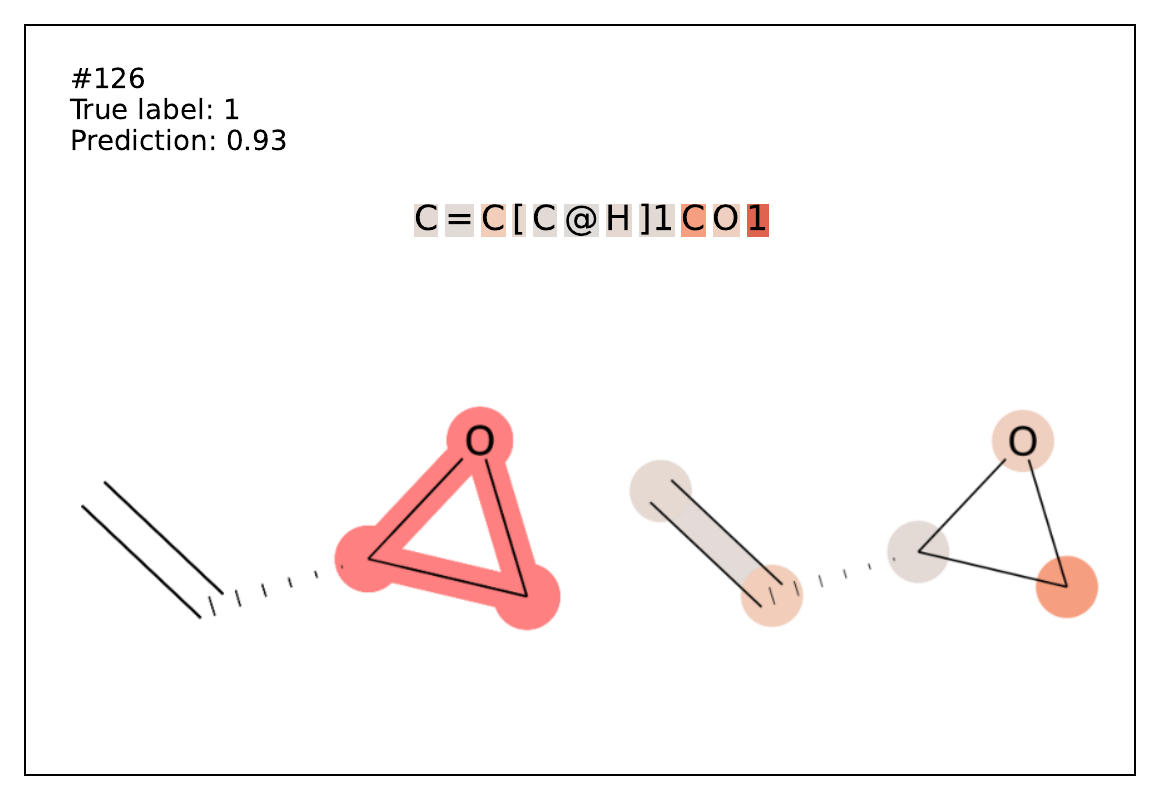} 
\end{subfigure}\begin{subfigure}[b]{0.33\textwidth} 
  \centering 
  \includegraphics[width=\textwidth]{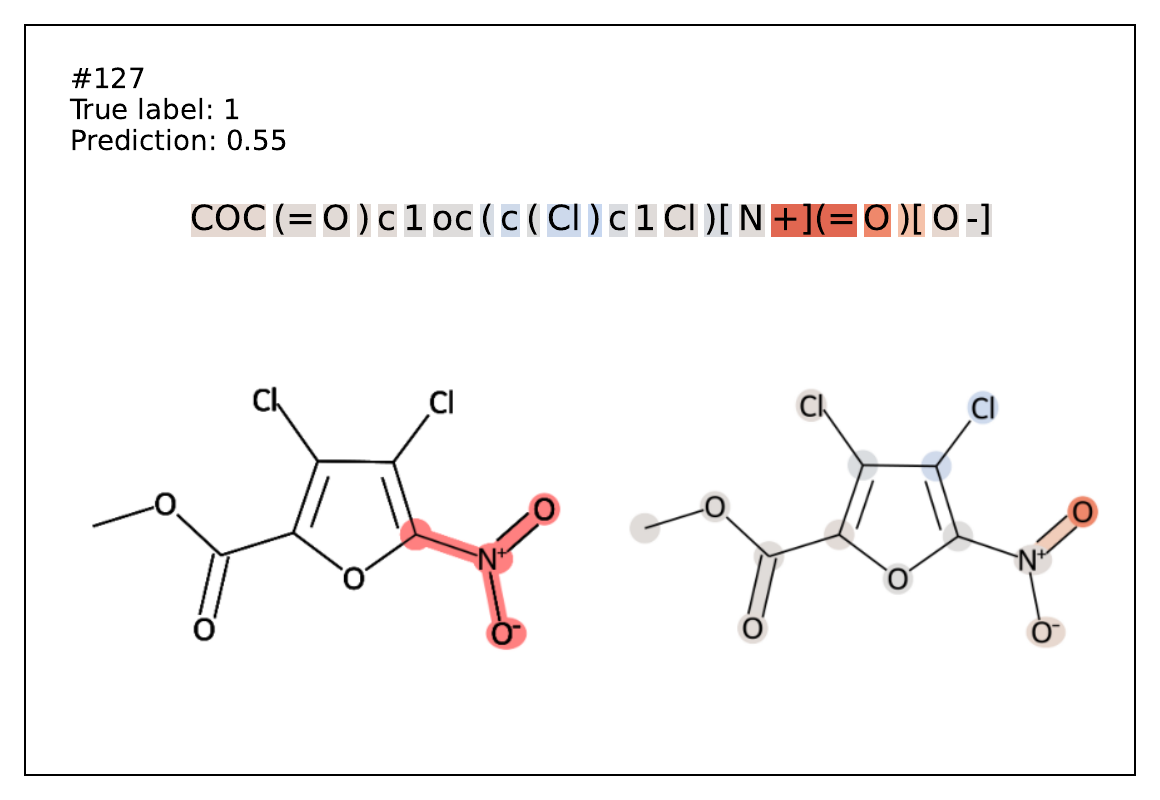} 
\end{subfigure} 
\begin{subfigure}[b]{0.33\textwidth} 
  \centering 
  \includegraphics[width=\textwidth]{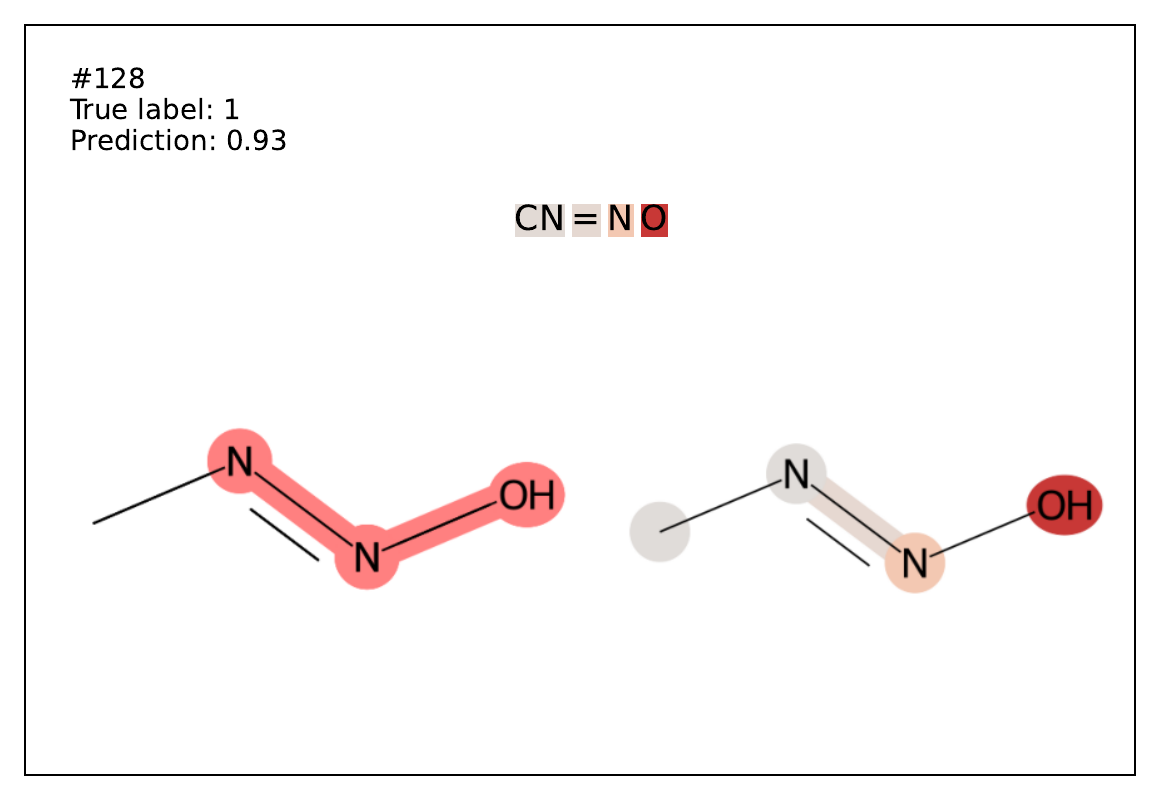} 
\end{subfigure}\begin{subfigure}[b]{0.33\textwidth} 
  \centering 
  \includegraphics[width=\textwidth]{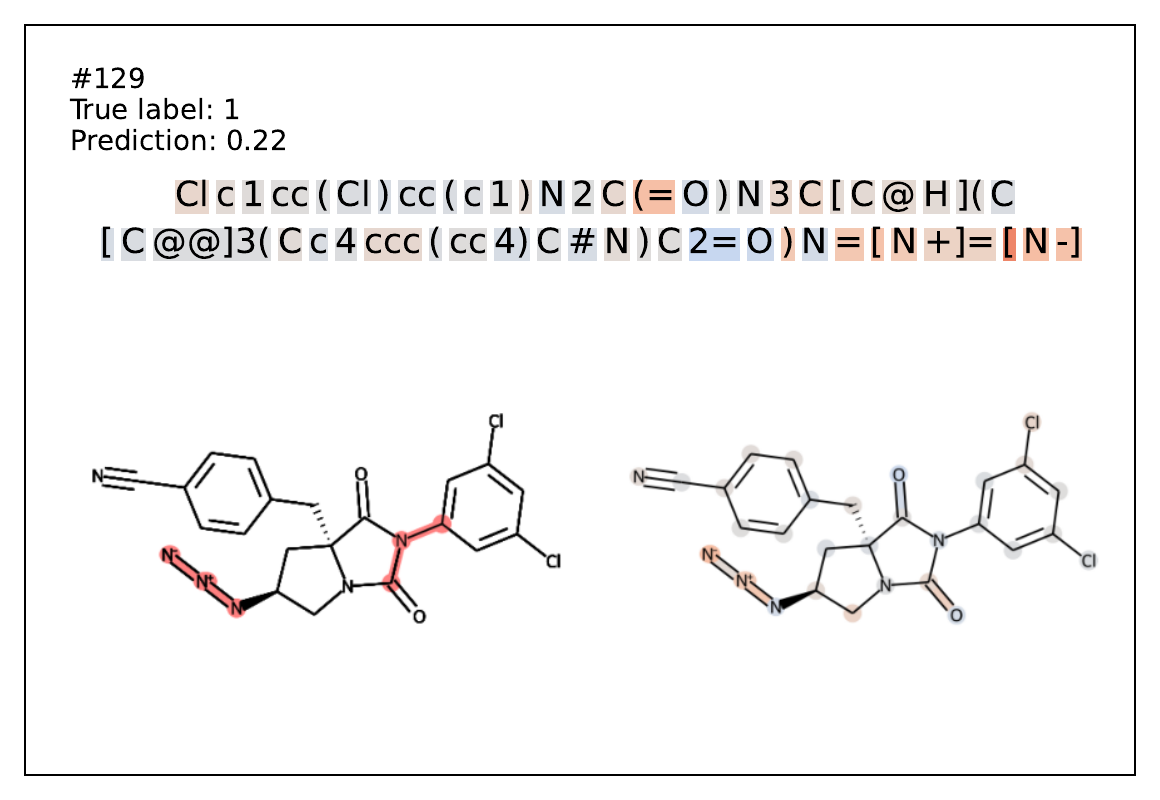} 
\end{subfigure}\begin{subfigure}[b]{0.33\textwidth} 
  \centering 
  \includegraphics[width=\textwidth]{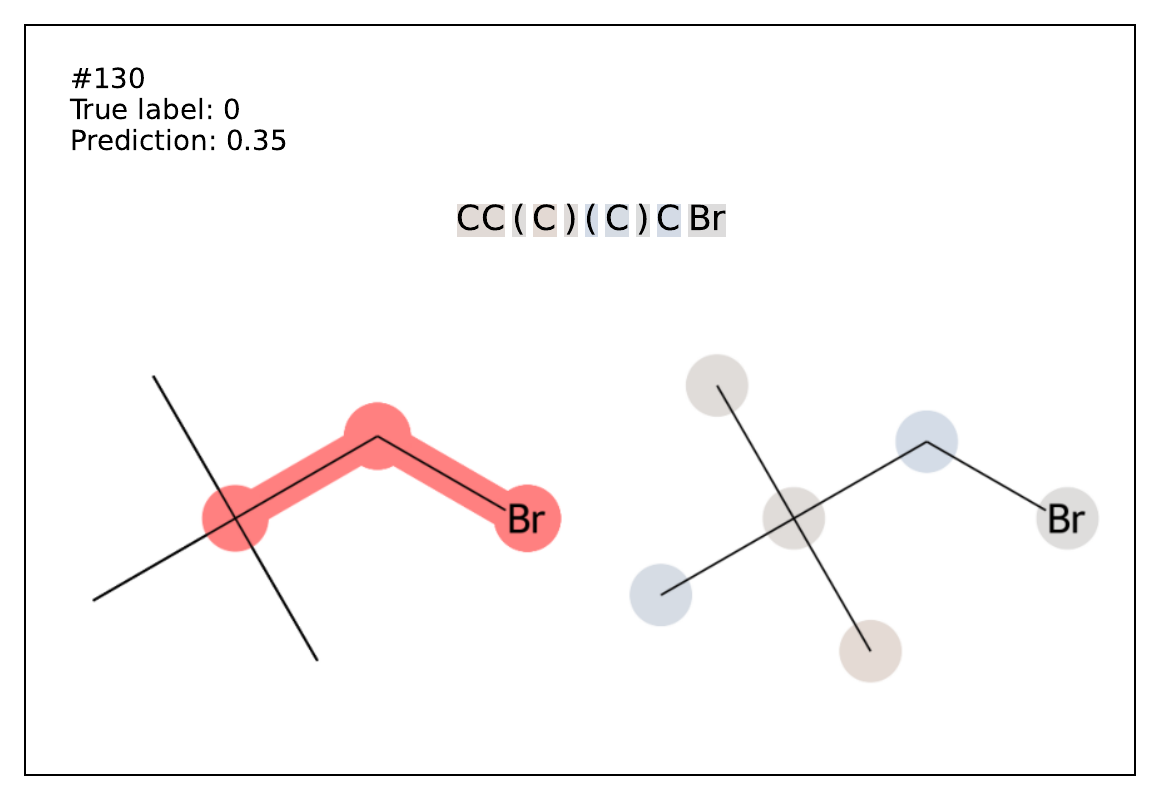} 
\end{subfigure} 
\begin{subfigure}[b]{0.33\textwidth} 
  \centering 
  \includegraphics[width=\textwidth]{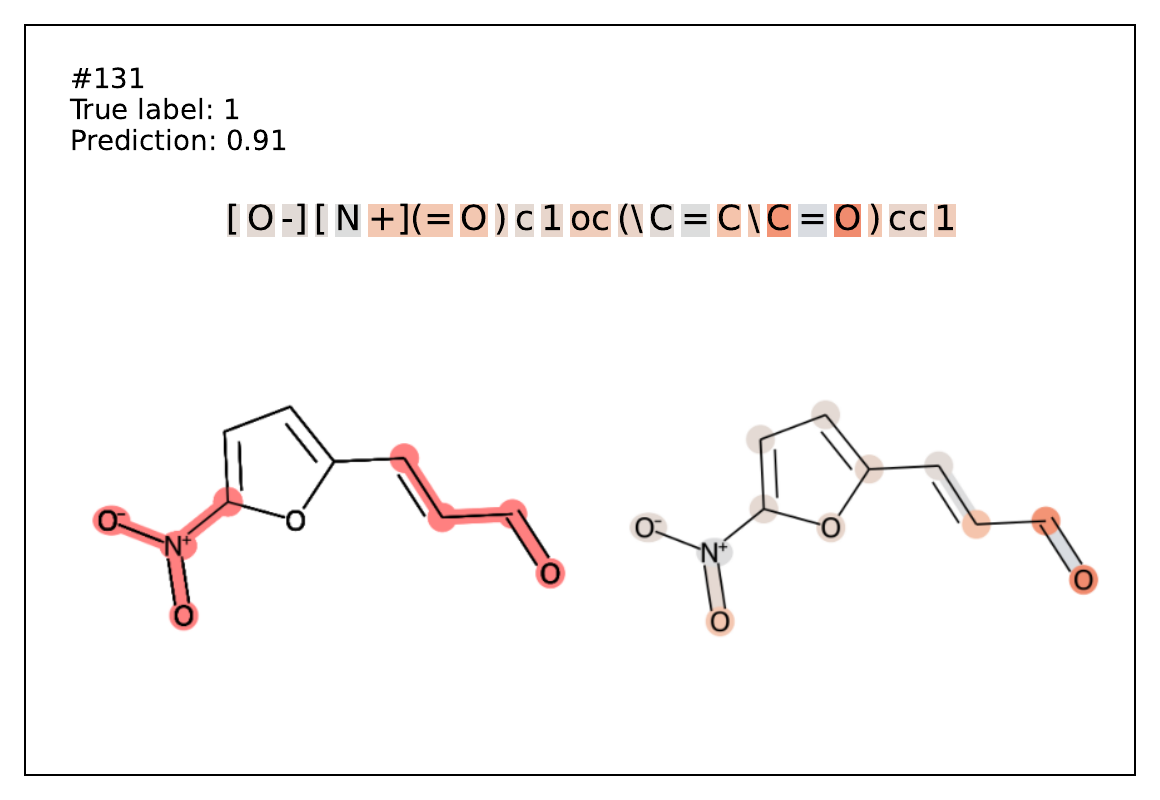} 
\end{subfigure}\begin{subfigure}[b]{0.33\textwidth} 
  \centering 
  \includegraphics[width=\textwidth]{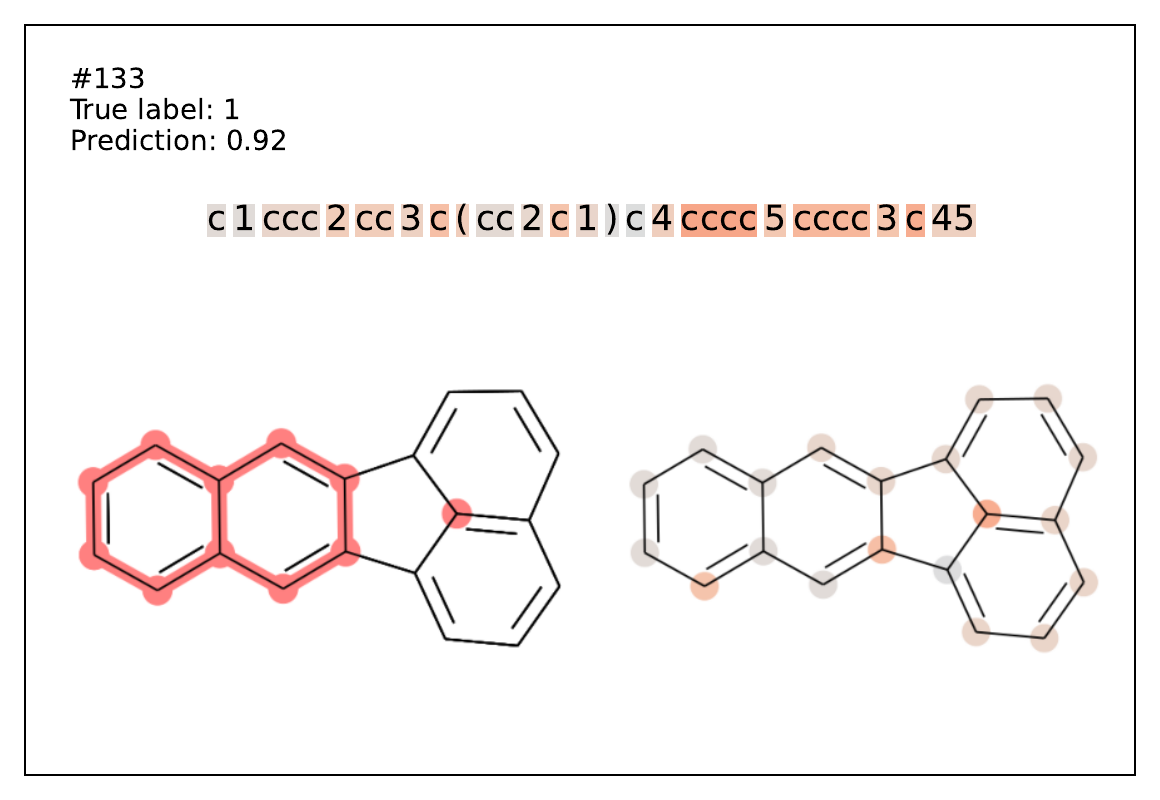} 
\end{subfigure}\begin{subfigure}[b]{0.33\textwidth} 
  \centering 
  \includegraphics[width=\textwidth]{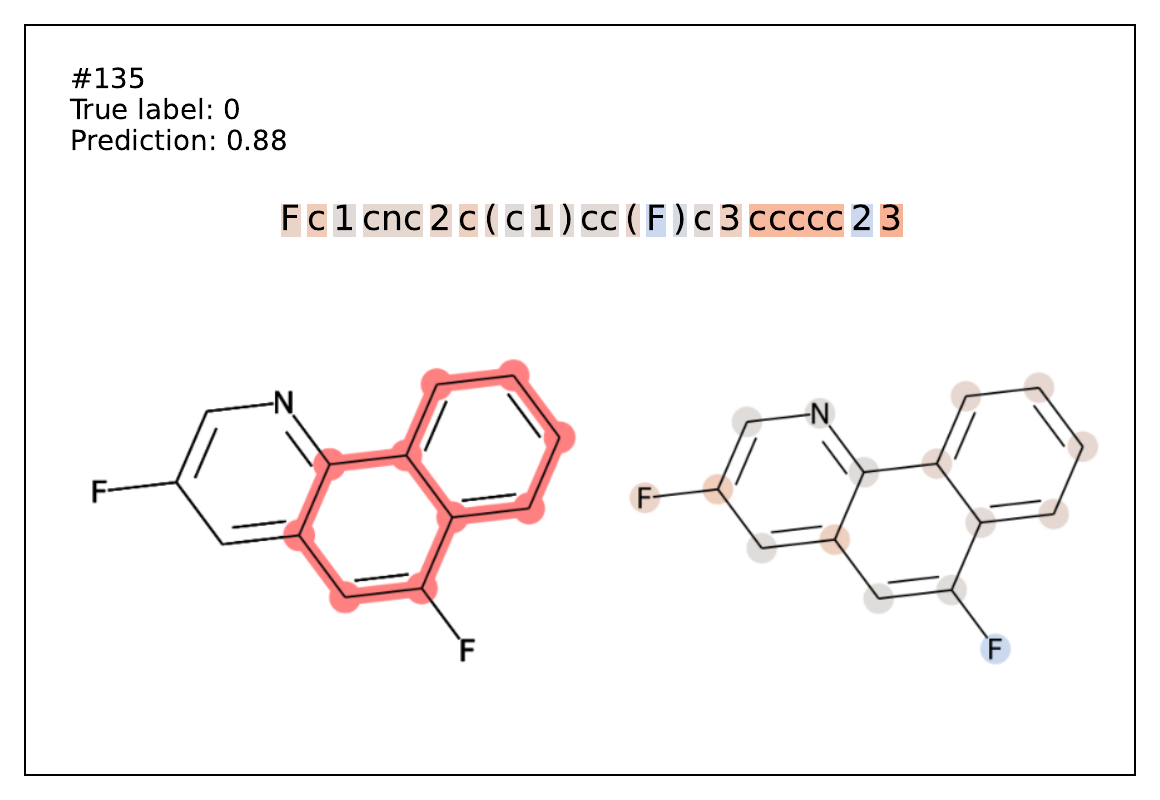} 
\end{subfigure} 
\begin{subfigure}[b]{0.33\textwidth} 
  \centering 
  \includegraphics[width=\textwidth]{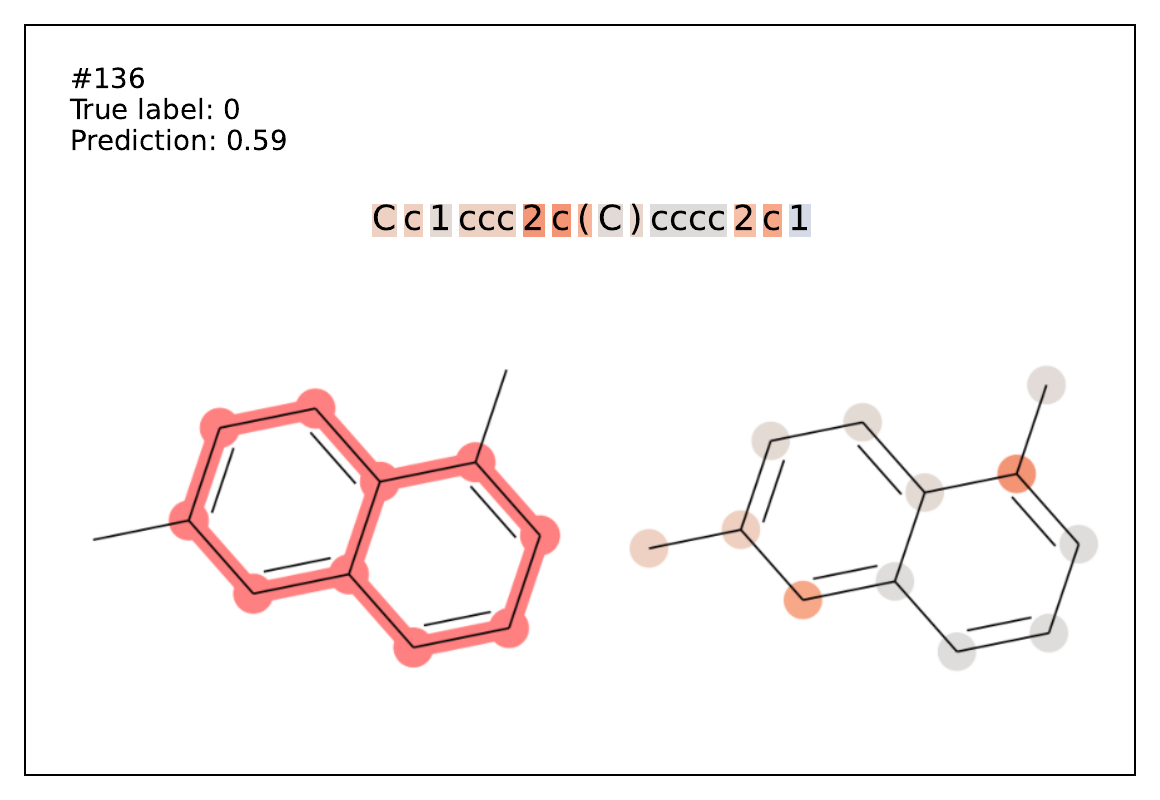} 
\end{subfigure}

\caption{Explaining predictions of the fine-tuned model on Ames dataset. See Section \ref{sec:captum}. Part 5/5}
\label{fig:captum-ames-5}
\end{figure}

In Figures \ref{fig:captum-ames-1} to \ref{fig:captum-ames-5} we show the structural alerts of more than a hundred molecules from the test set of Ames dataset. Chemicals under the numbers 10, 20, 21, 37, 40, 61, 71, 90, 93, 110, 117, and 127 are nitroaromatic compounds representing organic molecules that consist of at least one nitro group (-NO2) attached to an aromatic ring. These compounds and their residuals are well known for their carcinogenic and mutagenic potential. The nitro group in these molecules is characterized as structural alerts, i.e., molecular substructures related to the chemical's carcinogenic and mutagenic properties. 

In most cases, Integrated Gradients recognized two components of the structural alert: the positively charged nitrogen atom and a double bonded oxygen atom. In some cases, only a positive charge was highlighted along with the oxygen atom. It is essential to mention that a positively charged nitrogen atom plays a crucial role in the toxic potential of this structural group since this positive charge determines the electrophilic nature of the molecule. An electrophile is a chemical species that form bonds with nucleophiles by accepting an electron pair. Whereas DNA is abundantly equipped with nucleophilic sites, reaction with electrophiles results in diverse chemical-DNA interactions.%

However, in all cases, the most highlighted atom was a double-bonded oxygen atom. Two explanations may arise from this observation. First is that Integrated Gradients ``detects'' it as a nitroso group (N=O), e.g., as in the compounds under the numbers 42, 59, and 69, where they represent a separate structural alert group. Second, the activation of most nitroaromatic compounds' mutagenic properties is linked to the nitroreduction. As a result of nitroreduction, nitroso and superoxide species are produced.%
In both cases, double-bonded oxygen atoms play a vital role in forming toxic intermediates. So, in this case, it can be assumed that the algorithm recognizes target sites, such as positively charged nitrogen or oxygen atoms, responsible for producing DNA-damaging compounds.

The same pattern is observed in the case of compounds under the numbers 18, 38, 57, 60, 124, and 126. These compounds are characterized by the bearing of at least one epoxy group consisting of an oxygen atom joined by single bonds to two adjacent carbon atoms, which form the three-membered epoxide ring.%
The epoxy group is a well-known structural alert. Compounds bearing epoxy groups are alkylating agents in which carbon atoms represent electrophilic sites that react with nucleophilic DNA to form covalent bonds via nucleophilic substitution reaction, thus acting as direct genotoxins and carcinogens. Integrated Gradients recognized this structural alert in all cases, primarily highlighting the oxygen atom joined to one of the two carbon atoms. Most often, the electrophilic carbon atom responsible for chemical-DNA interaction was the most highlighted area, assuming that Integrated Gradients highlighted the most active site of the molecule from the genotoxicity point of view. Interestingly, the highlighting of the oxygen atom is always accompanied by one or two equally highlighted carbon atoms and never appears separately, only oxygen or carbon atoms. Presumably, the method highlights the bond of these atoms, which is the second most important characteristic of epoxy groups, since, during the nucleophilic substitution reaction, the three-membered ring is opened via one of the cleavages of the O-C bond, forming the favorable for the molecule unstrained acyclic intermediate. %

It is worth noting that in the case of nitroaromatic compounds, the components of the structural alerts have been detected, along with the high probability ($>0.8$) of correct prediction. In contrast, in the case of epoxy compounds, the structural alerts have been detected correctly, but the probability level of correct predictions was around 0.22-0.48, except for compounds 124 and 126 (probability levels 0.891 and 0.932).

An example where the attributions given by Integrated Gradients is not correlated with the structural alerts is the case of the naphthalene group in polycyclic aromatic hydrocarbons (compounds under the numbers 8, 28, 31, 48, 58, 62, 61, 76, 99, 113, and 108). The naphthalene group is a bicyclic aromatic hydrocarbon representing a structural alert in all mentioned compounds. Although the attributions of Integrated Gradients were correct with the high probability level in all cases, the structural alert has not been recognized. Here, the activated neurons have been dispersed distributed throughout polycyclic aromatic hydrocarbons, rarely matching the part of the structural alert.

%% file: sections/appendix-esol.tex
\begin{figure}
\centering
\begin{subfigure}[b]{0.33\textwidth} 
  \centering 
  \includegraphics[width=\textwidth]{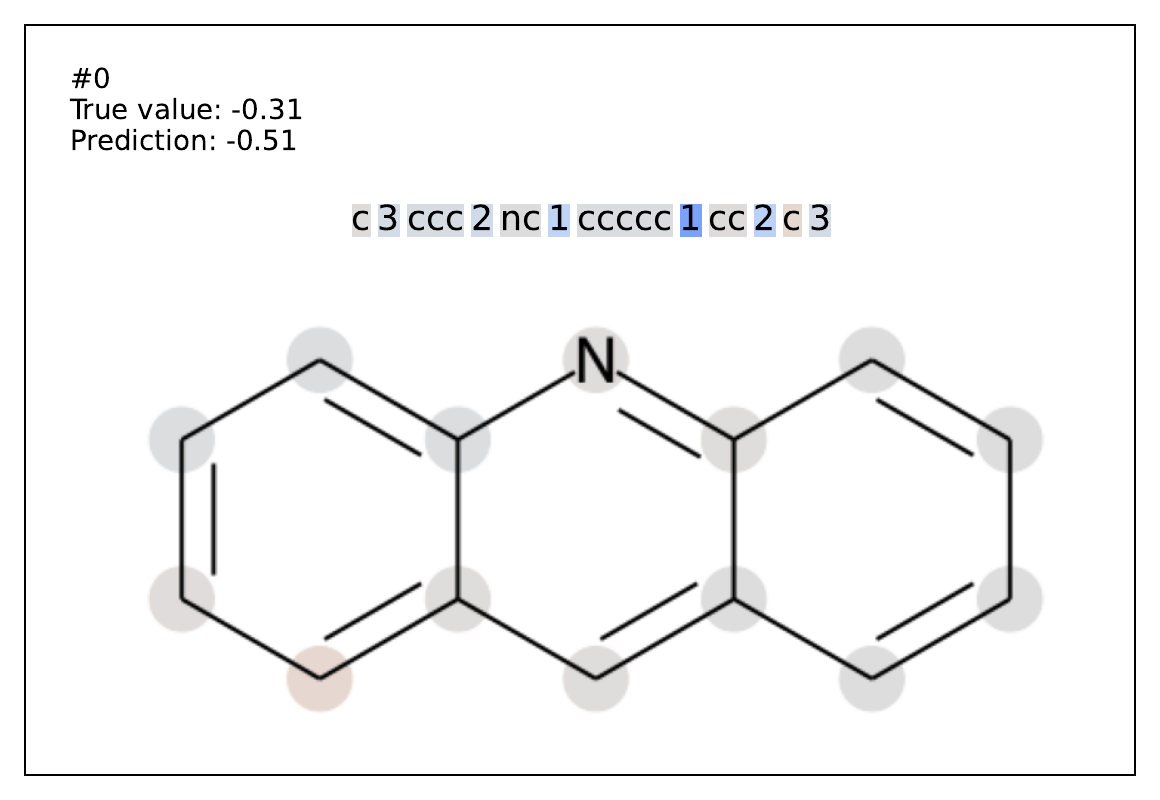} 
\end{subfigure}\begin{subfigure}[b]{0.33\textwidth} 
  \centering 
  \includegraphics[width=\textwidth]{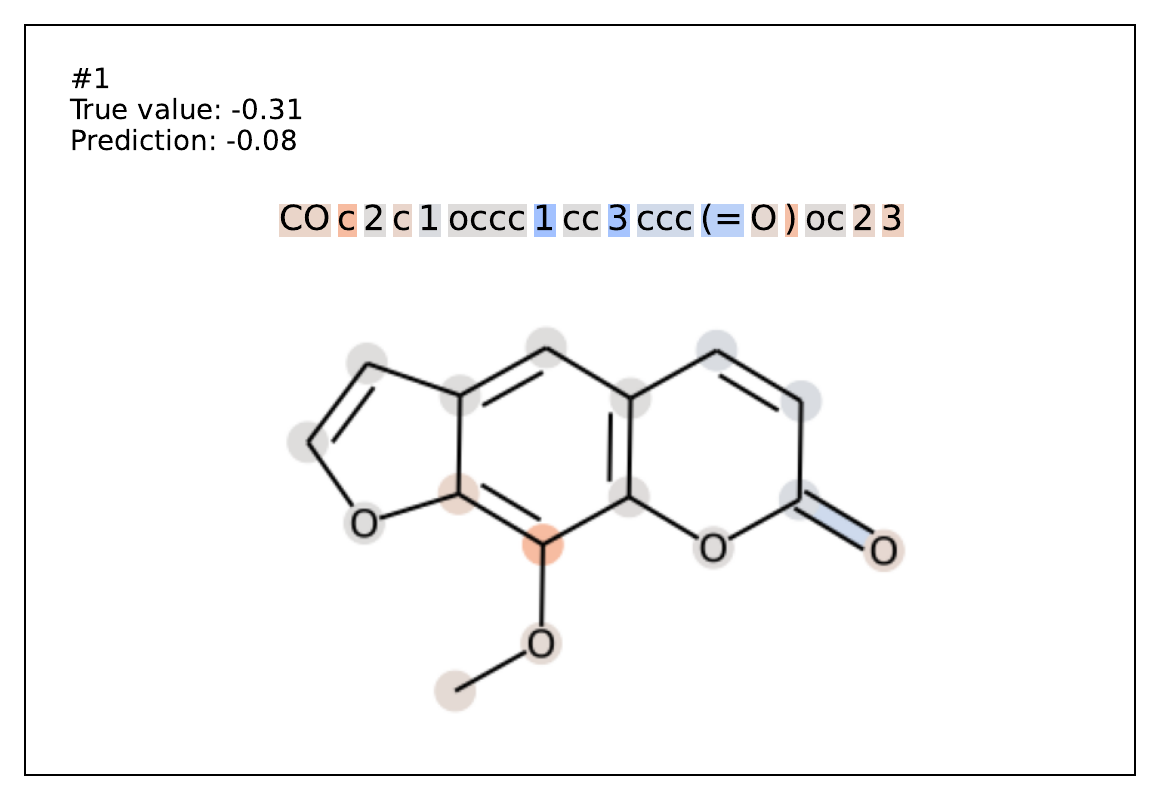} 
\end{subfigure}\begin{subfigure}[b]{0.33\textwidth} 
  \centering 
  \includegraphics[width=\textwidth]{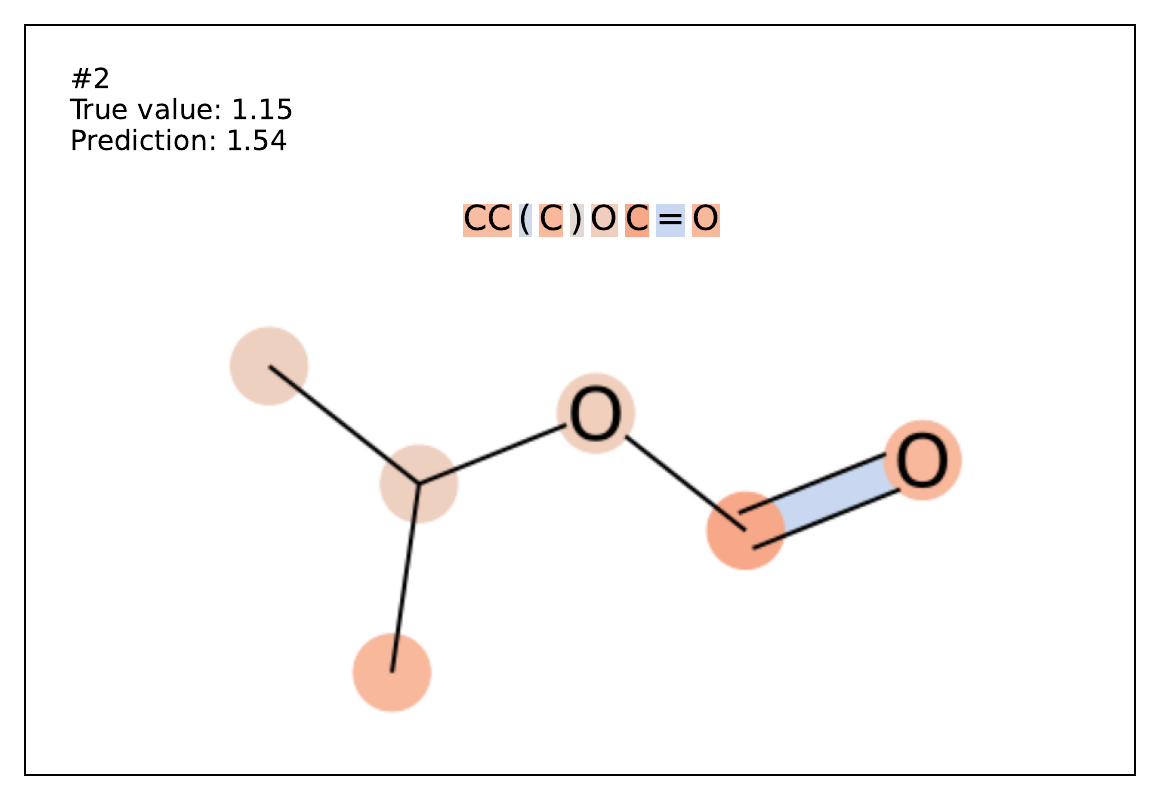} 
\end{subfigure}
\begin{subfigure}[b]{0.33\textwidth} 
  \centering 
  \includegraphics[width=\textwidth]{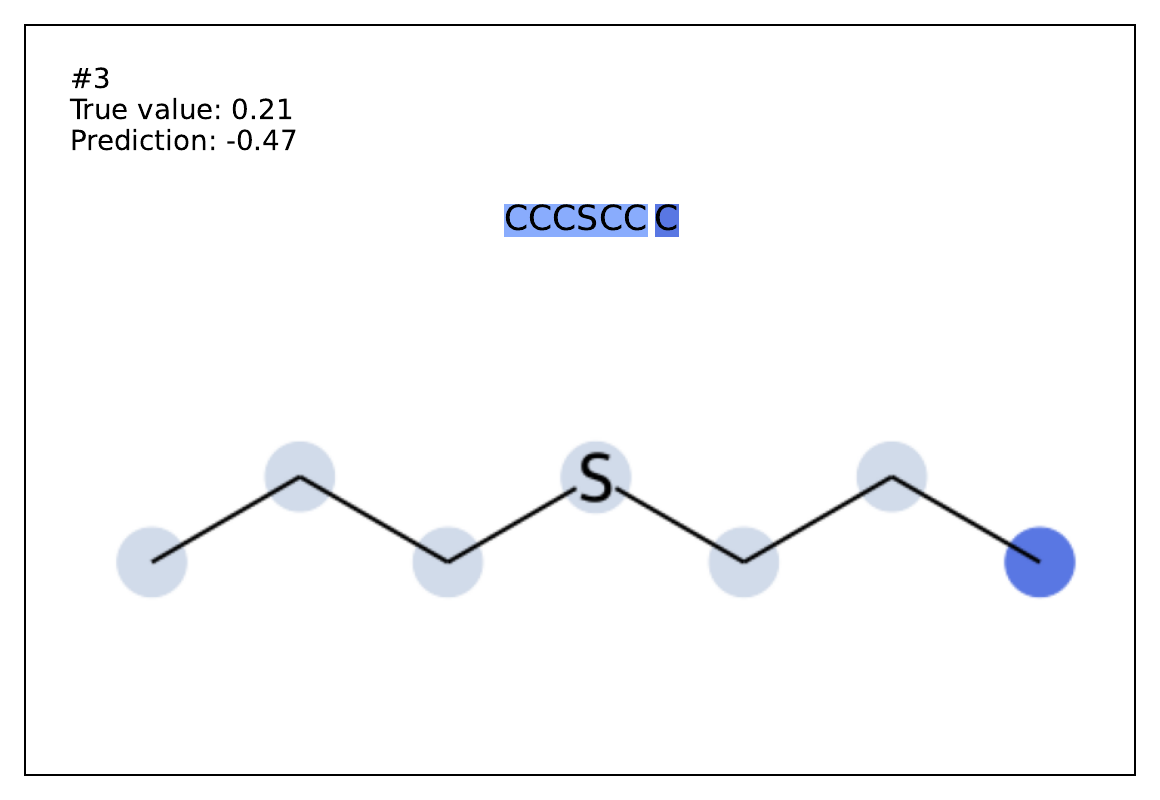} 
\end{subfigure}\begin{subfigure}[b]{0.33\textwidth} 
  \centering 
  \includegraphics[width=\textwidth]{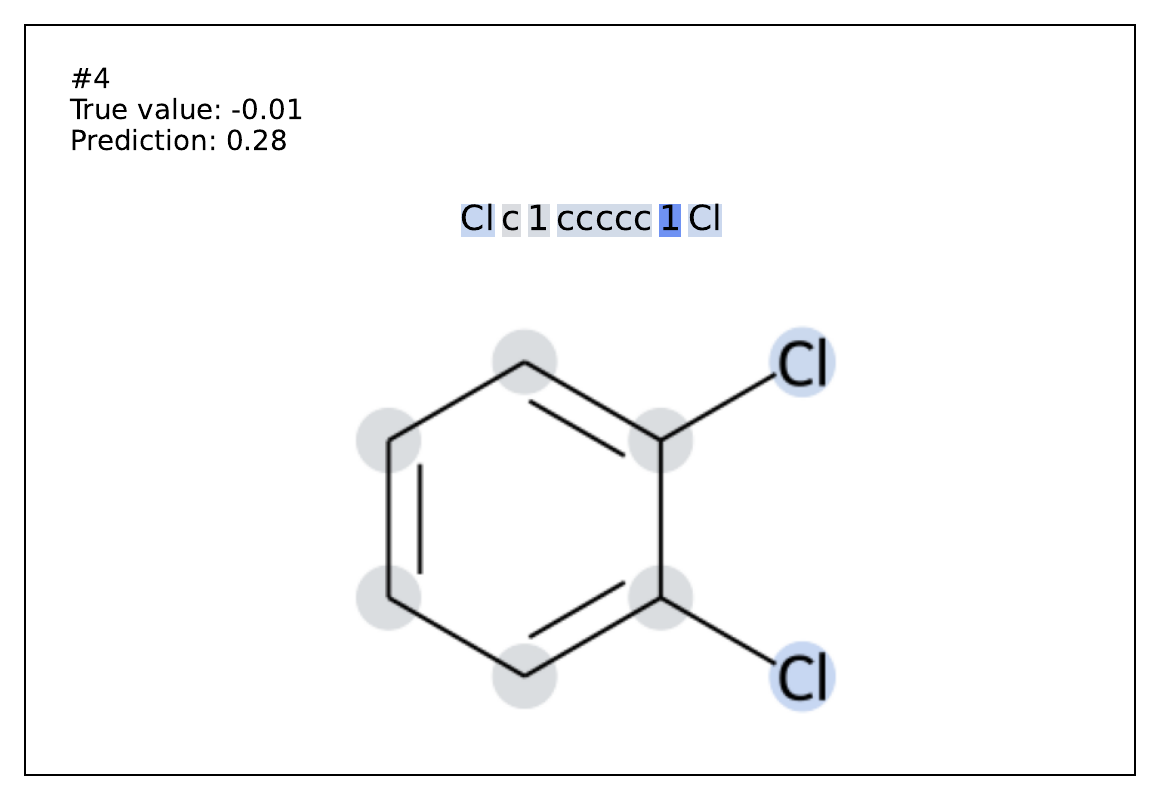} 
\end{subfigure}\begin{subfigure}[b]{0.33\textwidth} 
  \centering 
  \includegraphics[width=\textwidth]{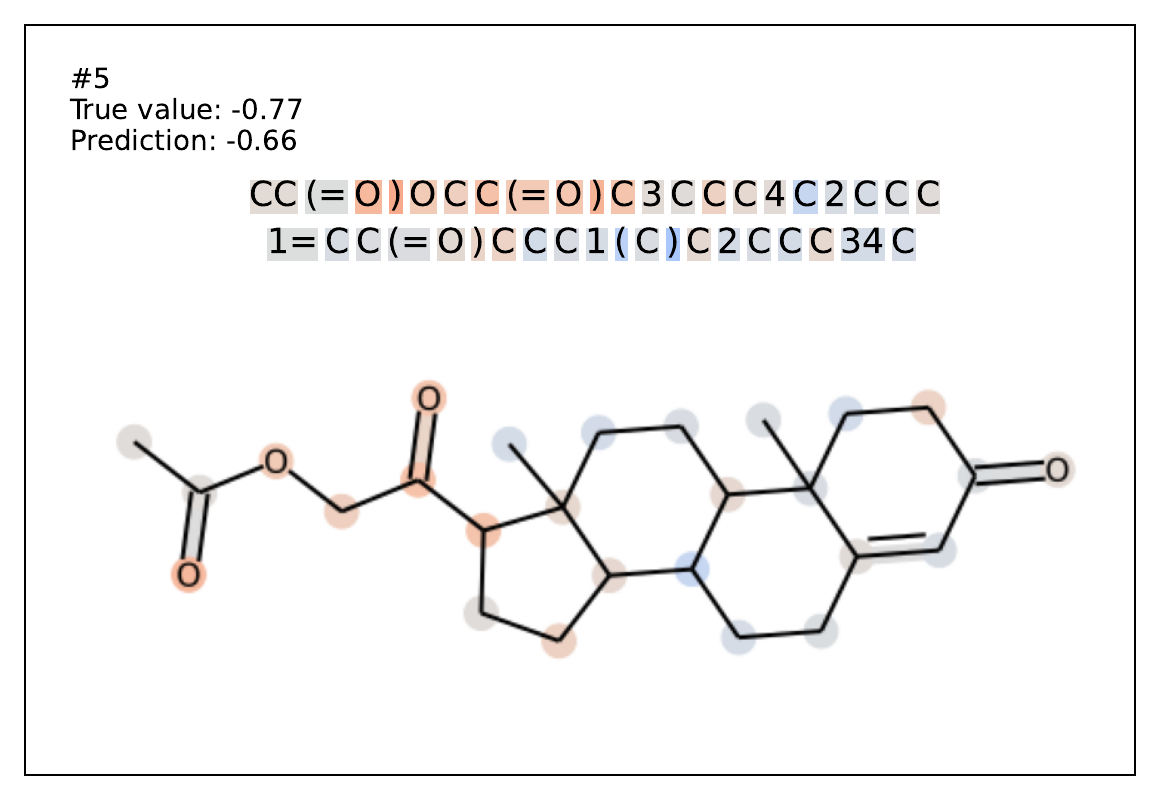} 
\end{subfigure}
\begin{subfigure}[b]{0.33\textwidth} 
  \centering 
  \includegraphics[width=\textwidth]{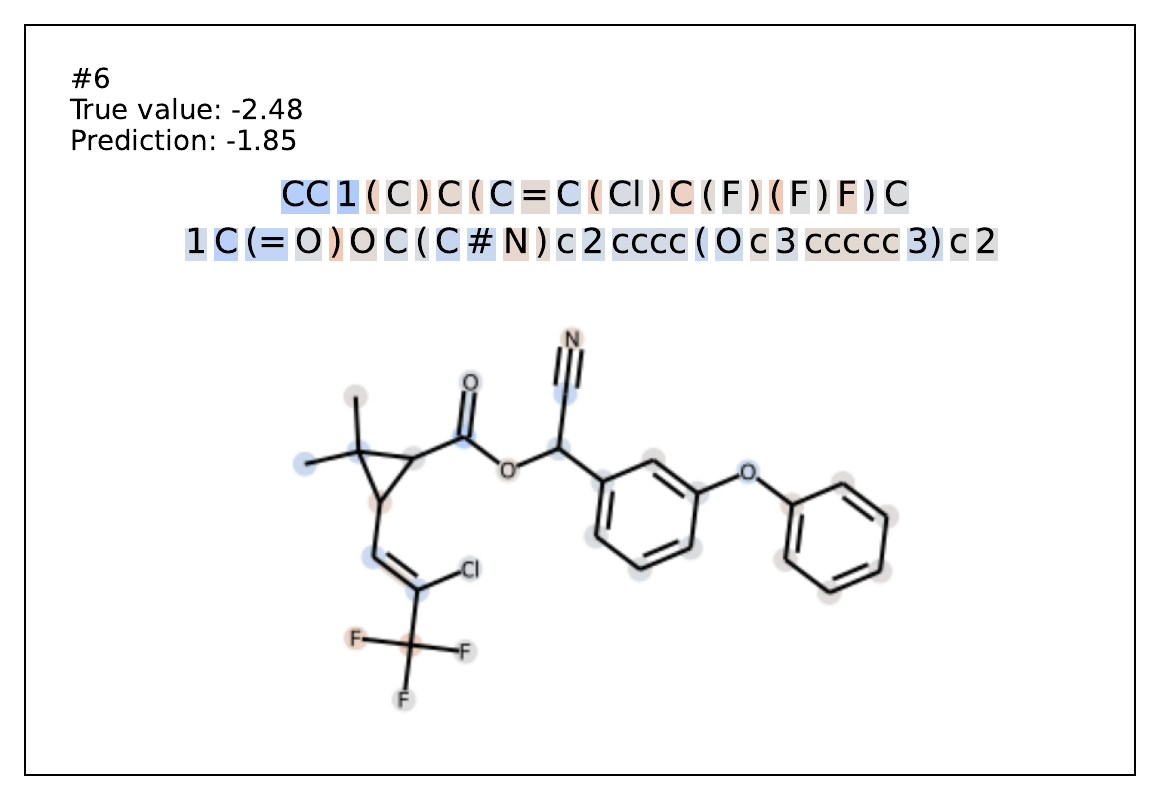} 
\end{subfigure}\begin{subfigure}[b]{0.33\textwidth} 
  \centering 
  \includegraphics[width=\textwidth]{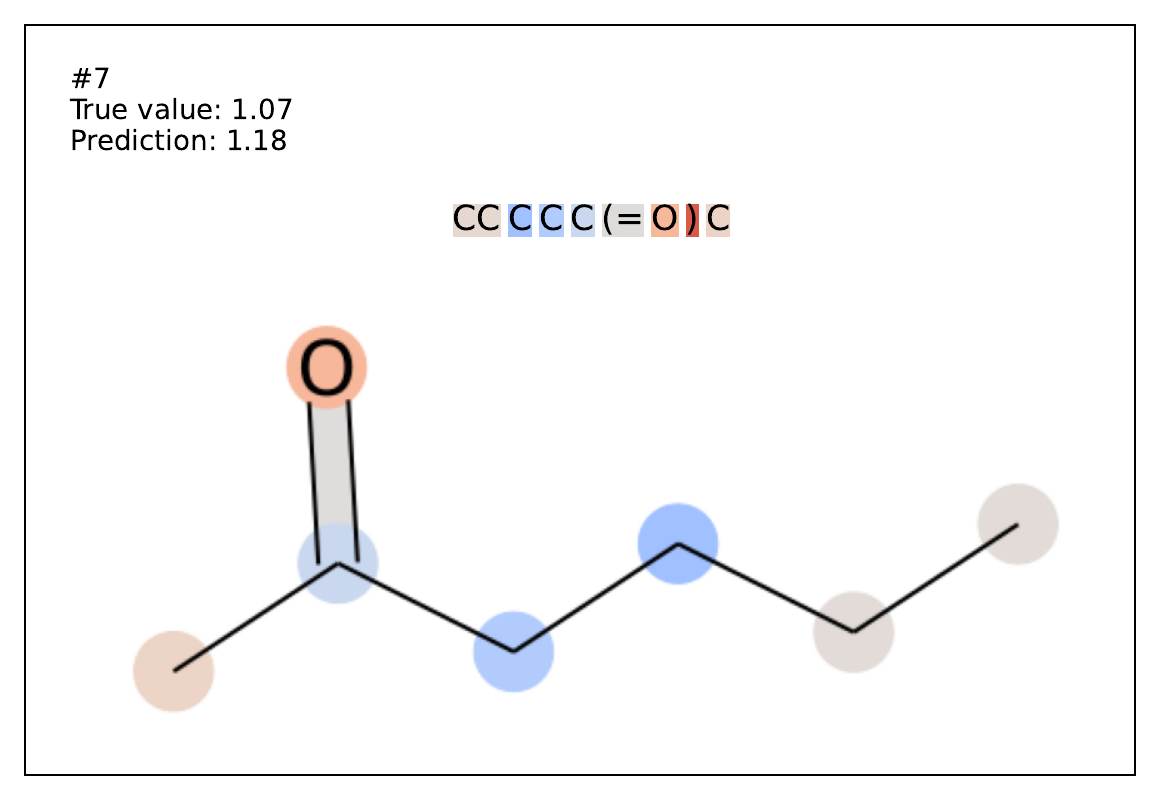} 
\end{subfigure}\begin{subfigure}[b]{0.33\textwidth} 
  \centering 
  \includegraphics[width=\textwidth]{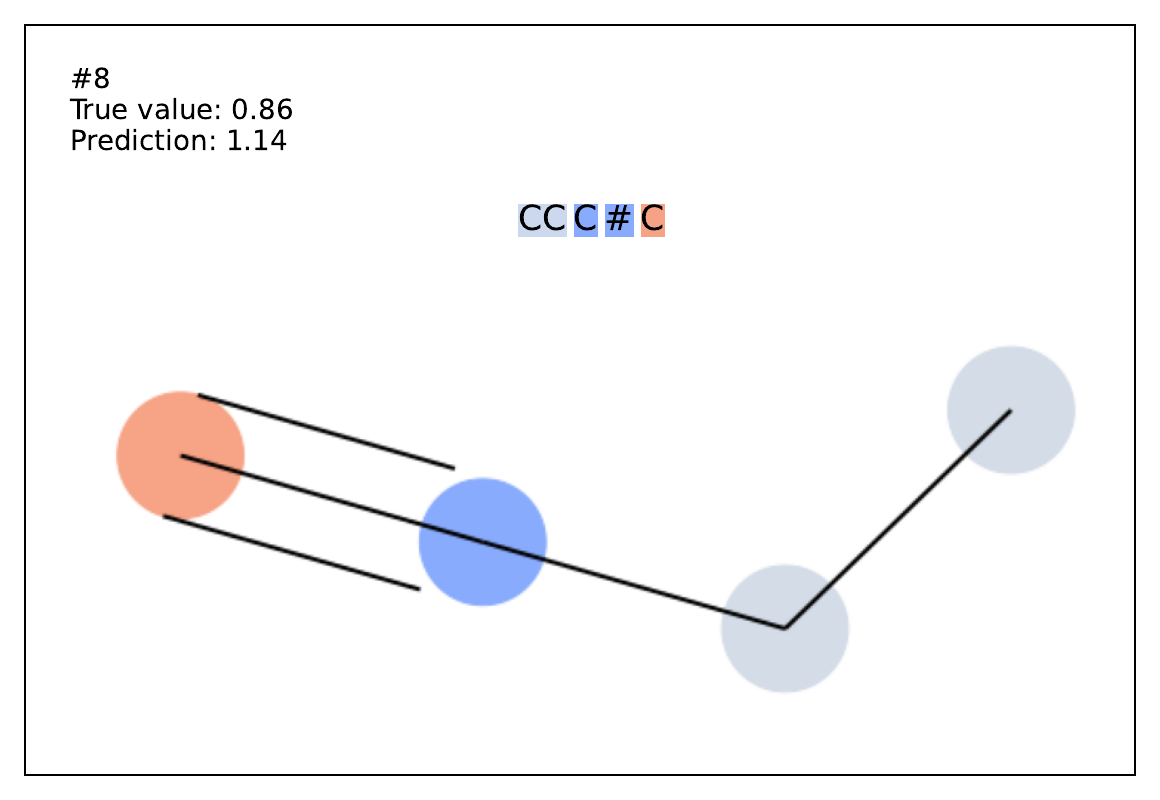} 
\end{subfigure}
\begin{subfigure}[b]{0.33\textwidth} 
  \centering 
  \includegraphics[width=\textwidth]{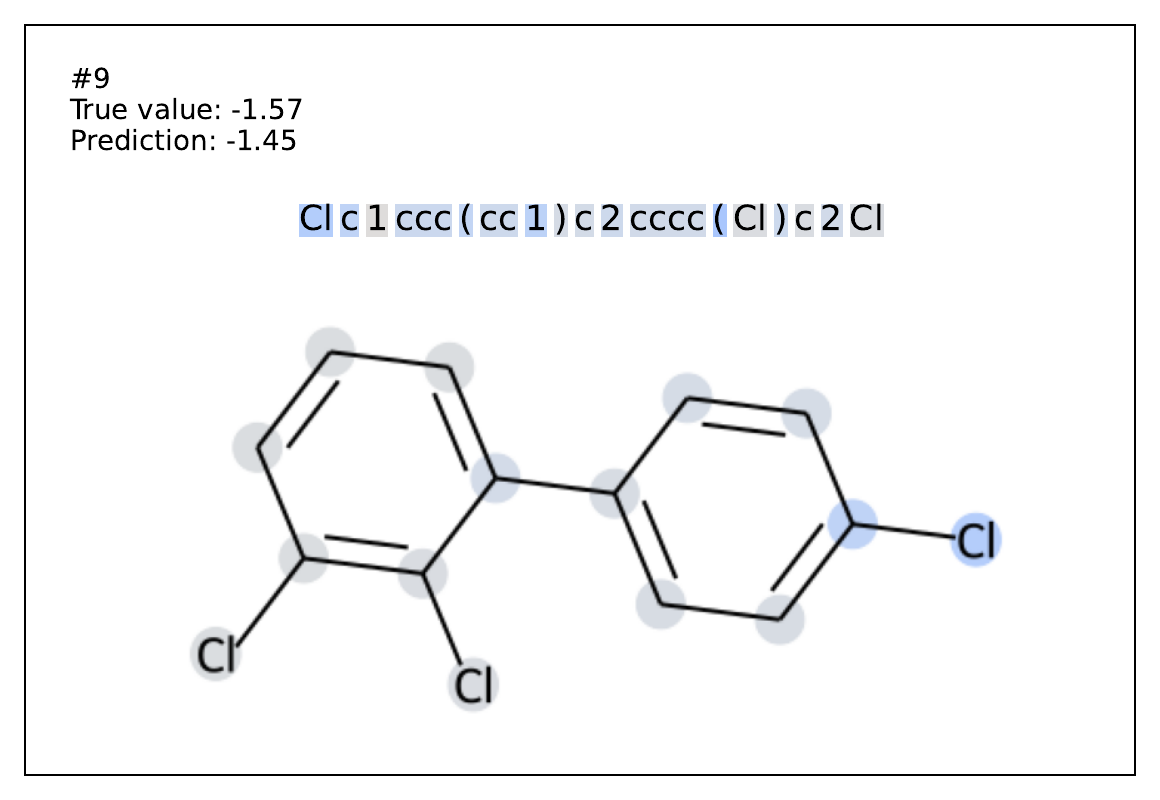} 
\end{subfigure}\begin{subfigure}[b]{0.33\textwidth} 
  \centering 
  \includegraphics[width=\textwidth]{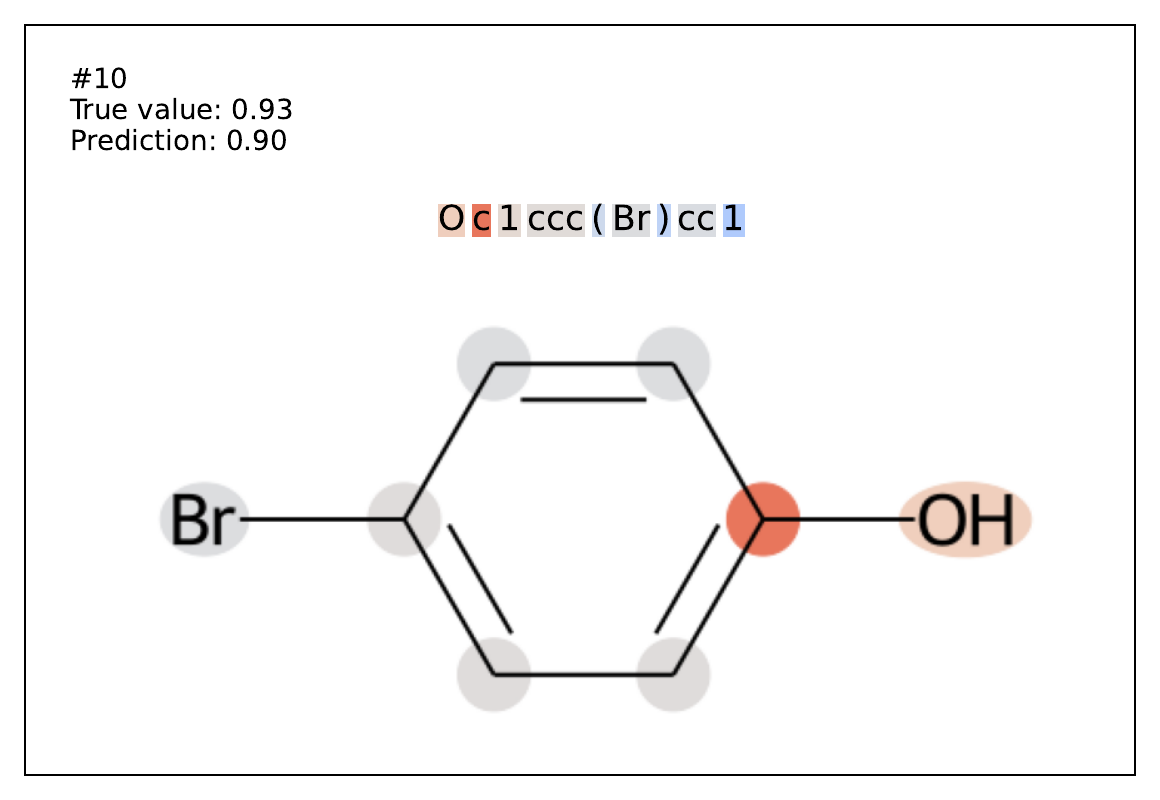} 
\end{subfigure}\begin{subfigure}[b]{0.33\textwidth} 
  \centering 
  \includegraphics[width=\textwidth]{figures/esol/esol12.pdf} 
\end{subfigure}
\begin{subfigure}[b]{0.33\textwidth} 
  \centering 
  \includegraphics[width=\textwidth]{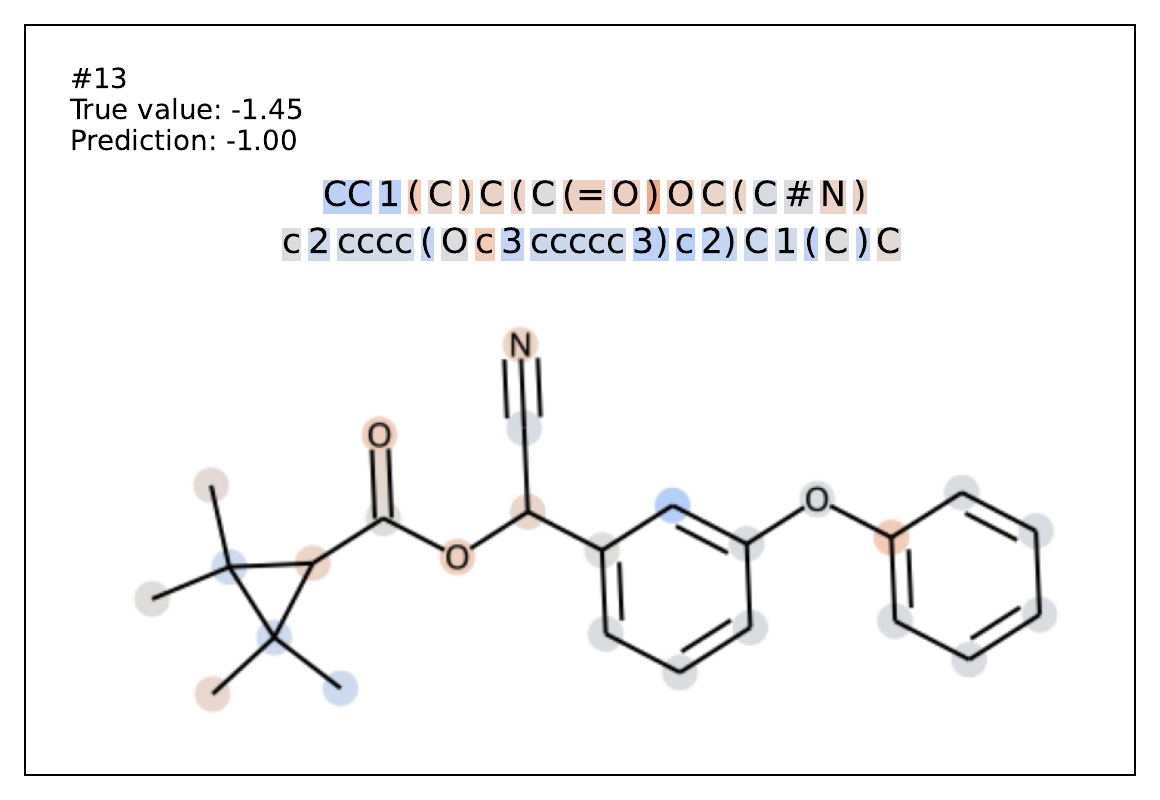} 
\end{subfigure}\begin{subfigure}[b]{0.33\textwidth} 
  \centering 
  \includegraphics[width=\textwidth]{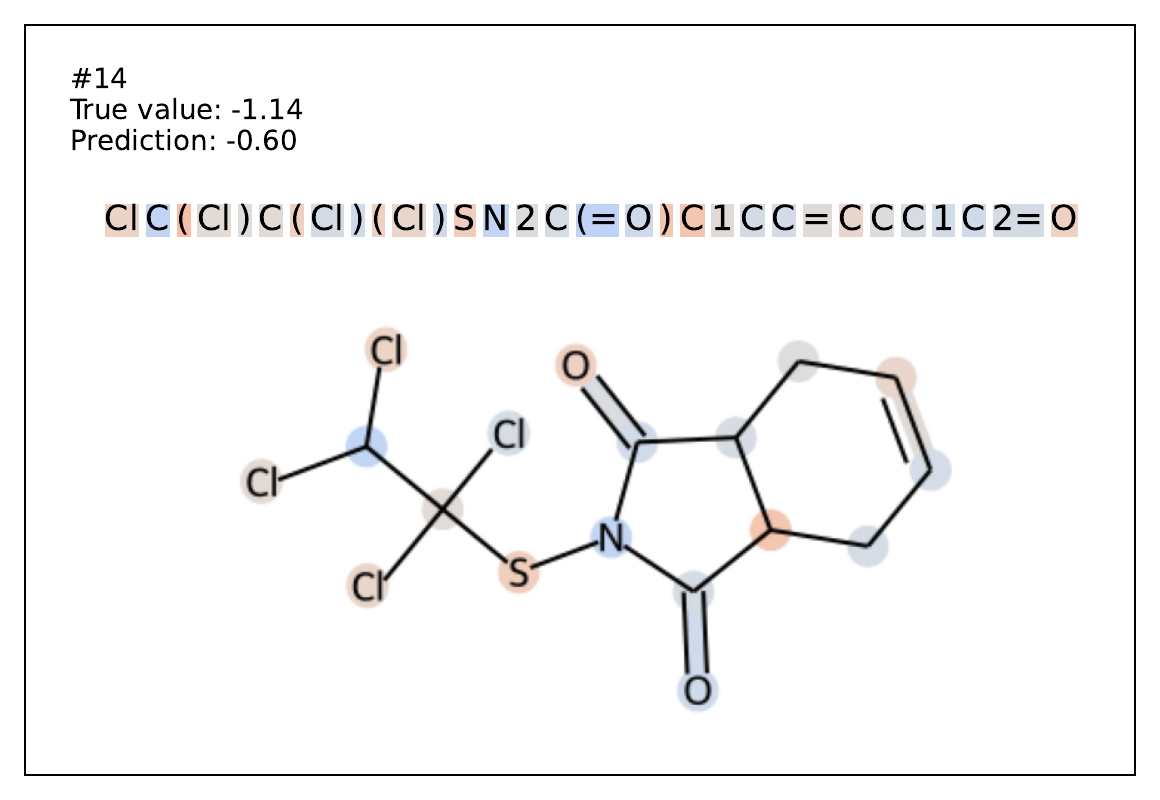} 
\end{subfigure}\begin{subfigure}[b]{0.33\textwidth} 
  \centering 
  \includegraphics[width=\textwidth]{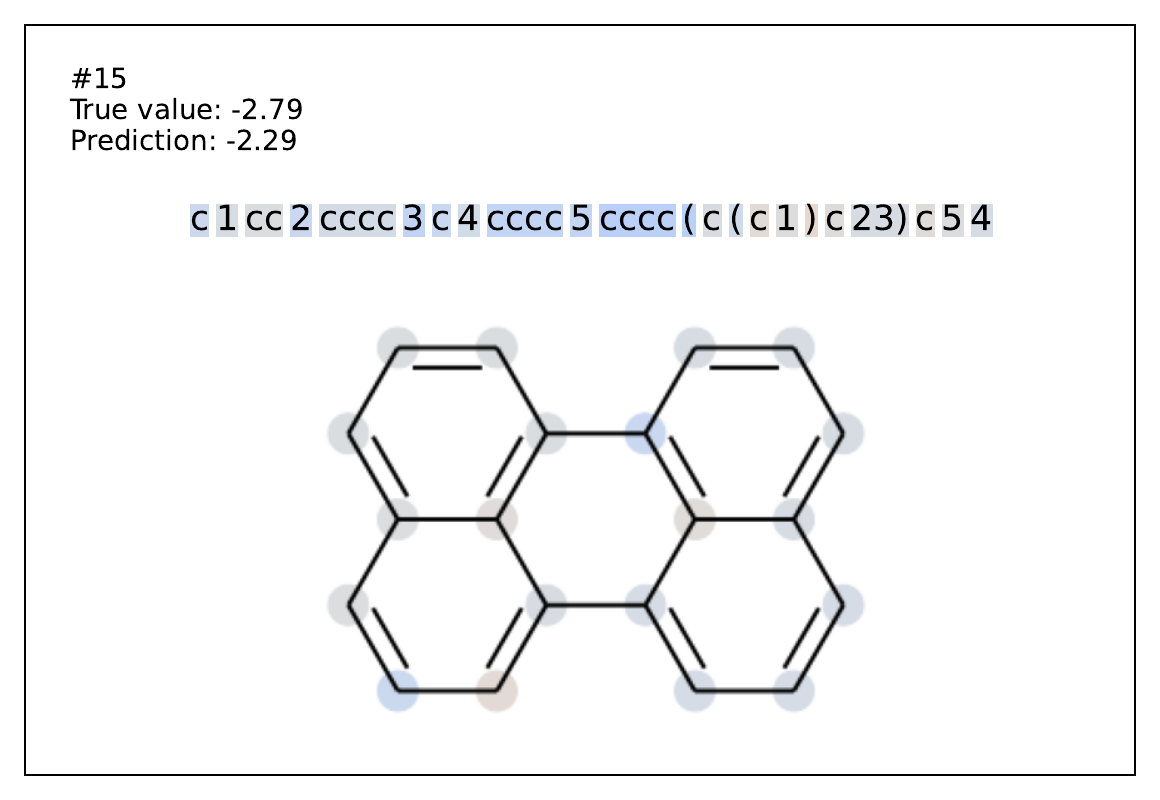} 
\end{subfigure}
\begin{subfigure}[b]{0.33\textwidth} 
  \centering 
  \includegraphics[width=\textwidth]{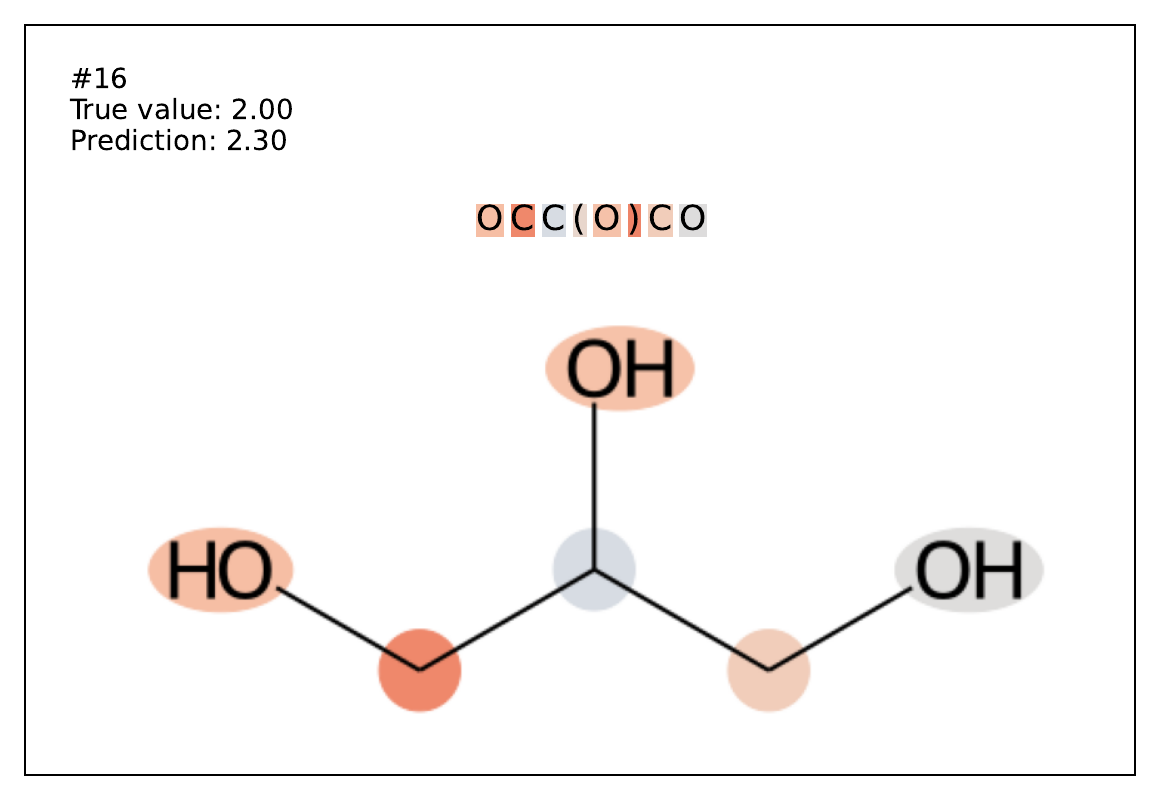} 
\end{subfigure}\begin{subfigure}[b]{0.33\textwidth} 
  \centering 
  \includegraphics[width=\textwidth]{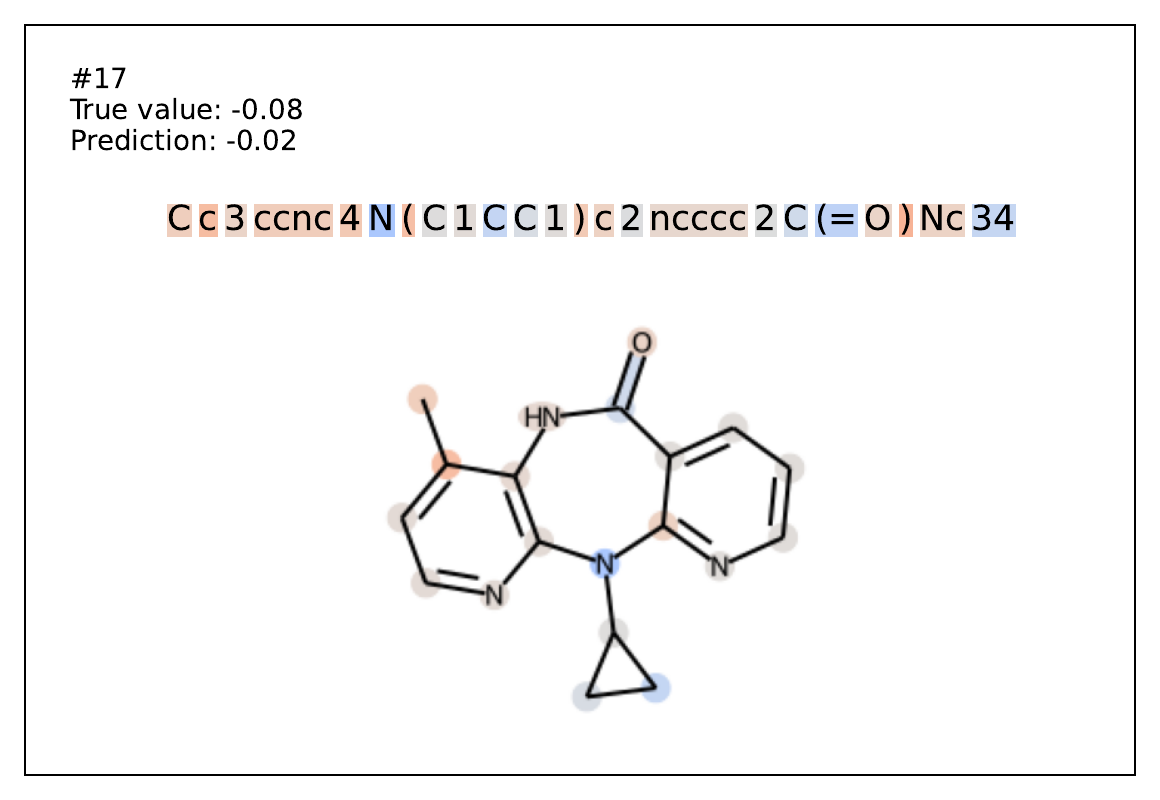} 
\end{subfigure}\begin{subfigure}[b]{0.33\textwidth} 
  \centering 
  \includegraphics[width=\textwidth]{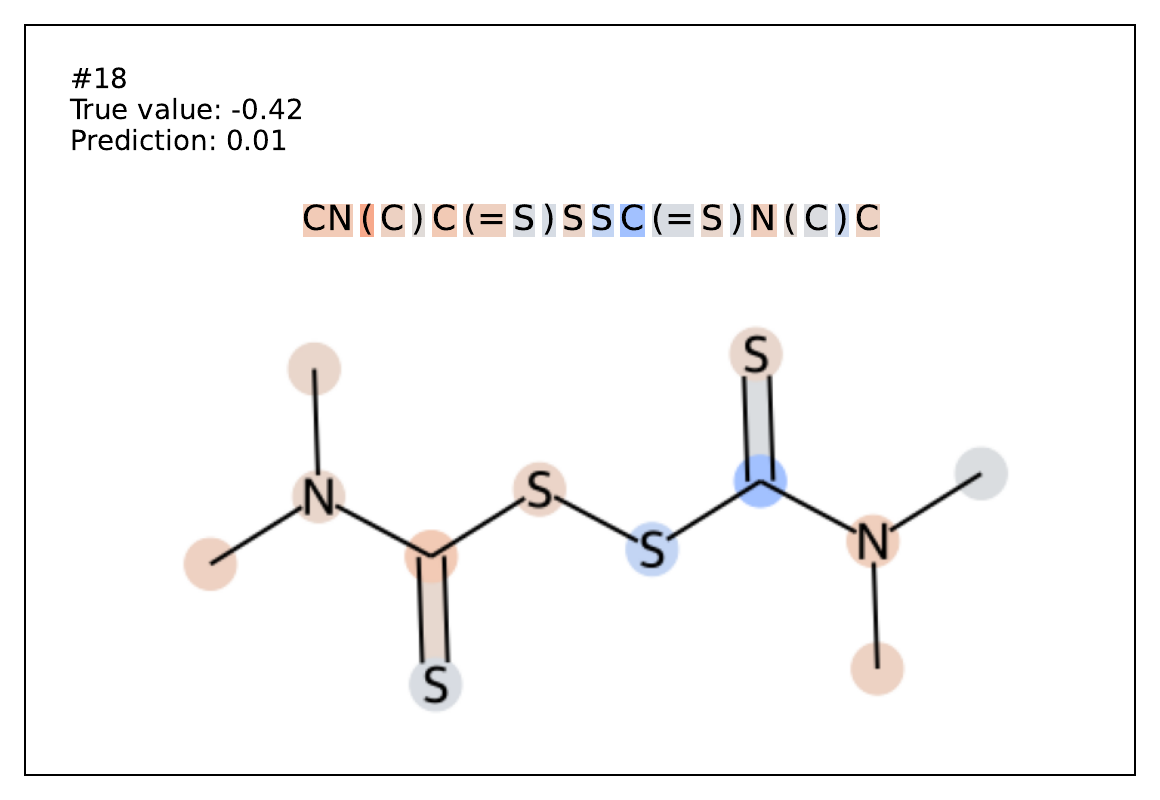} 
\end{subfigure}
\begin{subfigure}[b]{0.33\textwidth} 
  \centering 
  \includegraphics[width=\textwidth]{figures/esol/esol19.pdf} 
\end{subfigure}\begin{subfigure}[b]{0.33\textwidth} 
  \centering 
  \includegraphics[width=\textwidth]{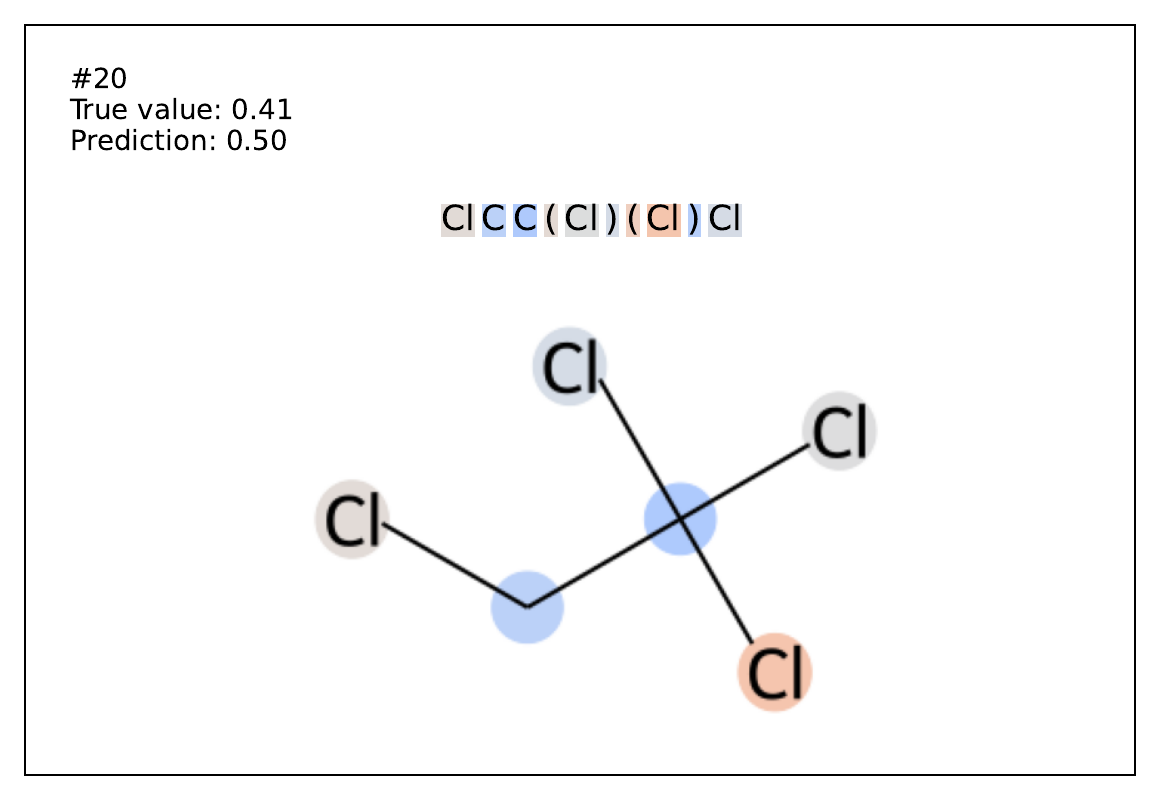} 
\end{subfigure}\begin{subfigure}[b]{0.33\textwidth} 
  \centering 
  \includegraphics[width=\textwidth]{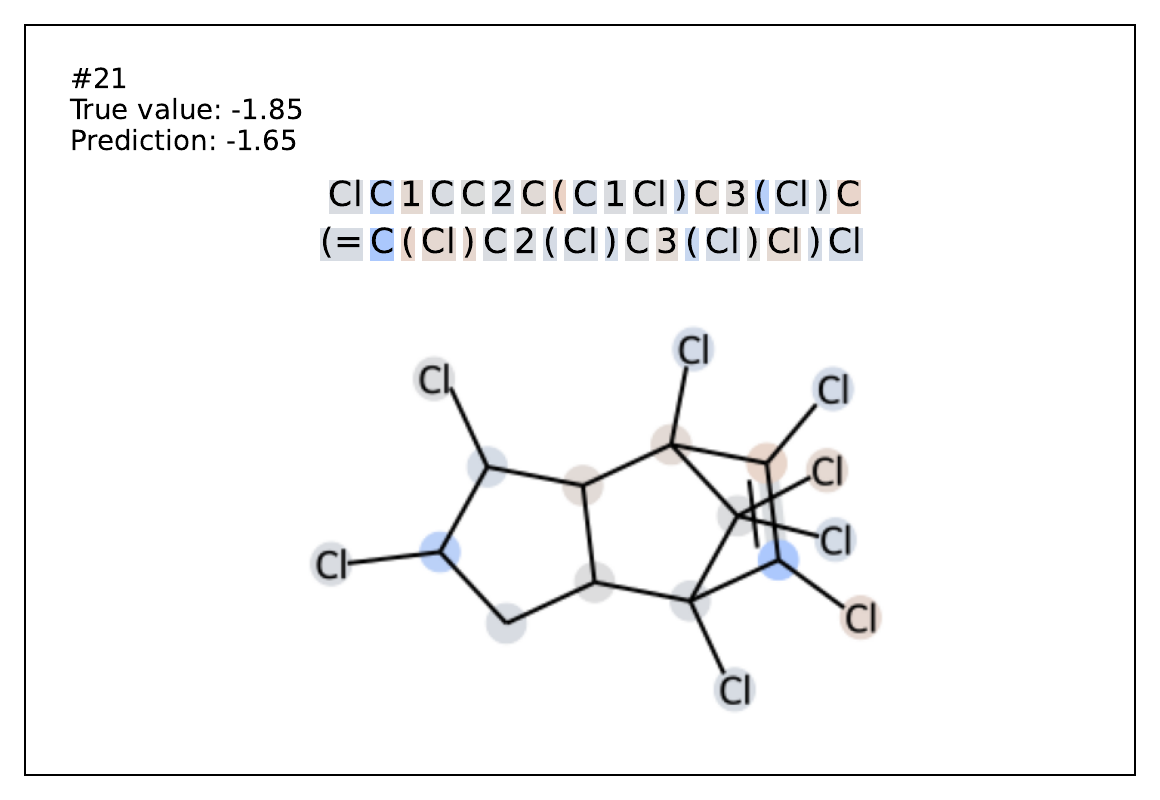} 
\end{subfigure}

\caption{Explaining predictions of the fine-tuned model on ESOL dataset. See Section \ref{sec:captum}. Part 1/3}
\label{fig:captum-esol-1}
\end{figure}

\begin{figure}
\centering

\begin{subfigure}[b]{0.33\textwidth} 
  \centering 
  \includegraphics[width=\textwidth]{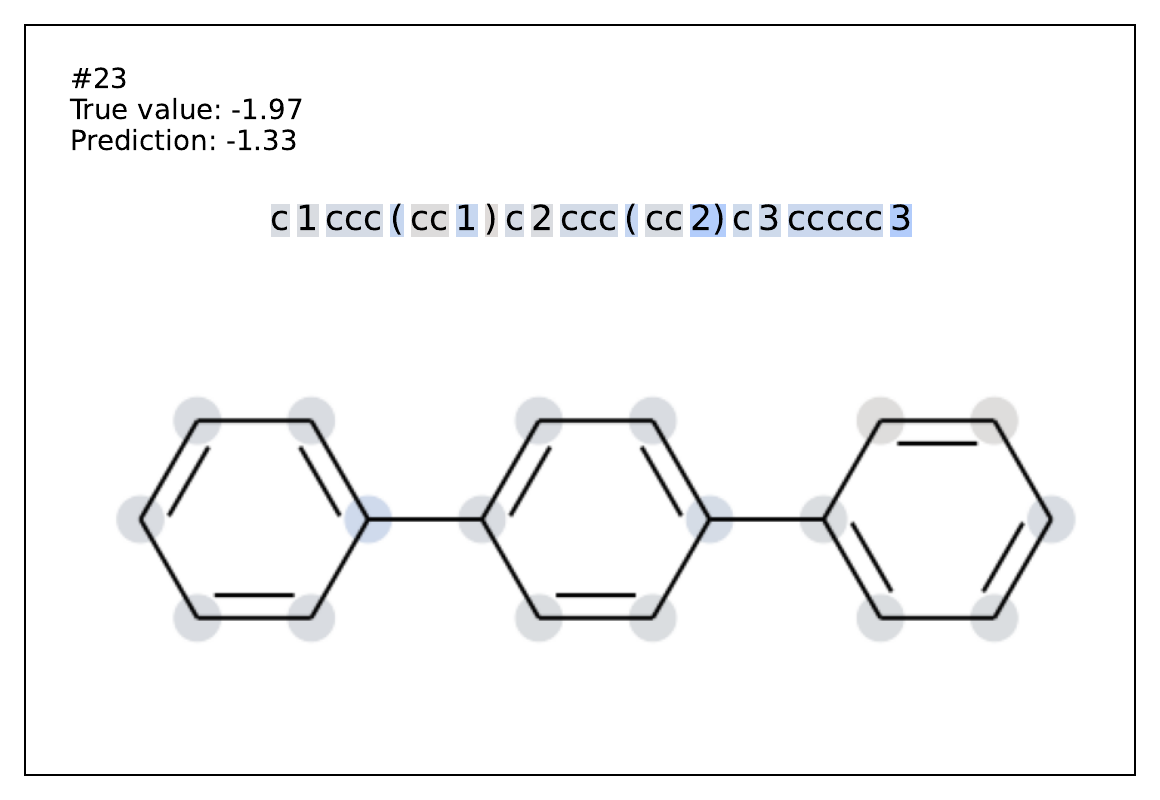} 
\end{subfigure}\begin{subfigure}[b]{0.33\textwidth} 
  \centering 
  \includegraphics[width=\textwidth]{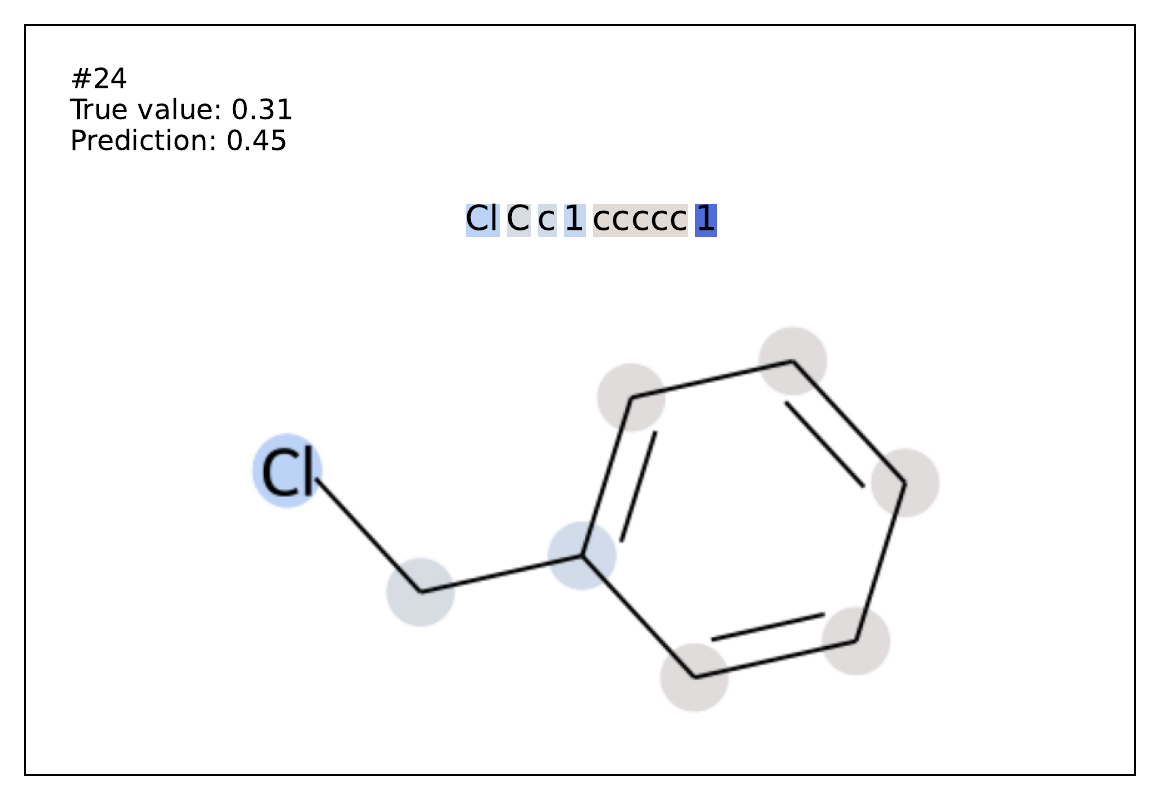} 
\end{subfigure}\begin{subfigure}[b]{0.33\textwidth} 
  \centering 
  \includegraphics[width=\textwidth]{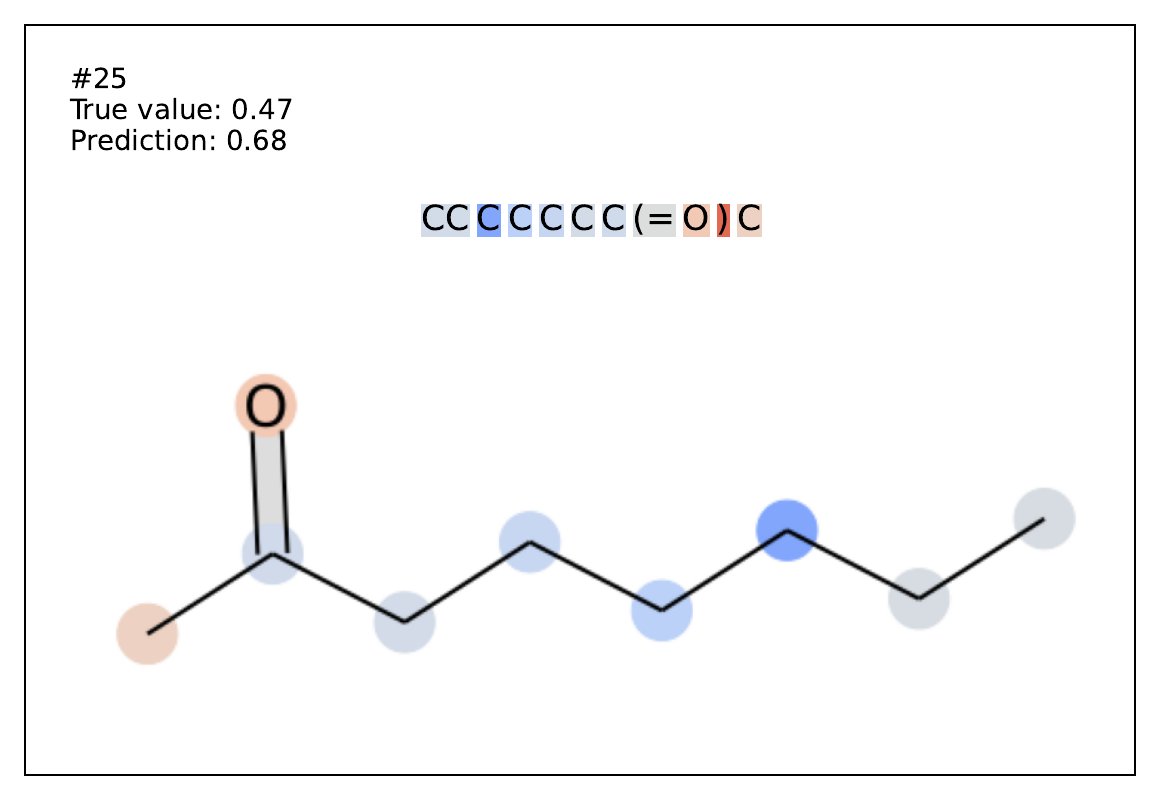} 
\end{subfigure}
\begin{subfigure}[b]{0.33\textwidth} 
  \centering 
  \includegraphics[width=\textwidth]{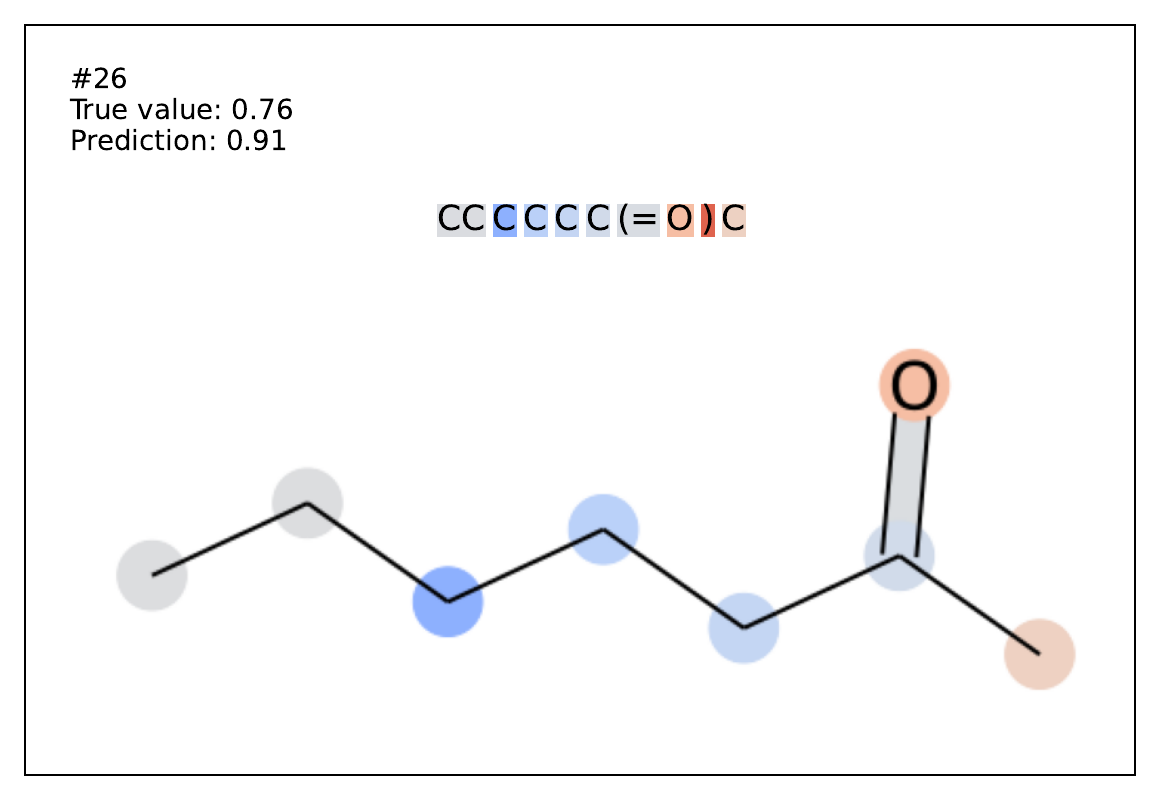} 
\end{subfigure}\begin{subfigure}[b]{0.33\textwidth} 
  \centering 
  \includegraphics[width=\textwidth]{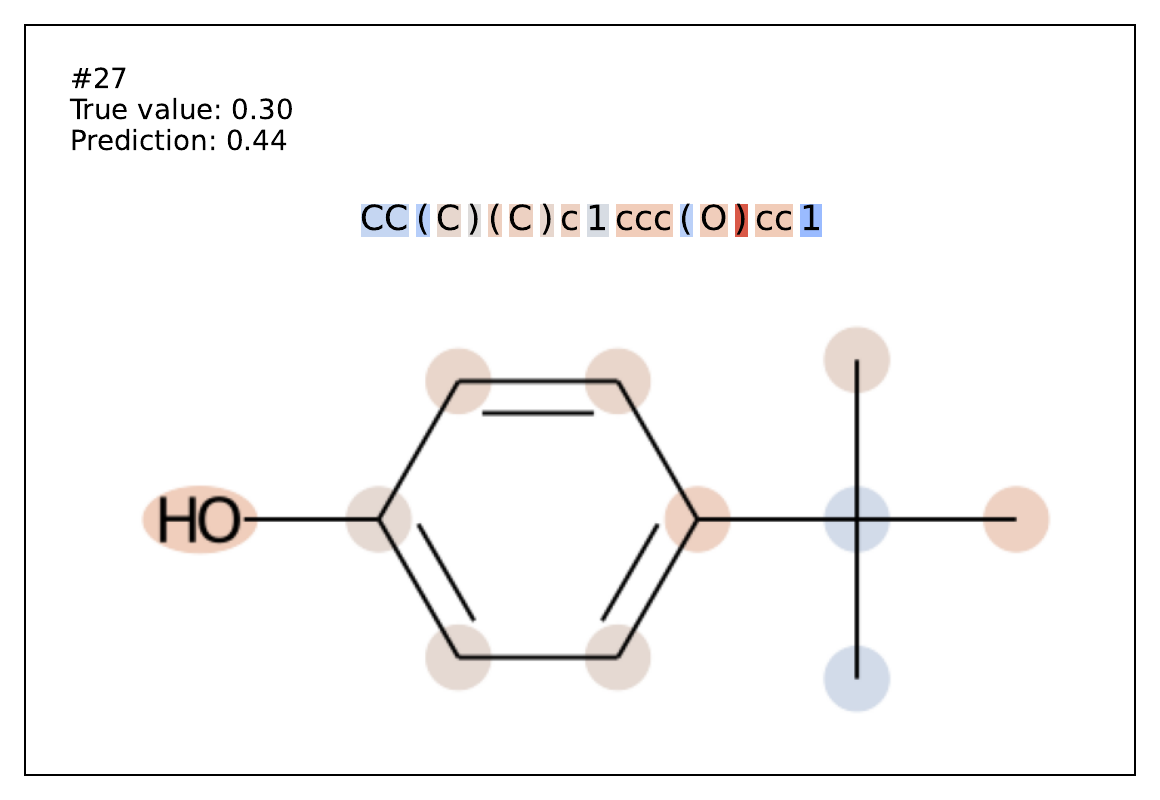} 
\end{subfigure}\begin{subfigure}[b]{0.33\textwidth} 
  \centering 
  \includegraphics[width=\textwidth]{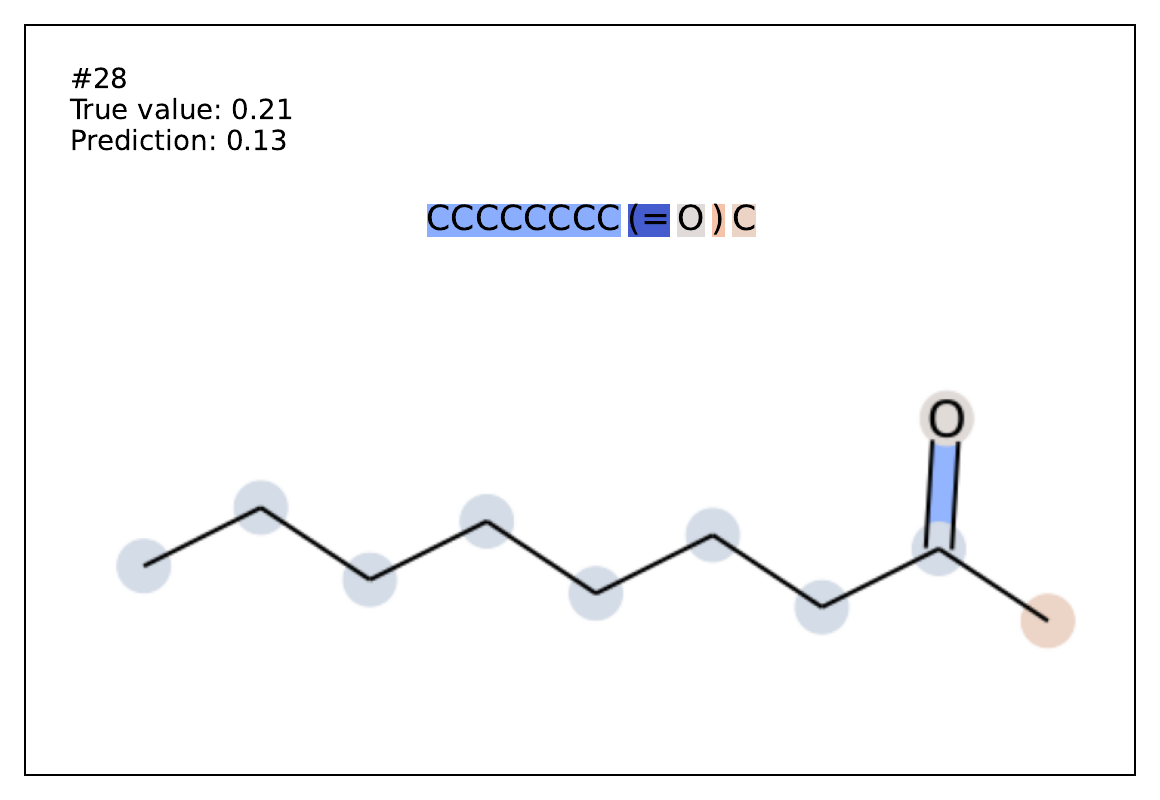} 
\end{subfigure}
\begin{subfigure}[b]{0.33\textwidth} 
  \centering 
  \includegraphics[width=\textwidth]{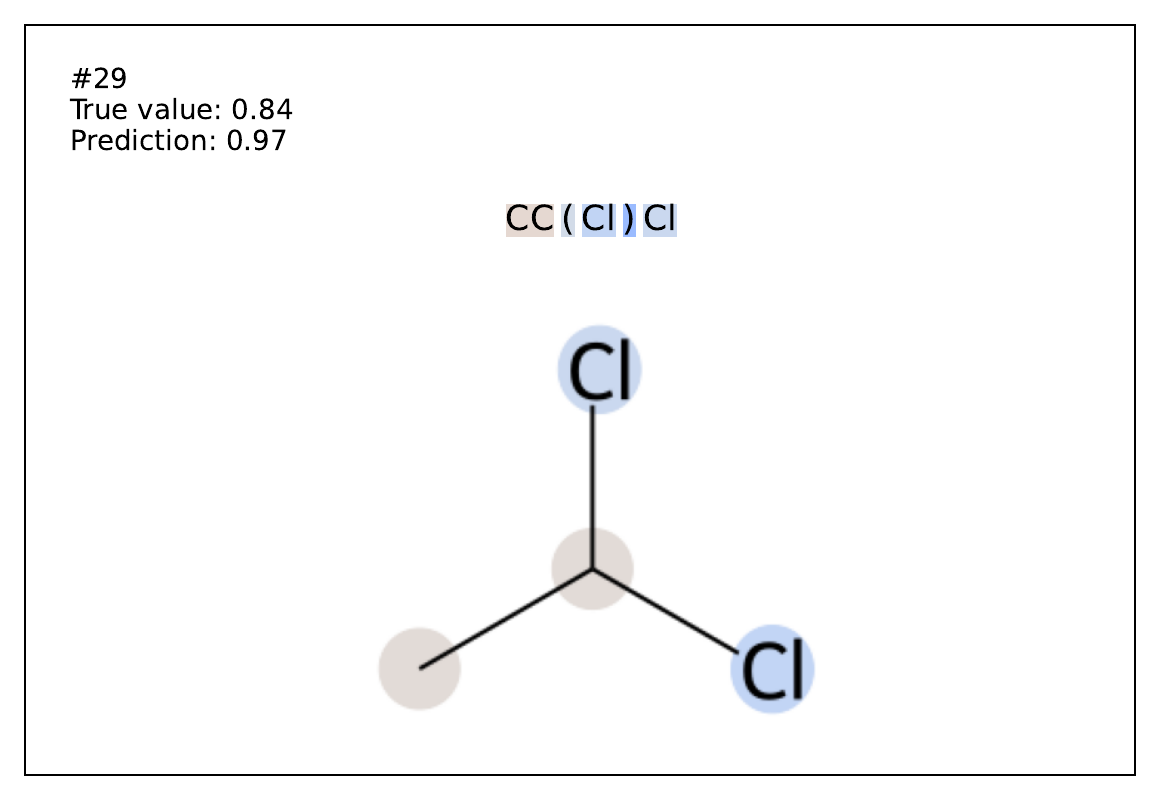} 
\end{subfigure}\begin{subfigure}[b]{0.33\textwidth} 
  \centering 
  \includegraphics[width=\textwidth]{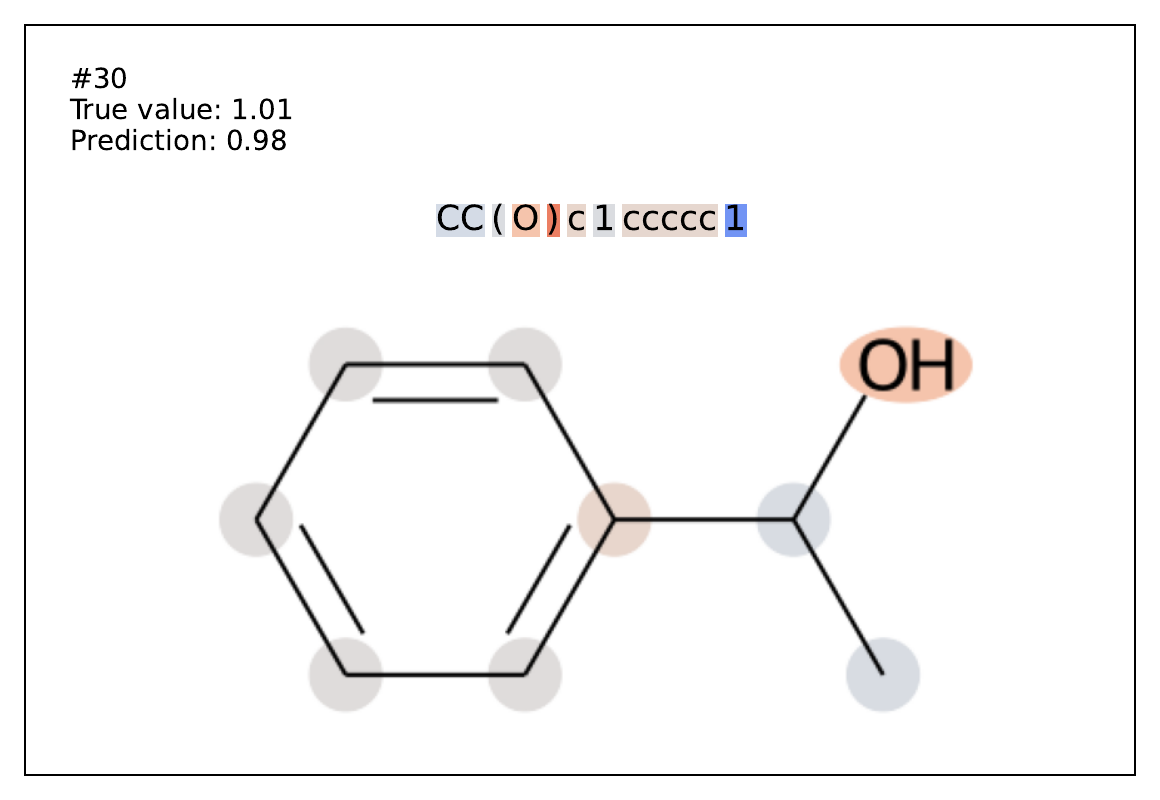} 
\end{subfigure}\begin{subfigure}[b]{0.33\textwidth} 
  \centering 
  \includegraphics[width=\textwidth]{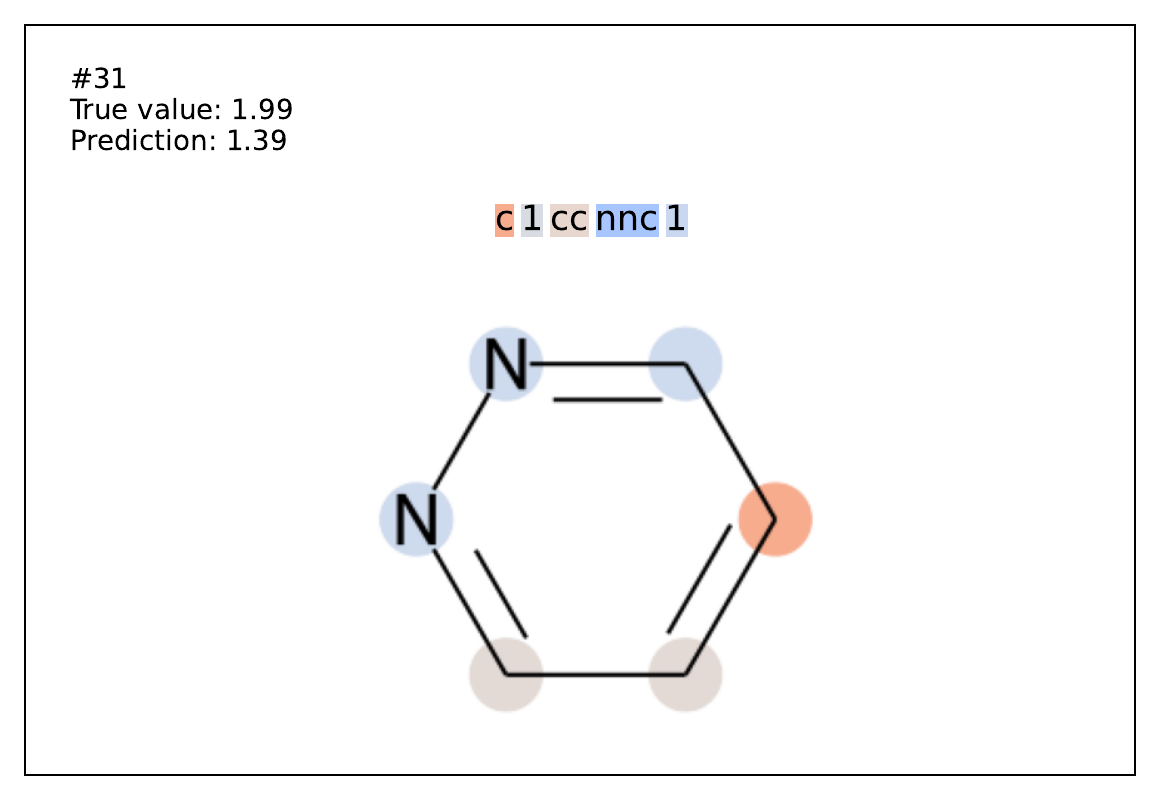} 
\end{subfigure}
\begin{subfigure}[b]{0.33\textwidth} 
  \centering 
  \includegraphics[width=\textwidth]{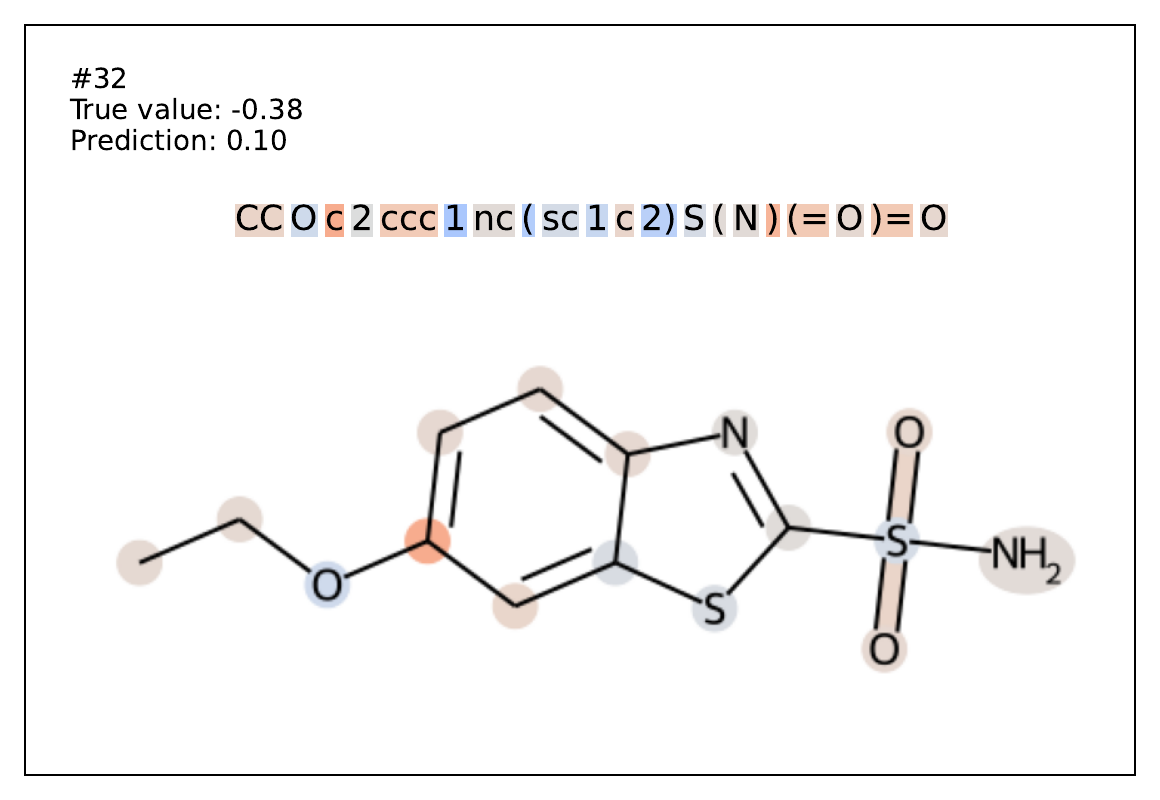} 
\end{subfigure}\begin{subfigure}[b]{0.33\textwidth} 
  \centering 
  \includegraphics[width=\textwidth]{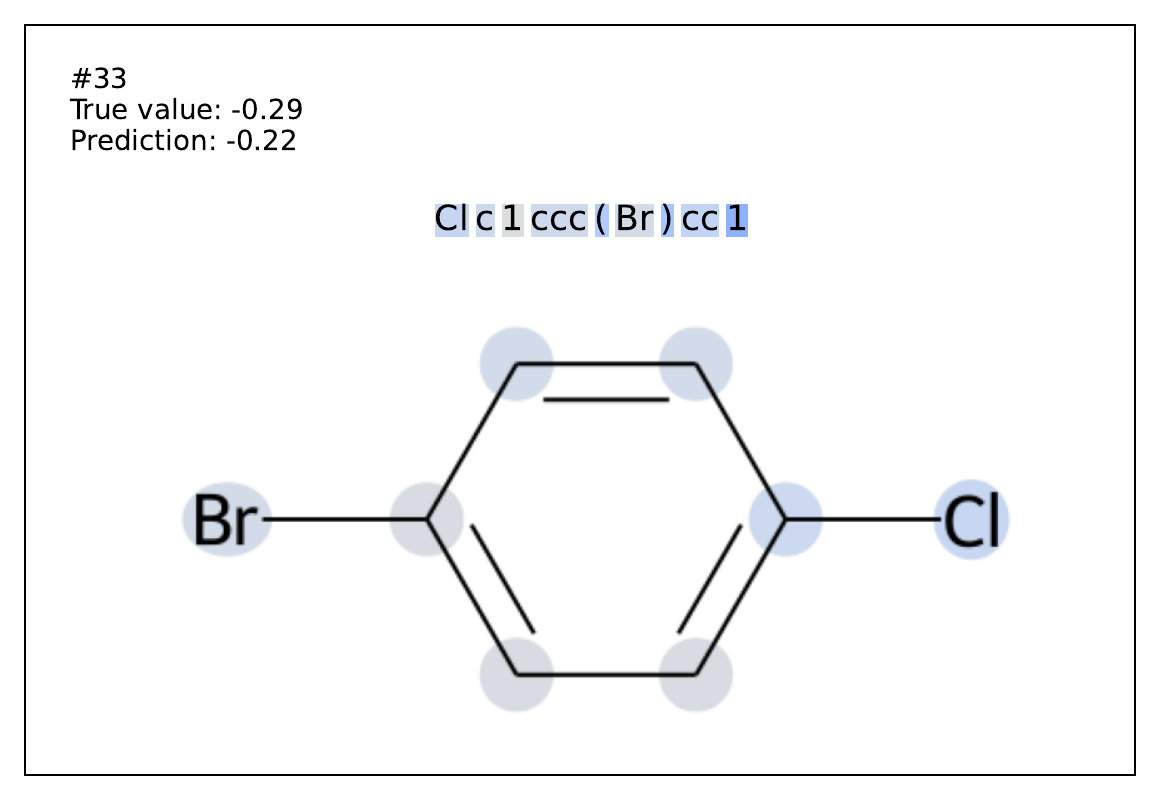} 
\end{subfigure}\begin{subfigure}[b]{0.33\textwidth} 
  \centering 
  \includegraphics[width=\textwidth]{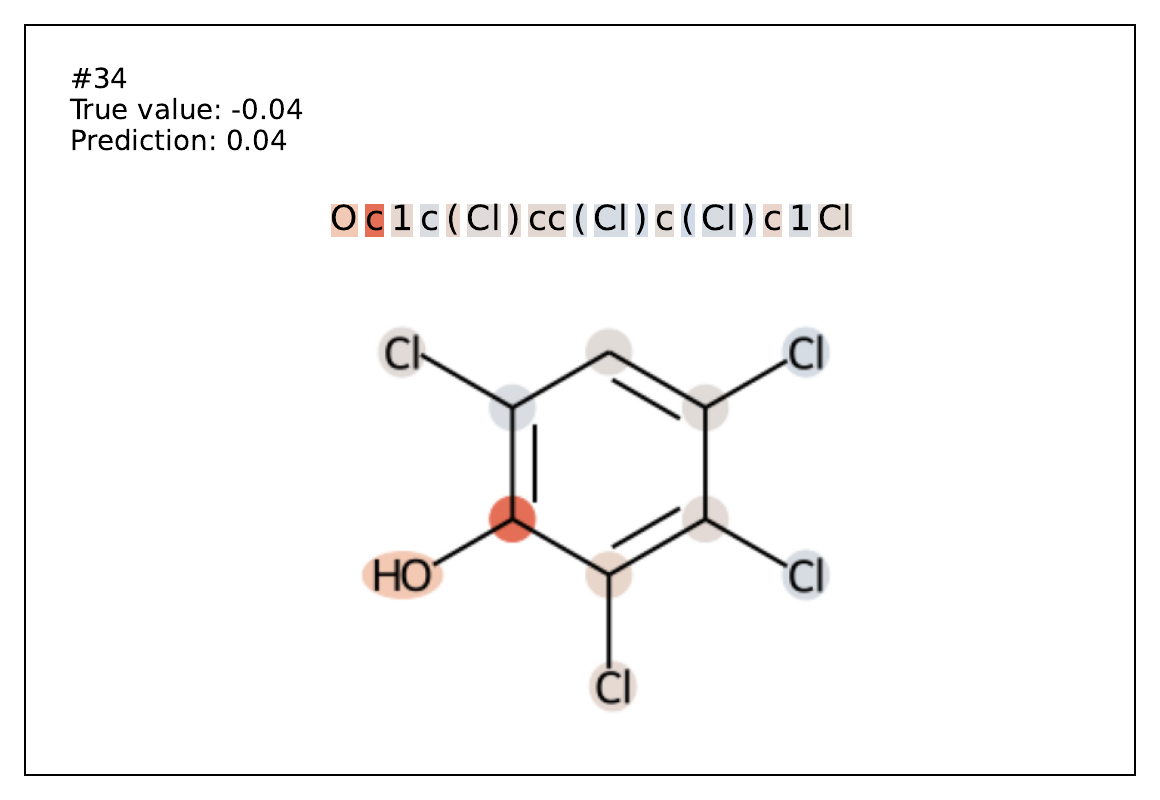} 
\end{subfigure}
\begin{subfigure}[b]{0.33\textwidth} 
  \centering 
  \includegraphics[width=\textwidth]{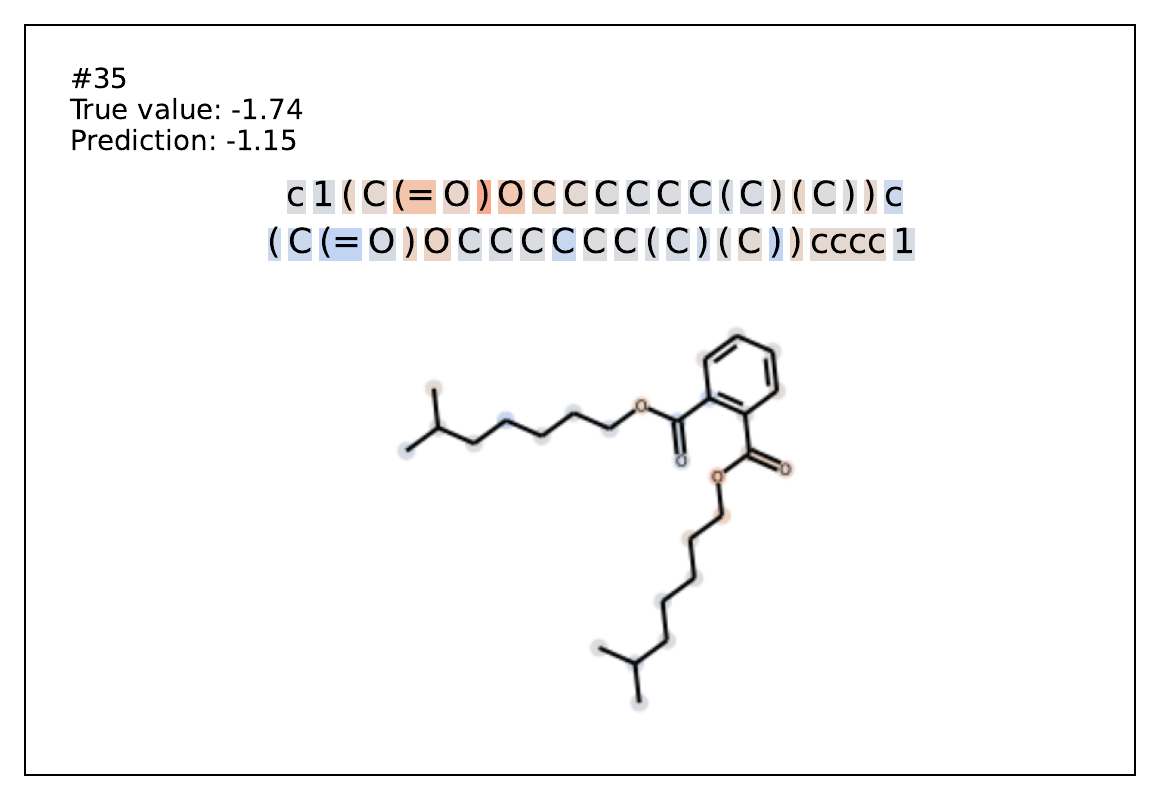} 
\end{subfigure}\begin{subfigure}[b]{0.33\textwidth} 
  \centering 
  \includegraphics[width=\textwidth]{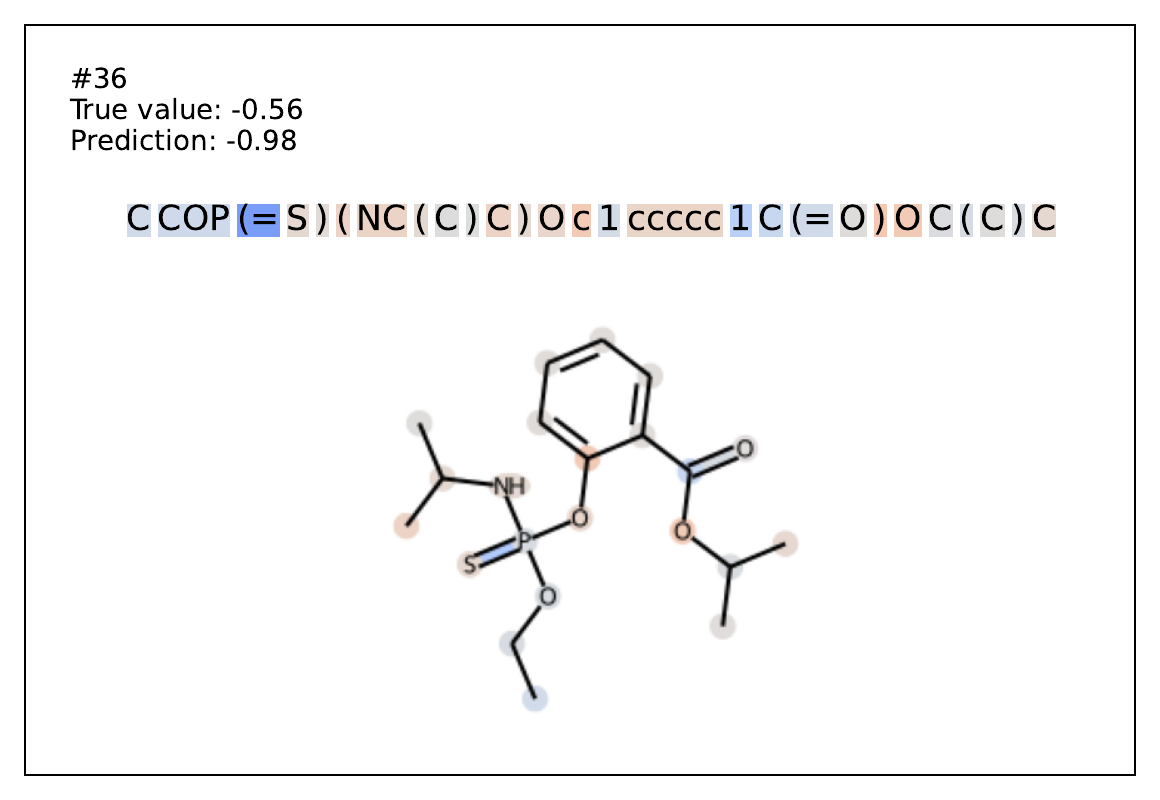} 
\end{subfigure}\begin{subfigure}[b]{0.33\textwidth} 
  \centering 
  \includegraphics[width=\textwidth]{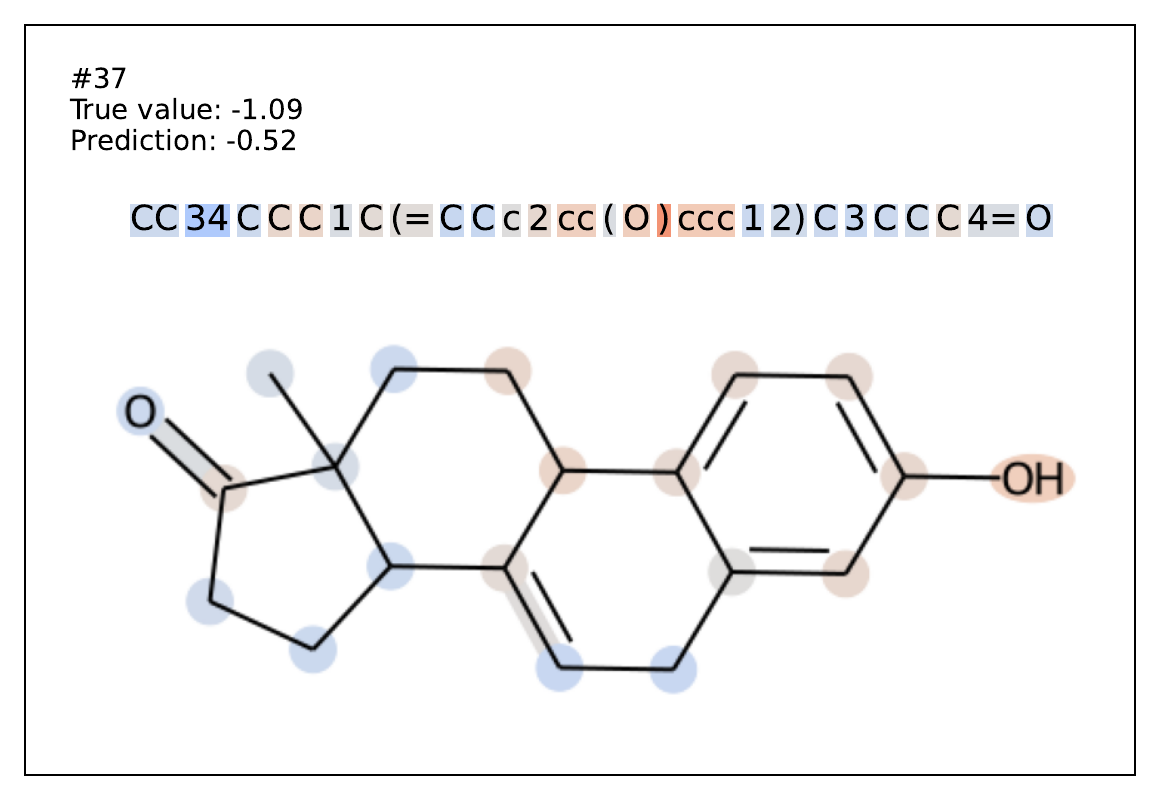} 
\end{subfigure}
\begin{subfigure}[b]{0.33\textwidth} 
  \centering 
  \includegraphics[width=\textwidth]{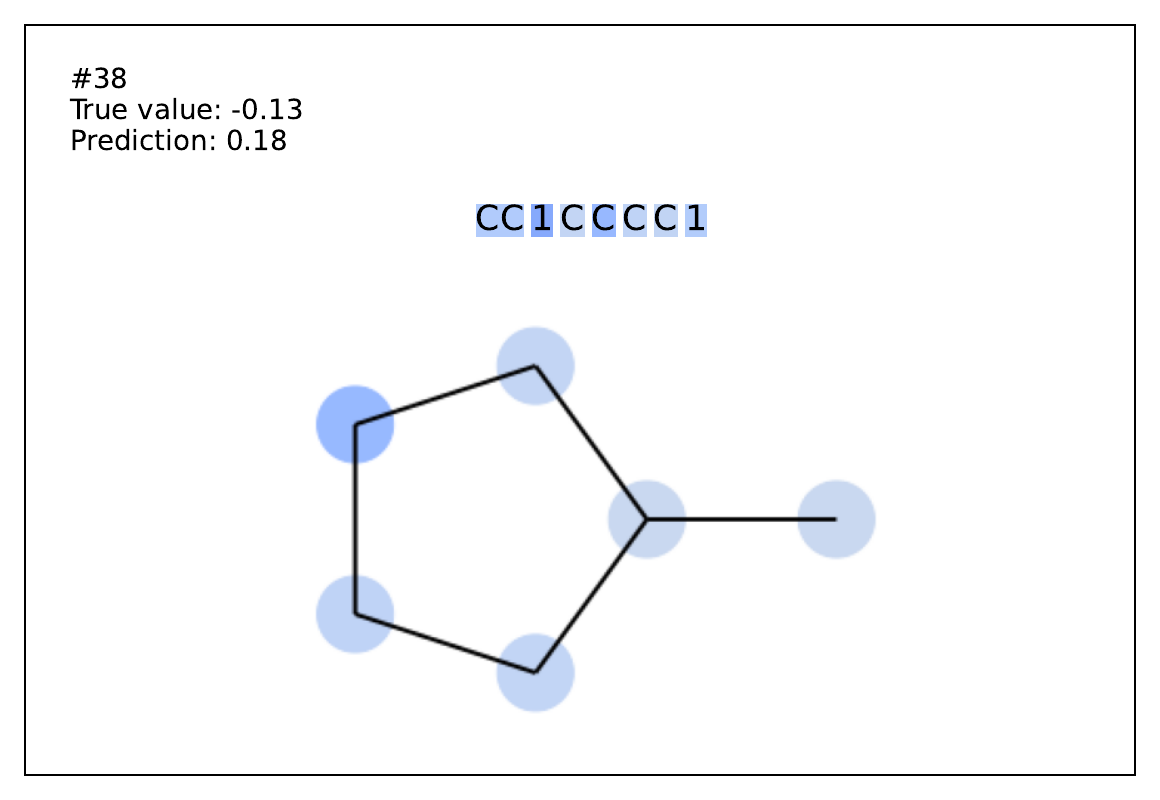} 
\end{subfigure}\begin{subfigure}[b]{0.33\textwidth} 
  \centering 
  \includegraphics[width=\textwidth]{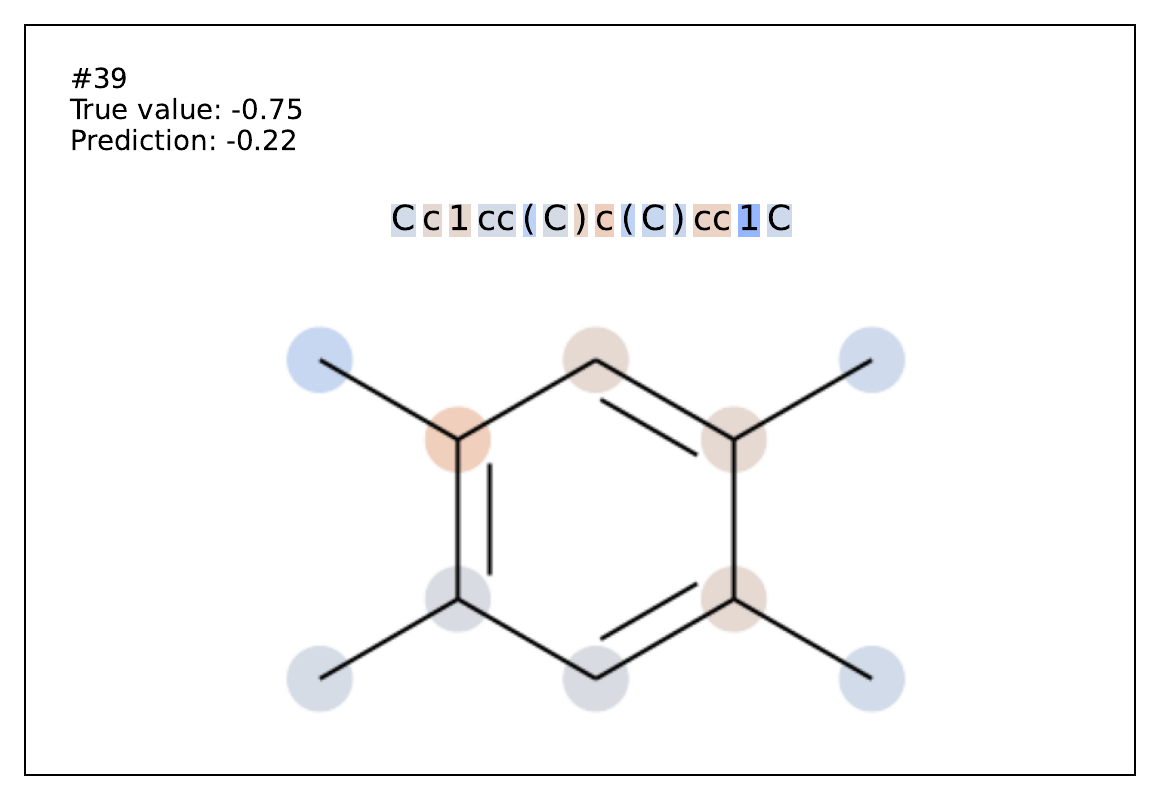} 
\end{subfigure}\begin{subfigure}[b]{0.33\textwidth} 
  \centering 
  \includegraphics[width=\textwidth]{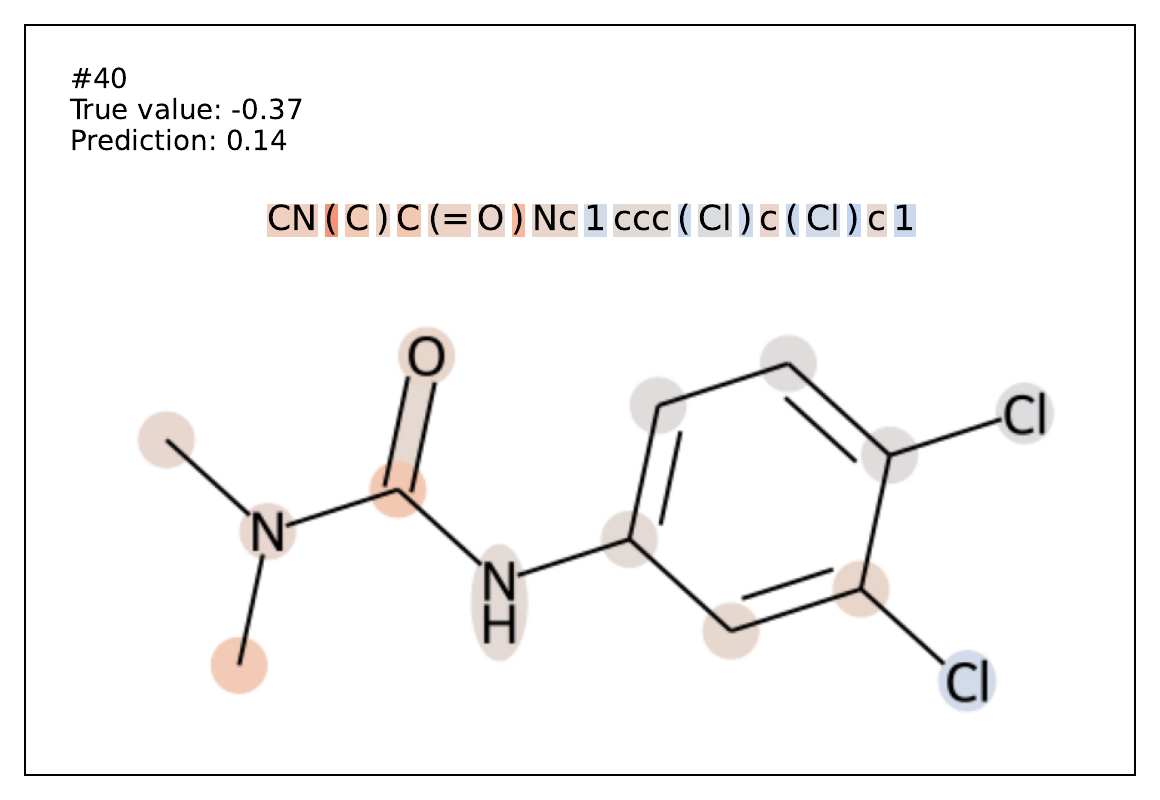} 
\end{subfigure}
\begin{subfigure}[b]{0.33\textwidth} 
  \centering 
  \includegraphics[width=\textwidth]{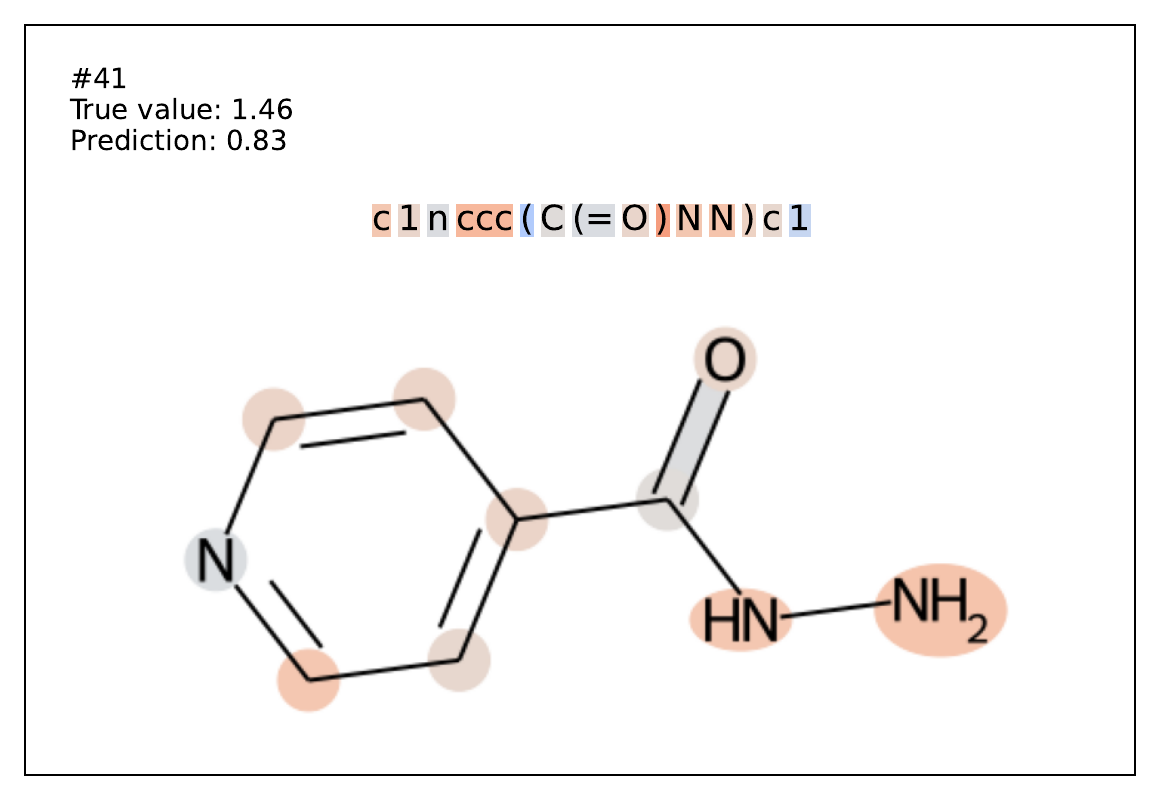} 
\end{subfigure}\begin{subfigure}[b]{0.33\textwidth} 
  \centering 
  \includegraphics[width=\textwidth]{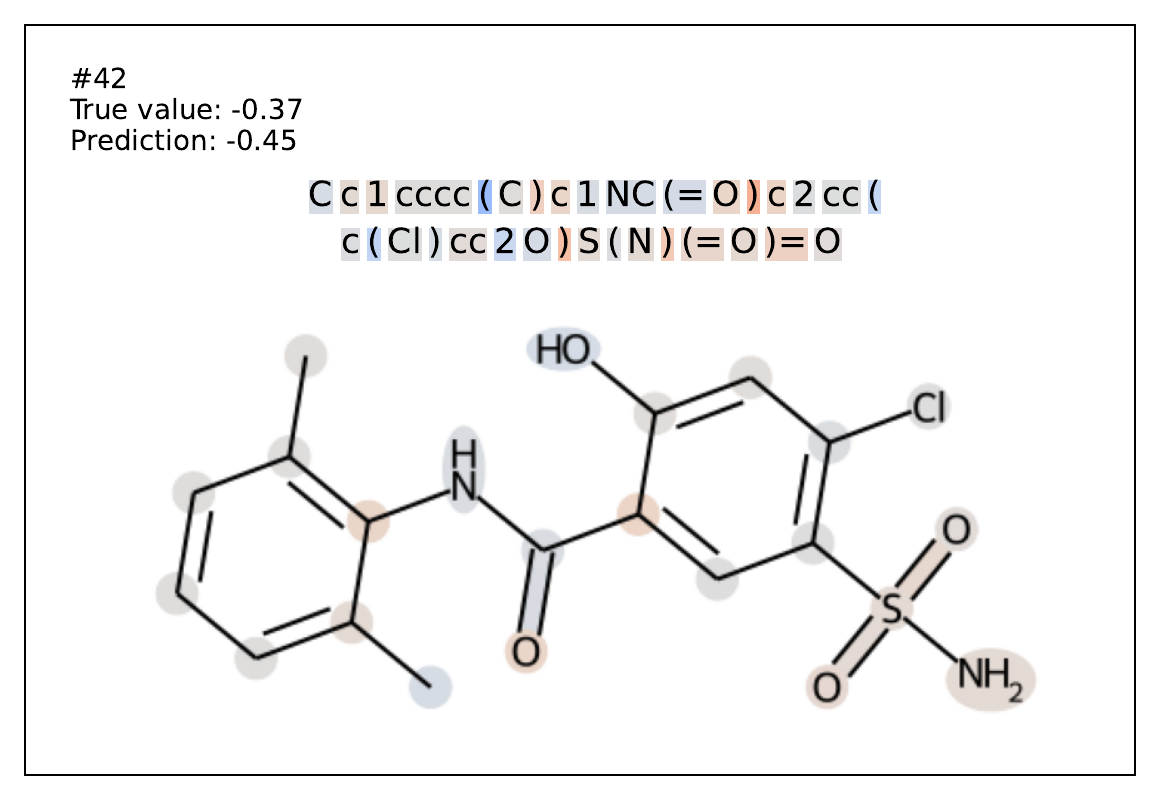} 
\end{subfigure}\begin{subfigure}[b]{0.33\textwidth} 
  \centering 
  \includegraphics[width=\textwidth]{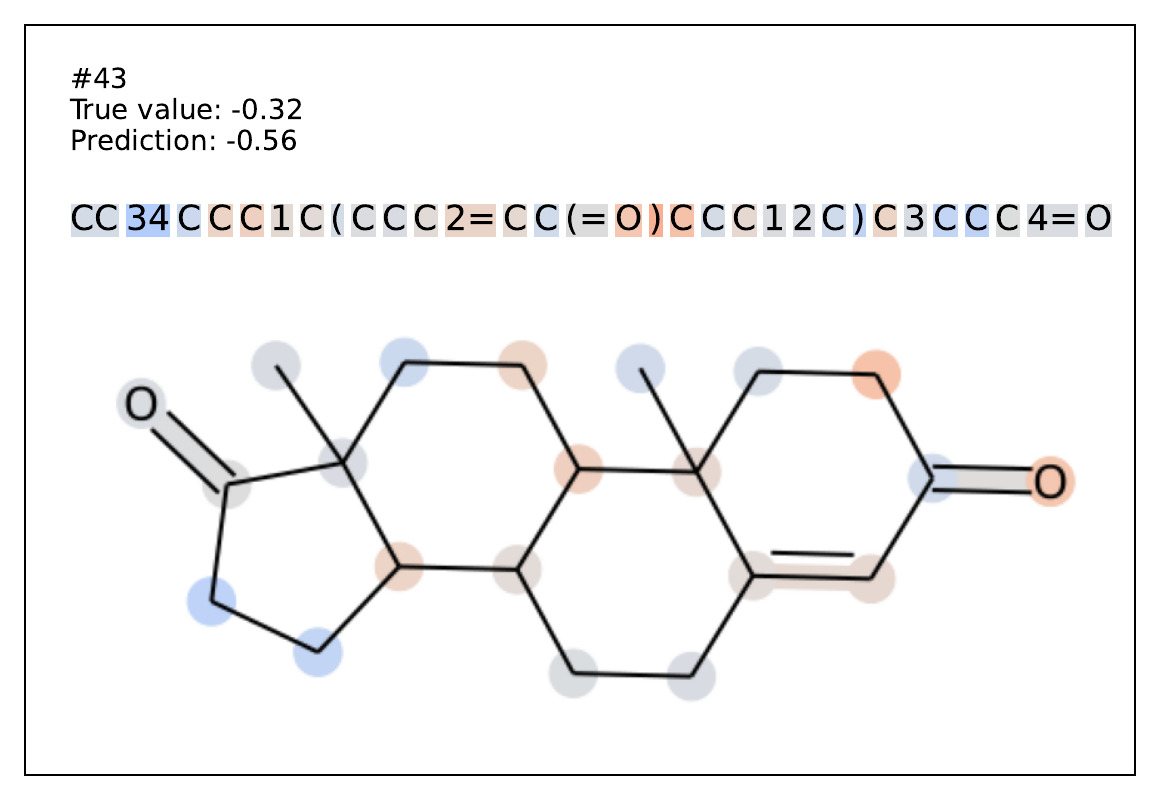} 
\end{subfigure}

\caption{Explaining predictions of the fine-tuned model on ESOL dataset. See Section \ref{sec:captum}. Part 2/3}
\label{fig:captum-esol-2}
\end{figure}

\begin{figure}[h]
\centering
\begin{subfigure}[b]{0.33\textwidth} 
  \centering 
  \includegraphics[width=\textwidth]{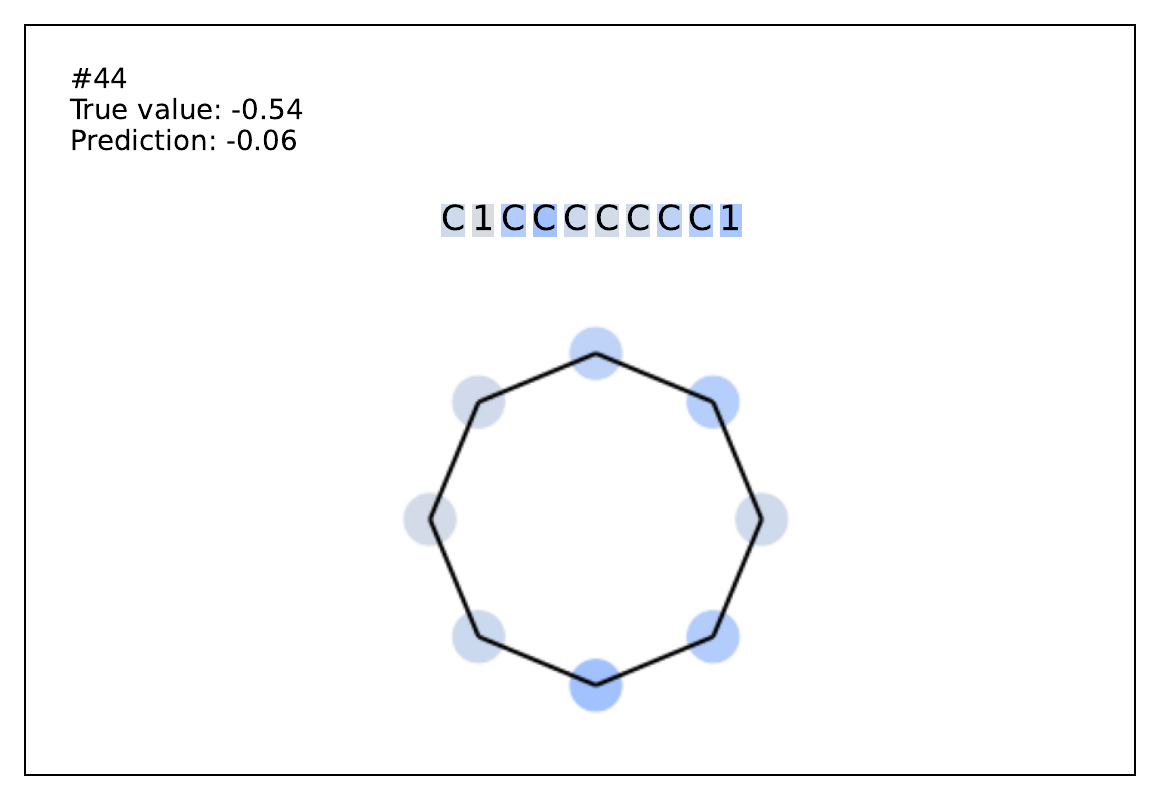} 
\end{subfigure}\begin{subfigure}[b]{0.33\textwidth} 
  \centering 
  \includegraphics[width=\textwidth]{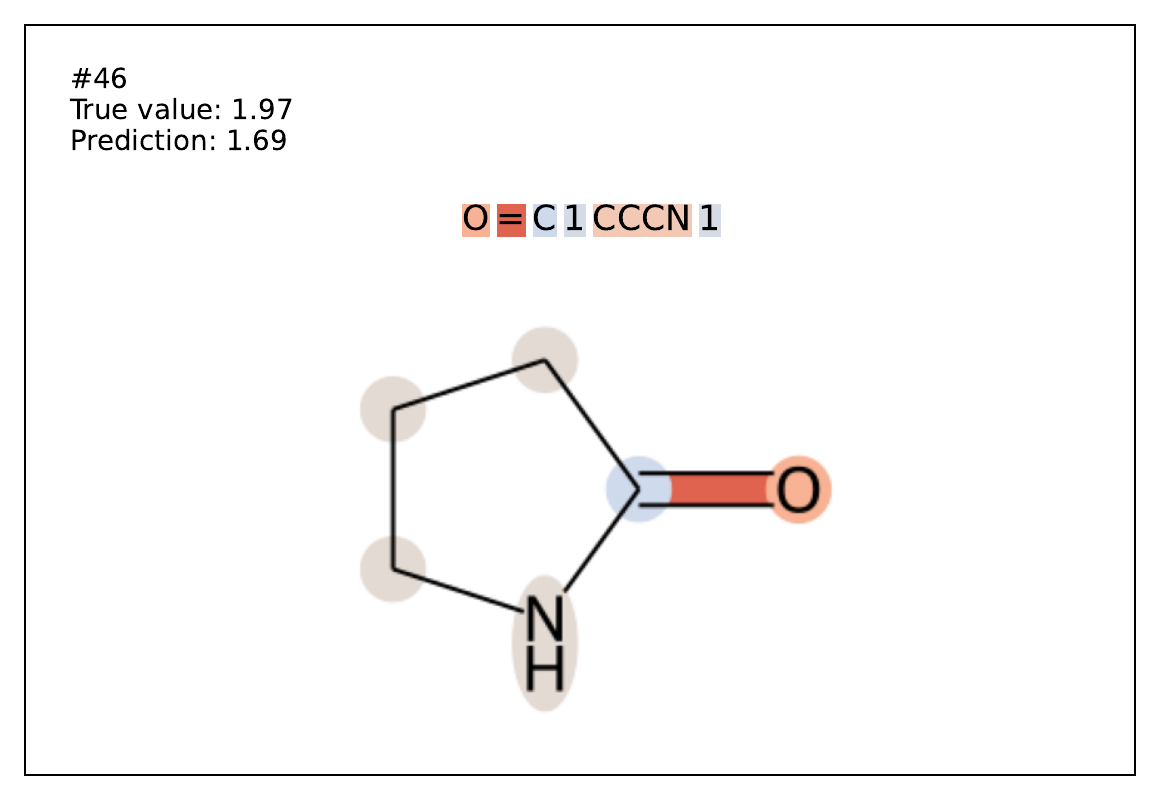} 
\end{subfigure}\begin{subfigure}[b]{0.33\textwidth} 
  \centering 
  \includegraphics[width=\textwidth]{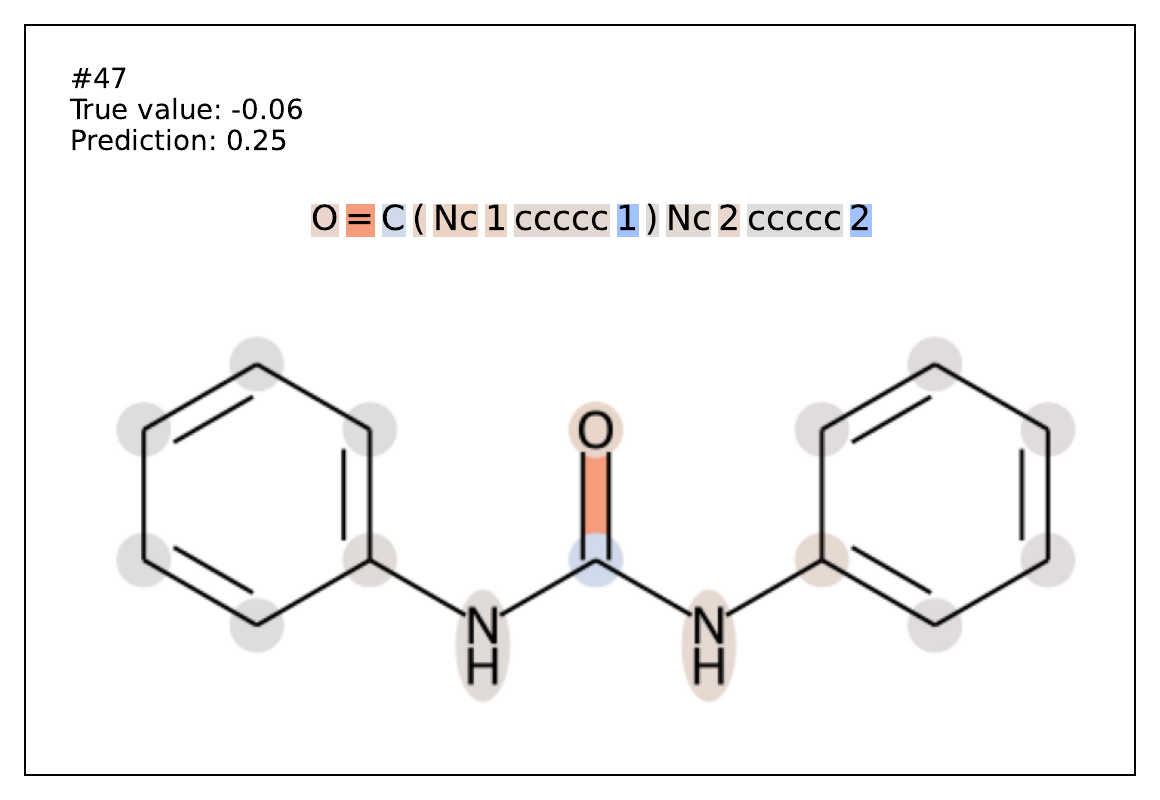} 
\end{subfigure}
\begin{subfigure}[b]{0.33\textwidth} 
  \centering 
  \includegraphics[width=\textwidth]{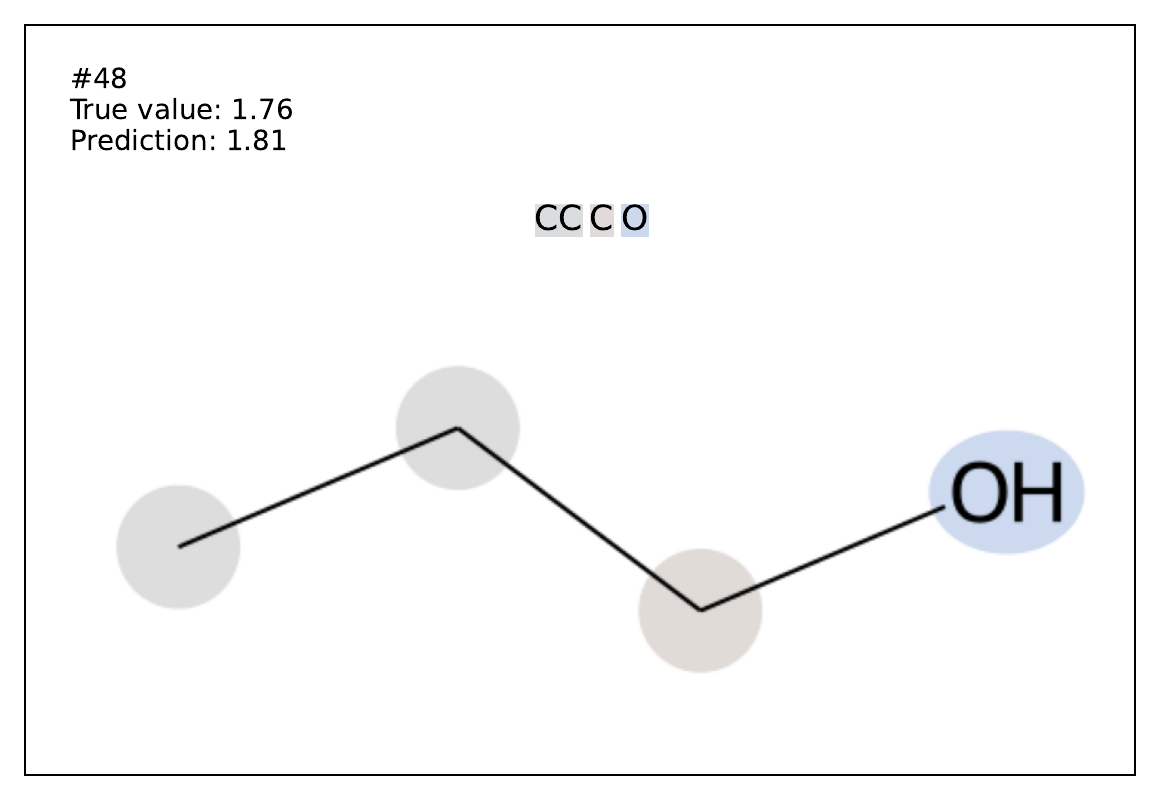} 
\end{subfigure}\begin{subfigure}[b]{0.33\textwidth} 
  \centering 
  \includegraphics[width=\textwidth]{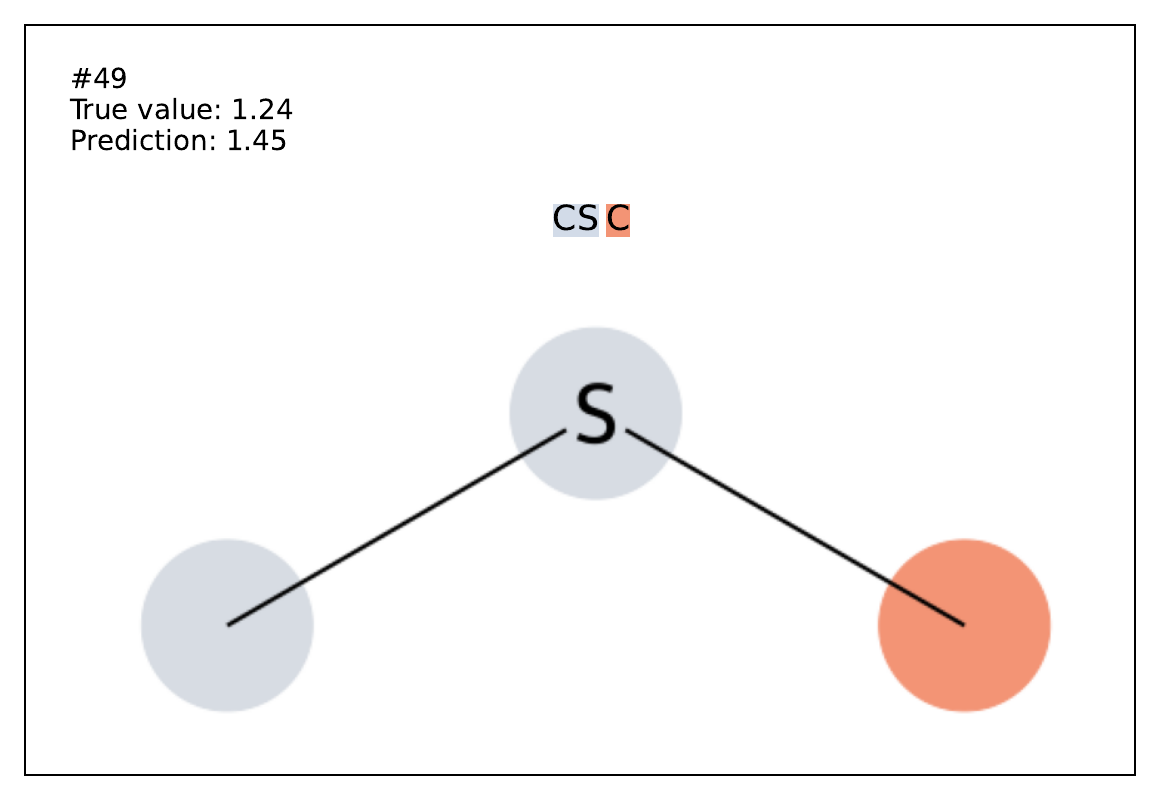} 
\end{subfigure}\begin{subfigure}[b]{0.33\textwidth} 
  \centering 
  \includegraphics[width=\textwidth]{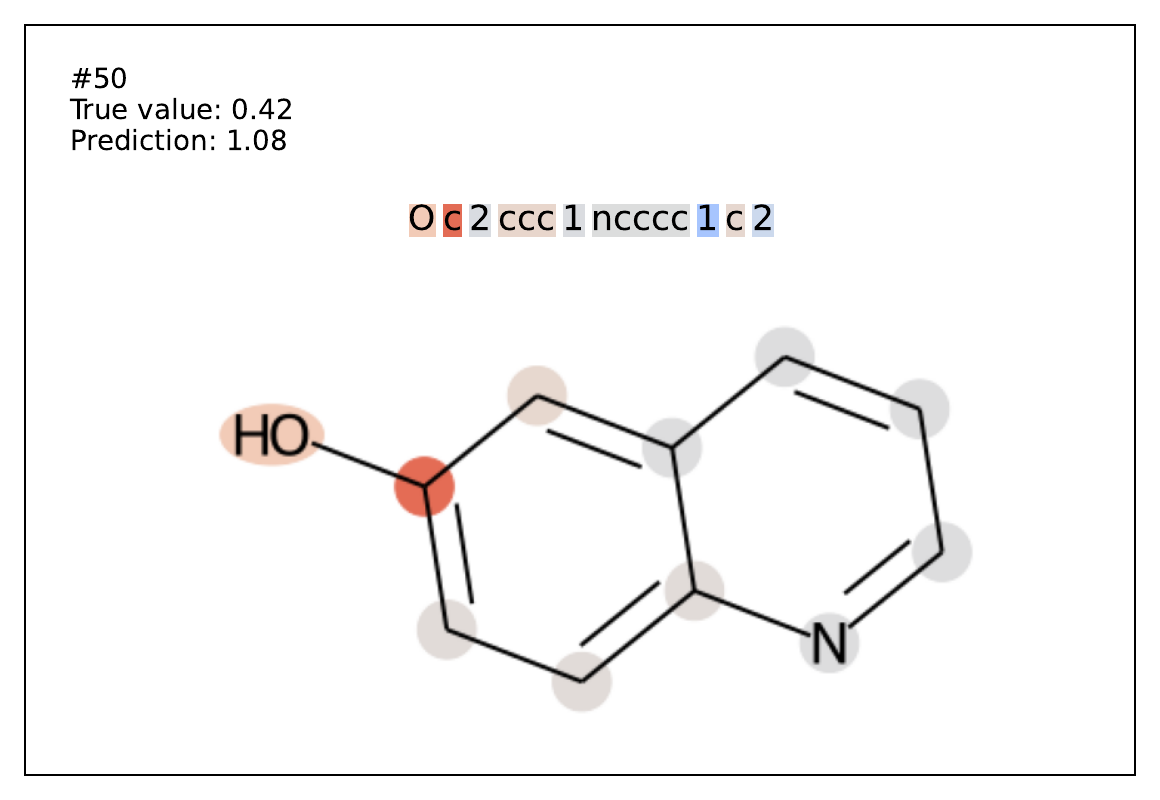} 
\end{subfigure}
\begin{subfigure}[b]{0.33\textwidth} 
  \centering 
  \includegraphics[width=\textwidth]{figures/esol/esol51.pdf} 
\end{subfigure}\begin{subfigure}[b]{0.33\textwidth} 
  \centering 
  \includegraphics[width=\textwidth]{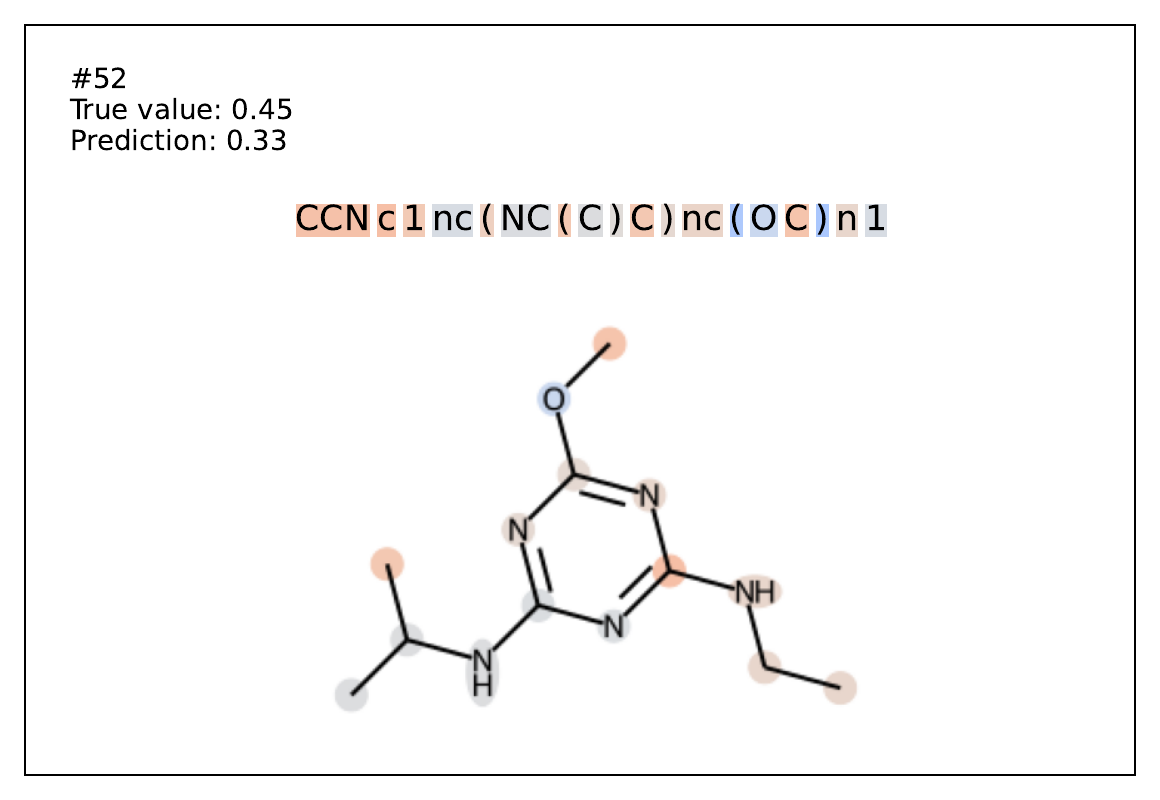} 
\end{subfigure}\begin{subfigure}[b]{0.33\textwidth} 
  \centering 
  \includegraphics[width=\textwidth]{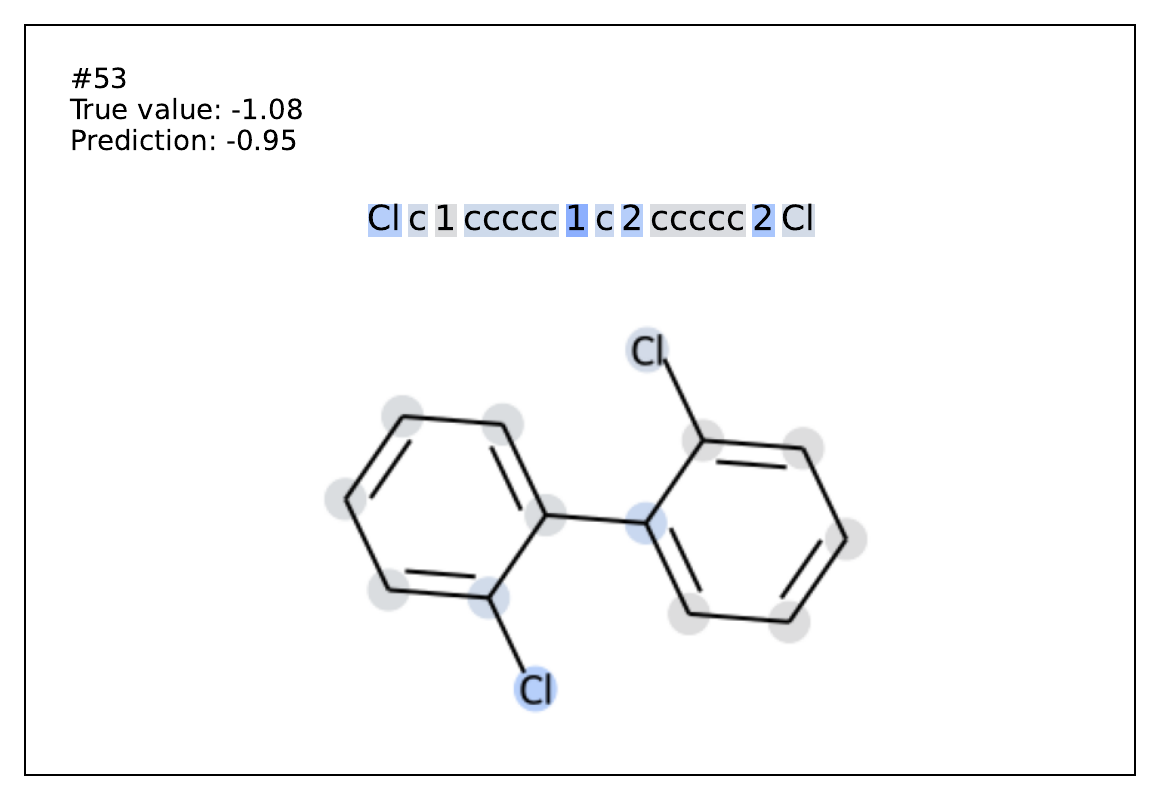} 
\end{subfigure}
\begin{subfigure}[b]{0.33\textwidth} 
  \centering 
  \includegraphics[width=\textwidth]{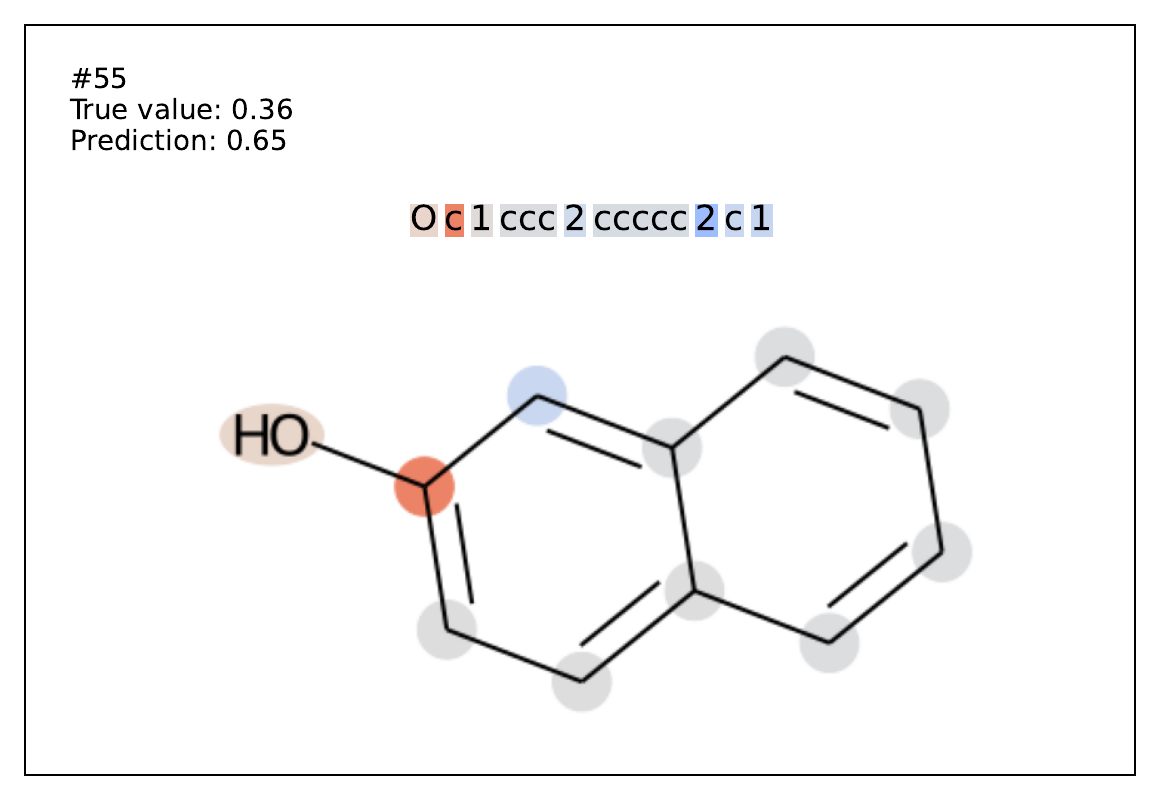} 
\end{subfigure}\begin{subfigure}[b]{0.33\textwidth} 
  \centering 
  \includegraphics[width=\textwidth]{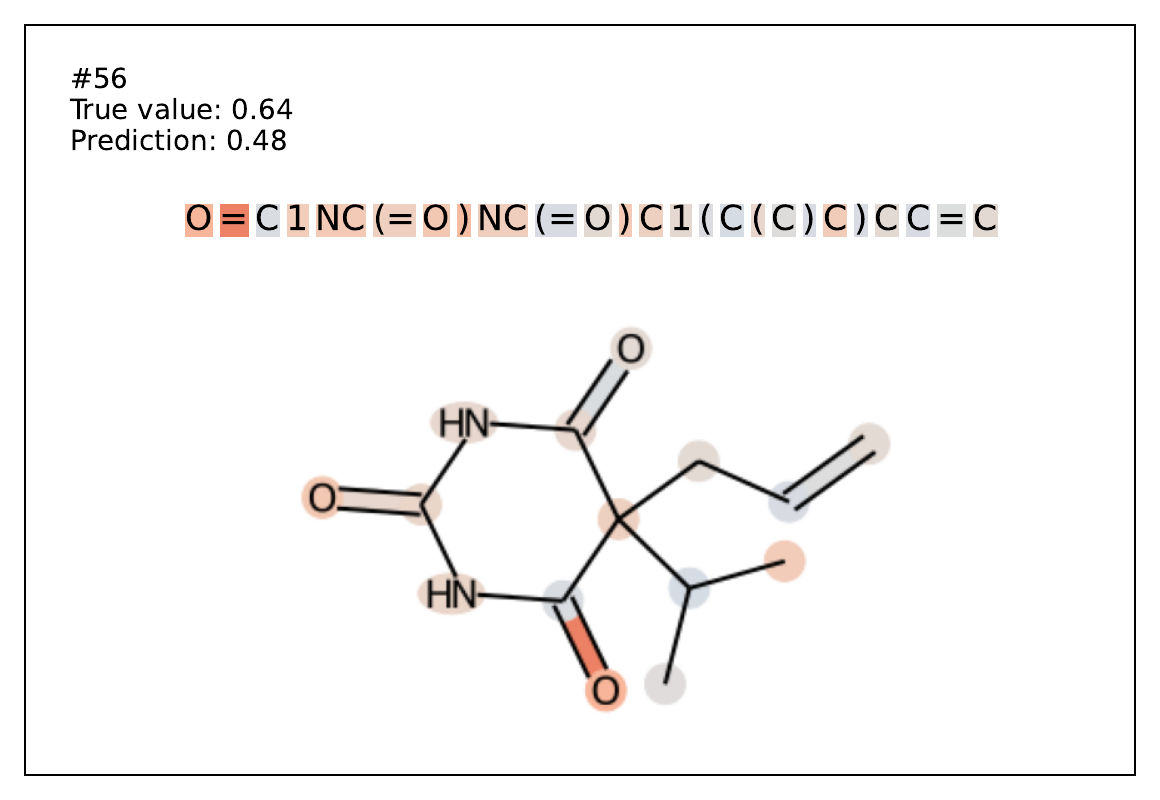} 
\end{subfigure}\begin{subfigure}[b]{0.33\textwidth} 
  \centering 
  \includegraphics[width=\textwidth]{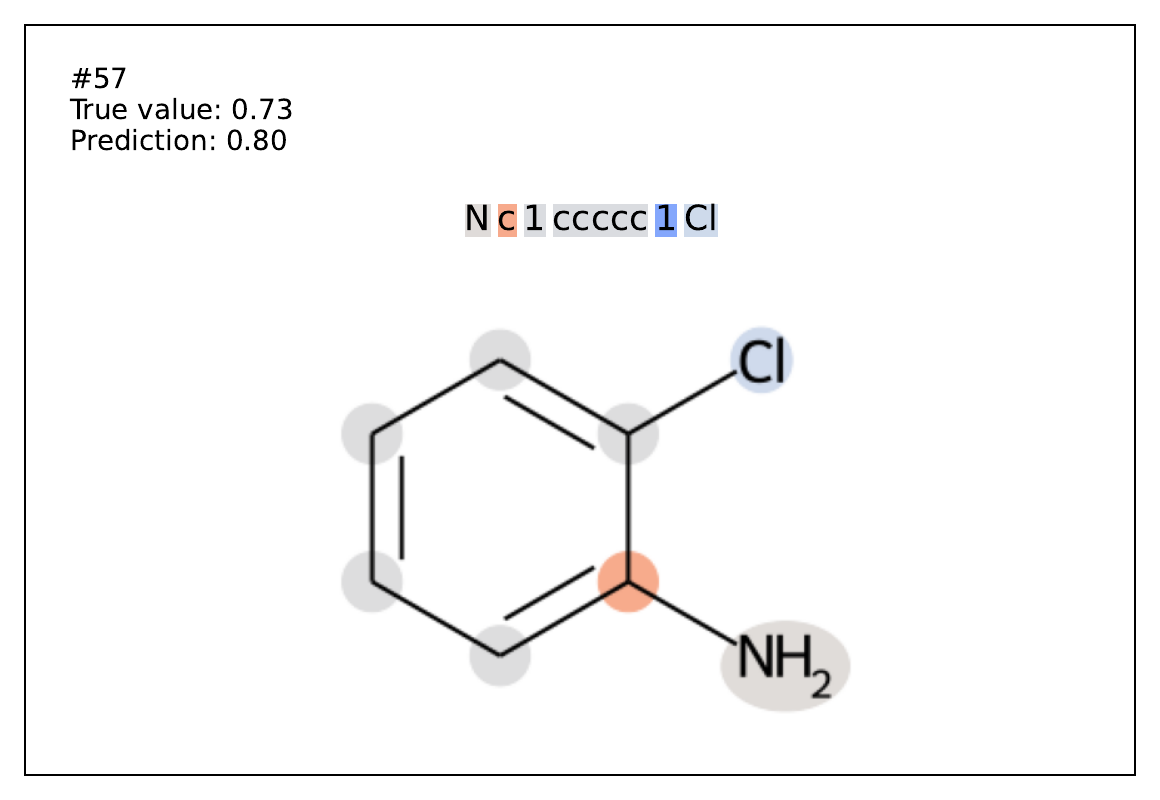} 
\end{subfigure}
\begin{subfigure}[b]{0.33\textwidth} 
  \centering 
  \includegraphics[width=\textwidth]{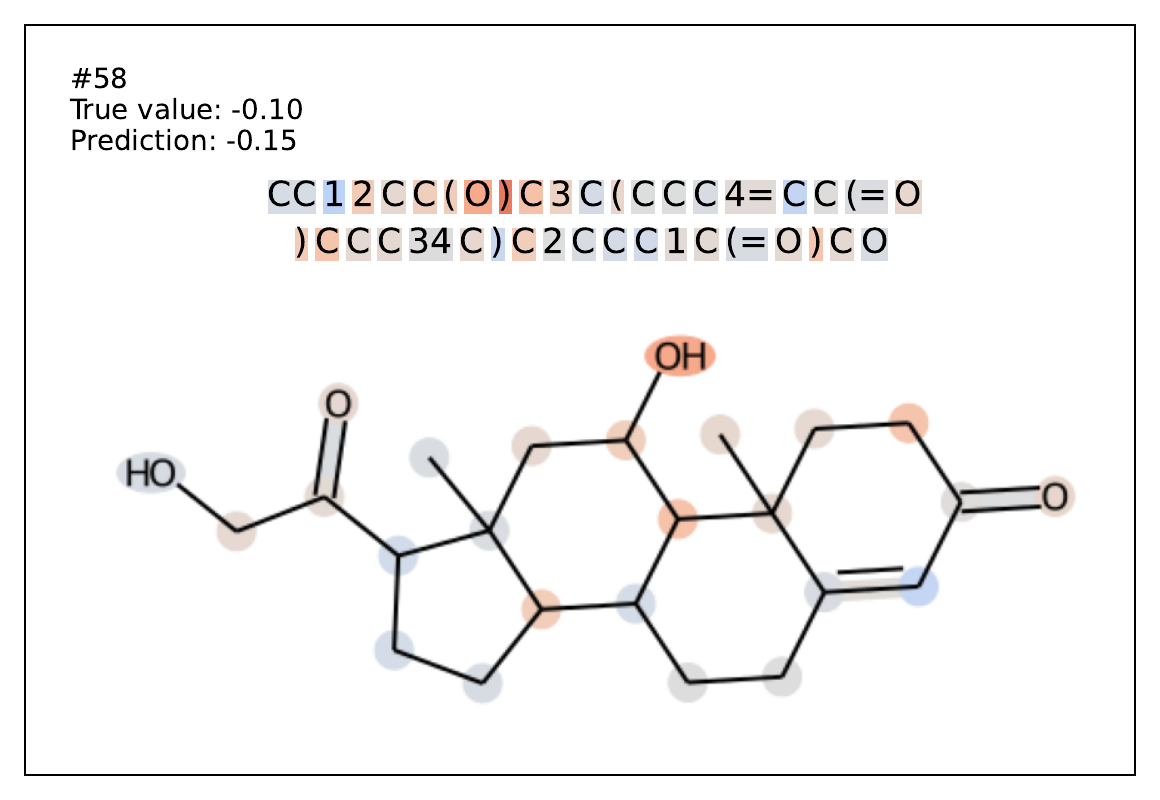} 
\end{subfigure}\begin{subfigure}[b]{0.33\textwidth} 
  \centering 
  \includegraphics[width=\textwidth]{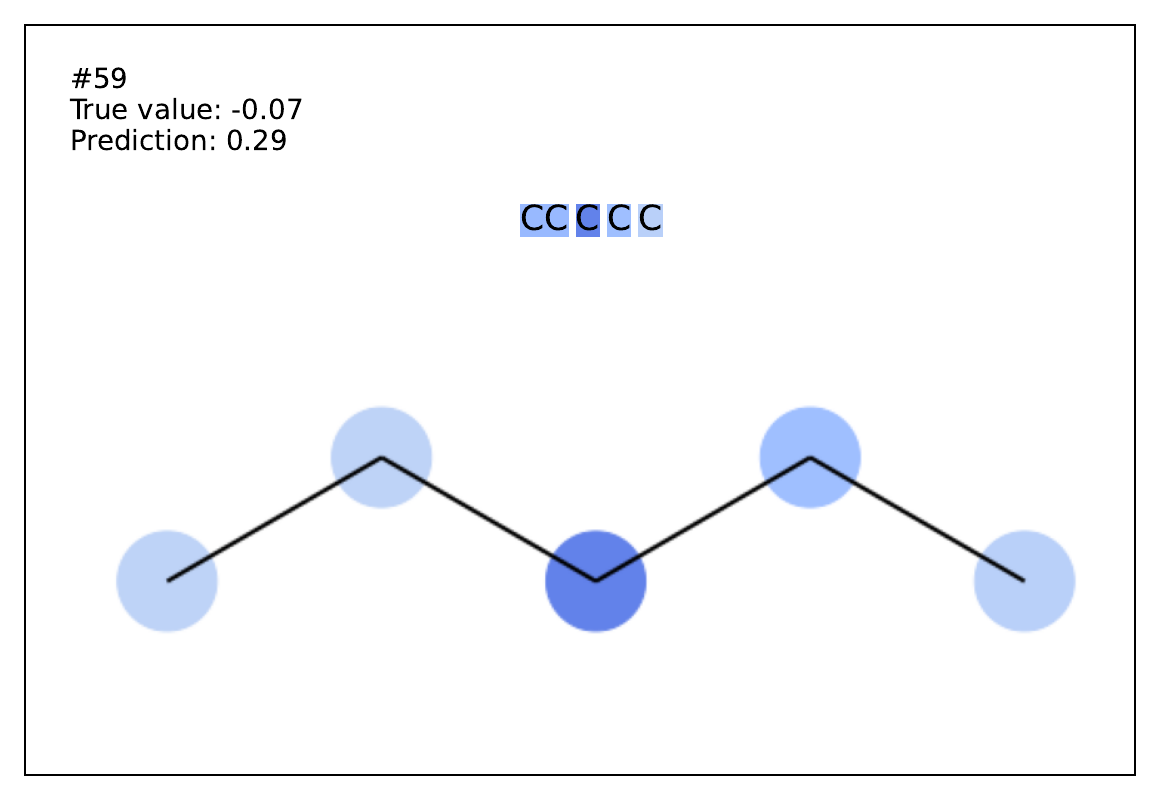} 
\end{subfigure}\begin{subfigure}[b]{0.33\textwidth} 
  \centering 
  \includegraphics[width=\textwidth]{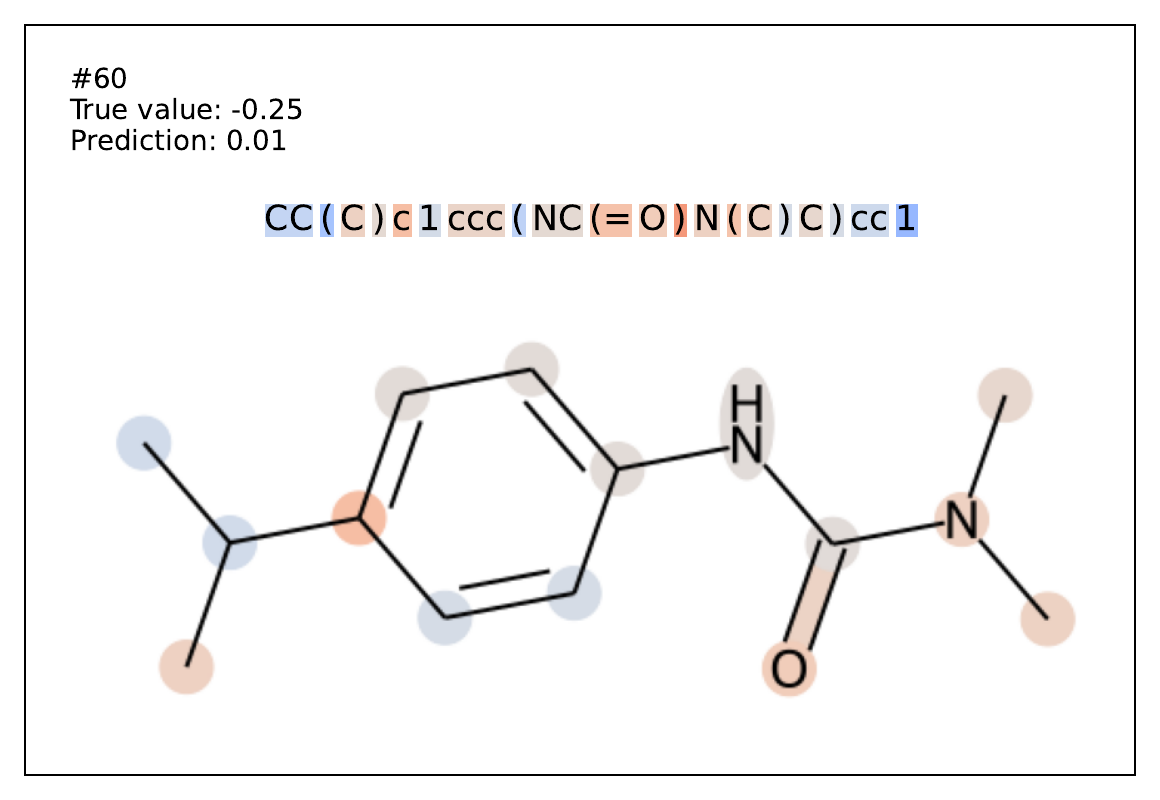} 
\end{subfigure}

\caption{Explaining predictions of the fine-tuned model on ESOL dataset. See Section \ref{sec:captum}. Part 3/3}
\label{fig:captum-esol-3}
\end{figure}

In \Cref{fig:captum-esol-1,fig:captum-esol-2,fig:captum-esol-3} we show the attributions on $60$ molecules from the test set of ESOL dataset. The images with indices 10, 16, 19, 27, 30, 37, 50, 51 and 55 contain hydroxyl groups, which are properly highlighted in red by the Integrated Gradient method. Same with carbonyl groups, as seen on molecules with indices 2, 5, 7, 12,%
25, 26, 46 and 47.

It is usually assumed that the noise in aqueous solubility datasets is roughly 0.7 \citet{ml-solubility-noise}. It is noteworthy that in all instances, the errors of the predicted values are less than that and, hence, can be assumed to be accurate. %

In one case, out of all molecules bearing the hydroxyl group, in compound \#48, the functional group was not highlighted and even marked as a negative contributor. Even so, the prediction value of solubility was correct. 

One more interesting example where the hydroxyl group has been recognized, but the prediction was incorrect (LogS ± 0.76) is compound \#34. In this compound, in addition to the hydroxyl group, four alkyl chloride (C-Cl) groups exist and contribute to the molecule's solubility (have a lower level of polarity). The lower proportion of alkyl chloride-bearing compounds in the training set may lead to underestimating these functional group contributions to the molecule's solubility. 

In the case of carbonyl groups, there are more examples where the Integrated Gradients method didn't highlight the functional groups, for instance, compounds under the numbers 13, 14, 17, 35, 36, 40, 41, 42 and 43. Note that logS values of all these compounds were in the range of 0 to $-2$, which is the range of less soluble compounds, compared to the solubility of aforementioned compounds with recognized carbonyl groups (in the range of 0 and higher, highly soluble).
Additionally, all compounds with unrecognized carbonyl groups are complex, polycyclic aromatic hydrocarbons, which, as mentioned in the Ames dataset, was more problematic for the analysis by the Integrated Gradients method.

Other interesting examples are in the aliphatic compounds, which are hydrocarbon compounds containing carbon and hydrogen joined together in straight chains with carbonyl groups (Compounds \# 7, 25, 26, 28) or hydroxyl groups (Compounds \# 19, 48, 51). A general rule for these compounds is that the longer the carbon chain, the lower the solubility in polar solvents such as water. Here, we see that the increasing amount of CH2 groups in a carbon chain enhances the hydrophobic effect, which decreases the solubility value. Hence, compounds \#28 and \#51 have the lowest solubility compared to the abovementioned examples. It is remarkable that the model somehow followed the rule, and predicted values also exhibit the same pattern. As to the Integrated Gradients method, the negative contribution of CH2 groups is constantly highlighted in all compounds, whereas peripheric CH2 groups have been marked as small contributions to solubility. 

There are other examples where the predictions were accurate, but the underlying chemistry is more complicated, and the highlighted substructures sometimes are unexplainable. 